\documentclass{article}
\usepackage{lmodern} 

\usepackage[preprint]{neurips_2025}

\usepackage[utf8]{inputenc} 
\usepackage[T1]{fontenc}    
\usepackage{hyperref}       
\usepackage{url}            
\usepackage{booktabs}       
\usepackage{amsfonts}       
\usepackage{nicefrac}       
\usepackage{microtype}      
\usepackage{xcolor}         
\usepackage{hyperref}
\usepackage{amsmath}
\usepackage{amssymb}
\usepackage{mathtools}
\usepackage{amsthm}
\usepackage{wrapfig}
\usepackage{caption}
\usepackage{xcolor}
\usepackage{enumitem} 
\usepackage{changepage} 

\usepackage{graphicx,subfigure}
\usepackage{array}
\usepackage{subcaption}

\usepackage[most]{tcolorbox} 

\usepackage[capitalize,noabbrev]{cleveref}
\usepackage[colorinlistoftodos,bordercolor=orange,backgroundcolor=orange!20,linecolor=orange,textsize=scriptsize]{todonotes} 
\usepackage{multirow}
\usepackage{algorithm,algpseudocode}
\usepackage{listings} 
\usepackage{soul}
\newcommand{\rom}[1]{\uppercase\expandafter{\romannumeral #1\relax}}

\newcounter{hypothesis}

\crefname{hypothesis}{Hypothesis}{Hypotheses}
\Crefname{hypothesis}{Hypothesis}{Hypotheses}

\newcommand{\takeaway}[1]{
    \refstepcounter{takeaway}
    \phantomsection\label{#1}
    \textbf{Takeaway \thetakeaway.}
}
\newcounter{takeaway}

\crefname{takeaway}{Takeaway}{Takeaways}
\Crefname{takeaway}{Takeaway}{Takeaways}

\newcommand{\prettybox}[1]{
\begin{tcolorbox}[
  colback=orange!20,
  colframe=orange!80!black,
  fonttitle=\bfseries,
  boxrule=0.8pt,
  arc=3mm,
  left=4pt,
  right=4pt,
  top=4pt,
  bottom=4pt,
  width=\textwidth,
  before skip=4pt,
  after skip=4pt,
  ]
#1
\end{tcolorbox}
}

\definecolor{pycharmBackground}{RGB}{245,245,220}
\definecolor{pycharmKeyword}{RGB}{255,0,0}
\definecolor{pycharmSpecialPurple}{RGB}{153,0,153}
\definecolor{pycharmMethodBlue}{RGB}{30,144,255}
\definecolor{pycharmComment}{RGB}{111,114,116}
\definecolor{pycharmString}{RGB}{0,100,0}
\definecolor{pycharmBasicText}{RGB}{0,0,0}
\lstdefinestyle{pycharmStyle}{
  backgroundcolor=\color{pycharmBackground},   
  commentstyle=\color{pycharmComment}\itshape,
  keywordstyle=\color{pycharmKeyword},
  stringstyle=\color{pycharmString},
  basicstyle=\fontsize{6.5}{6.5}\ttfamily\color{pycharmBasicText},
  breakatwhitespace=true,         
  breaklines=true,                 
  captionpos=b,                    
  keepspaces=true,                 
  numberstyle=\tiny\color{pycharmComment},
  numbersep=2pt,                  
  showspaces=false,                
  showstringspaces=false,
  showtabs=false,                  
  tabsize=2,
  captionpos=t,
  emphstyle={\color{pycharmSpecialPurple}},
  linewidth=0.98\columnwidth,
  frame=none,    
  xrightmargin=0pt,
  xleftmargin=0.23cm,
  numbers=left,
  aboveskip=0.4cm,
  belowskip=0.4cm,
  morekeywords={from,class,def,if,else,return,isinstance,torch,super,dict,ValueError,closure,function},
  keywordstyle=[2]\color{pycharmSpecialPurple},
  keywords=[2]{isinstance,torch,super,dict,ValueError,closure,function},
  moredelim=[is][\color{pycharmMethodBlue}]{.}{(},
}
\def\beq{\begin{equation}}
\def\eeq{\end{equation}}
\def\ba{\begin{array}}
\def\ea{\end{array}}

\providecommand{\hh}{\boldsymbol{h}}

\providecommand{\mm}{\boldsymbol{m}}

\providecommand{\qq}{\boldsymbol{q}}
\providecommand{\rr}{\boldsymbol{r}}

\renewcommand{\ss}{\boldsymbol{s}}

\providecommand{\vv}{\boldsymbol{v}}
\providecommand{\ww}{\boldsymbol{w}}
\providecommand{\xx}{\boldsymbol{x}}
\providecommand{\yy}{\boldsymbol{y}}
\providecommand{\zz}{\boldsymbol{z}}

\newcommand{\bg}{\boldsymbol{g}}

\newcommand{\xixi}{\boldsymbol{\xi}}
\newcommand{\el}{\boldsymbol{l}}

\title{Benchmarking Optimizers\\ for Large Language Model Pretraining}

\author{%
  Andrei Semenov \\
  EPFL \\
  \texttt{andrii.semenov@epfl.ch} \\
  \And
  Matteo Pagliardini \\
  EPFL \\
  \texttt{matteo.pagliardini@epfl.ch} \\
  \And
  Martin Jaggi \\
  EPFL \\
  \texttt{martin.jaggi@epfl.ch} \\
}

\begin{document}

\maketitle

\definecolor{hotpink}{RGB}{255, 105, 180}
\begin{center}
    \href{https://github.com/epfml/llm-optimizer-benchmark}{\textcolor{hotpink}{https://github.com/epfml/llm-optimizer-benchmark}}
\end{center}

\begin{abstract}
    The recent development of Large Language Models (LLMs) has been accompanied by an effervescence of novel ideas and methods to better optimize the loss of deep learning models.
    Claims from those methods are myriad: from faster convergence to removing reliance on certain hyperparameters. However, the diverse experimental protocols used to validate these claims make direct comparisons between methods challenging. 
    This study presents a comprehensive evaluation of recent optimization techniques across standardized LLM pretraining scenarios, systematically varying model size, batch size, and training duration.
    Through careful tuning of each method, we provide guidance to practitioners on which optimizer is best suited for each scenario.
    For researchers, our work highlights promising directions for future optimization research. Finally, by releasing our code and making all experiments fully reproducible, we hope our efforts can help the development and rigorous benchmarking of future methods.
\end{abstract}

\section{Introduction}
\label{sec:intro}

Over the past five years, Large Language Models (LLMs) \cite{deepseekai2024deepseekv3technicalreport,openai2024gpt4technicalreport,geminiteam2024geminifamilyhighlycapable,grattafiori2024llama3herdmodels} have shown growth in performance and size, demonstrating proficiency in various downstream tasks \cite{snell2024scalingllmtesttimecompute,brown2020languagemodelsfewshotlearners,wei2023chainofthoughtpromptingelicitsreasoning}.
The success of LLM pretraining hinges on three key pillars: high-quality data \cite{penedo2024fineweb-2,li2024datacomp}, architectural innovations \cite{jiang2024mixtralexperts,deepseekai2024deepseekv3technicalreport}, and scalable optimization techniques~\cite{jaghouar2024intellect1technicalreport,shah2024flashattention3fastaccurateattention,charles2025communicationefficientlanguagemodeltraining}.

Among these, the choice of optimizer has remained notably consistent in recent years, with \texttt{Adam(W)} \cite{kingma2017adammethodstochasticoptimization,loshchilov2019decoupledweightdecayregularization} dominating deep learning for nearly a decade.
However, recent advances  \cite{jordan2024muon,liu2025muonscalablellmtraining,vyas2024soapimprovingstabilizingshampoo,pagliardini2024ademamixoptimizerbetterfaster,pethick2025trainingdeeplearningmodels,frans2025stablewhiteningoptimizerefficient,defazio2024roadscheduled} challenge this status quo, offering alternatives that surpass \texttt{AdamW} in speed, communication efficiency \cite{ahn2025dioncommunicationefficientoptimizerlarge} or final downstream performance on various benchmarks \cite{Dahl2023AlgoPerf,karpathy2022}, particularly for autoregressive language modeling \cite{Radford2018ImprovingLU}. 
Despite these innovations, current benchmarks and ablation studies \cite{zhao2024deconstructingmakesgoodoptimizer,morwani2025connectionsschedulefreeoptimizersademamix,kaddour2023traingainrevisitingefficient} remain narrow in scope, often examining only isolated aspects of optimizer design~\cite{kasimbeg2025farawaytrulyhyperparameterfree}.
This lack of systematic comparison makes it difficult to obtain trustworthy insights for practitioners or identify the next promising research directions.

\begin{figure*}[h]
    \vspace{-2.6em}
    \centering
    \begin{minipage}{0.443\linewidth}
    \includegraphics[width=\linewidth]{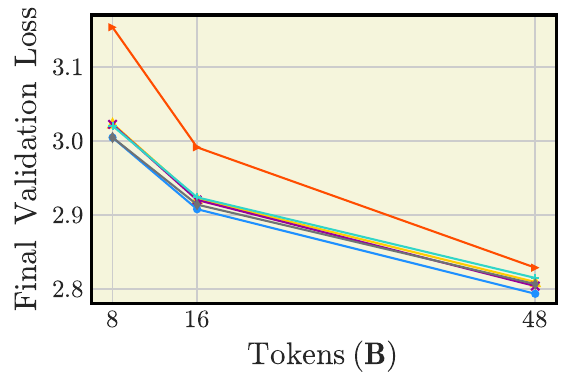}
    \end{minipage}
    \hfill
    \begin{minipage}{0.543\linewidth}
    \includegraphics[width=\linewidth]{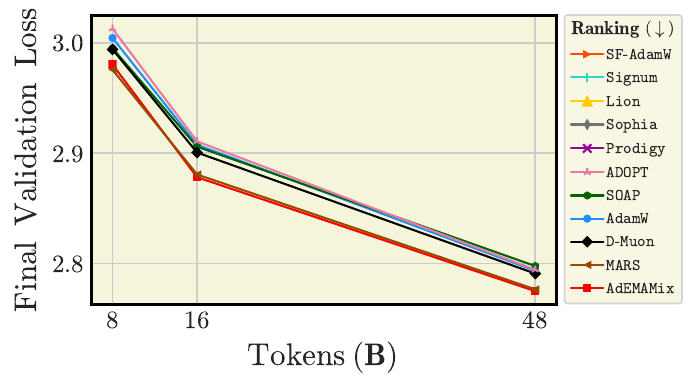}
    \end{minipage}
    \caption{\textbf{Ranking of optimizers for $\mathbf{720M}$ Llama-based models.}
    We plot the \textit{final validation loss} obtained by the best-tuned optimizers on the FineWeb dataset.
    We use a batch size of $1\mathbf{M}$ tokens and train multiple methods beyond and below the Chinchilla optimal duration, which is $14.4\mathbf{B}$ for model of this size.
    \texttt{AdEMAMix} and \texttt{MARS} are the best optimizers in this setup, with a noticable gap in performance compared to other methods.
    We also plot the \texttt{AdamW} baseline in both figures to distinguish the group of methods that consistently perform worse than \texttt{AdamW} from the group of optimizers that outperform it for some training durations.
    See \S~\ref{sec:setup} and \cref{sec:ap_tuning} for  a detailed description of our experimental setup, including hyperparameters.
    }
    \label{fig:benchmark-720}
\end{figure*}

In this work, our goal is to revisit the problem of benchmarking optimizers for LLM pretraining. 
We do so through standardized experiments which vary important parameters such as batch size, model size, and the number of training iterations. 
This allows us to formulate an up-to-date list of best-performing methods for the  community of researchers and practitioners. 
We demonstrate the efficiency of each considered method through careful tuning, and present insightful ablations along the way. Furthermore, we provide a set of best practices for LLM pretraining that are applicable regardless of the optimizer chosen.

We summarize our contributions as follows:

\textbf{(\underline{Contribution 1})} We conduct the first large-scale, controlled benchmark of $11$ different optimization methods across diverse LLM training scenarios.
A fair comparison is ensured by precise accounting for compute costs, and extensive hyperparameter tuning. 
We identify optimal optimizer choices in several relevant training regimes, for both dense and Mixture of Experts (MoE) architectures.

\textbf{(\underline{Contribution 2})} We perform comprehensive ablations of critical training hyperparameters---including warmup duration, initialization schemes, gradient clipping, final learning rates, and learning rate scheduler choices---providing actionable insights for optimizing LLM training in practice.

\textbf{(\underline{Contribution 3})} We open-source our full benchmarking toolkit, including training scripts, 
\begin{wrapfigure}{r}{0.46\linewidth}
    \vspace{-1.2em}
    \includegraphics[width=\linewidth]{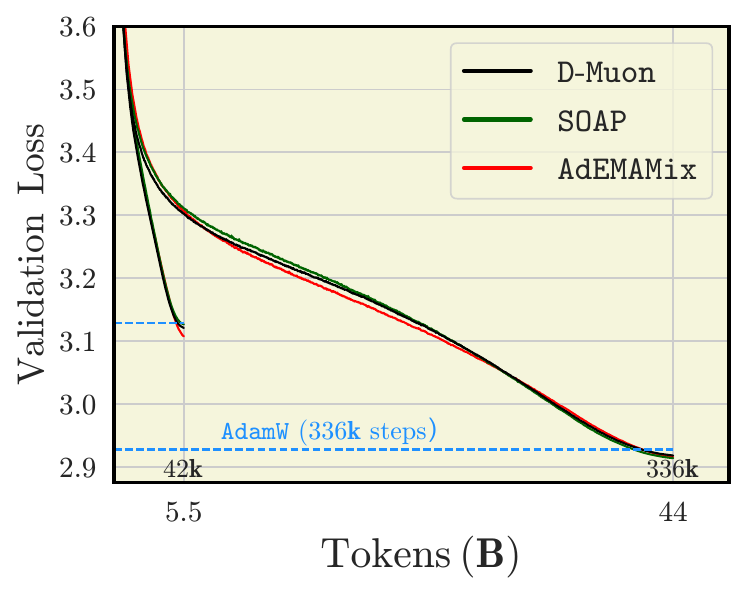}
    \caption{\textbf{Training dynamics of leading optimizers on $\mathbf{520M}$ MoE model pretraining.}
    We use a batch size of $131\mathbf{k}$ tokens, and train models for both short runs, i.e., less than Chinchilla optimal duration, and for extended runs beyond this regime.
    The dashed blue lines correspond to the final validation loss of \texttt{AdamW} baselines trained for both $42\mathbf{k}$ and $336\mathbf{k}$ steps.
    }
    \label{fig:moe-benchmarking-losses}
    \vspace{-1.3em}
\end{wrapfigure}
evaluation pipelines, and hyperparameter configurations, to enable reproducible research and facilitate future optimizer development.

\textbf{For practitioners}, our work provides an evidence-based answer to the burning question: ``\textit{Is \texttt{Adam} still the most effective optimizer in the age of LLMs, or can we achieve better performance at scale with novel optimizers?}''.

\textbf{For researchers}, our work delivers a unified benchmarking framework for LLM pretraining, along with extensive ablation studies which systematically evaluate both popular and overlooked optimizer designs—revealing previously unexplored tradeoffs between efficiency, stability, and final model performance.
Overall, our findings not only challenge long-held assumptions about optimizer selection but also establish a foundation for future advances in large-scale model training. By bridging the gap between theoretical innovation and practical deployment, this work aims to accelerate progress in both research and industry applications of LLM training.
\vspace{-1em}
\section{Background \& Related Work}
\label{sec:related_work}

\textbf{Optimizers.}
While computer vision models often show comparable performance between~\texttt{SGD}~\cite{sgd} and \texttt{AdamW}~\cite{zhang2020adaptivemethodsgoodattention}, the landscape differs dramatically in LLM training~\cite{srećković2025batchsizeproblemrevisiting}. 
Recent work~\cite{zhang2024transformersneedadamhessian} demonstrates that adaptive methods like \texttt{AdamW} provide substantially better optimization characteristics for transformer-based language models.
The question of why \texttt{AdamW} works so well has been a long-standing topic of research \cite{balles2020dissectingadamsignmagnitude,orabona2020neural,zhang2020adaptive,kunstner2024heavytailedclassimbalanceadam,Kunstner_2024}.
Modern methods often inherit \texttt{AdamW}'s core ideas in their structure, such as \texttt{ADOPT} \cite{taniguchi2024adoptmodifiedadamconverge} and \texttt{AdEMAMix} \cite{pagliardini2024ademamixoptimizerbetterfaster}. \texttt{ADOPT} has been motivated by solving long-standing convergence issues in \texttt{AdamW}. 
By normalizing the second-order moment prior to the momentum update, they eliminate the non-convergence issues of \texttt{AdamW} on smooth non-convex functions.
Meanwhile \texttt{AdEMAMix} extends \texttt{AdamW} with an additional slower momentum buffer, i.e., a slower exponential moving average (EMA), which allows the use of much larger momentum values, accelerating convergence.

One interpretation of \texttt{AdamW}'s effectiveness lies in its sign-based update~\cite{kunstner2023noisemainfactorgap}: without the exponential moving average (EMA), \texttt{AdamW} resembles \texttt{signSGD}~\cite{bernstein2018signsgd}.
Recent works \cite{zhao2024deconstructingmakesgoodoptimizer,karimireddy2019error} has shown that \texttt{Signum} (\texttt{signSGD} with momentum), can perform comparably to \texttt{AdamW}.
The community also discussed \texttt{Lion} \cite{chen2023symbolicdiscoveryoptimizationalgorithms}, a method with a similar sign-based structure. \texttt{Signum} and \texttt{Lion} offer memory benefits due to the use of only a single instead of Adam's two buffers for optimizer states.

Another family of methods stems from \texttt{AdamW}'s approximate second-order structure. 
This idea has given rise to \texttt{Sophia}
\cite{liu2024sophiascalablestochasticsecondorder}, where the diagonal of the Fisher information matrix is used as the second moment estimate. 
Exploiting the matrix structure of model weights and optimizer states has led to methods such as \texttt{SOAP}~\cite{vyas2024soapimprovingstabilizingshampoo}, \texttt{Muon}~\cite{jordan2024muon} and \texttt{Scion}~\cite{pethick2025trainingdeeplearningmodels}, including their extentions~\cite{liu2025muonscalablellmtraining,riabinin2025gluon,ahn2025dioncommunicationefficientoptimizerlarge}.

The parameter-free concept \cite{pmlr-v49-orabona16} has led to the development of \texttt{Schedule-Free AdamW} (\texttt{SF-AdamW})~\cite{defazio2024roadscheduled} and \texttt{Prodigy} \cite{mishchenko2024prodigyexpeditiouslyadaptiveparameterfree}. 
These optimizers do not require a decreasing learning rate schedule, making them relevant for continual training.
Last but not least, \texttt{MARS} \cite{yuan2024marsunleashingpowervariance}, builds upon this line of research and incorporates a variance reduction mechanism in its update rule.

\vspace{-1em}
\begin{figure*}[h]
    \centering
    \subfigure[Batch size $32\times512$ tokens.]{
        \begin{minipage}{\linewidth}
            \centering
            \includegraphics[width=0.316\linewidth]{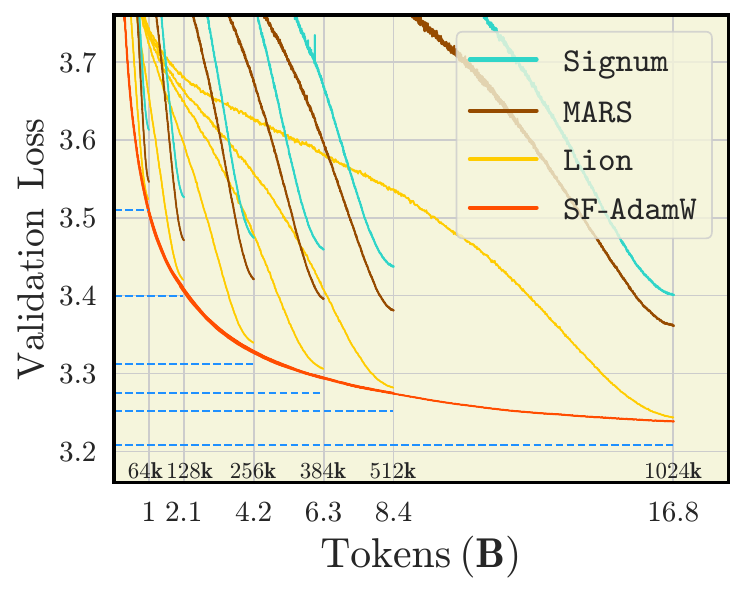}
            \hfill
            \includegraphics[width=0.316\linewidth]{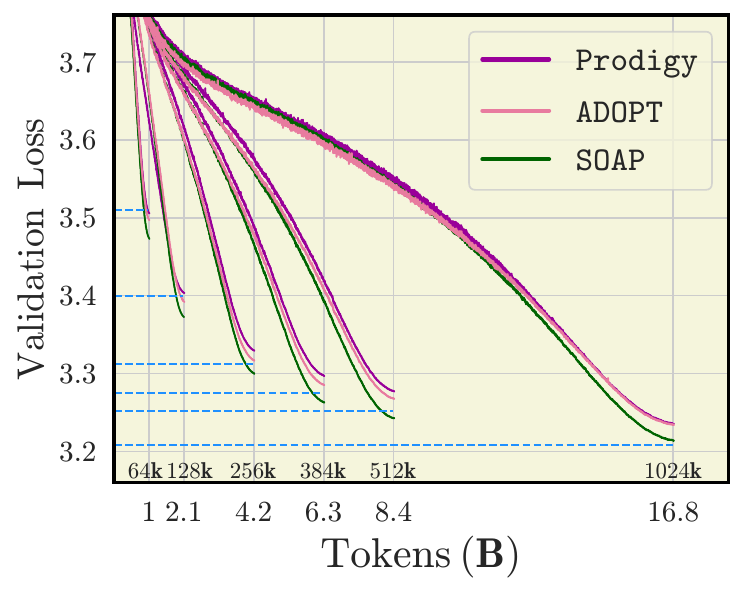}
            \hfill
            \includegraphics[width=0.316\linewidth]{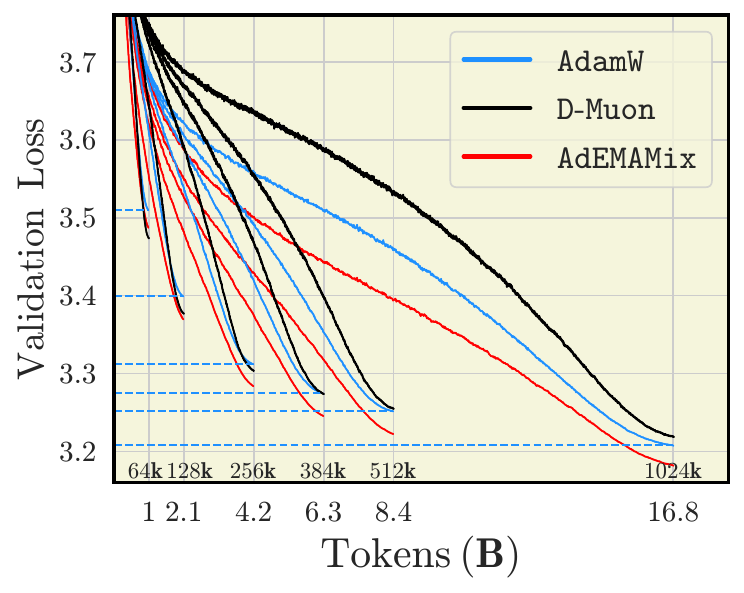}
        \end{minipage}
    }
    \subfigure[Batch size $256\times512$ tokens.]{
        \begin{minipage}{\linewidth}
            \centering
            \includegraphics[width=0.316\linewidth]{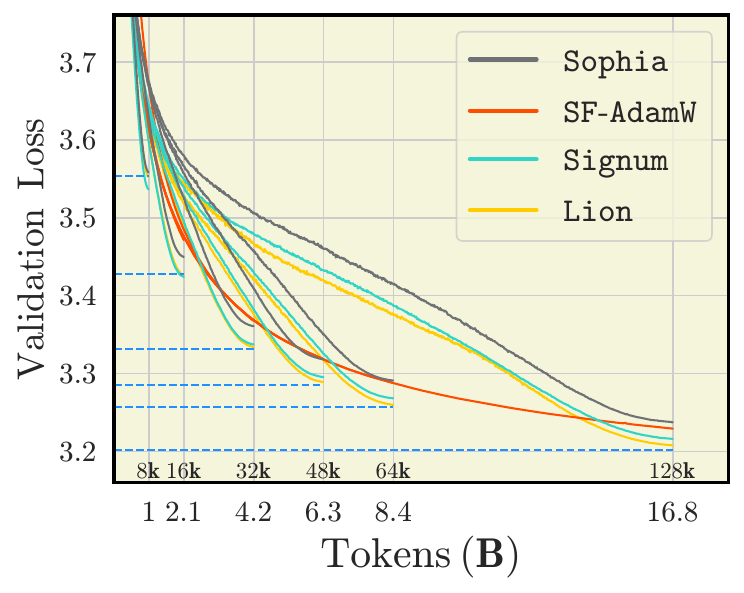}
            \hfill
            \includegraphics[width=0.316\linewidth]{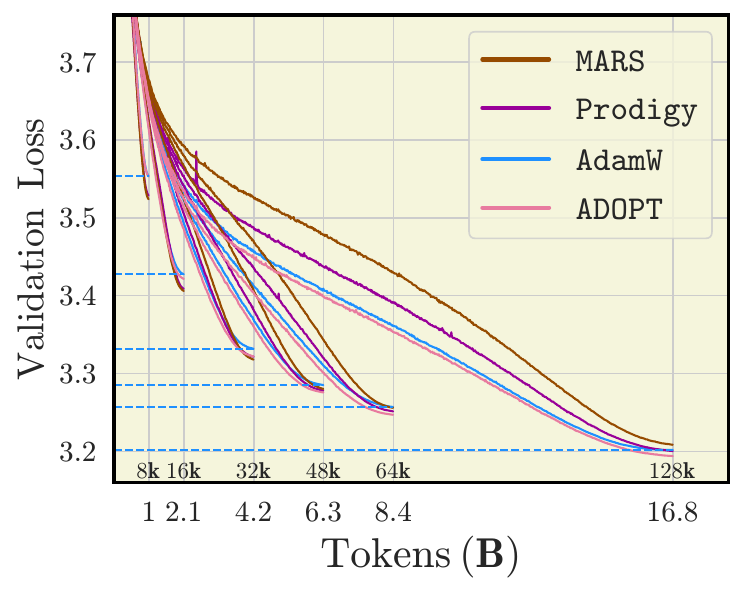}
            \hfill
            \includegraphics[width=0.316\linewidth]{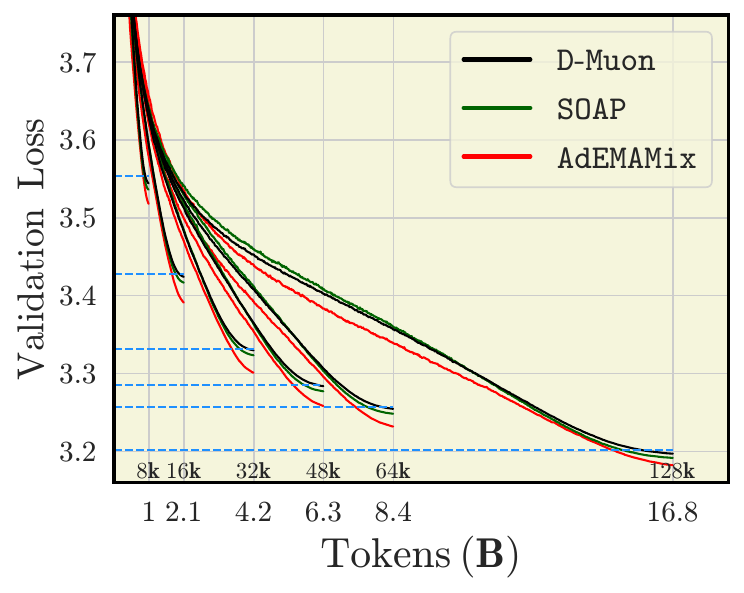}
        \end{minipage}
    }
    \caption{\textbf{Comparing optimizers for training a $\mathbf{124M}$ parameter LLM.}
    We plot the \textit{validation loss} dynamics obtained by considered optimizers.
    In \textbf{(a)}, we train methods with a ``small'' batch size of $16\mathbf{k}$ tokens for $\{64, 128, 256, 384, 512, 1024\}\mathbf{k}$ iterations.
    In \textbf{(b)}, we train methods with nearly $8\times$ larger batch size of $131\mathbf{k}$ tokens for $\{8, 16, 32, 48, 64, 128\}\mathbf{k}$ iterations.
    Thus, in both settings, we result in the same number of tokens models see during the training: $\{1, 2.1, 4.2, 6.3, 8.4, 16.8\}\mathbf{B}$.
    We observe that: (\rom{1}) many methods outperform \texttt{AdamW} in the short runs for $1\mathbf{B}$ or $2.1\mathbf{B}$ tokens; (\rom{2}) as training on more tokens, \texttt{AdamW} narrows the gap with \texttt{SOAP} and \texttt{D-Muon}, while \texttt{AdEMAMix} emerges as the best-performing method; (\rom{3}) \texttt{Signum}, \texttt{MARS}, \texttt{Lion}, \texttt{Prodigy} benefit from the increased batch size.
    }
    \vspace{-1em}
    \label{fig:benchmarking-124m-losses}
\end{figure*}

\textbf{Benchmarks.} 
To a large extent, the benchmarking setup determines the final conclusions.
Some benchmarks are designed for short speedruns in terms of training or validation loss \cite{modded_nanogpt_2024}, while others focus on a downstream target metric after training \cite{zhao2024deconstructingmakesgoodoptimizer,Dahl2023AlgoPerf,schmidt2021descendingcrowdedvalley}.
Methods that perform well in short speedruns might not be optimal for longer training horizons as in real LLM training runs (see \cref{fig:benchmark-124}~\textbf{(a)}, or \cref{fig:benchmarking-210m-losses,fig:benchmarking-720m-losses}~\textbf{(b)}).
''\textit{But what constitutes a sufficiently long horizon?}'' 
''\textit{What should be the compute budget for LLM training?}''
These are questions explored by scaling laws \cite{kaplan2020scalinglawsneurallanguage}.
Early benchmarks for optimizers and other ablation studies often rely on Chinchilla scaling laws \cite{hoffmann2022trainingcomputeoptimallargelanguage} with a ratio of roughly $20$ tokens per parameter needed for pretraining.
However, recent research \cite{li2025farseerrefinedscalinglaw,porian2024resolvingdiscrepanciescomputeoptimalscaling,sardana2024chinchillaoptimalaccountinginferencelanguage} argues that this is far from sufficient for production-ready models.

Another important issue is the choice of loss function.
Recent setups have used an auxiliary $z$-loss \cite{yang2023baichuan,chowdhery2022palmscalinglanguagemodeling} in addition to cross-entropy, which requires further investigation.
We believe that this choice is influenced by the use of the OLMo~\cite{olmo20242olmo2furious} codebase, which we also address in our work.

Additionally, we found that previous setups for comparing optimizers do not align with recent best practices regarding weight decay, learning rate decay, and overall hyperparameter tuning. 
All of these questions are revisited in our work.

\section{Experimental Setup}
\label{sec:setup}

\textbf{Notations.}
We use the following notations. 
Let $\gamma$ be the learning rate, $\lambda$ the weight decay coefficient, and $T$ the total number of iterations. 
Momentum-related parameters are represented by the symbol $\beta$.

\textbf{Optimizers.} Here is a list of the optimizers we considered in our work.
For each algorithm, we write in parentheses the optimizer-specific hyperparameters we tuned: \texttt{AdamW}($\beta_1$, $\beta_2$), \texttt{ADOPT}($\beta_1$, $\beta_2$), \texttt{AdEMAMix}($\beta_1$, $\beta_2$, $\beta_3$, $\alpha$), \texttt{Lion}($\beta_1$, $\beta_2$), \texttt{Signum}($\beta$), \texttt{Muon}($\gamma^\texttt{M}$, $\beta$, $\beta_1$, $\beta_2$), \texttt{D-Muon}($\beta$, $\beta_1$, $\beta_2$)~\cite{liu2025muonscalablellmtraining}, \texttt{SOAP}($\beta_1$, $\beta_2$) and preconditioning frequency, \texttt{Sophia}($\rho$, $\beta_1$, $\beta_2$), \texttt{SF-AdamW}($\beta_1$, $\beta_2$), \texttt{Prodigy}($\beta_1$, $\beta_2$), \texttt{MARS}($\eta$, $\beta_1$, $\beta_2$).
When an optimizer has several momentum variants e.g. Nesterov \cite{Nesterov1983AMF} or Polyak \cite{Polyak1964SomeMO}, we try both.
When optimizers use the Newton-Schulz orthogonalization~\cite{bernstein2024oldoptimizernewnorm,functionsmatrices}, we vary the number of steps for this procedure.
In addition, we tune the learning rate $\gamma$ extensively for all methods. 
We also try different gradient clipping levels, warmup steps, weight decay values, weights initialization, and learning rate schedulers. 
A summary of the hyperparameters tested and selected for each model size is in \cref{sec:ap_tuning}. 
All optimizers are described in depth in \cref{sec:ap_optimizers}.

\begin{figure*}[h]
    \centering
    \subfigure[Llama $124\mathbf{M}$ parameters.]{
        \includegraphics[width=0.316\linewidth]{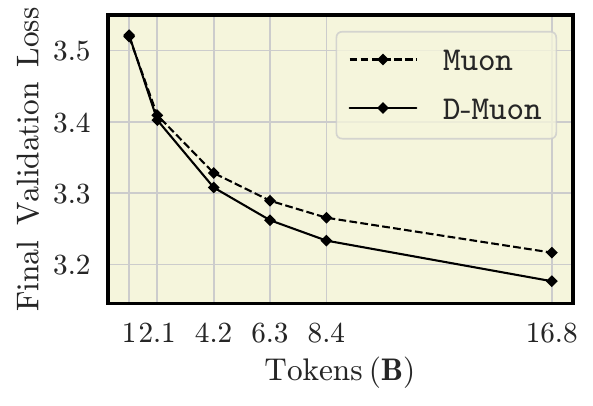}
    }
    \hfill
    \subfigure[Llama $210\mathbf{M}$ parameters.]{
        \includegraphics[width=0.316\linewidth]{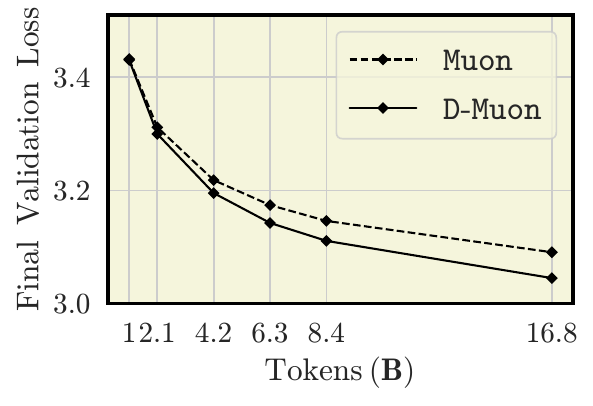}
    }
    \hfill
    \subfigure[Llama $720\mathbf{M}$ parameters.]{
        \includegraphics[width=0.316\linewidth]{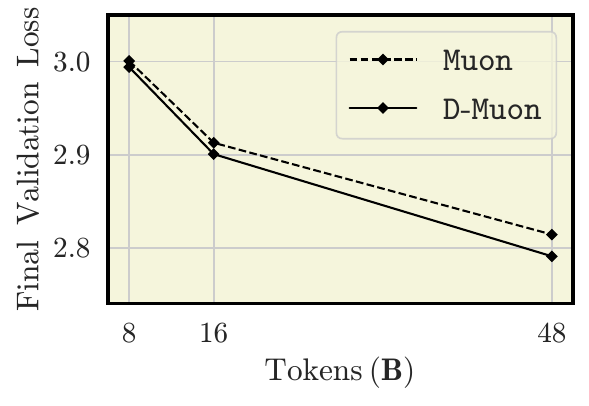}
    }
    \vspace{-2mm}
    \caption{\textbf{Weight decay in \texttt{Muon} \& \texttt{D-Muon}.}
    We compare two methods---basic \texttt{Muon}~\cite{jordan2024muon}, and \texttt{D-Muon}~\cite{liu2025muonscalablellmtraining} with a weight decay applied to all parameter groups.
    Across model sizes~\textbf{} used in our benchmarking of dense LLMs, we observe a major improvement of \texttt{D-Muon} over \texttt{Muon}.
    We relate this observation to our ablation on the importance of weight decay across different optimizers, and training horizons.
    See \cref{fig:muon-dmuon-val-loss} and \cref{sec:ap_wdablation} for more details.
    }
    \label{fig:muon-dmuon-final-val-loss}
\end{figure*}

\textbf{Models \& Data.}
For most experiments, we use a Llama-like transformer \cite{grattafiori2024llama3herdmodels} architecture with weight tying~\cite{press2017usingoutputembeddingimprove}, including SwiGLU activations \cite{shazeer2020gluvariantsimprovetransformer}, RMSNorm \cite{zhang2019rootmeansquarelayer}, and RoPE embeddings \cite{su2023roformerenhancedtransformerrotary}.
We experiment with four sizes of models: $124\mathbf{M}$, $210\mathbf{M}$, $583\mathbf{M}$, $720\mathbf{M}$.
In addition to our dense models, we also benchmark optimizers on a Llama-based $520\mathbf{M}$ MoE model, the corresponding setup is described in \S~\ref{sec:moe} and \cref{sec:ap_tuning}.
We train on a $100\mathbf{B}$ tokens\footnote{\href{https://huggingface.co/datasets/HuggingFaceFW/fineweb}{https://huggingface.co/datasets/HuggingFaceFW/fineweb}} subset of FineWeb \cite{penedo2024finewebdatasetsdecantingweb}. 
It consists of a cleaned and deduplicated
corpus for LLM pretraining, which we tokenize using the GPT-$2$ tokenizer prior to splitting into train and validation sequences.

\textbf{Iterations \& Batch size.}
Throughout our experiments, we use a sequence length of $512$ tokens. 
For clarity, we often report the batch size in tokens by writing $\text{\textit{batch size}}\times\text{\textit{sequence length}}$. 
For the $124\mathbf{M}$ model, we use batch sizes of $32\times512=16\mathbf{k}$, $256\times512=131\mathbf{k}$, and $512\times512=262\mathbf{k}$ tokens; for the $210\mathbf{M}$ model and $520\mathbf{M}$ MoE model, we use a batch size of $256\times512=131\mathbf{k}$; for the $583\mathbf{M}$ model, we leverage the batch size of $3936\times512=2\mathbf{M}$ tokens, finally, we use a batch size of $1984\times512=1\mathbf{M}$ tokens for the $720\mathbf{M}$ model. 
Depending on the model size, we vary the number of iterations---also measured in tokens for compatibility with scaling laws and to accommodate different batch size settings.
We train $124\mathbf{M}$ and $210\mathbf{M}$ models for equal durations of $\{1, 2.1, 4.2, 6.3, 8.4, 16.8\}\mathbf{B}$ tokens.
This corresponds to $T\in\{64, 128, 256, 384, 512, 1024\}\mathbf{k}$ iterations for a batch size of $32$, and $T\in\{8, 16, 32, 48, 64, 128\}\mathbf{k}$ iterations for a batch size of $256$. 
For $583\mathbf{M}$ models, we train on $13\mathbf{B}$ tokens, corresponding to $6.5\mathbf{k}$ iterations.
In the setup with $720\mathbf{M}$ model, we have $T\in\{8,16,48\}\mathbf{k}$ iterations for a batch size of $1\mathbf{M}$ tokens.
Thus, for all model scales, we include both Chinchilla optimal lengths of training and beyond. 
More details are available in \cref{sec:ap_modeldata}.

\textbf{Loss.}
We train using the classical cross-entropy next token prediction loss.
Some prior works introducing optimizers~\cite{vyas2024soapimprovingstabilizingshampoo}, benchmarkings~\cite{zhao2024deconstructingmakesgoodoptimizer}, or pretraining recipes for LLMs~\cite{jaghouar2024intellect1technicalreport,chowdhery2022palmscalinglanguagemodeling,yang2023baichuan,brandfonbrener2024losstolosspredictionscalinglaws}, use a $z$-loss regularizer in addition to cross-entropy. 
We found that this has little impact and, therefore, do not use $z$-loss.
An ablation showing results with and without $z$-loss is in \S~\ref{sec:results}.

\textbf{Hyperparameter Tuning.}
Training LLMs is a computationally intensive task~\cite{epoch2024datamovement}.
As a guidance, practitioners often rely on insights gathered at lower scales, scaling laws~\cite{openai2024gpt4technicalreport,deepseekai2024deepseekllmscalingopensource,sardana2024chinchillaoptimalaccountinginferencelanguage,li2025predictablescalei}, and other rules \cite{yang2022tensorprogramsvtuning,cerebras2024mupguide,blake2025umupunitscaledmaximalupdate,kumar2024scalinglawsprecision}. 
It is also commonplace to run experiments for only a shorter duration of training, as a way to test certain hyperparameters prior to extending the training horizon to more iterations. 
Because a full grid search over every hyperparameter, for each setting and optimizer, would be too costly, we resort to a similar approach. 
More precisely, for each model size, batch size, and optimizer, we extensively tune optimization hyperparameters for a number of training tokens which are near-Chinchilla optimal, e.g., we pick $\{2.1, 16\}\mathbf{B}$ tokens for tuning $\{124, 720\}\mathbf{M}$ models~(see \cref{sec:ap_tuning}).
We then keep those hyperparameters when we increase the number of iterations. 
While we found that the sensitivity to several hyperparameters can change as we increase the training horizon---see \cref{fig:ap_retuning_betas}---we found this approach simple and yet effective. 
The hyperparameters being considered depend on the optimizer. 
We proceeded from small to large model scale, and used insights gathered at smaller scales to guide the hyperparameter search at larger scales. 
Our hyperparameter sweeps are summarized in \cref{sec:ap_tuning}. 
We present the clarifications regarding the connection between the number of iterations and tokens for different batch size settings, as well as the Chinchilla optimal training durations for our models in Tables~\ref{tab:training_horizons_smallbs},~\ref{tab:training_horizons_largebs},~\ref{tab:training_horizons_2mbs},~\ref{tab:training_horizons_1mbs}, and~\ref{tab:training_horizons_moe}. 
As learning rate schedulers, we compare cosine~\cite{loshchilov2017sgdrstochasticgradientdescent}, linear and warmup-stable-decay (WSD) \cite{hu2024minicpmunveilingpotentialsmall,zhai2022scalingvisiontransformers,hägele2024scalinglawscomputeoptimaltraining}. 
Unless specified, we use a cosine scheduler. 
Results with WSD and linear schedulers are discussed in \S~\ref{sec:results}.
Recent works also emphasize the importance of sufficiently decaying the learning rate \cite{bergsma2025straightzerolinearlydecaying,schaipp2025surprisingagreementconvexoptimization,hägele2024scalinglawscomputeoptimaltraining,li2025predictablescalei,deepseekai2024deepseekv3technicalreport}. 
As such, we take care to decay to $0.01\times\gamma_{\max}$ instead of the often used $0.1\times\gamma_{\max}$~\cite{hoffmann2022trainingcomputeoptimallargelanguage,touvron2023llamaopenefficientfoundation,biderman2023pythiasuiteanalyzinglarge,workshop2023bloom176bparameteropenaccessmultilingual,olmo20242olmo2furious,groeneveld2024olmoacceleratingsciencelanguage,zhao2024deconstructingmakesgoodoptimizer}. 
To give an idea of how much effort was put into tuning each method, across all model sizes, batches and iterations, we trained a total of $2900$ models, and have spent roughly $30 000$ GPU hours.
See more details in \cref{sec:ap_implementation,sec:ap_tuning}.

\section{Results}
\label{sec:results}

We structure our story starting with smaller models and batch sizes, and gradually scaling up to larger configurations. In some instances, we complement the core benchmarking results with additional ablations and possible best-practices.

\subsection{Benchmarking \& Ablations at Small Scale: Training Models of $\mathbf{124M}$ Parameters}
\label{sec:smallscalebench}

\begin{figure*}[h]
    \centering
    \subfigure[Batch size $32\times512$ tokens.]{
        \includegraphics[width=0.43\linewidth]{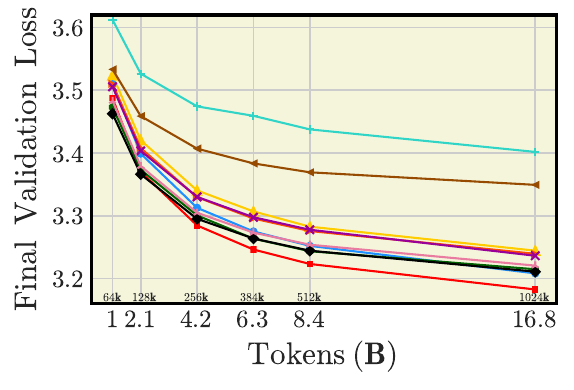}
    }
    \hfill
    \subfigure[Batch size $256\times512$ tokens.]{
        \includegraphics[width=0.523\linewidth]{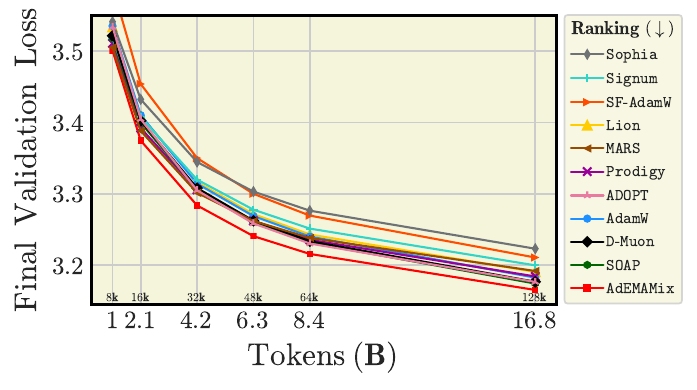}
    }
    \caption{\textbf{Ranking of optimizers for $\mathbf{124}\textbf{M}$ models with ``small'' and ``large'' batch sizes.}
    In both \textbf{(a)} and \textbf{(b)}, we show the \textit{final validation loss} for different training durations, corresponding to different numbers of tokens.
    Above each token number, we write the number of training iterations corresponding.
    In \textbf{(a)}, we use a ``small'' batch size of $32\times512$ tokens.
    In \textbf{(b)}, we use a larger batch size of $256\times512$ tokens.}
    \vspace{-1em}
    \label{fig:benchmark-124}
\end{figure*}

\textbf{Results with ``small'' batches.} 
We first report results when using batches of $32\times512$ tokens in Figures~\ref{fig:benchmark-124}~\textbf{(a)} and~\ref{fig:benchmarking-124m-losses}~\textbf{(a)}. 
We tune the hyperparameters by training for $2.1\mathbf{B}$ tokens ($128\mathbf{k}$ iterations) and then keep those hyperparameters for all other training durations. 
The best hyperparameters are reported in \cref{sec:ap_124mtuning}. 
We observe how, for the smallest number of iterations we considered ($1\mathbf{B}$ tokens $\equiv$ $64\mathbf{k}$), \texttt{SOAP}, \texttt{ADOPT}, \texttt{AdEMAMix}, \texttt{D-Muon}, \texttt{Prodigy}, and \texttt{SF-AdamW} all outperform \texttt{AdamW}, with \texttt{D-Muon} being the best. 
As we increase the number of iterations, \texttt{AdEMAMix} takes the lead while \texttt{AdamW} becomes a second, and closes the gap with \texttt{D-Muon} and \texttt{SOAP}. 
A sign-based methods such as \texttt{Lion} and \texttt{Signum} are expected to perform poorly when the batch size is small. 
Intuitively, this is due to the $\text{sign}(\cdot)$ operator being sensitive to gradient noise~\cite{tomihari2025understandingadamoutperformssgd,kornilov2025signoperatorcopingheavytailed}. 
As described in its original paper, \texttt{MARS} also performs poorly when the batch size is small. 
We found \texttt{Prodigy}, the basic \texttt{Muon} (see \cref{fig:muon-dmuon-final-val-loss,fig:muon-dmuon-val-loss}~\textbf{(a)}) and \texttt{SF-AdamW} to underperform in this setting compared to \texttt{AdamW}. 
On this scale, \texttt{Prodigy} suffers from the lack of bias correction of the learning rate, as well as being sensitive to $(\beta_1, \beta_2)$ (see \cref{fig:prodigy_betas}).
Importantly, when the batch size is sufficiently small, we observe that \texttt{Sophia} diverges when increasing the number of iterations, even if decreasing the learning rate (see~\cref{fig:failofsophia}).
Thus, we decided not to include \texttt{Sophia} at this stage of our benchmarking.

\textbf{Results with ``large'' batches.} 
We now report results when using batches of $256\times512$ tokens---$8\times$ larger than for our ``small'' batch setting. 
Results in Figures~\ref{fig:benchmark-124}~\textbf{(b)} and~\ref{fig:benchmarking-124m-losses}~\textbf{(b)} show how \texttt{Signum}, \texttt{MARS}, \texttt{Lion}, \texttt{Prodigy} greatly benefit from the increased batch size. 
Remarkably, we observe that the \texttt{Prodigy} method scales similarly to \texttt{AdamW}. 
We emphasize the possible community interest in this algorithm, as its effective learning rate---determined by two EMA sequences---emulates the learning rate behavior of \texttt{AdamW}. 
When the scheduler is applied and $\gamma_{\max}$ of \texttt{Prodigy}  is set to $1$ (its default value), these EMAs result in the maximal effective learning rate, which closely matches that of \texttt{AdamW}---see~\cref{fig:ap_prodigy_effective_lr}. 
For a small number of iterations (e.g. $T\in\{8\mathbf{k}, 16\mathbf{k}\}$ corresponding to $1\mathbf{B}$ and $2\mathbf{B}$ tokens), all methods outperform \texttt{AdamW} except for \texttt{SF-AdamW} and \texttt{Sophia}. 
As we increase the number of iterations \texttt{ADOPT}, \texttt{D-Muon}, \texttt{SOAP}, and \texttt{AdEMAMix} take the lead. 
In particular, \texttt{AdEMAMix} has a consistent lead over other methods. 
While we anticipated---in accordance with Vyas et al.~\cite{vyas2024soapimprovingstabilizingshampoo}---that \texttt{SOAP} would greatly benefit from the larger batch size, its behavior remains relatively consistent compared to our previous small batch setting.

\prettybox{
\takeaway{tkw:32bs_ranking}After experimenting with both ``small'' and ``large'' batch settings, we conclude that: (\rom{1}) \texttt{AdEMAMix} consistently achieves state-of-the-art performance and robust scaling with training duration; (\rom{2}) sign-based methods (\texttt{Signum}, \texttt{Lion}), and \texttt{MARS} greatly benefit from the increased batch size; (\rom{3}) \texttt{Sophia} diverges in the small-batch setting, when trained beyond the Chinchilla optimal horizon, even with sufficiently small learning rate; (\rom{4}) \texttt{SOAP} show a surprisingly consistent performance in both settings.
}

\begin{figure*}[h]
    \centering
    \includegraphics[width=\linewidth]{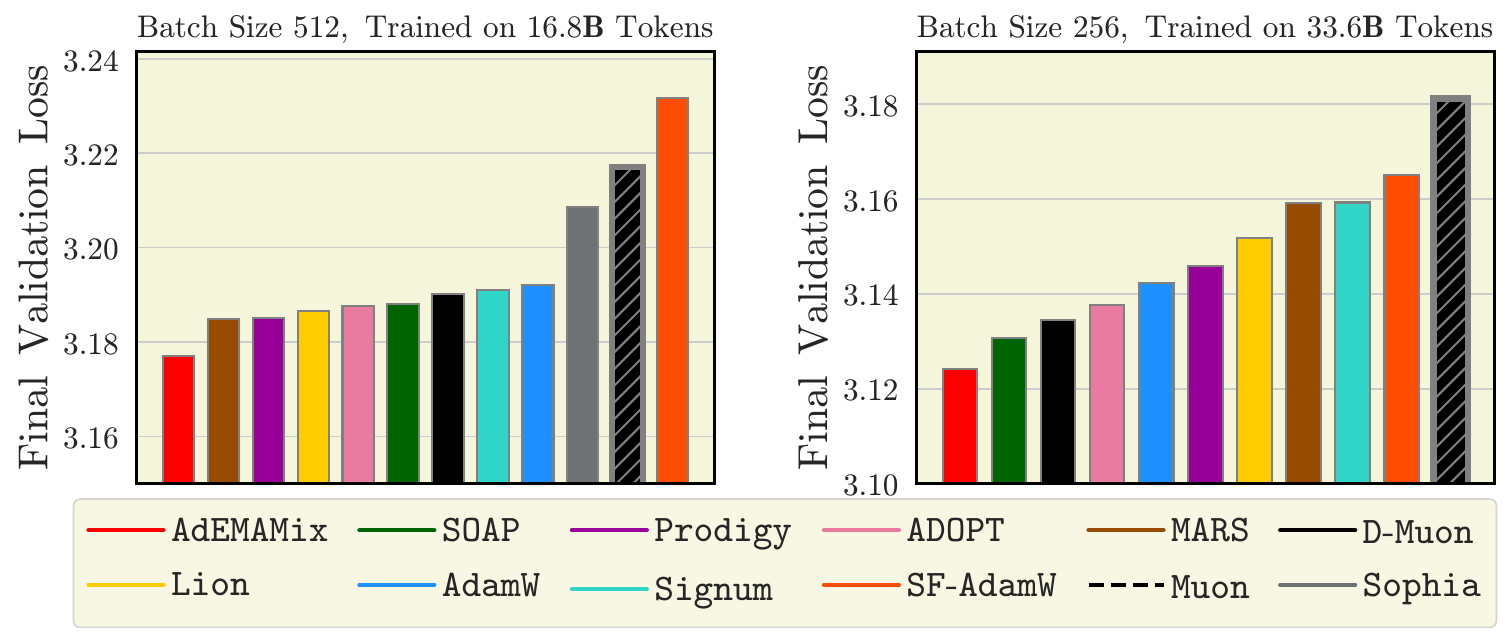}
    \caption{\textbf{Scaling batch size vs. scaling the number of iterations.}
    Our results demonstrate that: (\textit{left}) scaling the batch size significantly improves \texttt{MARS}, \texttt{Signum}, \texttt{Lion} and \texttt{Prodigy} making them as good as \texttt{AdamW} even for a long training for $16.8\mathbf{B}$ tokens.
    Which was not the case in \cref{fig:benchmark-124}~\textbf{(b)}, where we still observed a significant gap in performance; and (\textit{right}): indeed, with scaling of the number of iterations, the gap between \texttt{SOAP} and \texttt{AdEMAMix} narrow and, finally, increases.
    But, on the other hand, with increase of the \texttt{AdEMAMix} $\beta_3$ parameter, the performance gap with \texttt{SOAP} reappears.
    }
    \vspace{-1em}
    \label{fig:scalebs_scaletokens}
\end{figure*}

\textbf{Stability across training horizons.}
As mentioned in \S~\ref{sec:setup}, we tune hyperparameters training on $2.1\mathbf{B}$ tokens and keep those hyperparameters when extending the training horizon.
However, when increasing the length of training or scaling batch size, critical hyperparameters of optimizers such as learning rate, betas might change~\cite{busbridge2023scaleema}.
Thus, we additionally \textit{re-tune the methods} for $16.8\mathbf{B}$ length of training to show the best results.
We found that previously widely adopted~\cite{deepseekai2024deepseekv3technicalreport,wortsman2023smallscaleproxieslargescaletransformer,zhao2024deconstructingmakesgoodoptimizer,jaghouar2024intellect1technicalreport,hägele2024scalinglawscomputeoptimaltraining,li2025predictablescalei} for \texttt{AdamW} ($\beta_1=0.9$, $\beta_2=0.95$) parameters give worse results than ($\beta_1=0.9$, $\beta_2=0.999$).
We point that it would be beneficial to further increase the $\beta_2$ for \texttt{AdamW}-like optimizers when increasing the length of training.
The same applies to $\beta_3$ parameter of \texttt{AdEMAMix}, which we increase from $0.999$ to $0.9999$ when training on $16.8\mathbf{B}$ tokens and beyond (see~\cref{sec:ap_scalingiters} for a detailed ablation on that matter and references therein).
Importantly, from \cref{fig:benchmarking-124m-losses}~\textbf{(b)}, we see that \texttt{SOAP} and \texttt{D-Muon} narrow the gap with \texttt{AdEMAMix}.
It is interesting to see how the situation changes when the training horizon is extended to $33.6\mathbf{B}$ tokens ($\equiv 256\mathbf{k}$ iterations).
For this experiment, we use the batch size of ($256\times512$), and keep the re-tuned hyperparameters we found for $16.8\mathbf{B}$ tokens run, simply reusing them for longer training.
We report insights gathered from this ablation in \cref{fig:scalebs_scaletokens}~(\textit{right}).
As in the ``small'' batch ablation, we emphasize that \texttt{Sophia} exhibits convergence issues when extending the training run, and diverges shortly after $130\mathbf{k}$ steps~(\cref{fig:failofsophia-largebs}).
Regarding other optimizers, we observe a consistent behavior compared to the one from \cref{fig:benchmark-124}~\textbf{(b)}---all methods remain at the same position in our tier-list.
The results suggest that the best hyperparameters found at $16.8\mathbf{B}$ scale are also consistent w.r.t. doubling the number of steps. 
``But what can one say about scaling batch size while keeping the same amount of tokens seen?''

\textbf{Increasing the batch size further.}
We also run an experiment with batches of $512\times512=262\mathbf{k}$ tokens, training for $64\mathbf{k}$ iterations, thus, we keep the total amount of tokens to train on.
We show the results of this ablation in \cref{fig:scalebs_scaletokens}~(\textit{left}).
Noticeably \texttt{MARS} becomes the second best-performing method behind \texttt{AdEMAMix}, followed closely by \texttt{Prodigy}, \texttt{Lion}, \texttt{ADOPT}, and \texttt{SOAP}. 
Interestingly, \texttt{Signum} performs comparably to \texttt{AdamW}.
Our results with batches of $\{131, 262\}\mathbf{k}$ tokens show an evidence that sign-based methods greatly benefit from increased batch size, as noticed in many prior works~\cite{chen2023symbolicdiscoveryoptimizationalgorithms}.
Furthermore, the hyperparameter sweeps from~\cite{zhao2024deconstructingmakesgoodoptimizer,zhang2024doescriticalbatchsize} suggest that \texttt{Lion}, \texttt{Signum}, \texttt{AdamW} stay consistent w.r.t tuning all hyperparameters except for batch size, where they notice a worsens in performance at large batch sizes above ours $256\times512$, while we observe a quite opposite results in our setup.

\prettybox{
\takeaway{tkw:scalebs_scaletokens}(\rom{1}) Suprisingly, many methods, especially \texttt{MARS}, \texttt{Prodigy}, and sign-based ones, can outperform \texttt{AdamW} while trained on a sufficiently large batches.
(\rom{2}) We also found that in our setup, once optimizers are properly re-tuned for the maximal length of training considered, doubling of number of iterations does not affect the ranking of methods.
}

\begin{figure*}[h]
    \centering
    \subfigure[Use large $\lambda$ for short training.]{
        \includegraphics[width=0.316\linewidth]{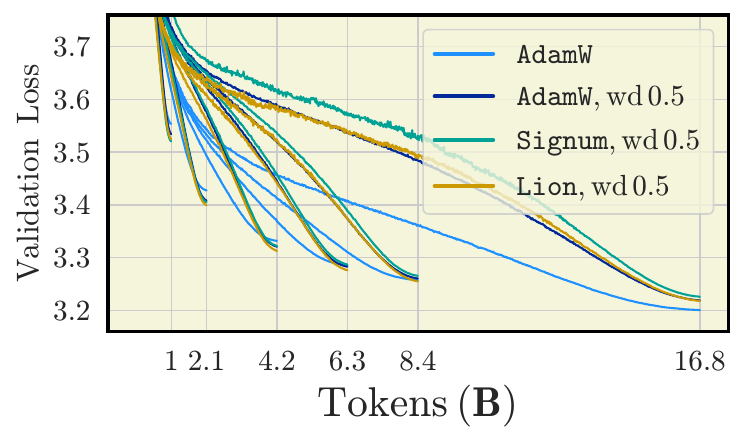}
    }
    \hfill
    \subfigure[Use $\lambda=0.1$ for long training.]{
        \includegraphics[width=0.316\linewidth]{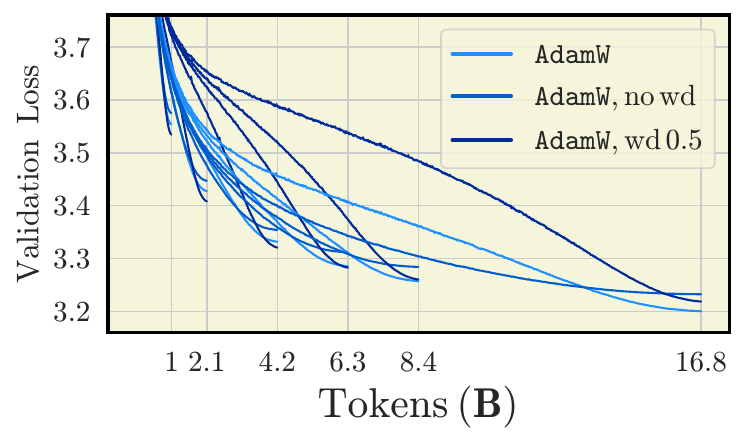}
    }
    \hfill
    \subfigure[Norm grows with smaller $\lambda$.]{
        \includegraphics[width=0.316\linewidth]{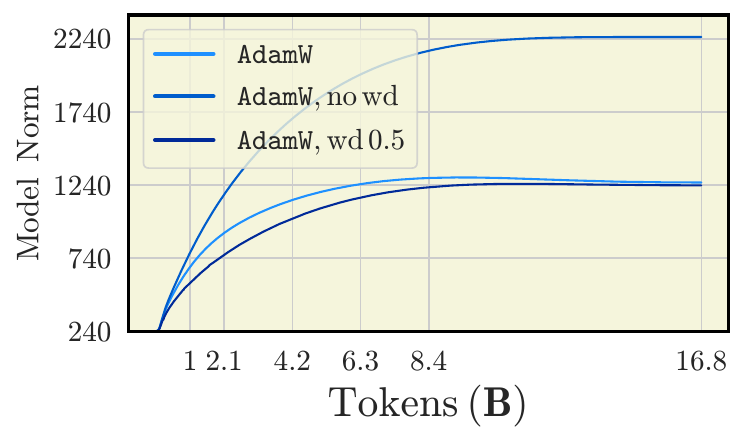}
    }
    \vspace{-2mm}
    \caption{\textbf{Larger weight decay achieves significantly better results when training on fewer tokens.}
    In \textbf{(a)} we observe that runs of \texttt{AdamW}, \texttt{Signum}, and \texttt{Lion} with the large weight decay of $0.5$ consistently outperform the baseline \texttt{AdamW} with weight decay of $0.1$ for all training durations except for the last one.
    Notably, \texttt{Signum} and \texttt{Lion} with large weight decay perform even better than \texttt{AdamW} with the same learning rate.
    In \textbf{(b)}, we also consider a setting without weight decay.
    We observe that this is suboptimal not only for \texttt{AdamW}, but also for the majority of other optimizers (see \cref{sec:ap_wdablation}), while the typical weight decay of $0.1$ remains the best for large training durations.
    Importantly, in \textbf{(c)}, we ablate the impact od weight decay on the model's $\ell_2$ norm.
    }
    \label{fig:wdablation_main}
\end{figure*}

\begin{figure*}[h]
    \centering
    \subfigure[Llama $124\mathbf{M}$ parameters.]{
        \includegraphics[width=0.316\linewidth]{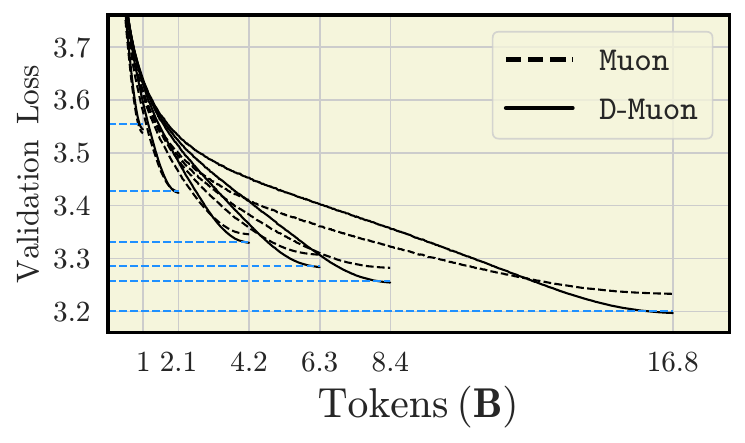}
    }
    \hfill
    \subfigure[Llama $210\mathbf{M}$ parameters.]{
        \includegraphics[width=0.316\linewidth]{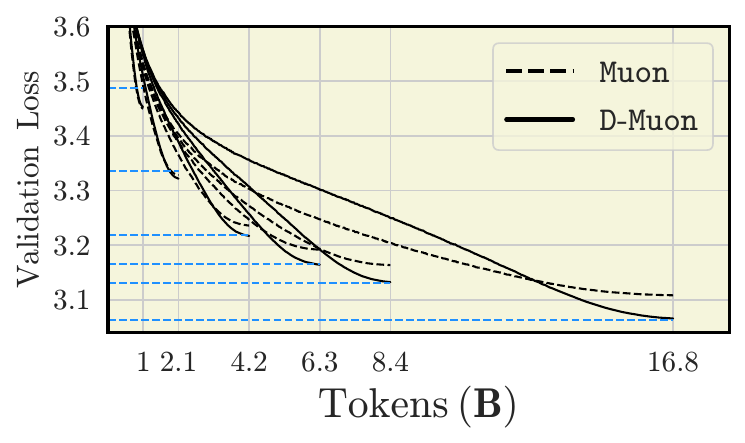}
    }
    \hfill
    \subfigure[Llama $720\mathbf{M}$ parameters.]{
        \includegraphics[width=0.316\linewidth]{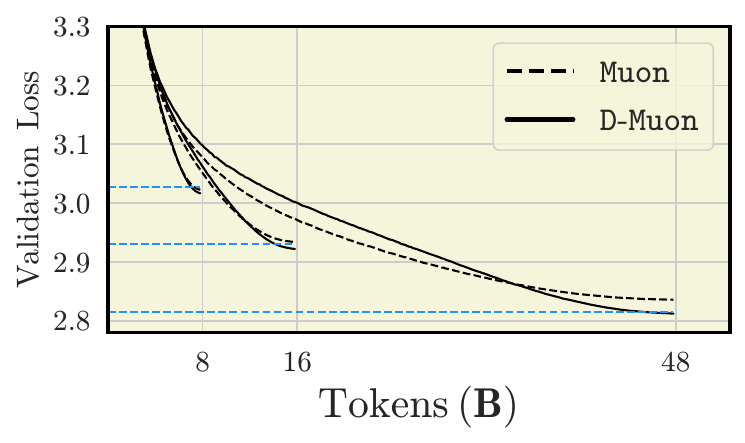}
    }
    \vspace{-2mm}
    \caption{\textbf{Importance of weight decay for \texttt{Muon}.}
    We complement our weight decay ablation with comparison of two version of \texttt{Muon}: one that uses a weight decay for all parameters (\texttt{D-Muon}), and another, with weight decay being applied only to embeddings, scalar parameters, and the final layer.
    For three scales of models~\textbf{(a,b,c)}, we show that \texttt{D-Muon} greatly outperforms the basic \texttt{Muon}.
    We emphasize that the main reason---weight decay, which supports our ablations in~\cref{fig:wdablation_main,fig:ap_wdablation}.
    }
    \label{fig:muon-dmuon-val-loss}
\end{figure*}

\textbf{Weight decay ablation.}
As recent frameworks for LLM pretraining or ablation studies omit weight decay as a default non-zero hyperparameter~\cite{olmo20242olmo2furious,groeneveld2024olmoacceleratingsciencelanguage,zhao2024deconstructingmakesgoodoptimizer}, some setups even mislead by not incorporating weight decay in their experiments~\cite{zhang2024doescriticalbatchsize,brandfonbrener2024losstolosspredictionscalinglaws,morwani2025connectionsschedulefreeoptimizersademamix}.
In this work, we demonstrate the importance of weight decay and its impact across different optimizers.
Surprisingly, increasing weight decay while keeping the learning rate constant proves to be an effective technique for training on shorter horizons~(\cref{fig:wdablation_main}~\textbf{(b,c)}). 
This approach is so effective that methods like \texttt{Signum} and \texttt{Lion} with high weight decay significantly outperform \texttt{AdamW} without weight decay (see \cref{fig:wdablation_main}~\textbf{(a)}).
Implementation details also warrant attention. 
Coupled weight decay ($\ell_2$ regularization)~\cite{tikhonov1943stability,shalevshwartzbendavid} is still used in some LLM pretraining settings~\cite{wortsman2023smallscaleproxieslargescaletransformer,brown2020languagemodelsfewshotlearners}, including the PyTorch \cite{paszke2019pytorchimperativestylehighperformance} optimizer implementations. 
Notably, the popular implementation of \texttt{Signum} becomes ineffective when weight decay is applied. 
Highlighting this oversight for the community, we contribute by demonstrating our implementation of \texttt{Signum} (\cref{alg:signumtorch}) with decoupled weight decay~\cite{loshchilov2019decoupledweightdecayregularization}.
The influence of weight decay on model weights is intriguing. 
As is known, model weights typically grow during training, but weight decay, by modifying the optimized function, significantly reduces the growth of the model's parameter norm~(\cref{fig:wdablation_main}~\textbf{(c)}). 
Such ablations of weight decay are also of interest to the community \cite{dangelo2024needweightdecaymodern,kosson2024rotationalequilibriumweightdecay}.

Regarding the ablation of weight decay for optimizers, we again select the best setup for each and conduct a sweep over weight decay values. 
Our results are presented in \cref{fig:wdablation_main} and in \cref{fig:ap_wdablation}.

For most of optimizers, we observe a consistent results: larger weight decay term of $0.5$ is preferable when training on less tokens, but when the length of training increases, the standard decoupled weight decay of $0.1$ in optimizers achieves better results.
At the same time, decreasing weight decay to $0$, leaves a huge gap with the widely accepted weight decay of $0.1$, and for optimizers this gap only increases with training horizon~(\cref{fig:ap_wdablation}), with one exception---the basic \texttt{Muon} algorithm~\ref{alg:muon}.
As weight decay is not used for two-dimensional parameters in \texttt{Muon}, but this issue was fixed in~\cite{liu2025muonscalablellmtraining} by introducing \texttt{D-Muon}, we complement our weight decay ablation by comparison of both variants in our benchmarking setup from \S~\ref{sec:setup} in~\cref{fig:muon-dmuon-final-val-loss,fig:muon-dmuon-val-loss}.
We report how much the algorithm with weight decay outperforms the basic variant.
Thus, showing that weight decay should definitely be applied across different optimizers.

With our weight decay ablation, we are ready to provide one more insight.

\prettybox{
\takeaway{tkw:weight_decay}The use of weight decay, particularly a large decoupled weight decay term ($0.5$ and above), can significantly impact the final loss value and optimizer behavior. 
However, for extended training horizons, a moderate, non-zero weight decay of $0.1$ proves to be a robust option.
}

\begin{figure*}[h]
    \vspace{-1em}
    \centering
    \subfigure[\texttt{Signum}, \texttt{Lion}, and \texttt{Sophia} diverge with large $\gamma$.]{
        \includegraphics[width=0.48\linewidth]{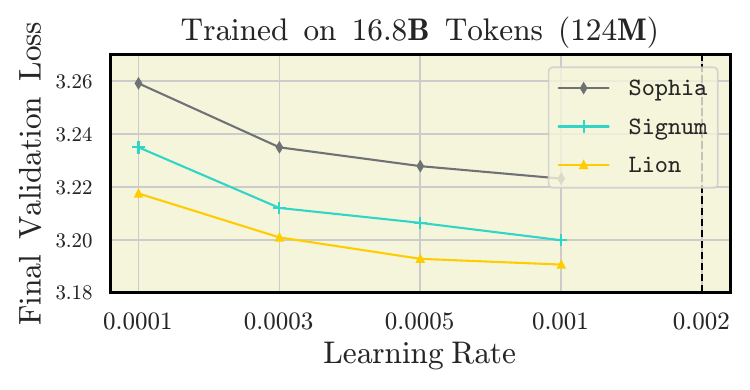}
    }
    \hfill
    \subfigure[Parabolic shape of $\gamma$ sensitivity for most optimizers.]{
        \includegraphics[width=0.48\linewidth]{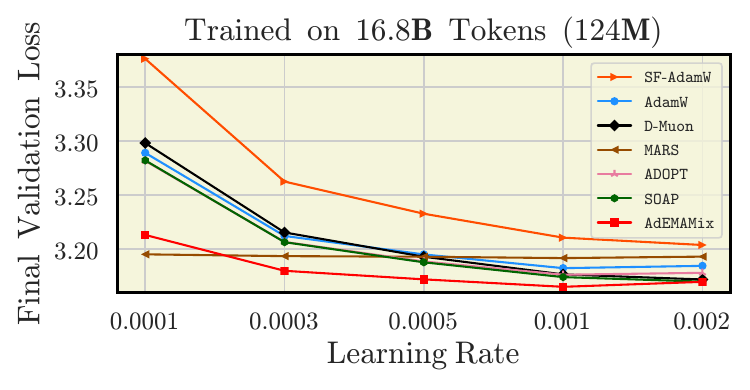}
    }
    \caption{\textbf{Optimal learning rate stability across optimizers.}
    The optimal learning rate determined during tuning on $2.1\mathbf{B}$ tokens remains consistent after a learning rate sweep on $16.8\mathbf{B}$ tokens for most optimizers.
    In \textbf{(a)}, we observe that sign-based methods and similar to them \texttt{Sophia} diverge with increasing learning rate.
    Interestingly, in \textbf{(b)}, \texttt{SF-AdamW}, \texttt{SOAP}, and \texttt{D-Muon} demonstrate their best performance with a large learning rate of $0.002$, while \texttt{MARS} maintains remarkably consistent performance across the entire learning rate sweep.
    See \cref{fig:ap_lrsensitivity} and \cref{sec:ap_124mablations} for more details.
    }
    \label{fig:lrsensitivity}
\end{figure*}

\textbf{Learning rate sensitivity.}
Since we tune optimizers at a shorter runs and then extrapolate, we pose the question whether the best learning rate we have found so far transfers to the larger training duration.
To verify this, we run $124\mathbf{M}$ model on $16.8\mathbf{B}$ tokens in $256\times512$ batch size setting, sweeping the learning rate across five typical values: $\{1e^{-4}, 3e^{-4}, 5e^{-4}, 1e^{-3}, 2e^{-3}\}$.
The best learning rate for each method at the moment of hyperparameter tuning on near Chinchilla optimal $2.1\mathbf{B}$ training duration we report in \cref{sec:ap_124mtuning}.
A summary of our results for larger number of tokens is provided in \cref{fig:lrsensitivity} and detailed results of the sweep are presented in \cref{sec:ap_lrsensitivity}.

\prettybox{
\takeaway{tkw:lr_sensitivity}For most optimizers, the learning rate $\gamma_{\max}$ selected near the Chinchilla optimal horizon transfers smoothly to our $8\times $longer run.
Notably, we found that: (\rom{1}) sign-based methods and \texttt{Sophia} diverge with larger $\gamma_{\max} = 2e^{-3}$; (\rom{2}) while \texttt{SF-AdamW}, \texttt{SOAP}, and \texttt{D-Muon} achieve better performance with such a large learning rate; (\rom{3}) \texttt{MARS} demonstrates a very consistent performance across $\gamma$ sweep, which is not typical for other optimizers.
}

\begin{figure*}[h]
    \vspace{-0.8em}
    \centering       
        \includegraphics[width=0.8\linewidth]{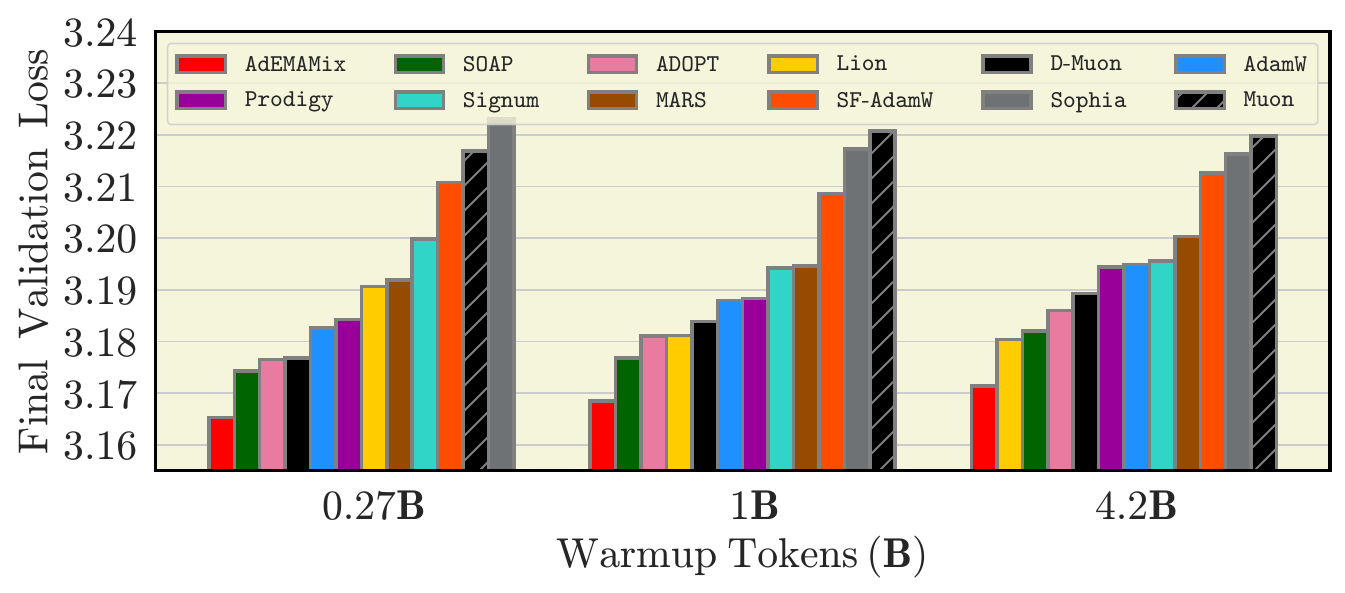}
    \caption{\textbf{Warmup ablation.}
    For $124\mathbf{M}$ model trained on the batches of $256\times512$ tokens, we perform a sweep over the linear warmup durations of $\{1.56\%, 6.25\%, 25\%\}$ of the length of training, which corresponds to $\{2, 8, 32\}\mathbf{k}$ steps, respectively.
    Clearly, sign-based optimizers, \texttt{Sophia}, and \texttt{SF-AdamW} benefit from the increased warmup.
    }
    \vspace{-1em}
    \label{fig:warmup}
\end{figure*}

\textbf{Warmup ablation.}
Another important ingredient of the pretraining is the learning rate warmup in the initial phase of training.
Recent studies have explored the necessity of warmup in modern deep learning, with some investigating its elimination~\cite{kosson2024analyzingreducingneed,xiong2020layernormalizationtransformerarchitecture} and others, ablating it to improve model performance and stability~\cite{zhang2024doescriticalbatchsize,gilmer2021losscurvatureperspectivetraining,wortsman2023smallscaleproxieslargescaletransformer}. 
We focus on the latter, examining how warmup affects optimizer setup and whether it can significantly enhance performance.
For each optimizer's best configuration for $16.8\mathbf{B}$ tokens run, we vary (a linear) warmup across three values: $\{0.27, 1, 4.2\}\mathbf{B}$ tokens, which corresponds to $\{2, 8, 32\}\mathbf{k}$ iterations.
Our choice of the largest warmup value is inspired by~\cite{zhang2024doescriticalbatchsize}.
We describe this experiment in \cref{sec:ap_warmupablation}.
Mainly, we observe that \texttt{Signum} and \texttt{SF-AdamW} perform better with a larger warmup of $8\mathbf{k}$ steps when training on $16.8\mathbf{B}$ tokens.
We also ablate the claim of Zhang et al.~\cite{zhang2024doescriticalbatchsize} that a warmup of $25\%$ of the Chinchilla optimal duration is the best.
However, our findings contradict this assertion (see \cref{fig:warmup_sweep_adamw}). 
We show that a moderate values of the warmup, generally, is better.
However, different optimizers could prefer different number of warmup steps. 
As such, \texttt{SF-AdamW}, \texttt{Sophia}, \texttt{Signum}, and \texttt{Lion} benefit from a large warmup, which is clearly depicted in \cref{fig:warmup}.
Surprisingly, with a warmup of $\{8, 32\}\mathbf{k}$ steps, \texttt{Lion} outperforms the \texttt{AdamW} baseline.

\prettybox{
\takeaway{tkw:warmup_ablation}As usual, a warmup duration in LLM pretraining is around $2\mathbf{k}$ steps.
However, we reveal that the warmup duration is optimizer-dependent and should be tuned: for \texttt{SF-AdamW}, \texttt{Sophia}, and \texttt{Signum}, longer warmup results in improved final performance, while \texttt{Lion} with increased warmup also surpasses strong baselines such as \texttt{AdamW}.
}

\textbf{Ablation on WSD, cosine, and linear $\gamma$-schedulers.}
Learning rate schedulers received a lot of attention recently \cite{shen2024powerschedulerbatchsize,schaipp2025surprisingagreementconvexoptimization,hägele2024scalinglawscomputeoptimaltraining}.
To study the connection between optimizers and learning rate schedulers, we conduct experiments comparing cosine~\cite{loshchilov2017sgdrstochasticgradientdescent} learning rate schedulers with WSD~\cite{hu2024minicpmunveilingpotentialsmall,zhai2022scalingvisiontransformers} and linear.
To compare with WSD, we consider optimally tuned (as in \S~\ref{sec:setup} and \cref{sec:ap_124mtuning}) cosine scheduler for each optimizer, and replicate the setup of H\"agele et al.~\cite{hägele2024scalinglawscomputeoptimaltraining}, which allows us to avoid adjusting additional hyperparameters (see details in \cref{sec:ap_tuning}).
To compare with the linear scheduler, we use the same maximal learning rate as for cosine.
Our findings, which demonstrate the superiority of the cosine scheduler\footnote{We emphasize that the difference between the two schedulers is generally less than $5\%$ of the total compute spent. However, this still represents a significant gap in our benchmarking setup, e.g., \texttt{SF-AdamW} may outperform \texttt{AdamW} when the latter employs WSD (see \cref{fig:owt2wsdcosine}).} across various optimization methods, are presented in \cref{fig:wsdvscosine}, and in the Appendix, \cref{fig:wsdcosine,fig:owt2wsdcosine}.
These results not only validate our initial preference but also provide insights into the interaction between learning rate schedules and different optimizers in large-scale language model training.

\begin{figure*}[h]
    \centering
    \subfigure[\texttt{Muon} ``prefers'' WSD.]{
        \includegraphics[width=0.316\linewidth]{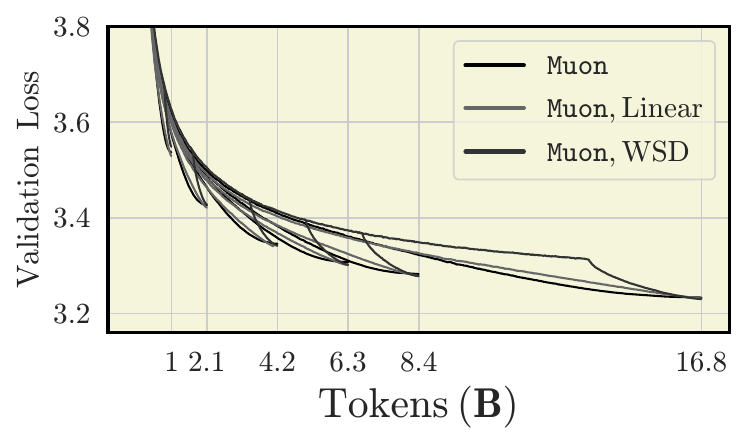}
    }
    \hfill
    \subfigure[No preference for \texttt{Sophia}.]{
        \includegraphics[width=0.316\linewidth]{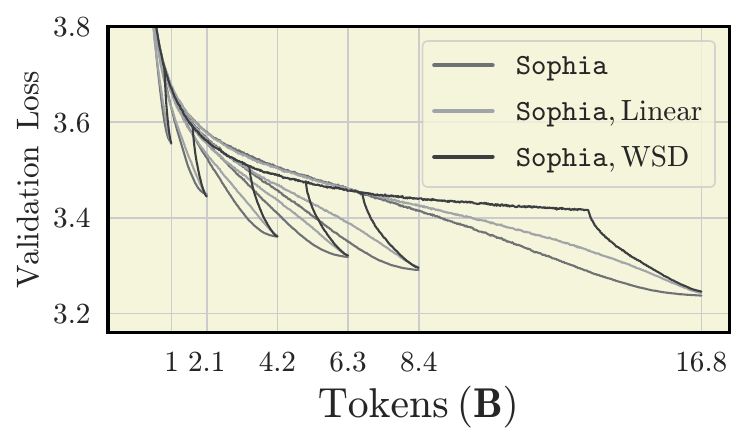}
    }
    \hfill
    \subfigure[Cosine \& \texttt{AdamW}.]{
        \includegraphics[width=0.316\linewidth]{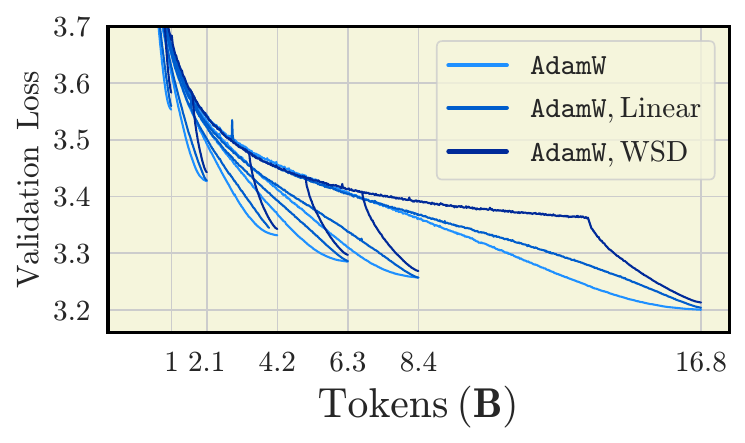}
    }
    \vspace{-2mm}
    \caption{\textbf{Comparisons between cosine, WSD, and the linear schedulers.}
    Notably, schedulers behave differently with respect to optimizer.
    In \textbf{(a)}, the \texttt{Muon} optimizer shows a preference for WSD across most training durations.
    \texttt{Sophia} exhibits an almost perfect match between all three schedulers.
    However, for \texttt{AdamW}, along with the majority of other optimizers studied (see \cref{fig:wsdcosine}), we get a better performance with cosine. 
    We also report a detailed comparison for all optimizers in \cref{sec:ap_124mablations}, and cover additional ablations on another dataset (see \cref{fig:owt2wsdcosine}).
    }
    \vspace{-1em}
    \label{fig:wsdvscosine}
\end{figure*}

\begin{figure*}[h]
    \vspace{-1em}
    \centering
    \subfigure[``Bump'' for \texttt{Lion}, due to large $\gamma$.]{
        \includegraphics[width=0.316\linewidth]{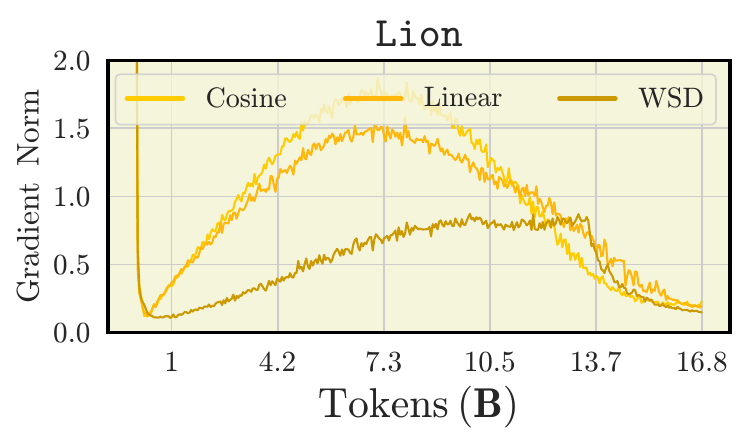}
    }
    \hfill
    \subfigure[Gradient norm increases.]{
        \includegraphics[width=0.316\linewidth]{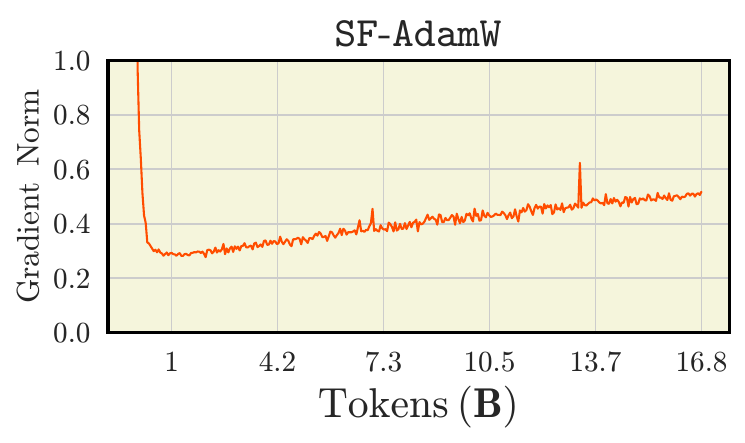}
    }
    \hfill
    \subfigure[``Bump'' for \texttt{Signum}.]{
        \includegraphics[width=0.316\linewidth]{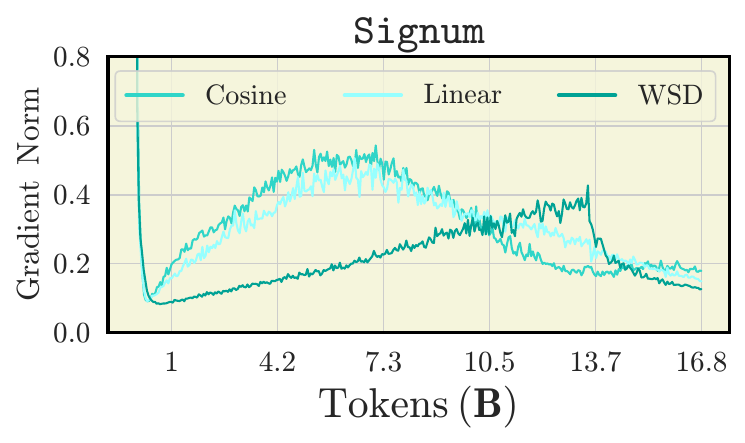}
    }
    \caption{\textbf{Gradient norm patterns for different schedulers.}
    In our experiments with learning rate schedulers~(\cref{fig:wsdvscosine}), we also track a gradient norm changes for all optimizers, showing how it is affected by the choice of the scheduler.
    We use our cosine baseline, linear scheduler with the same optimal learning rate as for cosine, and the WSD scheduler with a rule of thumb described in \cref{sec:ap_tuning}.
    We found that gradient norm evolution for majority of optimizers resembles the \texttt{SF-AdamW} pattern in~\textbf{(b)}.
    Exceptions are sign-based methods---\texttt{Signum}~\textbf{(c)}, and \texttt{Lion}~\textbf{(a)}.
    }
    \label{fig:grad-norms-main-part}
\end{figure*}

In addition to ablating schedulers, we emphasize a community interest in studying the training dynamics, especially the gradient norms patterns~\cite{defazio2025gradientsrapidlyincreasenear,defazio2024optimallineardecaylearning,kosson2024rotationalequilibriumweightdecay}.
As noticed in prior works, gradient norms tend to increase when training with certain values of $\gamma_{\max}$ and $\lambda$.
Worth mentioning that an alternative explanation exists~\cite{merrill2023effectsparameternormgrowth}, motivated by the ReLU networks, which suggests that along a fixed direction in parameter space, the gradient norm is roughly proportional to the parameter norm, which increases.
Regarding optimizers, we study their gradient norm patterns and report results in \cref{fig:grad-norms-main-part}, and in \cref{fig:ap_grad_norms_all}.

\prettybox{
\takeaway{tkw:schedulers}A choice of the learning rate scheduler is also optimizer-related.
For most methods, the cosine scheduler dominates. 
However, linear scheduler outperforms or matches cosine and WSD for sign-based methods, \texttt{SOAP}, and \texttt{MARS}.
WSD appears to be the best option for \texttt{Muon}.
We also study the gradient norm patterns for all optimizers and highlight it for sign-based method, who attain the completely different ''bump'' shape.
}

\subsection{Benchmarking \& Ablations at Medium Scale: Training Models of $\mathbf{210M}$ Parameters}
\label{sec:210m_benchmarking}

\begin{wrapfigure}{r}{0.5\linewidth} 
    \vspace{-1.8em}
    \includegraphics[width=1.05\linewidth]{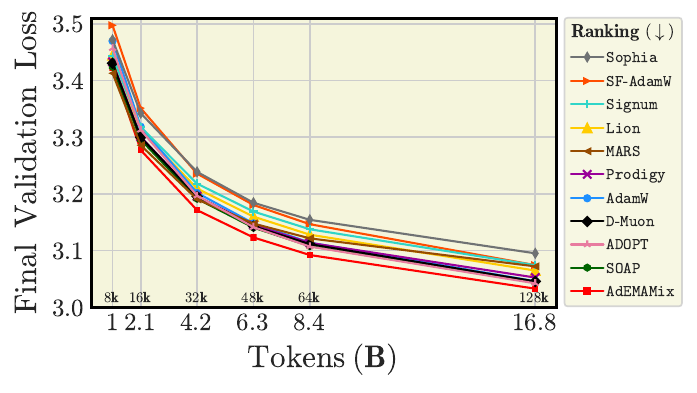}
    \caption{\textbf{Ranking of optimizers for $\mathbf{210M}$ models with the batch size of $\mathbf{256\times512}$ tokens.}
    Increasing a model size from $124\mathbf{M}$ to $210\mathbf{M}$ results in almost identical ranking of optimizers compared to \cref{fig:benchmark-124}~\textbf{(b)}. 
    At this scale, we observe a smooth transition in our benchmarking.
    }
    \label{fig:benchmark-210} 
\end{wrapfigure}

\begin{figure*}[h]
    \centering
    \subfigure[Underperforming \texttt{AdamW}.]{
        \includegraphics[width=0.316\linewidth]{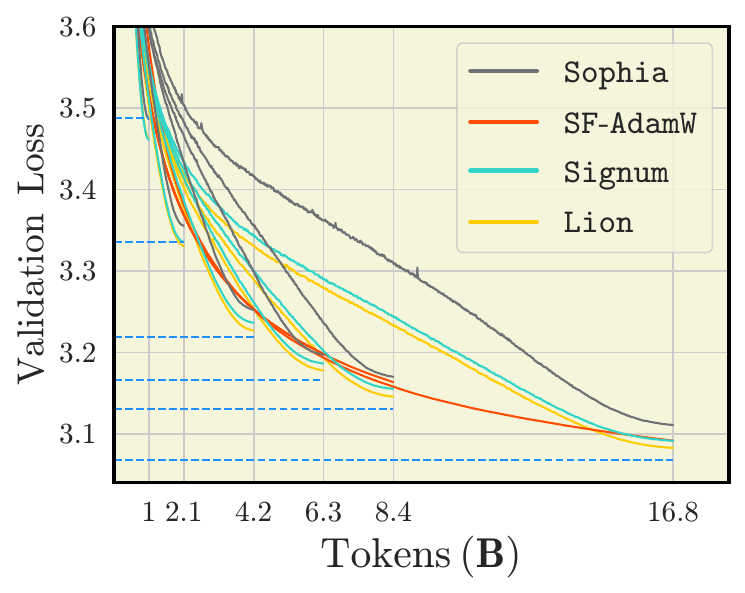}
    }
    \hfill
    \subfigure[Underperformers \& \texttt{D-Muon}.]{
        \includegraphics[width=0.316\linewidth]{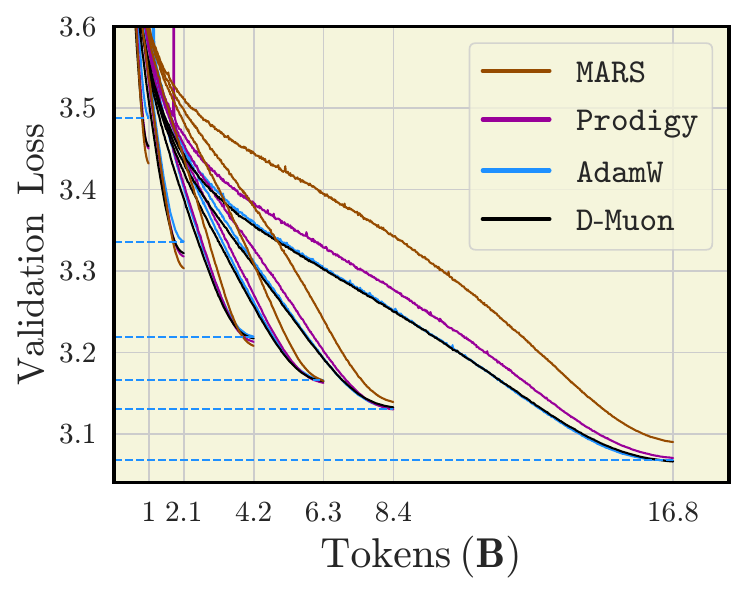}
    }
    \hfill
    \subfigure[Best optimizers.]{
        \includegraphics[width=0.316\linewidth]{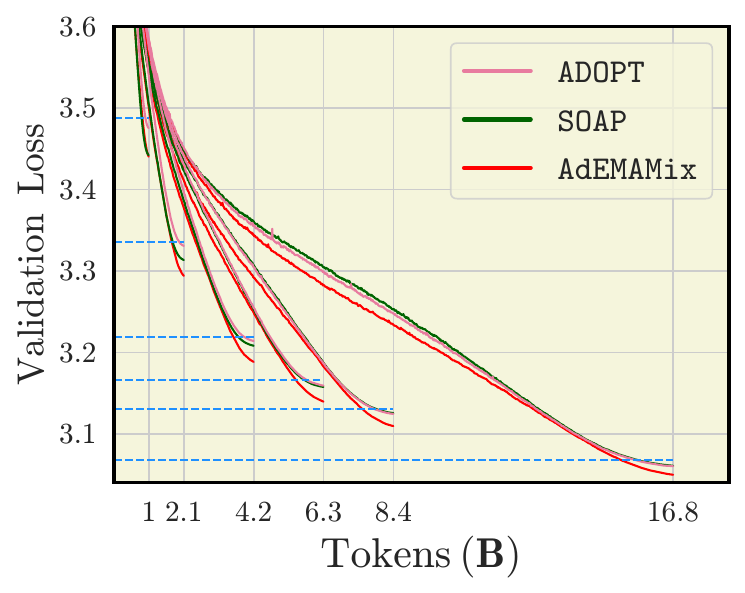}
    }
    \vspace{-1em}
    \caption{\textbf{Comparing optimizers for training a $\mathbf{210}\mathbf{M}$ parameter LLM.}
    We plot the training dynamics of: \textbf{(a,b)} optimizers that underperform \texttt{AdamW} for pretraining a $210\mathbf{M}$ model; \textbf{(c)} optimizers that outperform \texttt{AdamW} in this setup.
    A complete ranking of methods in this setting is in \cref{fig:benchmark-210}.
    }
    \label{fig:benchmarking-210m-losses}
\end{figure*}

\textbf{Results with a batch size of $\mathbf{256\times512}$.}
In this section, we verify if our selected hyperparameters from smaller $124\mathbf{M}$ allow accurate transfer to a slightly larger model.
We point out that the most important hyperparameters to be sweeped are learning rate and gradient clipping.
Regarding the learning rate, we observe that it only becomes a sensitive choice for sign-based methods, while the optimal hyperparameters for \texttt{AdamW} remain the same. 
After re-tuning the learning rate for sign-based optimizers (see \cref{sec:ap_210mtuning}), we replicate the setup from \S~\ref{sec:setup}: we stay in the ``large'' batch regime and train for the same number of steps (tokens) as in \cref{fig:benchmark-124}~\textbf{(b)}.
We report our benchmarking for $210\mathbf{M}$ models in \cref{fig:benchmark-210} and the training dynamics of optimizers in \cref{fig:benchmarking-210m-losses}.

\prettybox{
\takeaway{tkw:smooth_transition}We do not observe a much of a change in ranking of optimizers for $210\mathbf{M}$ model, compared to benchmarking on $124\mathbf{M}$.
At the same time, we replicated almost identical hyperparameters for all optimizers, except for the learning rate for sign-base methods.
We also point out that sign-based methods are more sensitive to the learning rate while scaling the model size.
As that, we changed the peak learning rate from $10^{-3}$ to $5\cdot 10^{-4}$ for \texttt{Lion} and \texttt{Signum}.
}

\begin{figure*}[h]
    \centering
    \subfigure[Cosine, $\gamma_\text{end}=0.01\times\gamma_{\max}$.]{
        \includegraphics[width=0.31\linewidth]{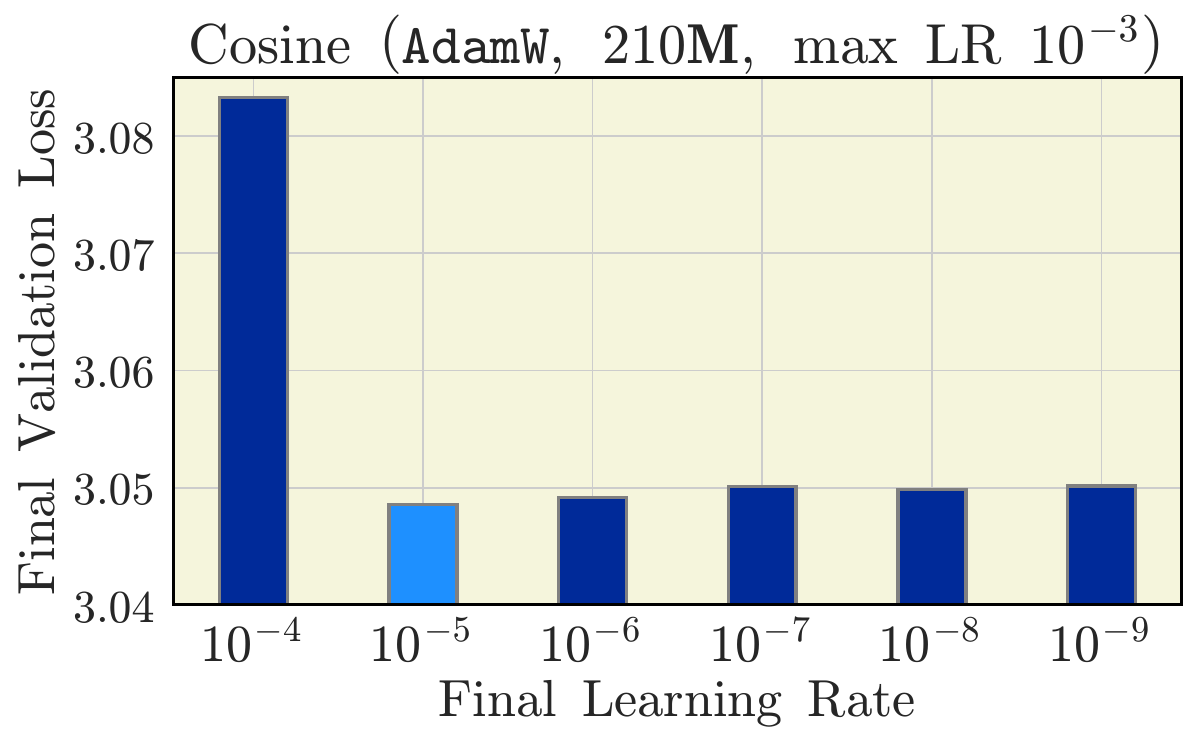}
    }
    \hfill
    \subfigure[Linear, $\gamma_\text{end}=0.001\times\gamma_{\max}$.]{
        \includegraphics[width=0.31\linewidth]{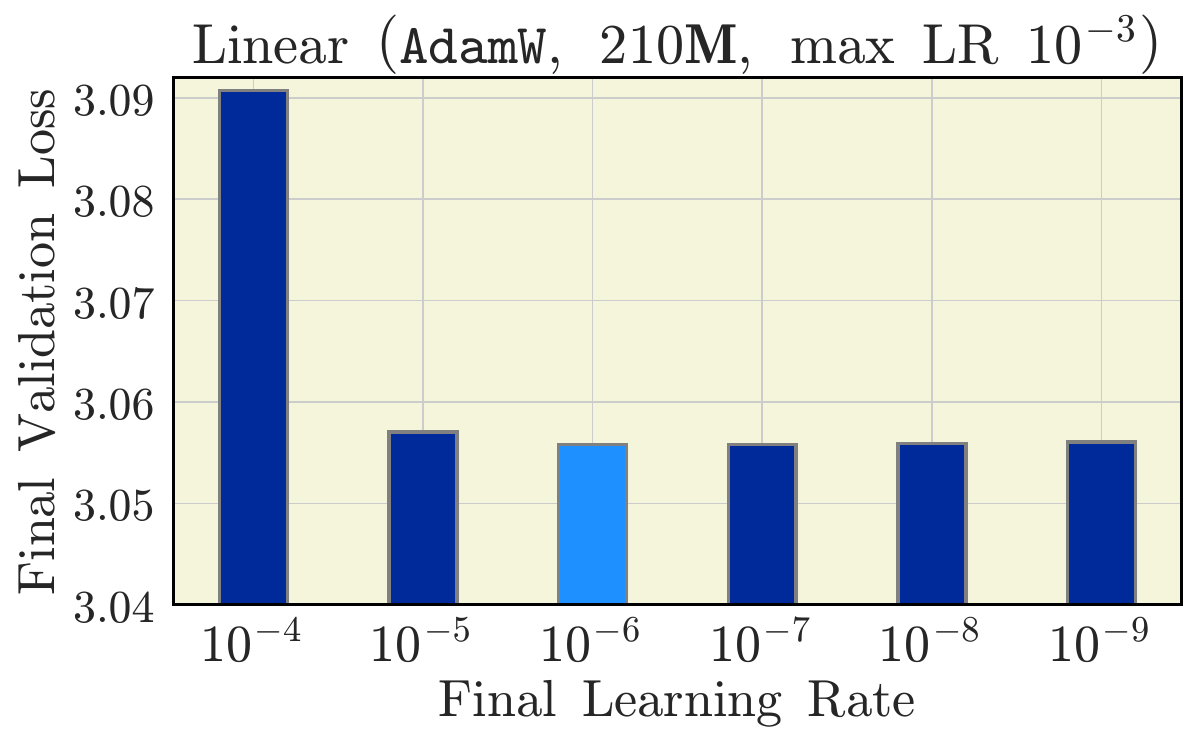}
    }
    \hfill
    \subfigure[WSD, $\gamma_\text{end}=0.01\times\gamma_{\max}$.]{
        \includegraphics[width=0.31\linewidth]{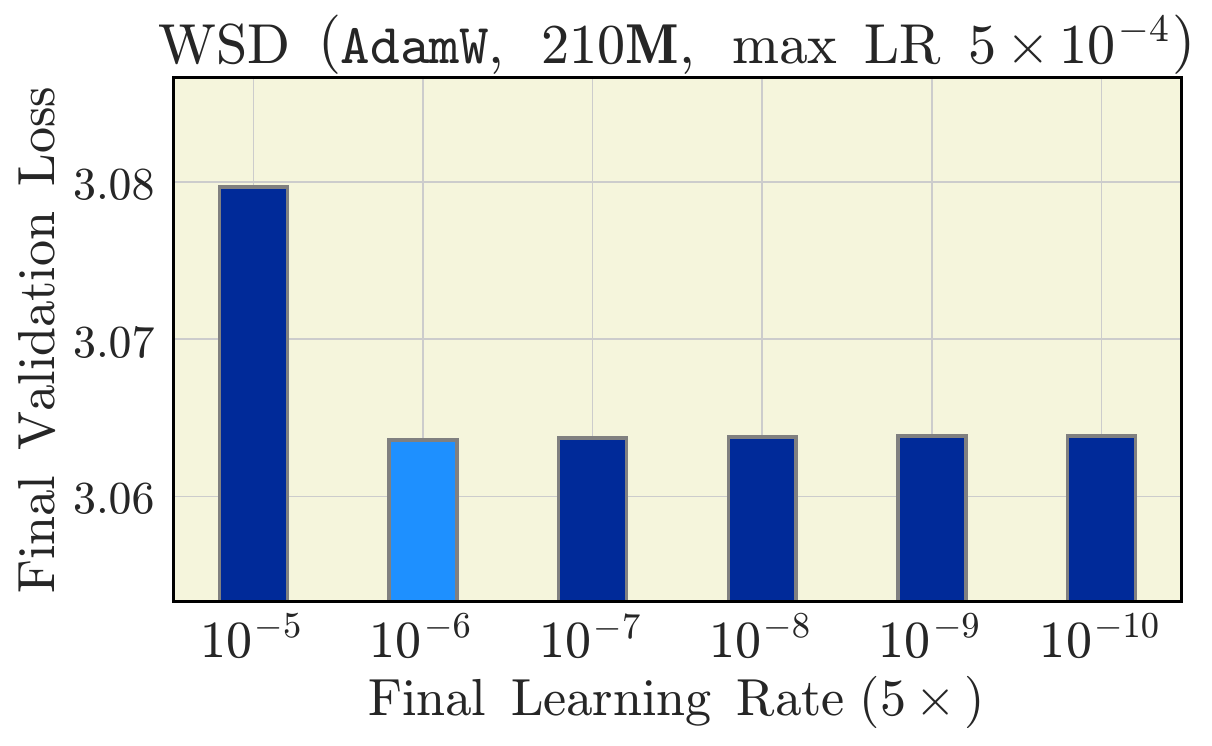}
    }
    \vspace{-2mm}
    \caption{\textbf{Decaying the learning rate down to $\mathbf{0.01} \boldsymbol{\times} \boldsymbol{\gamma}_{\boldsymbol{\max}}$ and beyond, instead of only to $\mathbf{10}\boldsymbol{\%}$.} 
    We run a $210\mathbf{M}$ Llama model, and observe a common pattern for different schedulers that decreasing the learning rate to moderate $0.01\times\gamma_{\max}$ value is a better choice than decreasing it down to zero. 
    Interestingly, the linear learning rate scheduler for models at a given scale, requires $0.001 \times \gamma_{\max}$.
    See \cref{fig:lrdecay-124m-appendix} for corresponding ablation for $124\mathbf{M}$ model.
    }
    \vspace{-1em}
    \label{fig:lrdecay}
\end{figure*}

\textbf{Decay the learning rate sufficiently.} 
Another important component of LLM pretraining---final learning rate value $\gamma_\text{end}$.
A widely adopted value in the literature regarding large-scale model training is $\gamma_\text{end} = 0.1\times \gamma_{\max}$~\cite{hoffmann2022trainingcomputeoptimallargelanguage,touvron2023llamaopenefficientfoundation,biderman2023pythiasuiteanalyzinglarge,workshop2023bloom176bparameteropenaccessmultilingual,olmo20242olmo2furious,groeneveld2024olmoacceleratingsciencelanguage,zhao2024deconstructingmakesgoodoptimizer}.
However, recent works question~\cite{bergsma2025straightzerolinearlydecaying,schaipp2025surprisingagreementconvexoptimization,hägele2024scalinglawscomputeoptimaltraining,li2025predictablescalei,deepseekai2024deepseekv3technicalreport} this heuristic, proposing to decay $\gamma$ to $0.01\times\gamma_{\max}$ or to smaller values. 
Thus, we ablate how does the learning rate decay rule combines with different schedulers and affects the overall optimizer's performance.
We study this effect on the \texttt{AdamW} method, and then apply the best-found heuristic to all other optimizers.
Interestingly, our ablation (\cref{fig:lrdecay}) suggests that the best style for cosine and WSD is to decay to $0.01\times\gamma_{\max}$, while for the linear schedule the best-performing run with decay to $0.001\times\gamma_{\max}$.
What is more important, is that the previous decay style to $10\%$ of the $\gamma_{\max}$ delivers much worse results compared to any decay we consider.
Building on this findings, we consistently use cosine decay down to $0.01\times \gamma_{\max}$.

\prettybox{
\takeaway{tkw:lr_decay}Decaying the learning rate further than $10\%$ of the maximal significantly improves the results.
However, for different schedulers, the best final learning rate is different.
}

\subsection{Scaling Up: Benchmarking models of $\mathbf{583M}$ and $\mathbf{720M}$ Parameters}
\label{sec:scaling_up}

\begin{wrapfigure}{l}{0.45\linewidth}  
    \vspace{-1.5em}
    \includegraphics[width=\linewidth]{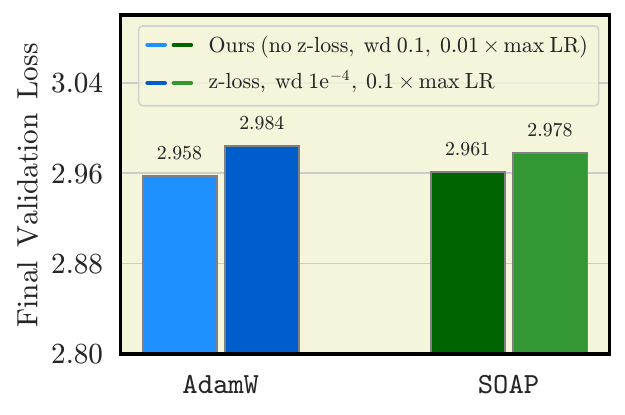}
    \caption{\textbf{Ablation of $z$-loss regularization.}
    Incorporating the $z$-loss regularizer does not improve the final loss or reduce the spikiness of the loss curves.
    Moreover, combining $z$-loss with small weight decay and decaying $\gamma$ down to $10\%$, further degrades overall performance.
    Notably, these changes can reverse the relative ranking of optimizers compared to the results reported by Vyas et al.~\cite{vyas2024soapimprovingstabilizingshampoo}.
    }
    \label{fig:z_loss_ablation} 
    \vspace{-1.3em}
\end{wrapfigure}

\textbf{Comparison between our setting and Vyas et al.~\citep{vyas2024soapimprovingstabilizingshampoo}.}
We pick two methods: \texttt{AdamW}, \texttt{SOAP}, and run experiments with a larger model of $583\mathbf{M}$ parameters, and a large batch size of $2\mathbf{M}$ tokens. 
The goal is to get closer to one of the settings described in \cite{vyas2024soapimprovingstabilizingshampoo}, i.e. train for the Chinchilla optimal amount of tokens and use the same batch size. 
Therefore, we train for $6500$ iterations, corresponding to $13\mathbf{B}$ tokens.
We found several key differences between our codebase and \cite{groeneveld2024olmoacceleratingsciencelanguage}, used by Vyas et al.~\citep{vyas2024soapimprovingstabilizingshampoo}: (\rom{1}) we decay the learning rate to $0.01 \times \gamma_{\max}$ instead of $0.1 \times \gamma_{\max}$, with $\gamma_{\max}$ being the maximum learning rate, (\rom{2}) we use typical weight decay values of e.g. $0.1$ instead of smaller values such as $0.01$ or $0.0001$, (\rom{3}) we do not use a $z$-loss in addition to ours. 
Our ablations in \cref{fig:lrdecay,fig:lrdecay-124m-appendix} already confirm that properly decaying the learning rate has an important effect on optimization.
Regarding $z$-loss and weight decay, we run an ablation to compare both settings and conclude that removing the $z$-loss and increasing the weight decay to $0.1$ improves the results.
We remind that hyperparameter choice in \cite{vyas2024soapimprovingstabilizingshampoo} has been suggested by popular codebases for LLM pretraining~\cite{groeneveld2024olmoacceleratingsciencelanguage,olmo20242olmo2furious,biderman2023pythiasuiteanalyzinglarge}.
In that view, we pose the following observation to practitioners.

\prettybox{
\takeaway{tkw:vyassetup}Hyperparameter choices commonly imposed by popular codebases---such as final learning rate, $z$-loss regularization, and weight decay---substantially affect both absolute performance and the relative ranking of optimizers at Chinchilla scale.
}

\textbf{Results on $\mathbf{720}\mathbf{M}$ parameter model \& $\mathbf{1M}$ batch size.}
To expand our ablations towards more practical scales, we train a $720\mathbf{M}$ parameter model with a batch size of $1\mathbf{M}$ tokens. 
As previously, we include both the Chinchilla optimal horizon and beyond, following the setup in \S~\ref{sec:setup}. 
Our goal is to characterize how optimizer performance evolves with increased model size, batch size, and total tokens seen.

We observe that sign-based methods and \texttt{Sophia} require careful re-tuning of the learning rate to converge on larger models.
Notably, despite increasing the training horizon in terms of tokens, with larger batch size, the number of steps is reduced compared to our runs in \S~\ref{sec:smallscalebench} and \S~\ref{sec:210m_benchmarking}; in this part of the benchmarking, we consider runs of $\{8, 16, 48\}\mathbf{k}$ iterations (the Chinchilla optimal duration at $\sim14.4\mathbf{k}$).
This reduction in steps necessitates re-tuning optimizer-related hyperparameters such as $\beta_2$.
We describe hyperparameter changes in \cref{sec:ap_720mtuning}.

\begin{figure*}[h]
    \centering
    \subfigure[Underperforming \texttt{AdamW} (\rom{1}).]{
        \includegraphics[width=0.316\linewidth]{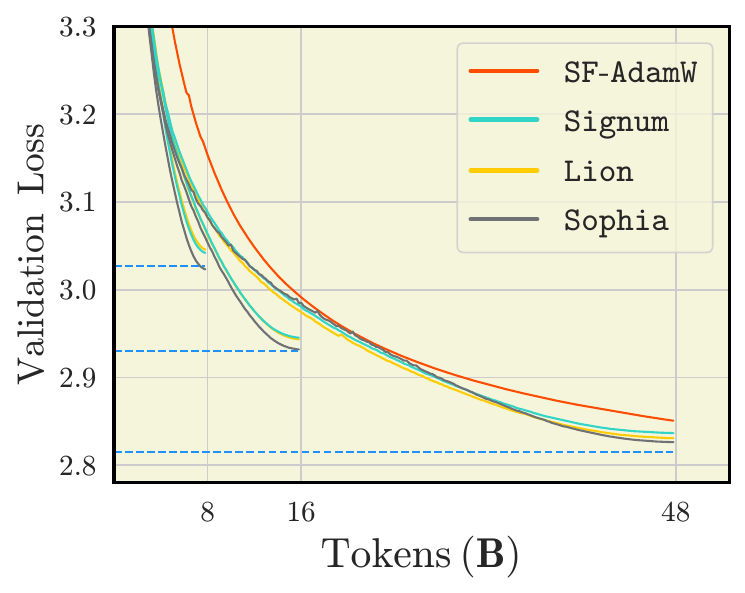}
    }
    \hfill
    \subfigure[Underperforming \texttt{AdamW} (\rom{2}).]{
        \includegraphics[width=0.316\linewidth]{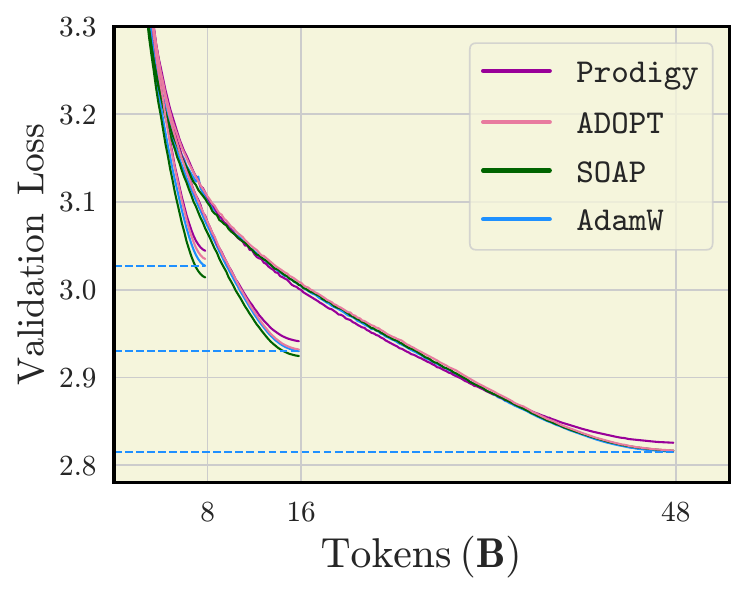}
    }
    \hfill
    \subfigure[Outperforming \texttt{AdamW}.]{
        \includegraphics[width=0.316\linewidth]{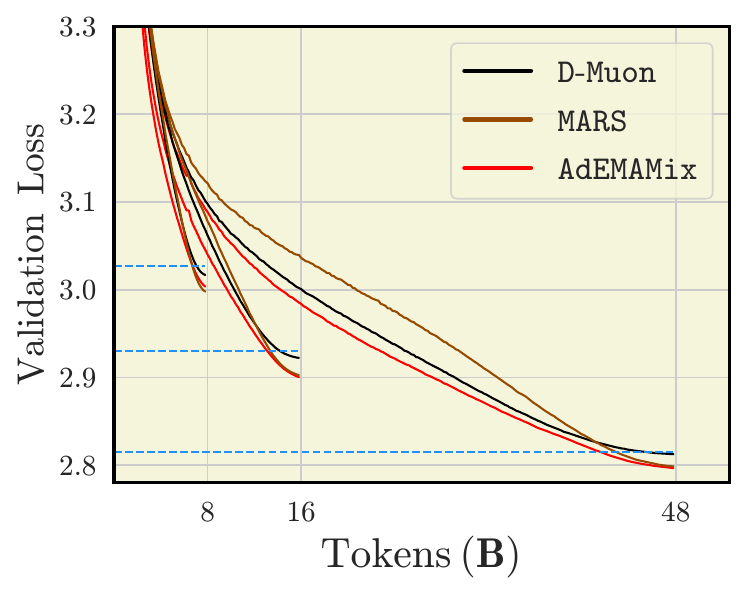}
    }
    \vspace{-2mm}
    \caption{\textbf{Comparing optimizers for training a $\mathbf{720M}$ parameter LLM.}
    We conduct runs with the batch size of $1\mathbf{M}$ tokens.
    While previous ablations (see~\cref{fig:scalebs_scaletokens}) reveal that sign-based methods can outperform \texttt{AdamW} at sufficiently large batches, this advantage does not persist when scaling model size.
    On another hand, \texttt{MARS}, that also benefits from the increased batch size, along with \texttt{AdEMAMix} dominates over other optimizers with a huge gap.
    }
    \label{fig:benchmarking-720m-losses}
\end{figure*}

Studying the training dynamics~(\cref{fig:benchmarking-720m-losses}), we find that \texttt{SF-AdamW}, and sign-based \texttt{Lion} and \texttt{Signum} scale poorly.
\texttt{Sophia} can outperform our \texttt{AdamW} for short runs of $8\mathbf{k}$ iterations, but then degrades significantly.
Interestingly, \texttt{MARS} greatly benefits from this setup, emerging the second best-performing optimizer, closely following \texttt{AdEMAMix}: as it benefits from large batch size~(see~\cref{fig:scalebs_scaletokens}~(\textit{left})), and does not degrade with increased model size unlike \texttt{Signum}, and \texttt{Lion}.
On another hand, \texttt{Prodigy} was proven to be more beneficial at larger batch size, however, this setup it occured to be less performant.
\texttt{D-Muon} is consistent across all settings we have tried, while \texttt{Muon} degrades when scaling model size (\cref{fig:muon-dmuon-val-loss}~\textbf{(c)}).

As in \cite{vyas2024soapimprovingstabilizingshampoo}, we find that \texttt{SOAP} outperforms \texttt{AdamW} at the Chinchilla optimal duration and below.
However, in longer training, \texttt{AdamW} narrows the gap and eventually surpasses \texttt{SOAP}.
Another claim regarding the \texttt{SOAP} optimizer---that it is more beneficial, when the batch size is sufficiently large---remains quite questionable: (\rom{1}) as~\cref{fig:z_loss_ablation}~(runs with $2\mathbf{M}$ batch size) suggests that the matter of \texttt{SOAP} being better than \texttt{AdamW} is conditioned by the setup choice, which when properly tuned turns that \texttt{AdamW} becomes better even at Chinchilla optimal duration; (\rom{2}) when considering $1\mathbf{M}$ batch size setup in \cref{fig:benchmark-720,fig:benchmarking-720m-losses}, the performance gain of \texttt{SOAP} over \texttt{AdamW} is less pronounced than in our settings with smaller batches for $124\mathbf{M}$ and $210\mathbf{M}$ models (see Figures~\ref{fig:benchmarking-124m-losses}~\textbf{(b)}~and~\ref{fig:benchmarking-210m-losses}~\textbf{(c)}).

\vspace{-1em}

\vspace{-1em}
\prettybox{
\takeaway{tkw:scalingupbenchmarking}(\rom{1}) At larger scale of model and batch size, \texttt{AdEMAMix} and \texttt{MARS} dominate, by far outperforming others---see~\cref{fig:benchmark-720}.
(\rom{2}) Despite training with large batches, \texttt{Signum} and \texttt{Lion} scale poorly.
(\rom{3}) \texttt{D-Muon} is consistent across all our benchmarking setups.
}

\begin{wrapfigure}{r}{0.5\linewidth}
    \vspace{-1.3em}
    \includegraphics[width=1.0\linewidth]{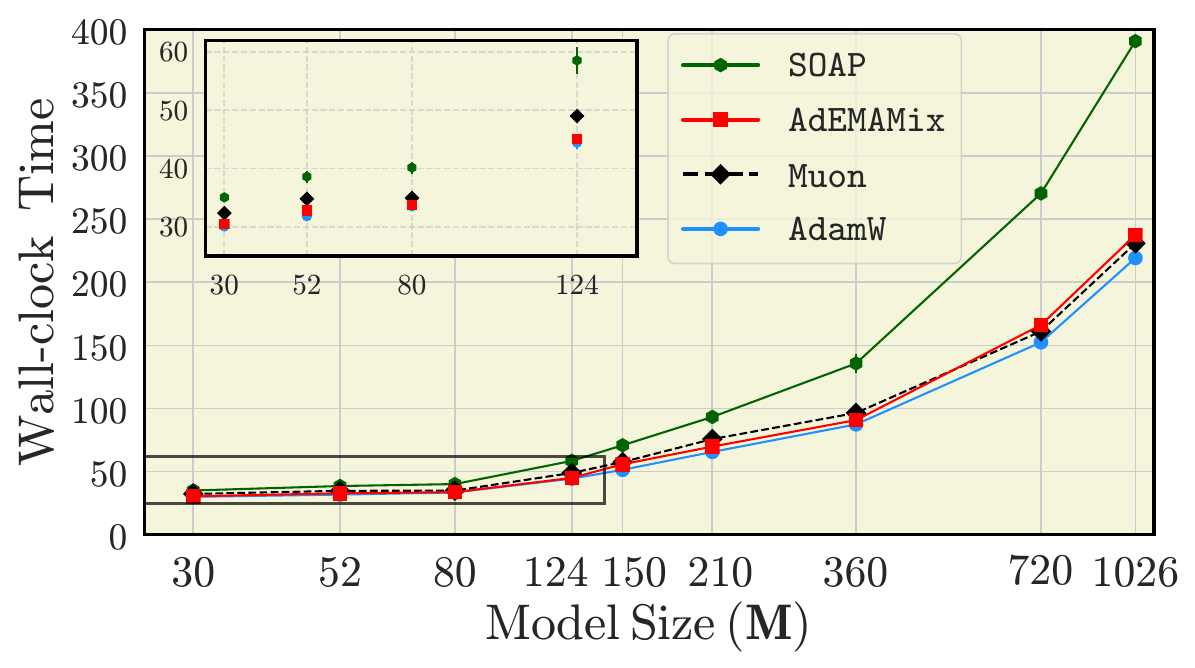}
    \caption{\textbf{Wall-clock time comparison.}
    \texttt{SOAP} slows down the most as model size increases. 
    }
    \label{fig:wall_time} 
    \vspace{-0.9em}
\end{wrapfigure}

\textbf{Wall-clock time comparison.}
After conducting experiments for models of different sizes, we are ready to present the wall-time comparison for each method.
For this purposes, we use a single GPU, and run each optimizer for $100$ iterations on a small batch size of $16$ without gradient accumulation and \texttt{torch.compile}.
In this ablation, we consider a wider range of model sizes ($30\mathbf{M}$--$1\mathbf{B}$).
We run each method $5$ times with different seeds, compute the standard deviation, and report the mean wall-clock time per $100$ iterations for each model configuration.
We observe that all methods take the roughly the same or very close time to complete $100$ iterations, with the exception of \texttt{SOAP}.
We point out that wall-clock time for all optimizers exhibits a linear dependence on the model size (``model size'' axis is rescaled in plots).
However, \texttt{SOAP} slows down faster and we may expect a slowdown further, due to its preconditioner matrices operations which are fast only for certain matrices that are smaller than a predefined size.
See details of this ablation in \cref{sec:ap_walltime}, and \cref{fig:ap_walltime,fig:ap_wall_time_individual}.

\prettybox{
\takeaway{tkw:wall_time}Most optimizers exhibit similar wall-time performance, with sign-based methods being slightly faster~(\cref{fig:ap_walltime}). \texttt{SOAP} is the main exception, showing a notable slowdown as model size increases.
}

\subsection{Extension to MoEs}
\label{sec:moe}
The goal of ablating optimizer performance on MoE models is to assess whether our previous benchmarking results transfer smoothly to a new type of architecture.
To show this smooth transition, we utilize an old batch size setup and keep untuned all optimizer-related hyperparameters found for the corresponding dense model---simulating a situation as one would do in practice without much time for running dozens of experiments on new architectures.

\textbf{Setup \& Comparison.}
Besides training dense Llama-like transformers, we also cover a comparison for MoE architectures \cite{shazeer2017outrageouslylargeneuralnetworks}.
Our variant of MoE is based on the Switch-Transformer implementation~\cite{fedus2022switchtransformersscalingtrillion}. 
We use a classical linear gating with softmax and top-$k$ routing ($k=2$) and $8$ experts. 
The activation functions remains the same as for the dense base model from \S~\ref{sec:setup}.
Given configuration of $124\mathbf{M}$ dense Llama model, we result in approximately $520\mathbf{M}$ parameter MoE model. 
In this setting, we train with a batch size of $256\times512$ for $T\in\{42, 336\}\mathbf{k}$ iterations ($\{5.5, 44\}\mathbf{B}$ tokens).
If we assume that Chinchilla scaling law is applicable to this model, then it results in $10.4\mathbf{B}$ tokens.
See \cref{sec:ap_520moetuning} for more details.

\begin{figure*}[h]
    \vspace{-1.2em}
    \centering
    \begin{minipage}{\linewidth}
    \includegraphics[width=\linewidth]{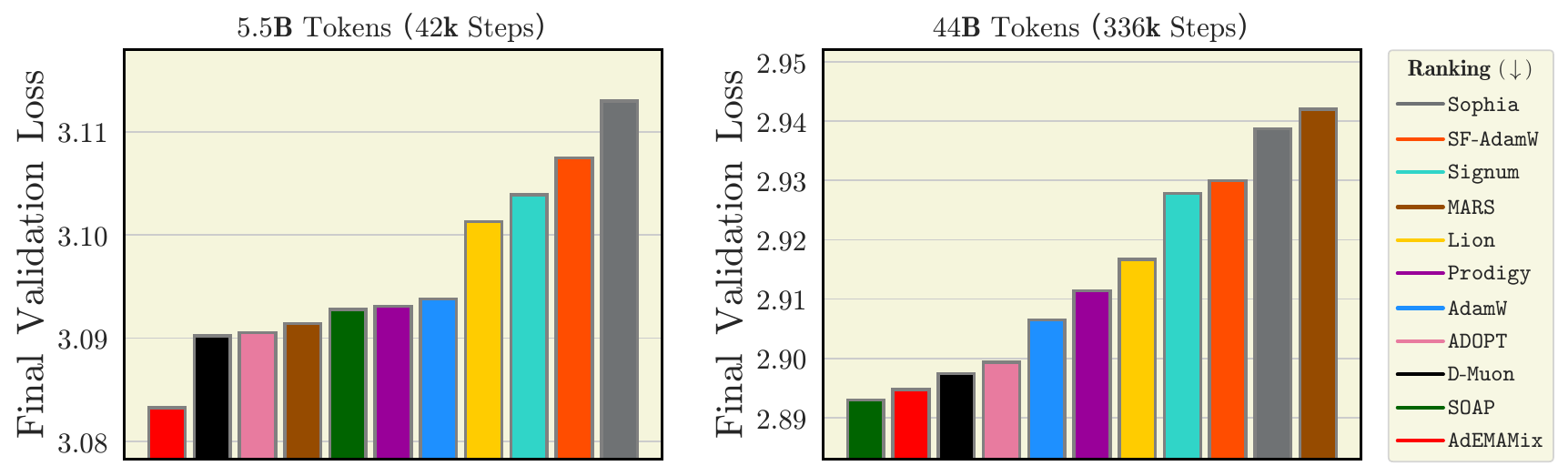}
    \end{minipage}
    \caption{\textbf{Ranking optimizers for $\mathbf{520M}$ MoE models with $\mathbf{256\times512}$ batch size.}
    We report results for models trained for both $42\mathbf{k}$ iterations~(\textit{left}), and $336\mathbf{k}$~(\textit{right}).
    MoE configuration correspond to one of the $124\mathbf{M}$ dense model.
    Optimizer rankings closely mirror those in~\cref{fig:benchmark-124}~\textbf{(b)}, indicating that our benchmarking results transfer smoothly from dense models to MoEs.
    We also see that \texttt{SOAP} outperforms \texttt{AdEMAMix} in $336\mathbf{k}$ steps run~(see also \cref{fig:moe-benchmarking-losses}), however, with re-tuned beta parameters we might expect the opposite results in longer training~(see Figures~\ref{fig:scalebs_scaletokens} and~\ref{fig:ap_retuning_betas}~\textbf{(b)}).
    }
    \label{fig:benchmark-moe}
\end{figure*}

\begin{figure*}[h]
    \vspace{-1.6em}
    \centering
    {
        \includegraphics[width=0.45\linewidth]{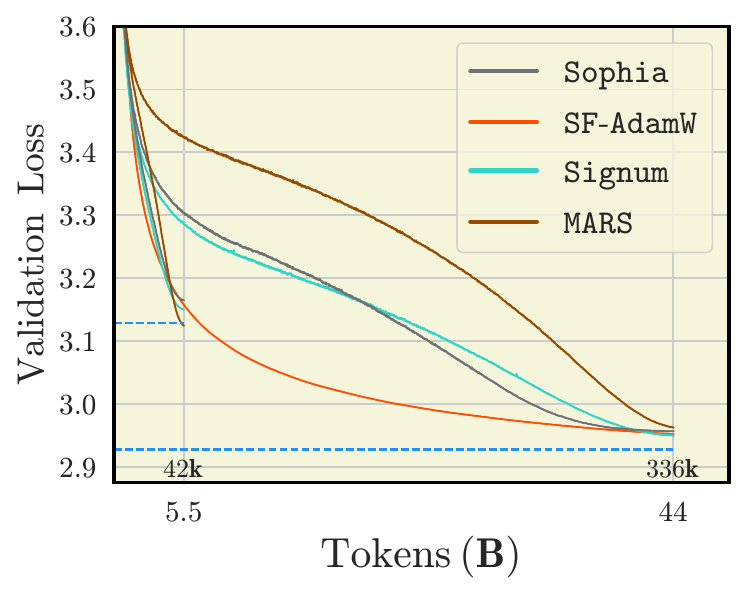}
    }
    \hfill
    {
        \includegraphics[width=0.45\linewidth]{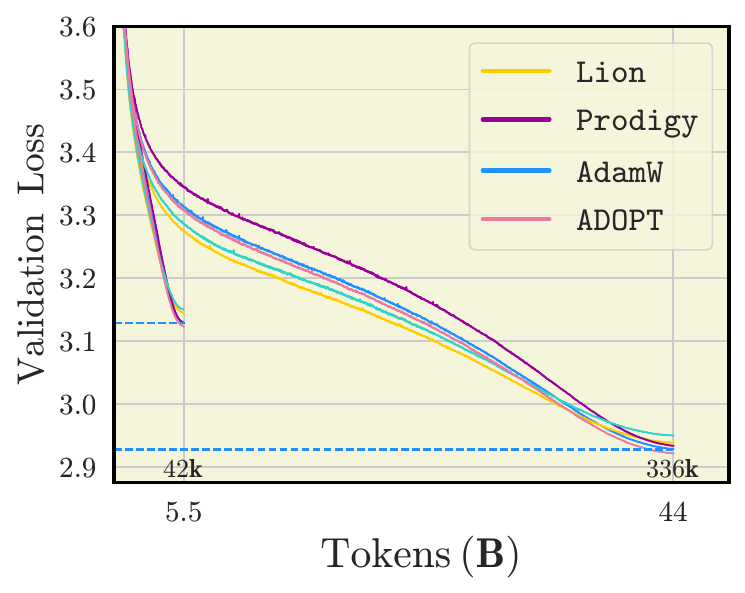}
    }
    \caption{\textbf{Comparing optimizers for training a $\mathbf{520M}$ parameter MoE.}
    Training dynamics of leading optimizers is in~\cref{fig:moe-benchmarking-losses}.
    Results closely remind those in \cref{fig:benchmarking-124m-losses}~\textbf{(a,b)}.
    The \texttt{AdamW} baseline by far outperforms \texttt{Sophia}, \texttt{SF-AdamW}, \texttt{MARS}, and sign-based methods for $44\mathbf{B}$ training horizon.
    Remarkably, in the same way as \texttt{Prodigy} followed \texttt{AdamW} in \cref{fig:benchmarking-124m-losses}~\textbf{(b)}, we observe a similar situation for the MoE model. 
    }
    \label{fig:benchmarking-moe-losses}
\end{figure*}

\prettybox{
\takeaway{tkw:moe_bench}Benchmarking results obtained for dense models transfer to corresponding MoEs.
}

\section{Discussion}
\label{sec:discussion}

\textbf{A summary of results.}
In this work, we benchmarked many interesting to community optimizers across architectural changes, model scales, training durations and batch sizes.
After an extensive hyperparameter tuning, we revealed ablations for each optimizers showing their sensitivity to certain of them.
We questioned flaws in popular code base for LLM pretraining---so important for careful benchmarking and the overall model performance.
Regarding the benchmarking setup, we built a rankings of optimizers in each setup considered, if consider the global result and the question of the effectiveness of \texttt{AdamW} for LLMs, we point that there are new reliable optimizers that would be beneficial at scale---\texttt{AdEMAMix}, \texttt{D-Muon}, \texttt{MARS}.
We point that methods such as \texttt{ADOPT} and \texttt{Prodigy} scale similarly to \texttt{AdamW}, and also worth a try for production purposes.

\textbf{Our advices on tuning each method.} 
Overall, we validate both widely used hyperparameters such as $\lambda=0.1$ and $T_\text{warmup}\approx 2\mathbf{k}$, and explore the sensitivity of optimizers to $\gamma$-schedulers, $\gamma$-decay, and optimizer-related hyperparameters.
Notably, a large weight decay ensures faster convergence when training for fewer iterations, and large warmup of $25\%$ of the total training duration $T$ is beneficial for sign-based methods, \texttt{Sophia}, and \texttt{SF-AdamW}.
For \texttt{Lion}---as mentioned in \cite{chen2023symbolicdiscoveryoptimizationalgorithms}---we find that the best value for $\beta_1$ is consistently $0.99$. 
The mechanism for \texttt{Lion} appears similar to \texttt{AdEMAMix}, suggesting that \texttt{Lion} could perform better with larger $\beta_1$, which would require schedulers. 
We also pose an interesting observation toward \texttt{Prodigy}: while it may not be so efficient with very small batch sizes, with scaling of the model size and the batch size, it becomes almost as competitive as \texttt{AdamW}.
\texttt{MARS} also benefits from large batches and continues to improve performance as the model size scales.
For \texttt{MARS}, when optimizing $1$D parameters with \texttt{AdamW}, we found that it is better to keep ($\beta_1$, $\beta_2$) of \texttt{AdamW}; for our largest models, $\beta_1=0.8$ performs slightly better than $\beta_1=0.9$.
Additionally, \texttt{MARS} betas determined for $2$D parameters in \cite{yuan2024marsunleashingpowervariance} also seem to be the best in our settings.
Basic \texttt{Muon} performs poorly at relatively small batch sizes ($32$, $256$) across different model sizes and training lengths; however, applying weight decay to $2$-dimensional parameters, as in \texttt{D-Muon},
resolves this and yields a robust optimizer across all benchmarking scenarios we considered.
\texttt{AdEMAMix} remains the best optimizer overall, scaling efficiently with bath size, model size, and training horizons.
Importantly, increasing $\beta_2$ for longer training substantially benefits \texttt{AdEMAMix} and other \texttt{AdamW}-like methods.
Moreover, \texttt{AdEMAMix} allows using a large weight decay term $\lambda$ during prolonged training, e.g., runs of $128\mathbf{k}$ iterations with $\lambda=0.5$ still slightly outperform those with $\lambda=0.1$.
Beyond optimizer-specific hyperparameters, we show that the choice of $\gamma$-scheduler also depends on the optimizer selected.
Regarding the learning rate, decaying $\gamma$ below $0.1\times\gamma_{\max}$ is important, as it significantly improves the optimization performance.

\textbf{Limitations.}
We conduct our benchmarking experiments on models of up to $720\mathbf{M}$ parameters, with long training runs of almost $50\mathbf{B}$ tokens.
The insights we find vary across scales, and training behavior may change further at practical scales and with extremely long training durations~\cite{wei2022emergentabilitieslargelanguage,tay2022scaleefficientlyinsightspretraining}.
Especially when certain optimizers are not widely supported by modern sharding frameworks~\cite{zhao2023pytorchfsdpexperiencesscaling,rajbhandari2020zeromemoryoptimizationstraining,deepspeed2020} at the moment.
Throughout this work, we study the loss dynamics, leaving aside downstream performances. 
Although these often scale reliably with loss~~\cite{du2025understandingemergentabilitieslanguage,gadre2024languagemodelsscalereliably}, there are also counterexamples~\cite{xu2025unveilingdownstreamperformancescaling,liu2022pretraininglossbetterdownstream}.
Bridging the gap between loss minimization and downstream task performance is important, as downstream abilities are ultimately the main metric of interest.
We leave a deeper investigation of this connection to future research.
We also do not cover previously explored \texttt{Adan}~\cite{xie2024adanadaptivenesterovmomentum}, \texttt{NAdam(W)}~\cite{dozat2016nadam}, \texttt{Shampoo}~\cite{gupta2018shampoopreconditionedstochastictensor} optimizers, as well as recently introduced \texttt{Scion}~\cite{pethick2025trainingdeeplearningmodels}, novel variations of \texttt{Muon}~\cite{qiu2025reparameterizedllmtrainingorthogonal,an2025asgoadaptivestructuredgradient,ahn2025dioncommunicationefficientoptimizerlarge}, and others~\cite{peng2024demodecoupledmomentumoptimization,defazio2025gradientsrapidlyincreasenear,guan2023adaplusintegratingnesterovmomentum,wang2025gradpowerpoweringgradientsfaster}.
In addition, it is important to come up with a unified benchmark of optimizers for memory-efficient pretraining~\cite{glentis2025minimalistoptimizerdesignllm,zhu2025apollosgdlikememoryadamwlevel,ma2025swansgdnormalizationwhitening,su2025galore2largescalellm}, as they become more popular and argue that they might even outperform the \texttt{AdamW} baseline.
We emphasize that there is still a huge branch of research on optimizers left to explore.

\newpage
\section*{Acknowledgements}
AS thanks Nikita Doikov for
discussions that lead to the idea of this project.
We thank Alexander H\"agele, Alejandro Hern\'andez-Cano, Philip Zmushko, Amirkeivan Mohtashami, and Imanol Schlag for helpful discussions
regarding the paper.
This work was supported by the Swiss State Secretariat for
Education, Research and Innovation (SERI) under contract
number $22.00133$, and by the Swiss National Supercomputing Centre (CSCS) under project ID a$06$ on Alps, as part of the Swiss AI Initiative.

\bibliography{ref}
\bibliographystyle{plain}

\newpage
\appendix
\tableofcontents
\newpage

\section{Optimizers we study}
\label{sec:ap_optimizers}

In this section, we describe all considered algorithms, presenting them in a unified formalism. 
We start with notation and then discuss the algorithms according to their logical grouping:

1. \texttt{Adam}-like methods: \texttt{AdamW} (\cref{alg:adamw}), \texttt{ADOPT} (\cref{alg:adopt}), and \texttt{AdEMAMix} (\cref{alg:ademamix}).

2. Sign-based methods: \texttt{Lion} (\cref{alg:lion}), \texttt{Signum} (\cref{alg:signum,alg:signumtorch}).

3. Approximate second-order optimizers: \texttt{Muon} (\cref{alg:muon}), \texttt{SOAP} (\cref{alg:soap}), and \texttt{Sophia} (\cref{alg:sophia}).

4. Learning rate / scheduler-free learning algorithms: \texttt{Schedule-Free AdamW} (\cref{alg:sfadamw}), \texttt{Prodigy} (\cref{alg:prodigy}).

5. \texttt{MARS} methods: (Algorithms~\ref{alg:marsadamw},~\ref{alg:marslion},~\ref{alg:marsshampoo}).

\textbf{Notation.}
In our work, we denote vectors and matrices in bold, and scalars in regular type.
Let $\mathcal{L}: \mathcal{D} \to \mathbb{R}$ be an empirical loss function parameterized by $\xx$ and mapping a batch of inputs $\xixi \subset \mathcal{D}$ to $\mathbb{R}$.
As $\bg = \nabla_{\xx} \mathcal{L}\left(\xx, \xixi\right)$, we denote a stochastic gradient of the loss w.r.t. parameters $\xx$.
For brevity, we omit $\xx$ in $\nabla$ and write $\nabla \mathcal{L}\left(\xx, \xixi\right)$.
We use the following standardized notation for specific symbols in our work: batch size---$|\xixi|$, learning rate---$\gamma$, weight decay---$\lambda$, momentum---$\beta$, iteration counter $t$ with the total number of iterations---$T$.
And basic notation for symbols in the algorithms: $\mm, \vv$---are first and second moment estimates, respectively, with their bias corrected versions $\hat{\mm}, \hat{\vv}$, and beta parameters---($\beta_1, \beta_2$).
We denote the dot product of two vectors $\zz$, $\yy$ as $\langle \zz, \yy \rangle$, while $\zz \odot \yy$ stands for their element-wise product.
All division and addition operations in the described algorithms are element-wise.

\subsection{\texttt{AdamW}, \texttt{ADOPT}, \texttt{AdEMAMix}}
\label{sec:ap_adamlike}

\textbf{\texttt{AdamW}.}
Our baseline---\texttt{Adam(W)}, has become a de facto optimizer for deep learning, demonstrating impressive performance across diverse domains---from tabular data to diffusion and language models.

The method originated from the ideas of \texttt{Adagrad} \cite{duchi11a} and \texttt{RMSProp} \cite{graves2014generatingsequencesrecurrentneural}, which utilize a second moment estimate $\vv$ in their update rule.
However, \texttt{Adam(W)} enhanced this prior scheme by incorporating momentum \cite{NEMIROVSKII198521,sutskever13}, establishing itself as a state-of-the-art method for a wide range of tasks.
All other algorithms we consider also employ a similar, if not identical, momentum scheme.

A key difference between \texttt{Adam} and \texttt{AdamW} is the use of decoupled weight decay~\cite{loshchilov2019decoupledweightdecayregularization} in the latter.
We adopt the decoupled weight decay scheme for all methods to ensure consistency, as correct weight decay is critical for optimizer comparison, hyperparameter tuning, and final performance.
The importance of the correct weight decay implementation is clearly observable, e.g., for \texttt{Signum}.

\begin{algorithm}[ht]
\caption{\texttt{AdamW}}
\label{alg:adamw}
\begin{algorithmic}[1]
\State {\bfseries Input:} Initial parameters $\xx_0$, number of iterations $T$, learning rate $\gamma_t$, weight decay $\lambda$, $\beta_1$, $\beta_2$, $\varepsilon$.
\State {\bfseries Initialize:} $\mm_0 \leftarrow \mathbf{0}$, $\vv_0 \leftarrow \mathbf{0}$
\For{$t \in [T]$}
\State $\bg_t \leftarrow \nabla \mathcal{L}(\xx_t, \xixi_t)$
\State $\mm_t \leftarrow \beta_1 \mm_{t-1} + (1 - \beta_1) \bg_t$
\State $\vv_t \leftarrow \beta_2 \vv_{t-1} + (1 - \beta_2) \bg_t \odot \bg_t$
\State $\hat{\mm}_t \leftarrow \mm_t / (1 - \beta_1^t)$, \; $\hat{\vv}_t \leftarrow \vv_t / (1 - \beta_2^t)$
\State $\xx_{t+1} \leftarrow \xx_t - \gamma_t \left(\frac{\hat{\mm}_t}{\sqrt{\hat{\vv}_t} + \varepsilon} + \lambda \xx_t \right)$
\EndFor
\State {\bfseries Return:} $\xx_T$
\end{algorithmic}
\end{algorithm}

\textbf{\texttt{ADOPT}.}
Recently, Taniguchi et al.~\cite{taniguchi2024adoptmodifiedadamconverge} proposed a modification of \texttt{Adam}, by removing the current gradient $\bg_t$ from the second moment estimate $\vv_t$ and altering the order of the momentum update $\mm_t$ and normalization.
As shown in \texttt{line 8} of \cref{alg:adopt}, the parameter update depends only on the previous value of the second moment estimate $\vv_{t-1}$. 
The authors analyze the convergence of \texttt{ADOPT} with the following update rule:
\begin{align*}
    \mm_t  & \leftarrow \beta_1 \mm_{t-1} + (1 - \beta_1) \frac{\bg_t}{\max\{\sqrt{\vv_{t-1}}, \varepsilon\}}, \\
    \xx_{t+1}  & \leftarrow \xx_t - \gamma_t \mm_t.
\end{align*}
However, the practical implementation differs in a few details.
To tackle instabilities caused by near-zero gradients during the early stages of training, the authors propose using a clipping on $\bg_t / \max\{\sqrt{\vv_{t-1}}, \varepsilon\}$, which we formalize as the $\mathtt{clamp}$ operation.
Given a vector $\bg$ and a positive scalar $c$, it is defined as:
\begin{align}
\label{eq:clamp}
    \mathtt{clamp}\left(\bg, c\right)^{(\rom{1})} = \min\left\{\max\left\{g^{(\rom{1})}, -c\right\}, c\right\}.
\end{align}
Thus, the element-wise $\mathtt{clamp}$ operation preserves $\bg_t$ from the division by near-zero values.

The authors theoretically claim that \texttt{ADOPT} achieves the optimal convergence bound for smooth non-convex objectives, regardless of the choice of the $\beta_2$ parameter.
We empirically investigate this claim and observe that, contrary to the theoretical results, there is a significant performance gap for different choices of $\beta_2$ in practice---see \cref{fig:adopt-beta2}.
Also, the effect of $\varepsilon$ in \cref{alg:adopt} is intriguing: in contrast to the typical $\varepsilon=10^{-8}$ value for \texttt{AdamW}, the authors pose that for \texttt{ADOPT} mechanism the smaller value of $10^{-6}$ is more suitable.
We notice that this also holds in practice for the method, and we provide the corresponding ablation in \cref{fig:adopt-eps}.

\begin{algorithm}[ht]
\caption{\texttt{ADOPT}}
\label{alg:adopt}
\begin{algorithmic}[1]
\State {\bfseries Input:} Initial parameters $\xx_0$, number of iterations $T$, learning rate $\gamma_t$, weight decay $\lambda$, $\beta_1$, $\beta_2$, $\varepsilon$.
\State {\bfseries Initialize:} $\mm_0 \leftarrow \mathbf{0}$, $\vv_0 \leftarrow \nabla \mathcal{L}(\xx_0, \xixi_0) \odot \nabla \mathcal{L}(\xx_0, \xixi_0)$
\For{$t \in [T]$}
\State $\bg_t \leftarrow \nabla \mathcal{L}(\xx_t, \xixi_t)$
\State $c_t \leftarrow t^{1/4}$ \Comment{Update clipping value schedule}
\State $\mm_t \leftarrow \beta_1 \mm_{t-1} + (1 - \beta_1) \mathtt{clamp}\left(\frac{\bg_t}{\max\{\sqrt{\vv_{t-1}}, \varepsilon\}}, c_t\right)$
\State $\vv_t \leftarrow \beta_2 \vv_{t-1} + (1 - \beta_2) \bg_t \odot \bg_t$
\State $\xx_{t+1} \leftarrow \xx_t - \gamma_t \left(\mm_t + \lambda \xx_t\right)$ \Comment{Update without $\vv_t$}
\EndFor
\State {\bfseries Return:} $\xx_T$
\end{algorithmic}
\end{algorithm}

\textbf{\texttt{AdEMAMix}.}
Another \texttt{Adam}-like optimizer we study is \texttt{AdEMAMix} \cite{pagliardini2024ademamixoptimizerbetterfaster}.
This work argues that using a single EMA to accumulate past gradients in the first moment estimate $\mm$ can be suboptimal, as it cannot simultaneously prioritize both immediate past and older gradients.
In \cref{alg:ademamix}, the authors incorporate two EMAs: one---\texttt{Adam}-like EMA for $\mm$ (fast), and another---a slow EMA $\mm^{\mathrm{slow}}$ (see \texttt{line 7}) with an additional $\beta_3$ parameter.
In the update rule, fast and slow EMAs are balanced with the constant factor $\alpha$ (see \texttt{line 10} of \cref{alg:ademamix}).
This algorithmic design enables \texttt{AdEMAMix} to benefit from older gradients, resulting in smoother loss curves during training.

To mitigate the effect of early instabilities, the authors use two additional schedulers for $\alpha$ and $\beta_3$ -- $\mathtt{alpha\_scheduler}$ and $\mathtt{beta\_scheduler}$, formalized in our work as follows:
\begin{align*}
    &\mathtt{alpha\_scheduler}(t, \alpha, T_\alpha) = \min\left\{\frac{t\alpha}{T_\alpha}, \alpha \right\}, \\
    &\mathtt{beta\_scheduler}(t, \beta_3, \beta_{\mathrm{start}}, T_{\beta_3}) = \min\left\{\exp\left(\frac{\log(\beta_{\mathrm{start}})\log(\beta_3)}{\left(1-\frac{t}{T_{\beta_3}}\right)\log(\beta_3)+ \frac{t}{T_{\beta_3}} \log(\beta_{\mathrm{start}})}\right), \beta_3\right\}.
\end{align*}

In all experiments, we set $\beta_\mathrm{start} = \beta_1$, and the warmup parameters equal to the length of training: $T_\alpha = T_{\beta_3} = T$.

Although these schedulers may seem at odds with the WSD scheduler~\cite{hu2024minicpmunveilingpotentialsmall}, setting $T_\alpha, T_{\beta_3}$ longer than the first WSD checkpoint does not noticeably harm performance. 
Thus, \texttt{AdEMAMix} can still be combined with recent findings regarding the WSD scheduler.

\begin{algorithm}[ht]
\caption{\texttt{AdEMAMix}}
\label{alg:ademamix}
\begin{algorithmic}[1]
\State {\bfseries Input:} Initial parameters $\xx_0$, number of iterations $T$, learning rate $\gamma_t$, weight decay $\lambda$, $\beta_1$, $\beta_2$, $\beta_3$, $\beta_{\mathrm{start}}$, $\alpha$, $\mathtt{beta\_scheduler}$, $\mathtt{alpha\_scheduler}$, warmup parameters $T_{\beta_3}$ and $T_\alpha$, $\varepsilon$.
\State {\bfseries Initialize:} $\mm_0 \leftarrow \mathbf{0}$, $\mm^{\mathrm{slow}}_0 \leftarrow \mathbf{0}$, $\vv_0 \leftarrow \mathbf{0}$
\For{$t \in [T]$}
\State $\beta_3(t) \leftarrow \mathtt{beta\_scheduler}(t, \beta_3, \beta_{\mathrm{start}}, T_{\beta_3})$, $\alpha(t) \leftarrow \mathtt{alpha\_scheduler}(t, \alpha, T_\alpha)$ \Comment{Update $\beta_3$ and $\alpha$ schedulers}
\State $\bg_t \leftarrow \nabla \mathcal{L}(\xx_t, \xixi_t)$
\State $\mm_t \leftarrow \beta_1 \mm_{t-1} + (1 - \beta_1) \bg_t$
\State $\mm^{\mathrm{slow}}_{t} \leftarrow \beta_3(t) \mm^{\mathrm{slow}}_{t-1} + (1 - \beta_3(t)) \bg_t$ \Comment{Update slow EMA}
\State $\vv_t \leftarrow \beta_2 \vv_{t-1} + (1 - \beta_2) \bg_t \odot \bg_t$
\State $\hat{\mm}_t \leftarrow \mm_t / (1 - \beta_1^t)$, \; $\hat{\vv}_t \leftarrow \vv_t / (1 - \beta_2^t)$
\State $\xx_{t+1} \leftarrow \xx_t - \gamma_t \left(\frac{\hat{\mm}_t + \alpha(t) \mm^{\mathrm{slow}}_t}{\sqrt{\hat{\vv}_t} + \varepsilon} + \lambda \xx_t \right)$
\EndFor
\State {\bfseries Return:} $\xx_T$
\end{algorithmic}
\end{algorithm}

\subsection{Sign-based methods: \texttt{Lion} and \texttt{Signum}}
\label{sec:ap_signbased}

Another branch of methods includes sign-based methods, represented by \texttt{Lion} and \texttt{Signum}.
To some extent, one can classify \texttt{Adam} as a sign-based optimizer also, but we mention only \texttt{Lion} and \texttt{Signum} as they explicitly incorporate the $\mathtt{sign}$ operation in the update rule.

These methods, particularly \texttt{Signum}, have been unfairly overlooked in the context of LLM pretraining.
However, our results demonstrate that, with sufficiently large batch sizes and at moderate model scales, these optimizers perform on par with \texttt{Adam}, and in some cases even outperform it.

\textbf{\texttt{Lion}.}
The first sign-based method we study is \texttt{Lion} \cite{chen2023symbolicdiscoveryoptimizationalgorithms}.
This optimizer is symbolically discovered in the program space of first-order optimization
primitives.
\texttt{Lion} updates its EMA of $\mm$ after updating the parameters and has additional term $(1 - \beta_1)\bg$ which adds to the momentum.
This interpolation $\beta_1 \mm_{t-1} + (1 - \beta_1)\bg_t$ (see \texttt{line 6} of \cref{alg:lion}) makes the symbolic-discovered idea behind \texttt{Lion} similar to the idea of the \texttt{AdEMAMix} optimizer.

\begin{algorithm}[ht]
\caption{\texttt{Lion}}
\label{alg:lion}
\begin{algorithmic}[1]
\State {\bfseries Input:} Initial parameters $\xx_0$, number of iterations $T$, learning rate $\gamma_t$, weight decay $\lambda$, $\beta_1$, $\beta_2$.
\State {\bfseries Initialize:} $\mm_0 \leftarrow \mathbf{0}$
\For{$t \in [T]$}
\State $\bg_t \leftarrow \nabla \mathcal{L}(\xx_t, \xixi_t)$
\State $\mm_t \leftarrow \beta_2 \mm_{t-1} + (1 - \beta_2) \bg_t$ \Comment{Update EMA of $\bg_t$}
\State $\xx_{t+1} \leftarrow \xx_t - \gamma_t \left(\mathtt{sign}\left(\beta_1\mm_{t-1} + (1 - \beta_1)\bg_t\right) + \lambda \xx_t \right)$
\EndFor
\State {\bfseries Return:} $\xx_T$
\end{algorithmic}
\end{algorithm}

\textbf{\texttt{Signum}.}
Another sign-based method, which is the adoptation of \texttt{signSGD}---\texttt{Signum} \cite{bernstein2018signsgd} (or, alternatively, \texttt{signSGD} with momentum).
This method differs from \texttt{Lion} in the interpolation term between the EMA of momentum and the current gradient, as well as in the \texttt{Signum}'s update rule, where a current EMA is used.

Importantly, while \texttt{Signum} is not yet as widespread for LLM pretraining and has largely remained a theoretical artifact, recent studies have begun to adopt \texttt{Signum} for scalable training~\cite{zhao2024deconstructingmakesgoodoptimizer}, primarily due to its memory efficiency compared to \texttt{AdamW}.

In this regard, we would like to highlight that many recent PyTorch \cite{paszke2019pytorchimperativestylehighperformance} implementations of the \texttt{Signum} optimizer are unlikely to be suitable for this method, which impairs its potential performance.

The main issue with open-source implementations is the use of decoupled weight decay in the PyTorch implementation of \texttt{SGDM} (\texttt{SGD} with momentum) \cite{sutskever13}.
Indeed, with decoupled weight decay, the update in \cref{alg:signum} transforms into:
\begin{align*}
    \xx_{t+1} \leftarrow \xx_t - \gamma_t \mathtt{sign}\left(\beta \mm_{t-1} + (1 - \beta)\bg_t - \lambda (1-\beta)\bg_t\right),
\end{align*}
which affects the sign of the update, potentially leading to the wrong optimization direction if the weight decay is sufficiently large.
See Figures~\ref{fig:wdablation_main}~\textbf{(a)}~and~\ref{fig:ap_wdablation} for the impact of the correct weight decay implementation for sign-based methods like \texttt{Signum} and \texttt{Lion}.

Another popular failure while using \texttt{Signum} is the incorrectly tractable PyTorch implementation of \texttt{SGD} with momentum.
It does not include such EMA as \texttt{line 5} in \cref{alg:signum}, on the other hand, in PyTorch, the momentum update depends on the dampening parameter $\tau$:
\begin{align*}
    \mm_t \leftarrow \beta \mm_{t-1} + (1 - \tau) \bg_t,
\end{align*}
where $\tau$ is zero by default.
Therefore, the typical update rule, reflecting the actual \texttt{Signum} behavior in practice, corresponds to the following update:
\begin{align*}
    \xx_{t+1} \leftarrow \xx_t - \gamma_t \left(\mathtt{sign}\left(\beta \mm_{t-1} + (1-\tau)\bg_t\right) + \lambda \xx_t\right),
\end{align*}
where the weight decay is decoupled and, consequently, does not affect $\mathtt{sign}$.

However, we found out that the PyTorch implementation of Nesterov momentum \cite{Nesterov1983AMF}
\begin{align*}
    \bg_t \leftarrow \bg_t + \beta \mm_t,
\end{align*}
improves \texttt{Signum}.
Since enabling Nesterov momentum requires zero dampening $\tau$, we revisited the description of \cref{alg:signum} and propose more practical, PyTorch-compatible version of \texttt{Signum} in \cref{alg:signumtorch}.
We study the role of dampening and Nesterov momentum in our variant of \texttt{Signum} in~\cref{fig:signum-configurations}.

\begin{figure}[ht]
\centering
\begin{minipage}[t]{0.47\textwidth}
\begin{algorithm}[H]
\caption{\texttt{Signum} (basic)}
\label{alg:signum}
\begin{algorithmic}[1]
\State {\bfseries Input:} Initial parameters $\xx_0$, number of iterations $T$, learning rate $\gamma_t$, weight decay $\lambda$, momentum $\beta$.
\State {\bfseries Initialize:} $\mm_0 \leftarrow \mathbf{0}$
\For{$t \in [T]$}
\State $\bg_t \leftarrow \nabla \mathcal{L}(\xx_t, \xixi_t)$
\State $\mm_t \leftarrow \beta \mm_{t-1} + (1 - \beta)\bg_t$
\State $\xx_{t+1} \leftarrow \xx_t - \gamma_t \left(\mathtt{sign}\left(\mm_t\right) + \lambda \xx_t\right)$
\EndFor
\State {\bfseries Return:} $\xx_T$
\end{algorithmic}
\end{algorithm}
\end{minipage}
\hfill
\begin{minipage}[t]{0.52\textwidth}
\begin{algorithm}[H]
\caption{\texttt{Signum} (our PyTorch variant)}
\label{alg:signumtorch}
\begin{algorithmic}[1]
\State {\bfseries Input:} Initial parameters $\xx_0$, number of iterations $T$, learning rate $\gamma_t$, weight decay $\lambda$, momentum $\beta$.
\State {\bfseries Initialize:} $\mm_0 \leftarrow \mathbf{0}$
\For{$t \in [T]$}
\State $\bg_t \leftarrow \mathcal{L}(\xx_t, \xixi_t)$
\State $\mm_t \leftarrow \beta \mm_{t-1} + \bg_t$
\State $\xx_{t+1} \leftarrow \xx_{t} - \gamma_t \left(\mathtt{sign}\left(\beta \mm_t + \bg_t\right) + \lambda \xx_t \right)$
\EndFor
\State {\bfseries Return:} $\xx_T$
\end{algorithmic}
\end{algorithm}
\end{minipage}
\end{figure}

Moreover, to prevent other researchers and practitioners from the possible wrong use of \texttt{Signum}, and to ensure reproducibility, we provide our Python code.
\begin{lstlisting}[language=Python,breaklines=true,showstringspaces=false,caption={\textbf{\texttt{Signum} code skeleton using PyTorch.}}]
from typing import Dict

import torch


class Signum(torch.optim.Optimizer):
    def __init__(
        self,
        params,
        lr=1e-3,
        momentum=0,
        dampening=0,
        weight_decay=0,
        nesterov=False,
        sign_update=True,
    ):
        if lr < 0.0:
            raise ValueError(f"Invalid learning rate: {lr}")
        if momentum < 0.0:
            raise ValueError(f"Invalid momentum value: {momentum}")
        if weight_decay < 0.0:
            raise ValueError(f"Invalid weight_decay value: {weight_decay}")

        defaults = dict(
            lr=lr,
            momentum=momentum,
            dampening=dampening,
            weight_decay=weight_decay,
            nesterov=nesterov,
            sign_update=sign_update,
        )
        if nesterov and (momentum <= 0 or dampening != 0):
            raise ValueError("Nesterov momentum requires a momentum and zero dampening")
        super().__init__(params, defaults)

    def __setstate__(self, state):
        super().__setstate__(state)
        for group in self.param_groups:
            group.setdefault("nesterov", False)

    @torch.no_grad()
    def _init_state(self, example, state=None):
        assert isinstance(example, torch.Tensor)
        assert isinstance(state, Dict) or state is None
        if state is None:
            state = {}
        state["step"] = 0
        state["momentum_buffer"] = torch.clone(example).detach()
        return state

    @torch.no_grad()
    def _compute_update(
        self, grad, state, lr, momentum, nesterov, dampening, sign_update, **kwargs
    ):
        if momentum != 0:  # Signum check
            buf = state["momentum_buffer"]
            buf.mul_(momentum).add_(grad, alpha=1 - dampening)

            if nesterov:
                grad = grad.add(buf, alpha=momentum)
            else:
                grad = buf

        if sign_update:
            grad = grad.sign()

        return grad * (-lr)

    @torch.no_grad()
    def step(self, closure=None):
        """Performs a single optimization step.

        Args:
            closure (Callable, optional): A closure that reevaluates the model
                and returns the loss.
        """
        loss = None
        if closure is not None:
            with torch.enable_grad():
                loss = closure()

        for group in self.param_groups:
            for p in group["params"]:
                if p.grad is None:
                    continue

                grad = p.grad
                state = self.state[p]

                if group["weight_decay"] != 0:
                    p.mul_(1 - group["lr"] * group["weight_decay"])

                if len(state) == 0:
                    self._init_state(example=p, state=state)
                    if not group["momentum"]:
                        state.pop("momentum_buffer", None)

                state["step"] += 1

                update = self._compute_update(
                    grad,
                    state,
                    group["lr"],
                    group["momentum"],
                    group["nesterov"],
                    group["dampening"],
                    group["sign_update"],
                )

                p.add_(update)

        return loss
\end{lstlisting}

\subsection{\texttt{Muon} \& \texttt{D-Muon}, \texttt{SOAP}, \texttt{Sophia}}
\label{sec:ap_muonsoapsophia}

The next page of the methods covers algorithms that rather aim to use more information about the problem's curvature (\texttt{SOAP} \cite{vyas2024soapimprovingstabilizingshampoo}, \texttt{Sophia} \cite{liu2024sophiascalablestochasticsecondorder}) or perform fast updates of matrix parameters involving higher order computations (\texttt{Muon} \cite{jordan2024muon}). 

Contrary to chronological order, we discuss them starting from the recent one---\texttt{Muon} and end up with \texttt{Sophia}.

\textbf{\texttt{Muon} \& \texttt{D-Muon}.}
Specifically designed for speedrun comparisons, \texttt{Muon} surpasses the \texttt{AdamW} baseline on the nanoGPT pretraining benchmark \cite{modded_nanogpt_2024}.
Claims from the \texttt{Muon} project extend to faster learning, lower memory usage and better sample-efficiency, with a small overhead in wall-clock time.

The reason why \texttt{Muon} is a good option for speedrun pretraining lies in its structure---\texttt{Muon} treats different parameters based on their tensor dimensionality.
One-dimensional ($1$D) parameters, large embedding layers, Layer Norm (or RMSNorm) parameters, and the output layer of LLM ($\mathtt{lm\_head}$) are optimized by \texttt{AdamW}.
And all parameters with two or more dimensions (e.g., Multi-Head Attention layers) are optimized by \cref{alg:muonnon1d}, which we call \texttt{MuonNon1D}.

Inspired by \texttt{Shampoo}'s preconditioners \cite{gupta2018shampoopreconditionedstochastictensor}, the authors of \texttt{MuonNon1D} incorporated an orthogonalization step to compute $\mathtt{SVD}$ transformation of the gradient matrix.
Before the orthogonalization step, \texttt{MuonNon1D} resembles \texttt{SGD} with Nesterov momentum.
To ensure a fast orthogonalization procedure, the authors, inspired by \cite{bernstein2024oldoptimizernewnorm}, use the Newton-Schulz procedure \cite{functionsmatrices}. 
As the number of Newton-Schulz iterations increases, the resulting matrix becomes closer to $\boldsymbol{U}\boldsymbol{V}^\top$ from $\mathtt{SVD}$ transformation.
The authors also mention that \texttt{Muon} can be considered an alternative method of smoothing spectral steepest descent \cite{Carlson2015PreconditionedSD}, offering a distinct set of memory and runtime trade-offs compared to \texttt{Shampoo}.

\begin{algorithm}[ht]
\caption{\texttt{MuonNon1D} (for non-$1$D parameters)}
\label{alg:muonnon1d}
\begin{algorithmic}[1]
\State {\bfseries Input:} Initial non-$1$D parameters $\xx_0$, number of iterations $T$, learning rate $\gamma_t$, momentum $\beta$, number of Newton-Schulz iterations $T_\mathrm{NS}$, $a$, $b$, $c$ coefficients.
\State {\bfseries Initialize:} $\mm_0 \leftarrow \mathbf{0}$
\For{$t \in [T]$}
\State $\bg_t \leftarrow \nabla \mathcal{L}(\xx_t, \xixi_t)$
\State $\mm_t \leftarrow \beta \mm_{t-1} + \bg_t$
\State $\bg_t \leftarrow \beta \mm_t + \bg_t$ \Comment{Practical implementation of Nesterov momentum}
\State {\bfseries Set:} $\ww_0 \leftarrow \bg_t / \|\bg_t\|_F$
\For{$n \in [T_{\mathrm{NS}}]$}
\State $\ww_{n+1} \leftarrow a \ww_n + b \ww_n\ww_n^\top + c \left(\ww_n\ww_n^\top\right)^2\ww_n$ \Comment{Newton-Schulz iteration}
\EndFor
\State $\xx_{t+1} \leftarrow \xx_t - \gamma_t \ww_{T_\mathrm{NS}}$
\EndFor
\State {\bfseries Return:} $\xx_T$
\end{algorithmic}
\end{algorithm}

\begin{algorithm}[ht]
    \caption{\texttt{Muon} (general scheme)}
    \label{alg:muon}
    \begin{algorithmic}[1]
        \State {\bfseries Input:} Initial parameters $\xx_0$, number of iterations $T$. \texttt{Muon}'s parameters: learning rate $\gamma_t^{\mathtt{M}}$, momentum $\beta$, number of Newton-Schulz iterations $T_\mathrm{NS}$, $a$, $b$, $c$ coefficients.
        \texttt{AdamW}'s parameters: learning rate $\gamma_t^{\mathtt{A}}$, weight decay $\lambda$, $\beta_1$, $\beta_2$, $\varepsilon$.
        \For{$t \in [T]$}
        \If{$\xx_t \in \{\mathtt{embeds}, \mathtt{scalar\_params}, \mathtt{lm\_head}\}$}
        \State $\xx_t^{\mathtt{A}} \leftarrow \xx_t$
        \State $\xx_{t+1}^{\mathtt{A}} \leftarrow \texttt{AdamW}\left(\xx_t^{\mathtt{A}}, \gamma_t^{\mathtt{A}}, \lambda, \beta_1, \beta_2, \varepsilon, T=1\right)$ \Comment{One iteration of \texttt{AdamW}}
        \Else
        \State $\xx_t^{\mathtt{M}} \leftarrow \xx_t$
        \State $\xx_{t+1}^{\mathtt{M}} \leftarrow \texttt{MuonNon1D}\left(\xx_t^{\mathtt{M}}, \gamma_t^{\mathtt{M}}, T_\mathrm{NS}, \beta, a, b, c, T=1\right)$ \Comment{One iteration of \texttt{MuonNon1D}}
        \EndIf
        \EndFor
        \State {\bfseries Return:} $\xx_T^{\mathtt{A}}, \xx_T^{\mathtt{M}}$
    \end{algorithmic}
\end{algorithm}

Importantly, we noticed that the original algorithmic description of \texttt{Muon} optimizer, provided in the official repository\footnote{\href{https://github.com/KellerJordan/modded-nanogpt}{https://github.com/KellerJordan/modded-nanogpt}}, differs from the actual one, presented in \cref{alg:muonnon1d}.
In the original code, as well as in our benchmarking, weight decay does not apply to the matrix parameters in the optimizer state of \texttt{MuonNon1D}, meaning that the only weight decay used during training is \texttt{AdamW}'s weight decay.
From this perspective, we observe that the gap between the final loss values for runs with weight decay of $0.1$ and $0$ almost disappears, while the run with a weight decay of $0.5$ becomes the worst, which is not the case for other optimizers.
See \cref{fig:wdablation_main,fig:ap_wdablation} regarding these ablations.

Noticeably, the weight decay issue was addressed in the recent paper~\cite{liu2025muonscalablellmtraining}, in which the authors also present a scheme for sharing the learning rate and weight decay between the matrix and non-matrix parameters of the model.
They do this via the RMS heuristic: since \texttt{AdamW} has the property of keeping its RMS updates close to $1$~\cite{hinton2012rmsprop}, particularly around $0.2$-$0.4$ in the practice of LLM training~\cite{liu2025muonscalablellmtraining,ai2025practicalefficiencymuonpretraining}, they scale the RMS update of \texttt{Muon} to this range.
With these adjustments, practitioners do not need to tune the learning rate and weight decay for $1$D and non-$1$D parameters separately, which is a significant bonus of the newer \texttt{Muon}-like algorithm.
We include this variation of \texttt{Muon} under the \texttt{D-Muon} naming.

Our ablations demonstrate that \texttt{D-Muon} scales better than the basic \texttt{Muon} in all settings we have considered so far (see~\cref{fig:benchmark-720,fig:benchmarking-720m-losses,fig:benchmark-210,fig:benchmarking-210m-losses,fig:benchmark-124,fig:benchmarking-124m-losses}).
We also report a detailed comparison of these two similar methods in \cref{fig:muon-dmuon-final-val-loss,fig:muon-dmuon-val-loss}, and discuss their close connection with the weight decay applied to non-$1$D parameters in the \texttt{D-Muon} algorithm.
Refer to this ablation in \S~\ref{sec:results}.

Another interesting aspect of \texttt{Muon} is the effect of the Newton-Schulz orthogonalization procedure~\cite{bernstein2024oldoptimizernewnorm,functionsmatrices} on optimization.
We show how the number of Newton-Schulz steps impacts the performance of \texttt{Muon} in~\cref{fig:muon_ns}.
Furthermore, we pose that improving the orthogonalization procedure in methods like \texttt{Muon}, \texttt{Scion}, \texttt{MARS-Shampoo} (see~\cref{alg:marsshampoo}) could substantially improve their overall performance.
Recent work has already begun to explore this avenue~\cite{amsel2025polarexpressoptimalmatrix,grishina2025acceleratingnewtonschulziterationorthogonalization}, but a deeper investigation remains an open research challenge.

\textbf{\texttt{SOAP}.}
Vyas et al.~\cite{vyas2024soapimprovingstabilizingshampoo} proposed a new, improved modification of \texttt{Shampoo} \cite{gupta2018shampoopreconditionedstochastictensor}.
\texttt{SOAP} reduces the computational overhead by optimizing only two-dimensional layers ($2$D) via \cref{alg:soapnon1d}, while running \texttt{AdamW} for $1$D layers.
At initialization, the preconditioners are computed via the eigenvector decomposition of the initial gradient matrices  $\mathtt{eigenbasis}\left(\nabla\mathcal{L}(\xx_0, \xixi_0)\nabla\mathcal{L}(\xx_0, \xixi_0)^\top\right)$: $\nabla\mathcal{L}(\xx_0, \xixi_0)\nabla\mathcal{L}(\xx_0, \xixi_0)^\top = \qq \Lambda \qq^{-1}$, where $\Lambda$ stands for the diagonal matrix whose diagonal elements are the corresponding eigenvalues.
For subsequent iterations, \texttt{SOAPNon1D} rotates gradients into this slowly changing basis, maintains second-moment statistics in that basis, and periodically updates the basis via $\mathtt{QR}$ decomposition (see \texttt{lines 15, 16} of \cref{alg:soapnon1d}) for all $2$D layers~(except for $\mathtt{embeds}$ and $\mathtt{lm\_head}$). 
This is the main computational part of the method.

A key idea behind the \texttt{SOAP} optimizer is:

1. Given the slowly changing coordinate basis provided by eigenvectors $\boldsymbol{l}$ and $\rr$, \texttt{SOAP} updates its second moment estimates in this basis; that is to say, it runs \texttt{AdamW} in another, a rotated space.

2. To update the eigenvectors of $\boldsymbol{l}$ and $\rr$, \texttt{SOAP} runs $\mathtt{QR}$ decomposition with the preconditioning frequency $\phi$.

\begin{algorithm}[ht]
\caption{\texttt{SOAPNon1D} (for non-$1$D parameters)}
\label{alg:soapnon1d}
\begin{algorithmic}[1]
\State {\bfseries Input:} Initial parameters $\xx_0$, number of iterations $T$, learning rate $\gamma_t$, weight decay $\lambda$, $\beta_1$, $\beta_2$, preconditioning frequency $\phi$, $\varepsilon$.
\State {\bfseries Initialize:} $\mm_0 \leftarrow \mathbf{0}$, $\vv_0 \leftarrow \mathbf{0}$
\State {\bfseries Initialize preconditioners:} $\qq_l, \qq_r \leftarrow \mathtt{eigenbasis}\left(\nabla\mathcal{L}(\xx_0, \xixi_0)\nabla\mathcal{L}(\xx_0, \xixi_0)^\top\right)$ 
\For{$t \in [T]$}
\State $\bg_t \leftarrow \nabla \mathcal{L}(\xx_t, \xixi_t)$
\State $\bg_t^\prime \leftarrow \qq_l^\top \bg_t\qq_r$ \Comment{Rotate $\bg_t$}
\State $\mm_t \leftarrow \beta_1 \mm_{t-1} + (1-\beta_1)\bg_t$
\State $\mm_t^\prime \leftarrow \qq_l^\top \mm_t \qq_r$ \Comment{Compute \texttt{Adam}'s statistics in rotational space}
\State $\vv_t \leftarrow \beta_2 \vv_{t-1} + (1 - \beta_2) \bg_t^\prime \odot \bg_t^\prime$
\State $\gamma_t \leftarrow \gamma_t \frac{\sqrt{1 - \beta_2^t}}{1 - \beta_1^t}$ \Comment{Optional: use bias correction}
\State $\xx_{t+1} \leftarrow \xx_t - \gamma_t\left(\qq_l\frac{\mm_t^\prime}{\sqrt{\vv_t} + \varepsilon}\qq_r^\top + \lambda \xx_t\right)$ \Comment{Perform update in original space}
\State $\el_t \leftarrow \beta_2 \el_{t-1} + (1 - \beta_2)\bg_t\bg_t^\top$ \Comment{Update preconditioners}
\State $\rr_t \leftarrow \beta_2 \rr_{t-1} + (1 - \beta_2)\bg_t^\top\bg_t$
\If{$t \equiv 1 \pmod{\phi}$}
\State $\qq_l \leftarrow \mathtt{QR}\left(\el_t\qq_l\right)$
\State $\qq_r \leftarrow \mathtt{QR}\left(\rr_t\qq_r\right)$
\EndIf
\EndFor
\State {\bfseries Return:} $\xx_T$
\end{algorithmic}
\end{algorithm}

In \cref{alg:soapnon1d}, setting both $\qq_l$ and $\qq_r$ to the identity matrix would result in \texttt{AdamW}.

The overall \texttt{SOAP} algorithm can be formalized as \cref{alg:soap}.

\begin{algorithm}[ht]
    \caption{\texttt{SOAP} (general scheme)}
    \label{alg:soap}
    \begin{algorithmic}[1]
        \State {\bfseries Input:} Initial parameters $\xx_0$, number of iterations $T$, learning rate $\gamma_t$, weight decay $\lambda$, $\beta_1$, $\beta_2$, preconditioning frequency $\phi$, $\varepsilon$.
        \For{$t \in [T]$}
        \If{$\xx_t \in \{\mathtt{embeds}, \mathtt{scalar\_params}, \mathtt{lm\_head}\}$}
        \State $\xx_t^{\mathtt{A}} \leftarrow \xx_t$
        \State $\xx_{t+1}^{\mathtt{A}} \leftarrow \texttt{AdamW}\left(\xx_t^{\mathtt{A}}, \gamma_t, \lambda, \beta_1, \beta_2, \varepsilon, T=1\right)$ \Comment{One iteration of \texttt{AdamW}}
        \Else
        \State $\xx_t^{\mathtt{S}} \leftarrow \xx_t$
        \State $\xx_{t+1}^{\mathtt{S}} \leftarrow \texttt{SOAPNon1D}\left(\xx_t^{\mathtt{S}}, \gamma_t, \lambda, \beta_1, \beta_2, \varepsilon, T=1\right)$ \Comment{One iteration of \texttt{SOAPNon1D}}
        \EndIf
        \EndFor
        \State {\bfseries Return:} $\xx_T^{\mathtt{A}}, \xx_T^{\mathtt{S}}$
    \end{algorithmic}
\end{algorithm}

\textbf{\texttt{Sophia}.}
Despite being named a second-order optimizer, \texttt{Sophia} \cite{liu2024sophiascalablestochasticsecondorder} performs an update that is quite similar to \texttt{Adam}'s.
It also leverages the diagonal preconditioner $\hh$, but not the curvature information of the optimization problem, which depends on the non-diagonal terms of the Hessian.
One should notice that \texttt{Sophia} was introduced with two types of preconditioner---Hutchinson~\cite{BEKAS20071214} and Gauss-Newton-Bartlett~\cite{martens2020newinsightsperspectivesnatural}.
Since the latter shows more promising performance, we only consider this type of preconditioner for \texttt{Sophia}.

Every $\phi$ iterations, \texttt{Sophia} updates its second moment estimate by computing the gradient $\hat{\bg}$ of the empirical loss $\mathcal{L}$ given $\mathtt{softmax}$ of the logits instead of the true logits.
Multiplying by the batch size, we obtain $\hat{\hh}$, after that, \texttt{Sophia} updates the EMA of $\hat{\hh}$.

Importantly, we found that the algorithmic description of \texttt{Sophia} in the original paper differs in minor details from the code implementation\footnote{\href{https://github.com/Liuhong99/Sophia}{https://github.com/Liuhong99/Sophia}}.
Indeed, the update rule in their work is formulated as follows:
\begin{align*}
    \xx_{t+1} \leftarrow \xx_t - \gamma_t \mathtt{clamp}\left(\frac{\mm_t}{\max\{\rho \hh_t, \varepsilon\}}, 1\right),
\end{align*}
where $\mathtt{clamp}$ is defined as in \cref{eq:clamp}.

On the other hand, the code from the official repository suggests:

\begin{lstlisting}[language=Python,breaklines=true,showstringspaces=false,caption={\textbf{\texttt{Sophia} update skeleton using PyTorch.}}]
# update step
step_t += 1

# Perform stepweight decay
param.mul_(1 - lr * weight_decay)

# Decay the first and second moment running average coefficient
exp_avg.mul_(beta1).add_(grad, alpha=1 - beta1)

else:
    step_size_neg = -lr

    ratio = (exp_avg.abs() / (rho * bs * hess + 1e-15)).clamp(None, 1)
    param.addcmul_(exp_avg.sign(), ratio, value=step_size_neg)
\end{lstlisting}
Therefore, the update rule of \texttt{Sophia} is misstated in the original paper and should be corrected to match \texttt{line 16} of \cref{alg:sophia}.

\prettybox{
\takeaway{tkw:sophia_update}The actual update rule of \texttt{Sophia} does not match its description in the original paper.
}

\begin{algorithm}[ht]
\caption{\texttt{Sophia}}
\label{alg:sophia}
\begin{algorithmic}[1]
\State {\bfseries Input:} Initial parameters $\xx_0$, number of iterations $T$, learning rate $\gamma_t$, weight decay $\lambda$, $\beta_1$, $\beta_2$, estimator frequency $\phi$, scaling factor $\rho$, $\varepsilon$.
\State {\bfseries Initialize:} $\mm_0 \leftarrow \mathbf{0}$, $\hh_0 \leftarrow \mathbf{0}$
\For{$t \in [T]$}
\State $\bg_t \leftarrow \nabla \mathcal{L}(\xx_t, \xixi_t)$
\State $\mm_t \leftarrow \beta_1 \mm_{t-1} + (1 - \beta_1) \bg_t$
\If{$t \equiv 1 \pmod{\phi}$}
\State $p_t \leftarrow \xixi_t$ \Comment{Obtain logits from batch}
\State $p_t \leftarrow \mathtt{softmax}\left(p_t\right)$ \Comment{Sample from logits}
\State $\hat{\mathcal{L}}(\xx_t, \xixi_t) \leftarrow p_t$ \Comment{Loss, where $p_t$ are labels}
\State $\hat{\bg}_t \leftarrow \nabla \hat{\mathcal{L}}(\xx_t, \xixi_t)$
\State $\hat{\hh}_t \leftarrow |\xixi_t| \hat{\bg}_t \odot \hat{\bg}_t$
\State $\hh_t \leftarrow \beta_2 \hh_{t - \phi} + (1 - \beta_2) \hat{\hh}_t$
\Else
\State $\hh_t \leftarrow \hh_{t-1}$
\EndIf
\State $\xx_{t+1} \leftarrow \xx_t - \gamma_t \left(\mathtt{sign}(\mm_t)\min\left\{\frac{|\mm_t|}{\rho \hh_t + \varepsilon}, 1\right\} + \lambda \xx_t\right)$
\EndFor
\State {\bfseries Return:} $\xx_T$
\end{algorithmic}
\end{algorithm}

\subsection{\texttt{Schedule-Free AdamW}, \texttt{Prodigy}}
\label{sec:ap_sfprodigy}

In this section, we outline two more players---\texttt{Schedule-Free AdamW} \cite{defazio2024roadscheduled} and \texttt{Prodigy} \cite{mishchenko2024prodigyexpeditiouslyadaptiveparameterfree}.
Both of them have a promising advantages and require less hyperparameter tuning which paves the road to parameter-free optimizers. 

\textbf{\texttt{Schedule-Free AdamW}.}
Defazio et al.~\cite{defazio2024roadscheduled} introduced the notion of schedule-free optimizers.
The underlying idea behind \texttt{Schedule-Free SGD} and \texttt{Schedule-Free AdamW} (\texttt{SF-AdamW}) is to eliminate learning rate schedulers by replacing them with iterate averaging.
Specifically, the schedule-free method uses interpolation between Polyak-Ruppert averaging~\cite{polyakavg1990,Ruppert1988EfficientEF} and Primal averaging~\cite{primalavgnesterov} for the momentum update, rather than the usual EMA (see \texttt{line 4} of \cref{alg:sfadamw}). 
To stabilize scalable training, the authors also propose an internal warmup mechanism (see \texttt{line 7} of \cref{alg:sfadamw}), which gradually increases the learning rate while ensuring \texttt{Adam}-style bias correction.

An interesting result we observe, is that \texttt{SF-AdamW} shows the best performance with a larger number of warmup iterations compared to other methods---see \cref{fig:warmup}. 

Another key point---training with \texttt{SF-AdamW} is sensitive to the choice of beta parameters.
Unlike in \texttt{AdamW}, these parameters play distinct roles in \texttt{SF-AdamW}: $\beta_1$ determines the interpolation between the $\zz_t$ and $\xx_t$ sequences, which acts as a form of schedule-free momentum.  
Specifically, the term $(1-\beta_1)\bg_t$ is immediately incorporated into the iterate sequence $\yy_t$, while the remainder of $\bg_t$ is gradually incorporated through averaging---a mechanism analogous to the momentum EMA, but with a longer delay for the residual contribution.
By contrast, $\beta_2$ controls the EMA of the second moment estimate with respect to $\yy_t$ (rather than directly with $\xx_t$; see \texttt{line 6} of \cref{alg:sfadamw}).

For \texttt{Adam} it is common to analyze in theory the case, when $\beta_2 = 1 - 1 / T$ \cite{reddi2019convergenceadam,NEURIPS2018_90365351,chezhegov2024gradientclippingimprovesadagrad}, i.e., the choice of the ``optimal'' $\beta_2$ parameter depends on the length of training.
Which, presumably, is also the case for \texttt{SF-AdamW}, making it not fully schedule-free.
Ha\"gele et al.~\cite{hägele2024scalinglawscomputeoptimaltraining} observed this sensitivity to beta parameters, and we go beyond this ablation also~(\cref{fig:sf_betas}).

Importantly, the authors mention that disabling gradient norm clipping is crucial for schedule-free runs; however, we do not observe this in practice and instead find the opposite effect---see \cref{fig:sfclipping}.

\begin{algorithm}[ht]
\caption{\texttt{SF-AdamW}}
\label{alg:sfadamw}
\begin{algorithmic}[1]
\State {\bfseries Input:} Initial parameters $\xx_0$, number of iterations $T$, learning rate $\gamma$, weight decay $\lambda$, $\beta_1$, $\beta_2$, warmup iterations $T_\mathrm{warmup}$, $\varepsilon$.
\State {\bfseries Initialize:} $\zz_0 \leftarrow \xx_0$, $\vv_0 \leftarrow \mathbf{0}$
\For{$t \in [T]$}
\State $\yy_t \leftarrow (1 - \beta_1) \zz_t + \beta_1 \xx_t$
\State $\bg_t \leftarrow \nabla \mathcal{L}(\yy_t, \xixi_t)$
\State $\vv_t \leftarrow \beta_2 \vv_{t-1} + (1 - \beta_2) \bg_t \odot \bg_t$
\State $\gamma_t \leftarrow \gamma \sqrt{1 - \beta_2 ^ t}\min\{1, t/T_\mathrm{warmup}\}$
\State $\zz_{t+1} \leftarrow \zz_t - \gamma_t \left(\bg_t / (\sqrt{\vv_t} + \varepsilon) + \lambda \yy_t\right)$
\State $c_{t+1} \leftarrow \frac{\gamma_t^2}{\sum_{i=0}^t}\gamma_i^2$
\State $\xx_{t+1} \leftarrow (1 - c_{t+1})\xx_t + c_{t+1}\zz_{t+1}$
\EndFor
\State {\bfseries Return:} $\xx_T$
\end{algorithmic}
\end{algorithm}

\textbf{\texttt{Prodigy}.}
Mishchenko et al.~\cite{mishchenko2024prodigyexpeditiouslyadaptiveparameterfree} extended the \texttt{D-Adaptation} framework.
Drawing inspiration from the \texttt{AgaGrad}~\cite{duchi11a} theory, the authors derived an alike step-size rule, giving rise to a new family of methods.
While studying the convergence (in the deterministic case) of several proposed algorithms that are based on the gradient descent and dual averaging, the authors also introduced an \texttt{Adam}-like version of their methods---the \texttt{Prodigy} optimizer~(\cref{alg:prodigy})---that effectively removes the need for hand-tuned learning rates through an intrinsic, adaptive step-size scheme.
The EMA of \texttt{Prodigy} specifically includes $d_t\bg_t$ sequence rather than the raw gradients $\bg_t$~(see \texttt{lines 5, 6, 8, 9} of \cref{alg:prodigy}).
The new term $d_t$ is determined by two additional EMA sequences, which are also responsible for the adaptive rescaling of the learning rate according to \texttt{line 10}.
Mishchenko et al.~\cite{mishchenko2024prodigyexpeditiouslyadaptiveparameterfree} evaluate \texttt{Prodigy} in practice on language models by running a shallow nanoGPT transformer on the Shakespeare (over-training regime) and BookWiki datasets.
We extend the experiments with \texttt{Prodigy} to a larger scale and a greater variety of LLM pretraining settings.

Crucially, \texttt{Prodigy} does not require extensive learning rate tuning. Typically, we initialize $\gamma = 1$, as suggested by the authors, and it remains remarkably stable, as demonstrated in our $\gamma$-sweeps~(\cref{fig:ap_lrsensitivity,fig:lrsensitivity}).
However, \texttt{Prodigy} is still be compatible with learning rate schedules, which we verify experimentally~(\cref{fig:wsdvscosine,fig:wsdcosine}).
We further show that, without any schedulers, $d_t$ sequence behaves similarly to the constant learning rate with warmup~(see~\cref{fig:ap_prodigy_effective_lr} and related ablations). 
Moreover, \texttt{Prodigy} scales reliably similar to \texttt{AdamW}, making it a promising choice for future development of parameter-free methods.

\begin{algorithm}[ht]
\caption{\texttt{Prodigy}}
\label{alg:prodigy}
\begin{algorithmic}[1]
\State {\bfseries Input:} Initial parameters $\xx_0$, number of iterations $T$, learning rate $\gamma$, weight decay $\lambda$, $\beta_1$, $\beta_2$, $\varepsilon$.
\State {\bfseries Initialize:} $d_0 \leftarrow 10^{-6}$, $\gamma \leftarrow 1$, $\mm_0 \leftarrow \mathbf{0}$, $\vv_0 \leftarrow \mathbf{0}$, $r_0 \leftarrow 0$, $\boldsymbol{s}_0 \leftarrow \mathbf{0}$ \Comment{Optional: use scheduler on $\gamma$}
\For{$t \in [T]$}
\State $\bg_t \leftarrow \nabla \mathcal{L}(\xx_t, \xixi_t)$
\State $\mm_t \leftarrow \beta_1 \mm_{t-1} + (1 - \beta_1)d_t\bg_t$
\State $\vv_t \leftarrow \beta_2 \vv_{t-1} + (1 - \beta_2) d^2_t \bg_t \odot \bg_t$
\State $\gamma_t \leftarrow \gamma \sqrt{1 - \beta_2^t}/(1 - \beta_1^t)$ \Comment{Optional: use bias correction}
\State $r_t \leftarrow \sqrt{\beta_2} r_{t-1} + (1 - \sqrt{\beta_2})\gamma_t d^2_t\langle \bg_t, \xx_0 - \xx_t\rangle$
\State $\boldsymbol{s}_t \leftarrow \sqrt{\beta_2}\boldsymbol{s}_{t-1} + (1 - \sqrt{\beta_2})\gamma_t d^2_t\bg_t$
\State $d_{t+1} \leftarrow \max\left\{d_t, \frac{r_t}{\|\boldsymbol{s}_t\|_1}\right\}$
\State $\xx_{t+1} \leftarrow \xx_t - \gamma_t d_t\left(\mm_t / \left(\sqrt{\vv_t} + d_t \varepsilon \right) + \lambda \xx_t\right)$
\EndFor
\State {\bfseries Return:} $\xx_T$
\end{algorithmic}
\end{algorithm}

\subsection{\texttt{MARS}}
\label{sec:ap_mars}

Very recently, Yuan, Liu et al.~\cite{yuan2024marsunleashingpowervariance} introduced \texttt{MARS}---a family of optimizers incorporating modern adaptive~\cite{loshchilov2019decoupledweightdecayregularization,chen2023symbolicdiscoveryoptimizationalgorithms} and approximate second-order methods~\cite{gupta2018shampoopreconditionedstochastictensor} methods with a variance reduction update style.

This optimization framework gave a rise to: \texttt{MARS-AdamW}, our main baseline, which we call simply \texttt{MARS}; \texttt{MARS-Lion}; and \texttt{MARS-Shampoo}.
We mainly include \texttt{MARS-AdamW} in our ablation studies, but also report results for the other two optimizers~(see \cref{fig:mars_types}).

The authors modified the variance reduction update by introducing a scaling parameter $\eta$, which we call variance reduction scaling in the outlined algorithms and experiments.
This parameter controls the scale of gradient correction---see \texttt{line 5} of Algorithms~\ref{alg:marsadamw},~\ref{alg:marslion},~and~\ref{alg:marsshampoo}.

Importantly, we follow only the approximate scheme of \texttt{MARS}-like optimizers, i.e., we evaluate the gradient $\bg_t$ in different stochasticity, meaning that
\begin{align*}
    \bg_t &= \nabla \mathcal{L}\left(\xx_t, \xixi_t\right), \\
    \bg_{t-1} &= \nabla \mathcal{L}\left(\xx_{t-1}, \xixi_{t-1}\right).
\end{align*}

In the same spirit as for \texttt{SOAP} and \texttt{Muon}, the authors use \texttt{MARS}-like algorithms for layers with two or more dimensions. 
\textit{For $1$D layers, embeds, scalar parameters and the final layer of neural network, this method utilizes} \texttt{AdamW}.
This design choice enables efficient and fast training with \texttt{MARS}.
Following the practices from the original work, we also use \texttt{MARS} only for $2$D layers.
Importantly, for \texttt{MARS}-based methods, one need to tune both the \texttt{AdamW}'s learning rate $\gamma_t^{\mathtt{A}}$, and the learning rate for $2$D parameters, which we denote as $\gamma_t^{\mathtt{M}}$ for compatibility with the \texttt{Muon} pseudocode~\ref{alg:muon}. 

\textbf{\texttt{MARS}~(\texttt{MARS-AdamW}).}
For the \texttt{AdamW}-like algorithm, the difference occurs in the computation of $\mm_t$ and $\vv_t$, where the variance reduction update $\boldsymbol{c}_t$ is used instead of the gradient.

\begin{algorithm}[ht]
\caption{\texttt{MARS} (\texttt{MARS-AdamW})}
\label{alg:marsadamw}
\begin{algorithmic}[1]
    \State \textbf{Input:} Initial parameters $\xx_0$, number of iterations $T$, learning rate $\gamma_t$, weight decay $\lambda$, $\beta_1$, $\beta_2$, variance reduction scaling $\eta$, $\varepsilon$.
    \State {\bfseries Initialize:}  $\mm_0\leftarrow \mathbf{0}$, $\vv_0\leftarrow \mathbf{0}$
    \For{$t \in [T]$}
        \State $\bg_t \leftarrow \nabla \mathcal{L}(\xx_t, \xixi_t)$
        \State $\boldsymbol{c}_t \leftarrow \bg_t + \eta \frac{\beta_1}{1 - \beta_1}\left(\bg_t - \bg_{t-1}\right)$
        \If{$\|\boldsymbol{c}_t\|_2 > 1$}
        \State $\boldsymbol{c}_t \leftarrow \boldsymbol{c}_t / \|\boldsymbol{c}_t\|_2$
        \EndIf
        \State $\mm_t \leftarrow \beta_1 \mm_{t-1} + (1-\beta_1)\boldsymbol{c}_t$
        \State $\vv_t \leftarrow \beta_2 \vv_{t-1} + (1-\beta_2) \boldsymbol{c}_t \odot \boldsymbol{c}_t$
        \State $\hat{\mm}_t \leftarrow \mm_t / (1 - \beta_1^t)$, \; $\hat{\vv}_t \leftarrow \vv_t / (1 - \beta_2^t)$
        \State $\xx_{t+1} = \xx_t - \gamma_t \left( \frac{\hat{\mm}_t}{\sqrt{\hat{\vv}_t} + \varepsilon}+ \lambda \xx_t\right)$
    \EndFor
    \State {\bfseries Return:} $\xx_T$
\end{algorithmic}
\end{algorithm}

We point out once again that, for LLM training, we only run~\cref{alg:marsadamw} for $2$D parameters, resulting in the following two updates at each iteration:
\begin{align*}
    \xx_{t+1}^{\mathtt{M}} &\leftarrow \texttt{MARS}\left(\xx_t^{\mathtt{M}}, \gamma_t^{\mathtt{M}}, \lambda^{\mathtt{M}}, \beta_1^{\mathtt{M}}, \beta_2^{\mathtt{M}}, \varepsilon, T=1\right) \;\;\;\;\,\, \text{for $2$D parameters}, \\
    \xx_{t+1}^{\mathtt{A}} &\leftarrow \texttt{AdamW}\left(\xx_t^{\mathtt{A}}, \gamma_t^{\mathtt{A}}, \lambda^\mathtt{A}, \beta_1^{\mathtt{A}}, \beta_2^{\mathtt{A}}, \varepsilon, T=1\right) \;\;\,\, \text{for $1$D parameters},
\end{align*}

i.e., in the same way as in \cref{alg:muon,alg:soap}.
The same holds for two more versions---\texttt{MARS-Lion} and \texttt{MARS-Shampoo}, which we discuss below.

\textbf{\texttt{MARS-Lion}.}
Similarly to the \texttt{Lion} algorithm, the authors use scaled gradient correction $\boldsymbol{c}_t$ with the current gradient.
Importantly, \cref{alg:marslion} does not leverage second moment estimates to update $2$D parameters.
Instead, the updates rely on the sign-based characteristic of \texttt{Lion} integrated with the variance reduction framework.

\begin{algorithm}[ht]
\caption{\texttt{MARS-Lion}}
\label{alg:marslion}
\begin{algorithmic}[1]
    \State \textbf{Input:} Initial parameters $\xx_0$, number of iterations $T$, learning rate $\gamma_t$, weight decay $\lambda$, $\beta_1$, variance reduction scaling $\eta$, $\varepsilon$.
    \State {\bfseries Initialize:}  $\mm_0\leftarrow \mathbf{0}$, $\vv_0\leftarrow \mathbf{0}$
    \For{$t \in [T]$}
        \State $\bg_t \leftarrow \nabla \mathcal{L}(\xx_t, \xixi_t)$
        \State $\boldsymbol{c}_t \leftarrow \bg_t + \eta \frac{\beta_1}{1 - \beta_1}\left(\bg_t - \bg_{t-1}\right)$
        \If{$\|\boldsymbol{c}_t\|_2 > 1$}
        \State $\boldsymbol{c}_t \leftarrow \boldsymbol{c}_t / \|\boldsymbol{c}_t\|_2$
        \EndIf
        \State $\mm_t \leftarrow \beta_1 \mm_{t-1} + (1-\beta_1)\boldsymbol{c}_t$
        \State $\xx_{t+1} = \xx_t - \gamma_t \left( \mathtt{sign}\left(\mm_t\right)+ \lambda \xx_t\right)$
    \EndFor
    \State {\bfseries Return:} $\xx_T$
\end{algorithmic}
\end{algorithm}

\textbf{\texttt{MARS-Shampoo}.}
The same holds for \texttt{MARS-Shampoo}.
A key point to note is that, to compute $\mathtt{SVD}$ of the first moment estimate, the authors also perform the Newton-Schulz steps \cite{bernstein2024oldoptimizernewnorm,functionsmatrices}.
In our experiments, we use $5$ iterations of this orthogonalization scheme for \texttt{MARS-Shampoo}.

\begin{algorithm}[ht]
\caption{\texttt{MARS-Shampoo}}
\label{alg:marsshampoo}
\begin{algorithmic}[1]
    \State \textbf{Input:} Initial parameters $\xx_0$, number of iterations $T$, learning rate $\gamma_t$, weight decay $\lambda$, $\beta_1$, variance reduction scaling $\eta$, $\varepsilon$.
    \State {\bfseries Initialize:}  $\mm_0\leftarrow \mathbf{0}$, $\vv_0\leftarrow \mathbf{0}$
    \For{$t \in [T]$}
        \State $\bg_t \leftarrow \nabla \mathcal{L}(\xx_t, \xixi_t)$
        \State $\boldsymbol{c}_t \leftarrow \bg_t + \eta \frac{\beta_1}{1 - \beta_1}\left(\bg_t - \bg_{t-1}\right)$
        \State $\mm_t \leftarrow \beta_1 \mm_{t-1} + (1-\beta_1)\boldsymbol{c}_t$
        \State $\boldsymbol{U}_t, \boldsymbol{\Sigma}_t, \boldsymbol{V}_t \leftarrow \mathtt{SVD}(\mm_t)$ \Comment{Use Newton-Schulz orthogonalization}
        \State $\xx_{t+1} = \xx_t - \gamma_t \left(\boldsymbol{U}_t \boldsymbol{V}_t^\top + \lambda \xx_t\right)$
    \EndFor
    \State {\bfseries Return:} $\xx_T$
\end{algorithmic}
\end{algorithm}

\section{Implementation}
\label{sec:ap_implementation}

Our code is based on an extension of nanoGPT\footnote{\href{https://github.com/karpathy/nanoGPT}{https://github.com/karpathy/nanoGPT}} and uses PyTorch \cite{paszke2019pytorchimperativestylehighperformance} as well as FlashAttention \cite{dao2022flashattentionfastmemoryefficientexact}.
We incorporate mixed-precision training \cite{micikevicius2018mixedprecisiontraining}, i.e., we train in \texttt{bfloat16} precision, except for normalization modules and softmax which we train in \texttt{float32}.
The optimizer states are also stored in \texttt{float32}.
The majority of experiments were performed using a cluster of A$100$-SXM$4$-$80$GB, H$100$-HBM$3$-$80$GB GPUs as well as GH$200$-$120$GB.
We trained both in a single GPU regime and in DDP \cite{li2020pytorchdistributedexperiencesaccelerating} (from $2$ to $8$ GPUs per one run).
We estimate that the full cost of all experiments for our project to roughly $30 000$ GPU hours.
To give an idea of how much effort was put into tuning each method, across all model sizes, batches and iterations, we trained a total of $2900$ models. 
This includes includes \textit{nearly}: $750$ \texttt{AdamW}, $145$ \texttt{ADOPT}, $238$ \texttt{AdEMAMix}, $158$ \texttt{Lion}, $165$ \texttt{Signum}, $231$ \texttt{Muon}, $135$ \texttt{D-Muon}, $354$ \texttt{SOAP}, $199$ \texttt{Sophia}, $133$ \texttt{SF-AdamW}, $195$ \texttt{Prodigy}, $217$ \texttt{MARS-AdamW}, $26$ \texttt{MARS-Lion}, and $20$ \texttt{MARS-Shampoo} models.
See~\cref{sec:ap_tuning} for details about hyperparameter tuning.

\section{Model \& Data}
\label{sec:ap_modeldata}

\textbf{Architecture details.}
In our project, we use the Llama-like family of models \cite{grattafiori2024llama3herdmodels}.
We implement the popular in the community decoder-only transformer with SwiGLU activation functions \cite{shazeer2020gluvariantsimprovetransformer}, RoPE embeddings \cite{su2023roformerenhancedtransformerrotary}, RMSNorm \cite{zhang2019rootmeansquarelayer}.
The vocabulary is based on the GPT2 \cite{radford2019language} tokenizer \footnote{\href{https://github.com/openai/tiktoken}{https://github.com/openai/tiktoken}} and contains $50304$ tokens.
Importantly, our variant of the Llama-based architecture employs weight tying~\cite{press2017usingoutputembeddingimprove}.

The number of parameters in our models is fully configurable, and we present the exact configurations used in our experiment in \cref{tab:llama_models}.

\begin{table*}[ht]
    \centering
    \caption{\textbf{Configurations for our Llama-like models.}}
    \label{tab:llama_models}
    \begin{tabular}{|c|c|c|c|c|}
    \hline
    \textbf{\# Parameters} & $\mathbf{124}\mathbf{M}$ & $\mathbf{210}\mathbf{M}$ & $\mathbf{583}\mathbf{M}$ & $720\mathbf{M}$ \\ 
    \hline
    Hidden size & $768$ & $768$ & $1920$ & $2048$\\ 
    \hline
    \# Attention heads & $12$ & $12$ & $15$ & $16$\\ 
    \hline
    \# Layers & $12$ & $24$ & $11$ & $12$ \\ 
    \hline
     Init. std & $0.02$ & $0.02$ & $0.02$ & $0.02$\\
    \hline
    Use bias & no & no & no & no\\
    \hline
    RMSNorm epsilon & $0.00001$ & $0.00001$ & $0.00001$ & $0.00001$ \\
    \hline
    Positional encoding & RoPE & RoPE & RoPE & RoPE \\
    \hline
    \end{tabular}
\end{table*}

\textbf{Dataset.}
Our main findings are obtained on the subset of FineWeb \cite{penedo2024finewebdatasetsdecantingweb} with $100\mathbf{B}$ tokens \footnote{\href{https://huggingface.co/datasets/HuggingFaceFW/fineweb}{https://huggingface.co/datasets/HuggingFaceFW/fineweb}}, cleaned and deduplicated
corpus for LLM pretraining, which we split into train and validation sequences.
During training, we evaluate the models with a fixed set of $32$ batches of our chosen sequence length ($512$ for almost all experiments, the same context length as training) to establish the validation loss curves.
At the end of the training, we compute the full validation loss and perplexity (this loss is reported as $\mathrm{Final\;Validation\;Loss}$ in the figures).
We also performed our initial results on the subset of the OpenWebText2 dataset \cite{pile}.

\section{Additional Results}
\label{sec:ap_additionalres}

In this section, we complement our results from the main part with extended experiments.
We start sequentially with smaller models of $124\mathbf{M}$ and $210\mathbf{M}$ parameters, ablating: \textbf{warmup}, \textbf{weight decay}, \textbf{learning rate schedulers}, \textbf{gradient norm patterns}, \textbf{learning rate decaying}, and other optimizer-related phenomena.
We finalize this section with the \textbf{wall-clock performance of optimizers}.
Details on hyperparameter searches are provided in \cref{sec:ap_tuning}.

\subsection{Ablations for $\mathbf{124M}$ model}
\label{sec:ap_124mablations}

At first, we systematically gather all ablations with $124\mathbf{M}$ parameter models.
As in \S~\ref{sec:smallscalebench}, we study: the effect of scaling the number of iterations and hyperparameter dependence on $T$; warmup; the importance of weight decay for optimizers; $\gamma$-sensitivity; a comparison of $\gamma$-schedulers; gradient norm patters during training; learning rate decaying; and optimizer-specific phenomena for \texttt{Sophia}, \texttt{SF-AdamW}, \texttt{Prodigy}, \texttt{Muon}, \texttt{Signum}, and \texttt{MARS}-based methods.

\textbf{Scaling the number of iterations.}
\label{sec:ap_scalingiters}
We stay in the setup from \S~\ref{sec:setup}, training $124\mathbf{M}$ model with batches of $256\times512$ tokens.
Our training runs in \cref{fig:benchmarking-124m-losses}~\textbf{(b)} demonstrate the the gap between \texttt{SOAP} and \texttt{AdEMAMix} narrows as the training horizon extends.
As for $124\mathbf{M}$ model, we tuned all optimizers on $2.1\mathbf{B}$ tokens length of training, some hyperparameters, particularly those sensitive to the training duration, may become suboptimal for longer runs of $16.8\mathbf{B}$ tokens.
For example, the beta parameter ($\beta_2$) of the second moment estimate in \texttt{AdamW}-like methods should arguably be re-tuned when the number of iterations is increased~\cite{orvieto2025searchadamssecretsauce,marek2025smallbatchsizetraining,busbridge2023scaleema}, which also makes theoretical claims~\cite{reddi2019convergenceadam,défossez2022simpleconvergenceproofadam,wang2023closinggapupperbound,NEURIPS2018_90365351,taniguchi2024adoptmodifiedadamconverge,chezhegov2024gradientclippingimprovesadagrad}.
Our extensive tuning on $2.1\mathbf{B}$ yielded the result that for \texttt{AdamW}, \texttt{SOAP}, and \texttt{AdEMAMix} optimizers, $\beta_2$ should be set to $0.999$.
Importantly, Pagliardini et al.~\cite{pagliardini2024ademamixoptimizerbetterfaster} suggest increasing $\beta_3$ of \texttt{AdEMAMix}, which controls the slow EMA~(see \texttt{line 7} of \cref{alg:ademamix}), for longer training.

As such, we conducted two experiments.

First, we keep the best hyperparameters found for $2.1\mathbf{B}$ tokens horizon, and extend them to $2\times$ loner duration than the maximum one ($16.8\mathbf{B}$ tokens), resulting in a total of $33.6\mathbf{B}$ tokens---it is interesting to observe whether the gap between \texttt{SOAP} and \texttt{AdEMAMix} finally closes in a longer run.
Secondly, we re-tune beta parameters of \texttt{SOAP} and \texttt{AdEMAMix} for $16.8\mathbf{B}$ and $33.6\mathbf{B}$ runs, and compare results.
Our re-tuned values are $\beta_2=0.9999$ for \texttt{SOAP}, and $\beta_3=0.9999$ for \texttt{AdEMAMix}.

\begin{figure}[h]
    \centering
    \subfigure[Training dynamics with different $\beta_3$.]{
        \includegraphics[width=0.48\linewidth]{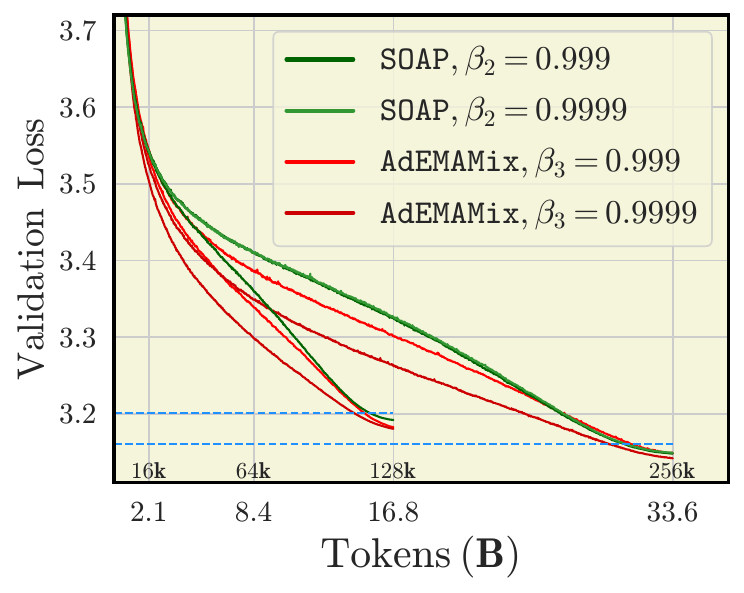}
    }
    \hfill
    \subfigure[Increase $\beta_3$ for \texttt{AdEMAMix} for longer training.]{
        \includegraphics[width=0.48\linewidth]{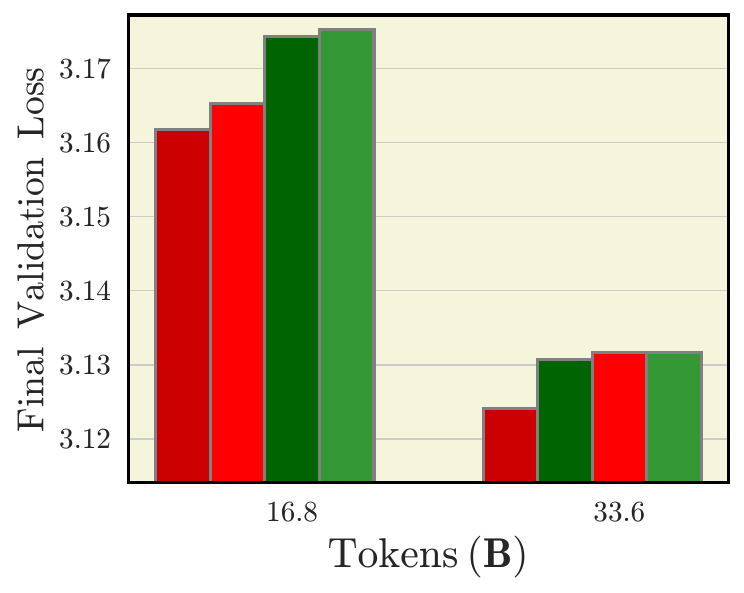}
    }
    \caption{\textbf{Re-tuning beta parameters is significant for longer training.}
    This experimental setup coincides with the one from \cref{fig:scalebs_scaletokens}---where we consider an impact of increasing batch size or number of iterations.
    We elaborate more on the impact of beta parameters for \texttt{AdEMAMix}: in \textbf{(a)}, we show a training dynamics of \texttt{AdEMAMix} and \texttt{SOAP} for $T\in\{128, 256\}\mathbf{k}$ steps, and \textbf{(b)} demonstrates the final loss.
    Throughout this experiment we use a batch size of $256\times512$ tokens and train a $124\mathbf{M}$ model.
    Our results reveal that increasing $\beta_3$ for \texttt{AdEMAMix} is crucial for long runs.
    Without these changes, \texttt{SOAP} with $\beta_2=0.999$ found via tuning on $16\mathbf{k}$ steps runs ends up outperforming \texttt{AdEMAMix} with $\beta_3=0.999$.
    Re-tuning $\beta_2$ of \texttt{SOAP} does not change the results much, and give almost identical loss curves, however, the training dynamics if \texttt{AdEMAMix} changes dramatically with $\beta_3$ as noticed in~\cite{pagliardini2024ademamixoptimizerbetterfaster}, and gives best results with larger $\beta_3=0.9999$.
    }
    \label{fig:ap_retuning_betas}
\end{figure}

This ablation is described in \cref{fig:ap_retuning_betas}~\textbf{(a,b)}.
We see that, indeed, without re-tuning, \texttt{SOAP} ends up outperforming \texttt{AdEMAMix} when extending the training horizon further to $33.6\mathbf{B}$ tokens ($\equiv 256\mathbf{k}$ iterations).
However, with re-tuning of $\beta_2$ for \texttt{SOAP} and $\beta_3$ for \texttt{AdEMAMix}, the latter optimizer still takes the lead.
Notably, in our experiments, $\beta_3=0.999$ is only better than $\beta_3=0.9999$ when the number of training iterations is less than $32\mathbf{k}$.
Surprisingly, given the many theoretical claims from works analyzing \texttt{Adam}, that $\beta_2$ depends on the number of iterations: $\beta_2 = \beta_2(T) = 1 - 1/T$, we do not observe this to be a common rule in practice, as many influential settings regarding LLM pretraining~\cite{deepseekai2024deepseekv3technicalreport,brown2020languagemodelsfewshotlearners,touvron2023llamaopenefficientfoundation,olmo20242olmo2furious} utilize a typical ($\beta_2=0.95$) even for very long training for trillions of tokens.
Therefore, we highlight this oversight in \cref{tkw:retuning_betas}, proposing to re-tune $\beta_2$ hyperparameter of \texttt{Adam}-like methods when changing of the training horizon.

\prettybox{
\takeaway{tkw:retuning_betas}We highlight the overlooked claim that $\beta_2$ parameters of \texttt{Adam}-like methods should be re-tuned with training durations.
One needs to increase $\beta_2$ for longer training.
This re-tuning significantly improves the optimizer performance.
}

\textbf{Warmup ablation.}
\label{sec:ap_warmupablation}
In this section, we supplement the experiments on warmup from \S~\ref{sec:smallscalebench}.
We study the impact of warmup on the final validation loss.
Replicating our setup~(\S~\ref{sec:setup}), we use the batch size of $256\times512$ tokens and reuse the best hyperparameters found through tuning, except for $T_\text{warmup}$.
For all methods, we sweep over $T_\text{warmup}\in\{1.56\%, 6.25\%, 25\%\}$ of the total training duration $T$ to examine each method's sensitivity to warmup.
Additionally, for \texttt{AdamW}, we extend this sweep to $T_\text{warmup}\in\{1.56\%, 5\%, 6.25\%, 10\%, 25\%\}$ of $T$.
We specifically consider $1.56\%$ and $6.25\%$ percentages because the former represents a typical number of warmup steps ($2000$) for models of our scale, while the latter ($6.25\%$ of $128000$ steps) aligns with the warmup strategy used in Llama~\cite{grattafiori2024llama3herdmodels}.

Contrary to the insights from~\cite{zhang2024doescriticalbatchsize}, we observe that $25\%$ of the Cinchilla optimal duration~($620\mathbf{M}$ tokens for $124\mathbf{M}$ model) is far from being the best batch size for pretraining.
We emphasize that their results were obtained for $85\mathbf{M}$ models and then extrapolated to larger scales.
However, in our setting, we found that the basic $2000$ steps were a more suitable warmup option for most optimizers; exceptions include sign-based methods (\texttt{Signum}, \texttt{Lion}), and \texttt{Sophia} with \texttt{SF-AdamW}.

\begin{figure}[h]
    \centering       
        \includegraphics[width=0.6\linewidth]{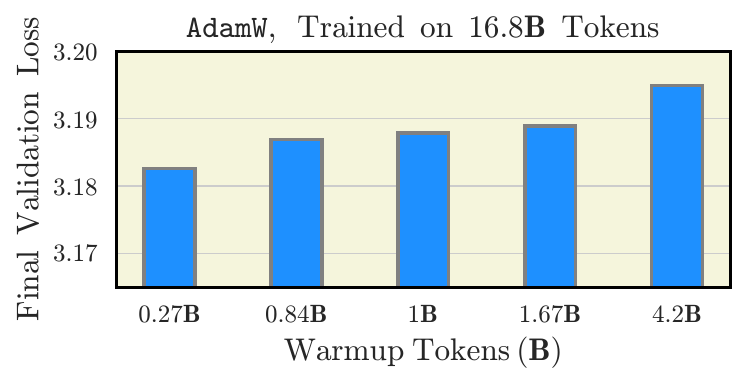}
    \caption{\textbf{Warmup sweep for \texttt{AdamW}.}
    We observe that the smaller yet reasonable warmup value is the best.
    However, this is not the case for other methods like \texttt{Signum}, \texttt{Lion}, \texttt{Sophia}, and \texttt{SF-AdamW}---see \cref{fig:warmup}.
    }
    \label{fig:warmup_sweep_adamw}
    \vspace{-2.4em}
\end{figure}

We provide the warmup sweep for \texttt{AdamW} in \cref{fig:warmup_sweep_adamw}.

\textbf{Weight decay ablation.}
\label{sec:ap_wdablation}
Prior work analyzing the importance of weight decay $\lambda$ in model training suggests tuning both $\lambda$ and the learning rate $\gamma$ so that their product $\lambda \gamma$ remains constant.
D{'}Angelo et al.~\cite{dangelo2024needweightdecaymodern} argue that, across different pairs of $\gamma$ and $\lambda$, the lowest error of the model is observed along the contour in the hyperparameter space where $\lambda \gamma = \text{const}$.
The authors also establish a connection between the quantity of $\lambda \gamma$ product and an effect of regularization and noise scale in the over-training regime, such as for small computer vision models trained over multiple epochs.
Kosson et al.~\cite{kosson2024rotationalequilibriumweightdecay} highlight that if $\gamma$ and $\lambda$ are chosen to result in constant product, the model achieves the same equilibrium rotation value, reflecting a similar effective update size for the weights.
While previous studies have analyzed the rule of keeping $\lambda \gamma = \text{const}$ primarily on image classification tasks with ResNet-based models~\cite{he2015deepresiduallearningimage}, Pagliardini et al.~\cite{pagliardini2024ademamixoptimizerbetterfaster} also used this heuristic when tuning hyperparameters for LLM pretraining.

\begin{figure*}[h]
    \centering
    \begin{minipage}{0.318\linewidth}
        \includegraphics[width=\linewidth]{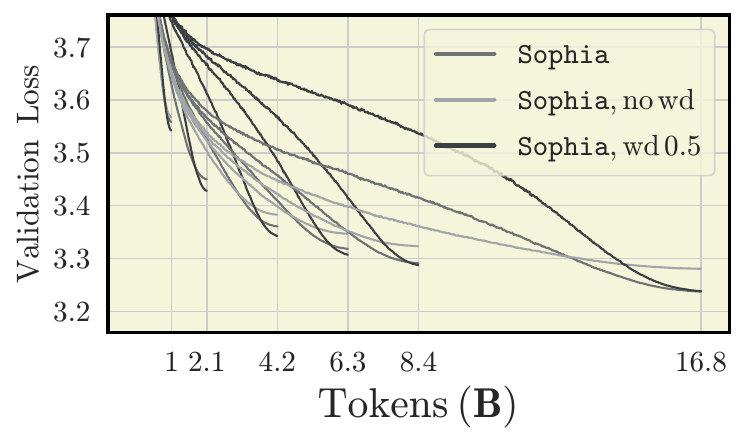}
    \end{minipage}
    \hfill
    \begin{minipage}{0.318\linewidth}
        \includegraphics[width=\linewidth]{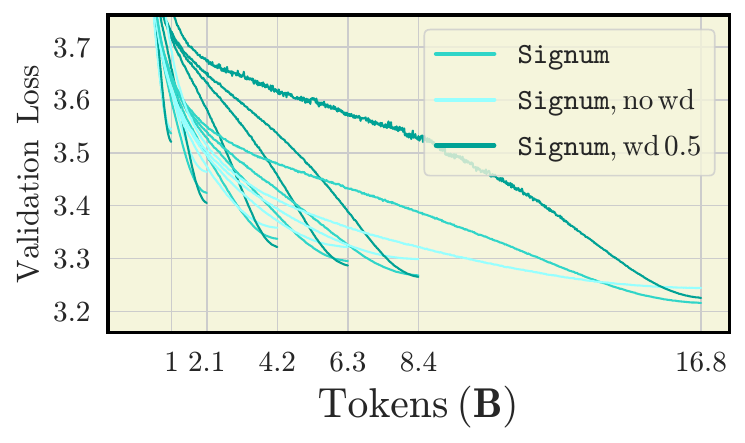}
    \end{minipage}
    \hfill
    \begin{minipage}{0.318\linewidth}
        \includegraphics[width=\linewidth]{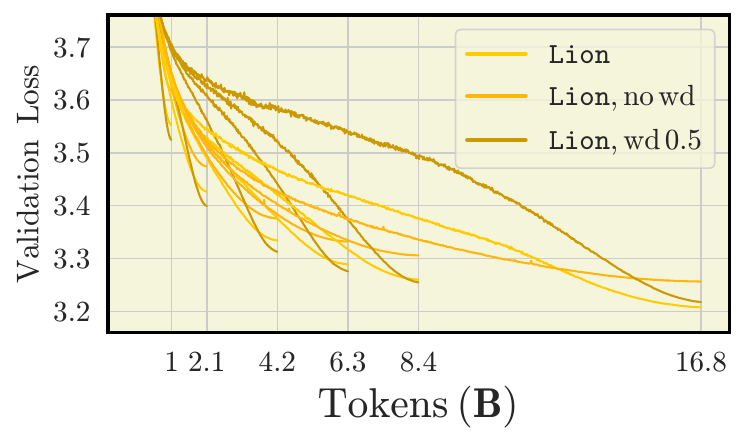}
    \end{minipage}
    \\
    \begin{minipage}{0.318\linewidth}
        \includegraphics[width=\linewidth]{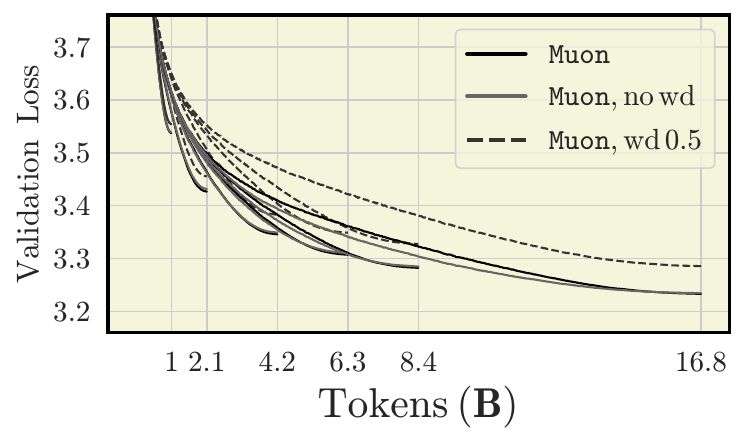}
    \end{minipage}
    \hfill
    \begin{minipage}{0.318\linewidth}
        \includegraphics[width=\linewidth]{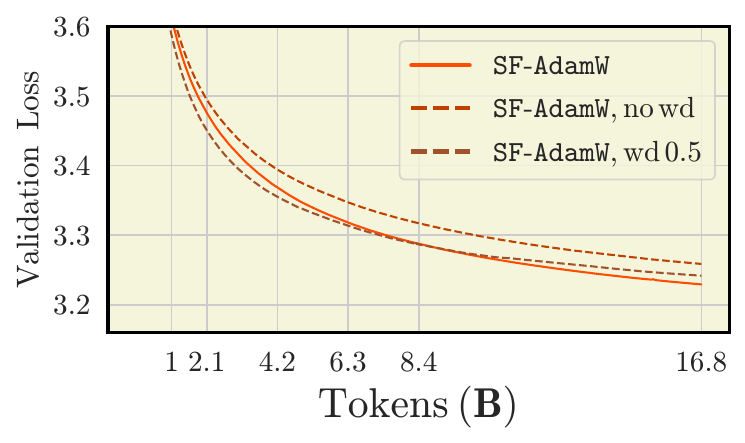}
    \end{minipage}
    \hfill
    \begin{minipage}{0.318\linewidth}
        \includegraphics[width=\linewidth]{figures/weight-decay/adamw_wd.pdf}
    \end{minipage}
    \\
    \begin{minipage}{0.318\linewidth}
        \includegraphics[width=\linewidth]{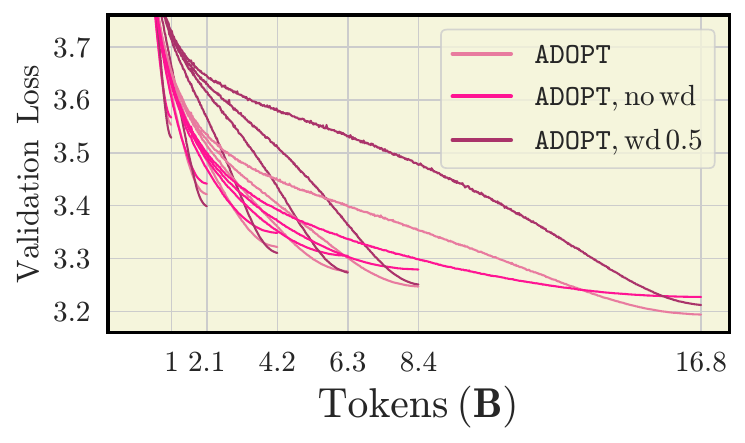}
    \end{minipage}
    \hfill
    \begin{minipage}{0.318\linewidth}
        \includegraphics[width=\linewidth]{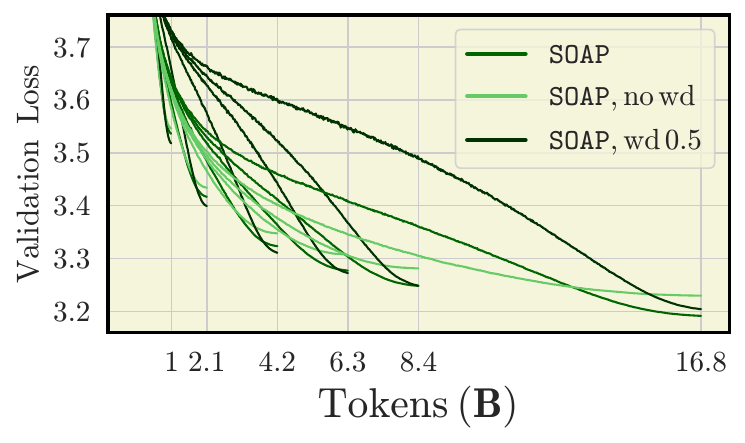}
    \end{minipage}
    \hfill
    \begin{minipage}{0.318\linewidth}
        \includegraphics[width=\linewidth]{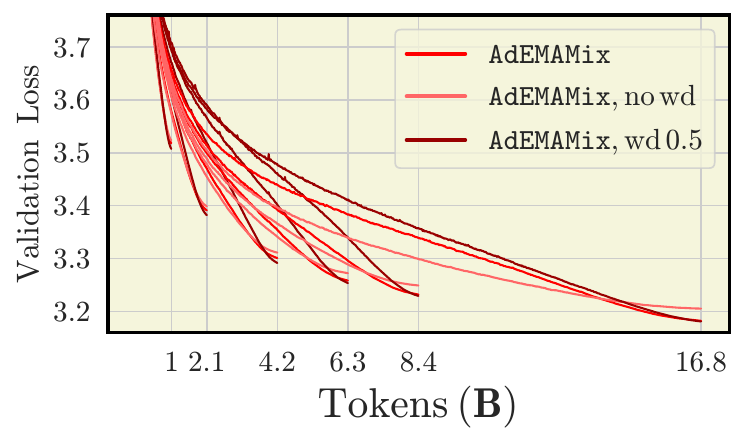}
    \end{minipage}
    \\
    \centering 
    \begin{minipage}{0.318\linewidth}
        \includegraphics[width=\linewidth]{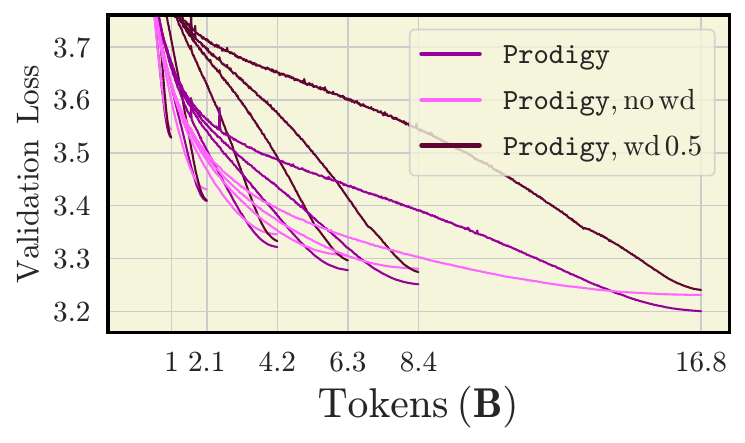}
    \end{minipage}
    \hfill
    \begin{minipage}{0.318\linewidth}
        \includegraphics[width=\linewidth]{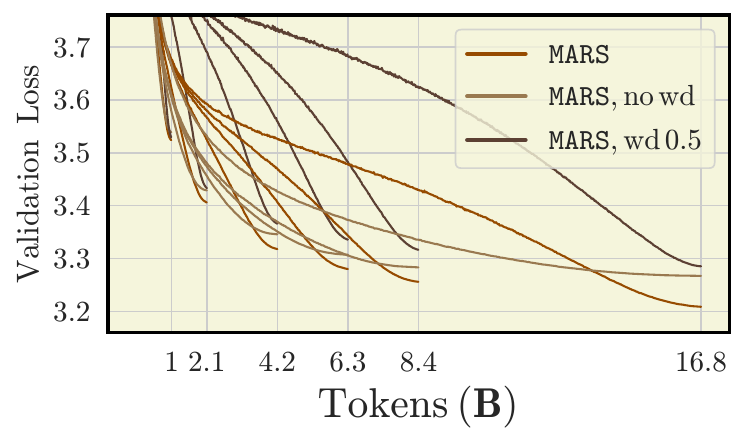}
    \end{minipage}
    \hfill
    \begin{minipage}{0.318\linewidth}
        \includegraphics[width=\linewidth]{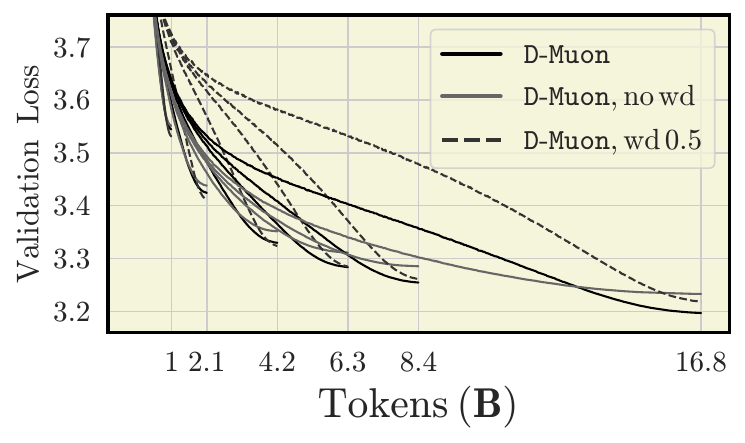}
    \end{minipage}
    \caption{\textbf{Larger weight decay achieves significantly better results when training on fewer tokens.}
    We observe that the majority of runs with the large weight decay of $0.5$ consistently outperform those with weight decay of $0.1$ for all training durations except for the long training on $16.8\mathbf{B}$ tokens.
    Notably, \texttt{Signum} and \texttt{Lion} with large weight decay perform even better than \texttt{AdamW} with the same learning rate---see~\cref{fig:wdablation_main}.
    We also consider a setting without weight decay.
    We observe that this is suboptimal for most of other optimizers, while the typical weight decay of $0.1$ remains the best for long training durations.
    An interesting pattern emerges for optimizers that treat one-dimensional and two-dimensional parameters differently, such as \texttt{Muon} and \texttt{MARS}.
    For these, runs with large weight decay ($0.5$) consistently underperform those with $0.1$ and, in some cases, even those without weight decay.
    For \texttt{Muon}, we attribute this effect to its algorithmic design, in which weight decay is not employed to optimize matrix parameters (see \cref{alg:muonnon1d}), in contrast to \texttt{D-Muon}, where the observed patterns are reliably similar to those seen with \texttt{AdamW}.
    For \texttt{MARS}, we only vary the weight decay corresponding to matrix parameters while keeping $0.1$ for all scalar, one-dimensional and final layer parameters.
    In this case, we conclude that the gap between large and small weight decay values narrows significantly faster.
    }
    \label{fig:ap_wdablation}
    \vspace{-2em}
\end{figure*}

In our study, which focuses solely on language modelling, we demonstrate that using a relatively large weight decay term with a fixed learning rate can significantly accelerate short training runs.
Throughout our weight decay ablation experiments, we fix the best value of $\gamma$ found via tuning on near-Chinchilla optimal $T$, and sweep the weight decay across $\lambda \in \{0, 0.1, 0.5\}$, where $\lambda=0.1$ is the standard value of the decoupled weight decay term in our work.
Our results are consistent across optimizers and training horizons: runs with large $\lambda$ dominate for a small number of iterations, but as the training length increases to $\{8.4, 16.8\}\mathbf{B}$ tokens, runs with a moderate $\lambda=0.1$ begin to outperform~(\cref{fig:ap_wdablation,fig:wdablation_main}). 
An important example is \texttt{Muon}.
As this optimizer does not use weight decay for $2$D parameters, we observe that runs with $\lambda=0.5$ underperform those with $\lambda \in \{0, 0.1\}$ even in short training on $\{1, 2.1, 4.2, 6.3\}\mathbf{B}$ tokens. 
However, when we consider an implementation of the \texttt{D-Muon} optimizer with learning rate and weight decay shared across all parameters, we again observe a similar pattern to that seen with other methods---larger weight decay dominates when training on fewer tokens.

We highlight these observations for practitioners and suggest that this approach may be useful for short training runs. 
Our main claim from this section is summarized in \cref{tkw:weight_decay}.

\begin{figure*}[h]
    \centering
        \begin{minipage}{0.318\linewidth}
        \includegraphics[width=\linewidth]{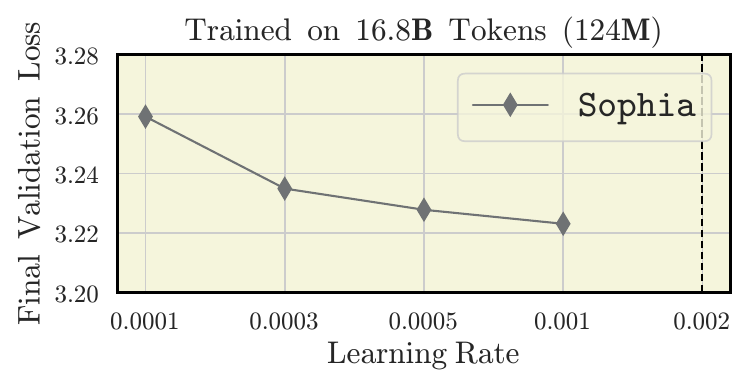}
    \end{minipage}
    \hfill
    \begin{minipage}{0.318\linewidth}
        \includegraphics[width=\linewidth]{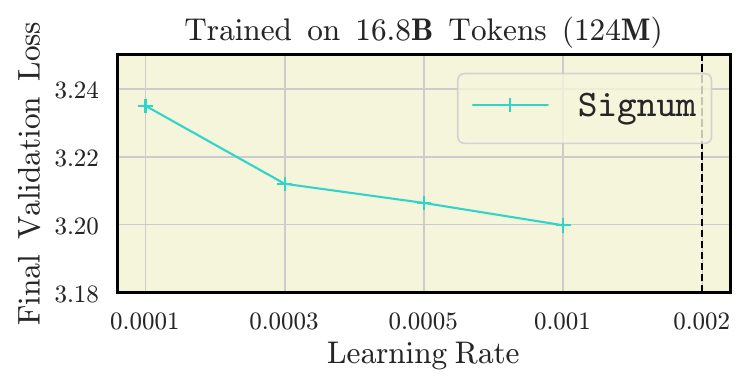}
    \end{minipage}
    \hfill
    \begin{minipage}{0.318\linewidth}
        \includegraphics[width=\linewidth]{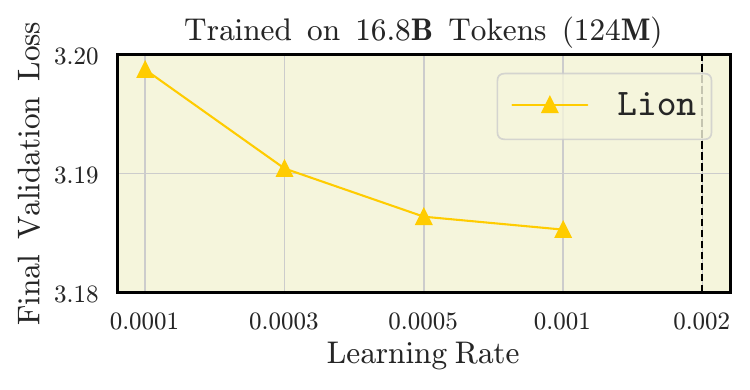}
    \end{minipage}
    \\
    \begin{minipage}{0.318\linewidth}
        \includegraphics[width=\linewidth]{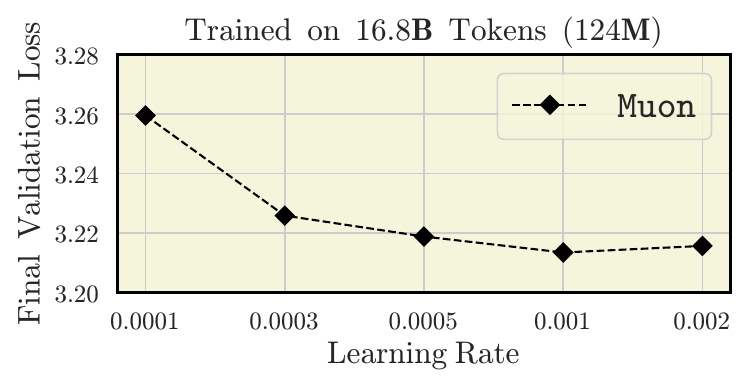}
    \end{minipage}
    \hfill
    \begin{minipage}{0.318\linewidth}
        \includegraphics[width=\linewidth]{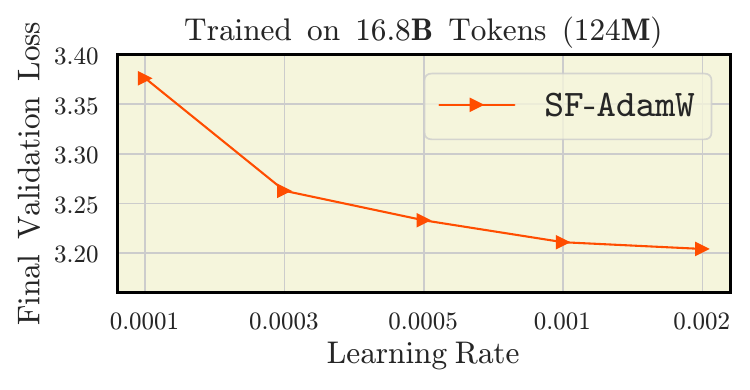}
    \end{minipage}
    \hfill
    \begin{minipage}{0.318\linewidth}
        \includegraphics[width=\linewidth]{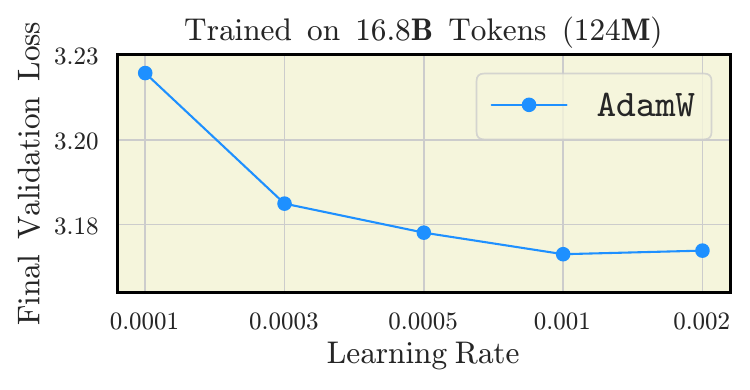}
    \end{minipage}
    \\
    \begin{minipage}{0.318\linewidth}
        \includegraphics[width=\linewidth]{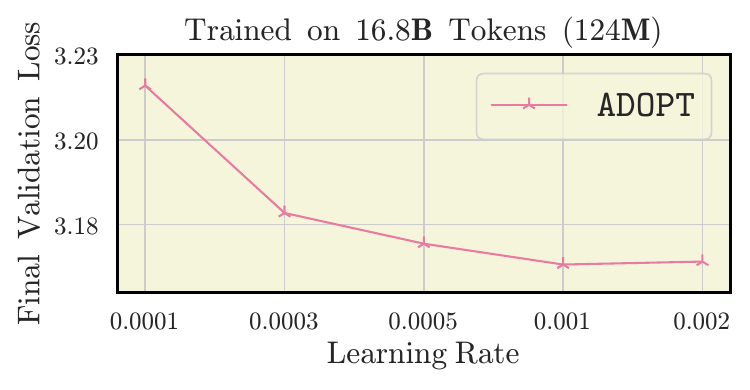}
    \end{minipage}
    \hfill
    \begin{minipage}{0.318\linewidth}
        \includegraphics[width=\linewidth]{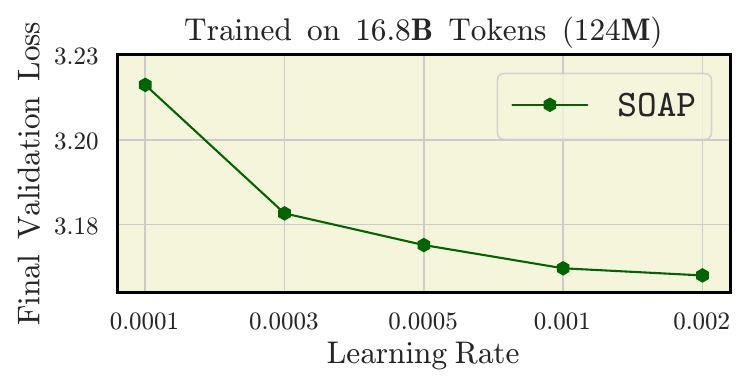}
    \end{minipage}
    \hfill
    \begin{minipage}{0.318\linewidth}
        \includegraphics[width=\linewidth]{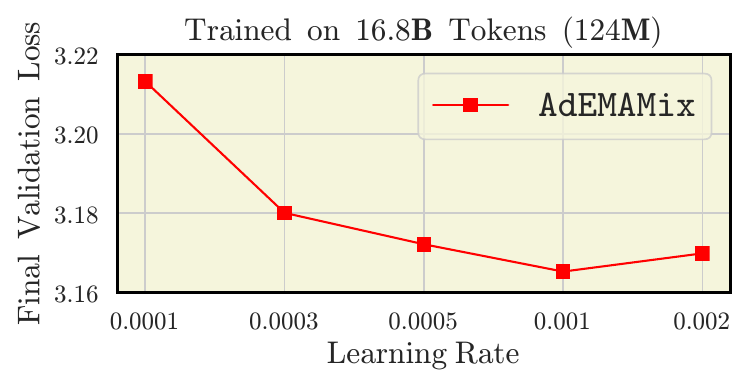}
    \end{minipage}
    \\
    \centering 
    \begin{minipage}{0.318\linewidth}
        \includegraphics[width=\linewidth]{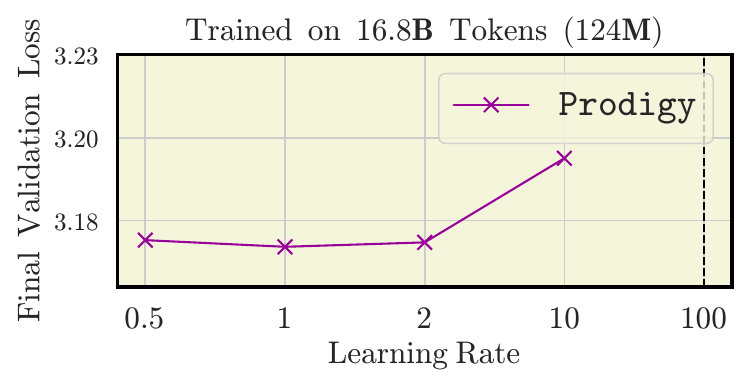}
    \end{minipage}
    \hfill
    \begin{minipage}{0.318\linewidth}
        \includegraphics[width=\linewidth]{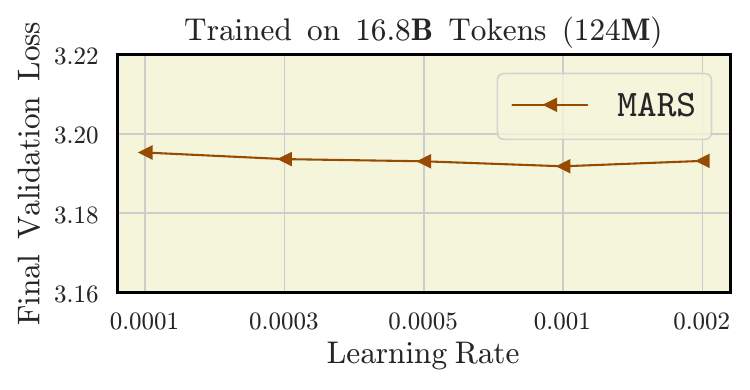}
    \end{minipage}
    \hfill
    \begin{minipage}{0.318\linewidth}
        \includegraphics[width=\linewidth]{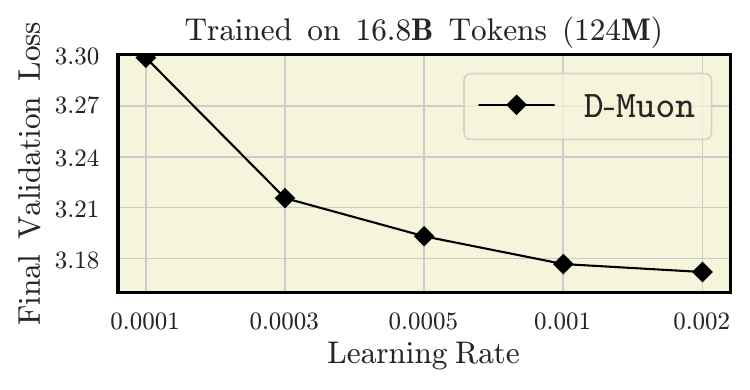}
    \end{minipage}
    \caption{\textbf{Learning rate sensitivity.}
    In the current setting, only \texttt{SOAP}, \texttt{SF-AdamW}, and \texttt{D-Muon} reach the better performance with the large learning rate of $0.002$.
    Conversely, \texttt{Sophia} and all sign-based methods (\texttt{Signum} and  \texttt{Lion}) diverge with this learning rate value.
    \texttt{MARS} and \texttt{Prodigy} show a remarkably consistent performance across the learning rate sweep.
    And, \texttt{Prodigy} diverges for sufficiently large value of $\gamma_{\max}$---see~\cref{fig:ap_prodigy_effective_lr} for more insights regarding the learning rate of \texttt{Prodigy}.
    }
    \label{fig:ap_lrsensitivity}
    \vspace{-1em}
\end{figure*}

\textbf{Learning rate sensitivity.}
\label{sec:ap_lrsensitivity}
In this part of the work, we meticulously replicate the learning rate sweep process and present comprehensive results. 
In line with our experimental setup (\S~\ref{sec:setup}), our aim is to determine the true impact of the learning rate and its transferability to longer training horizons.
For each optimizer, we only vary the learning rate while maintaining the best hyperparameters obtained during our initial tuning (see \cref{sec:ap_tuning,sec:ap_124mtuning}) on $2.1\mathbf{B}$ tokens for $124\mathbf{M}$ parameter model.
That is, the learning rate has been re-tuned for all optimizers on the training length of $16.8\mathbf{B}$ tokens. 
We do not present $\gamma$-sensitivity for \texttt{Prodigy} in the main part~(\S~\ref{sec:results}) because of the difference in axis scale: we sweep across $\gamma_{\max} \in \{0.5, 1, 2, 10, 100\}$ for this optimizer.
We show the results of the learning rate sweep in \cref{fig:ap_lrsensitivity}.

\begin{figure}[h]
    \centering
    \subfigure[Cosine scheduler outperforms WSD and \texttt{SF-AdamW}.]{
        \includegraphics[width=0.48\linewidth]{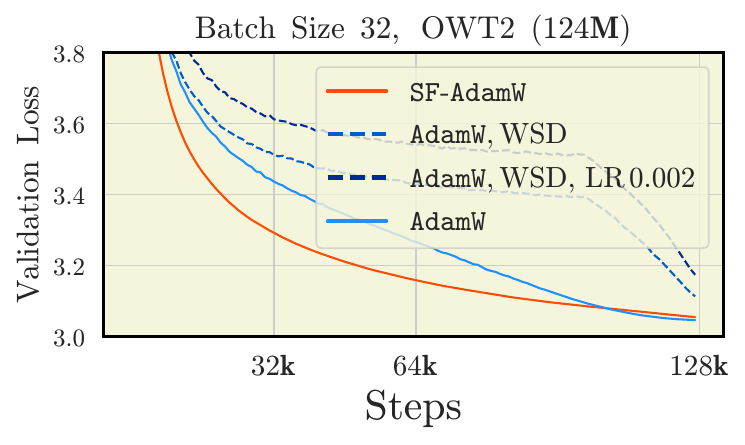}
    }
    \hfill
    \subfigure[Performance gap narrows with longer training.]{
        \includegraphics[width=0.48\linewidth]{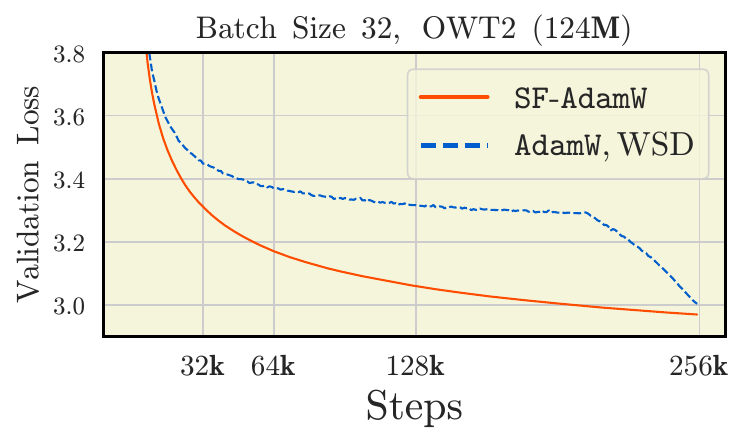}
    }
    \caption{\textbf{WSD scheduler underperforms both \texttt{AdamW} with cosine scheduler and \texttt{SF-AdamW}.}
    This is the only experiment we conduct on the OpenWebText$2$ (OWT$2$) dataset.
    We follow the small-batch setup, and replicate the best hyperparameters of each optimizer found through our tuning process.
    Once the learning rate and beta parameters of \texttt{SF-AdamW} and \texttt{AdamW} are properly tuned, we observe a surprisingly large performance gap between the WSD scheduler and its competitors.
    Figure \textbf{(b)} suggests that this gap may potentially diminish with extended training.
    }
    \label{fig:owt2wsdcosine}
    \vspace{-1.2em}
\end{figure}

\begin{figure*}[h]
    \centering
    \begin{minipage}{0.48\linewidth}
        \includegraphics[width=\linewidth]{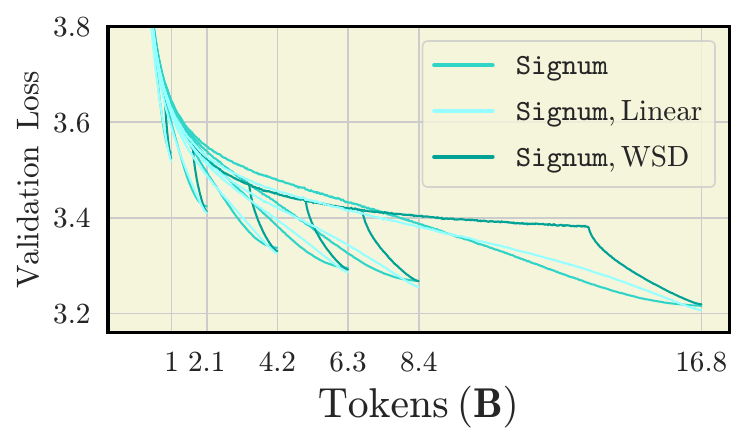}
    \end{minipage}
    \begin{minipage}{0.48\linewidth}
        \includegraphics[width=\linewidth]{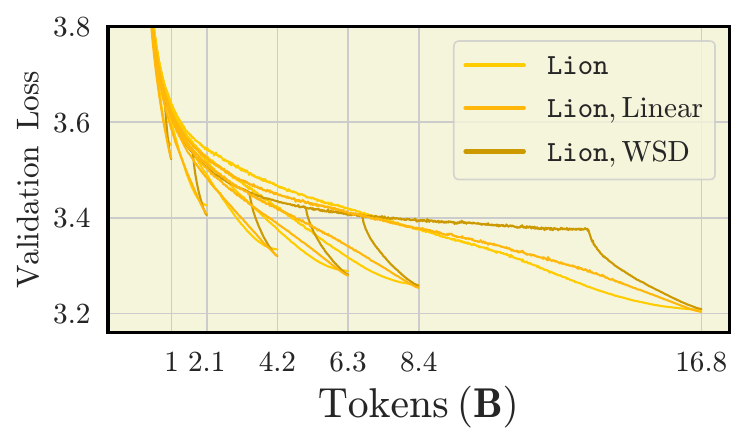}
    \end{minipage}
    \hfill
    \begin{minipage}{0.48\linewidth}
        \includegraphics[width=\linewidth]{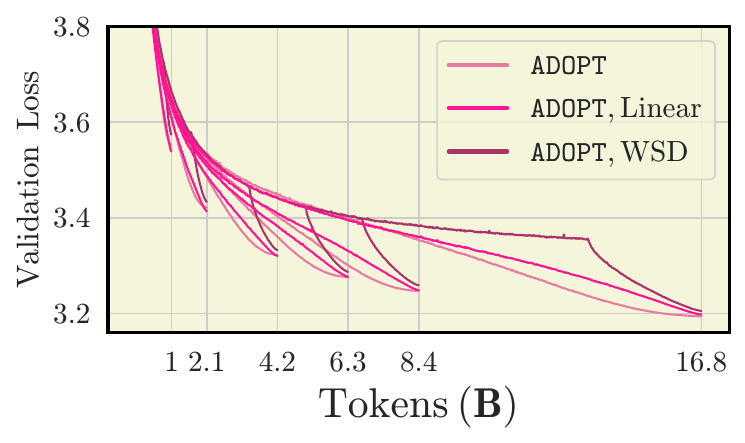}
    \end{minipage}
    \begin{minipage}{0.48\linewidth}
        \includegraphics[width=\linewidth]{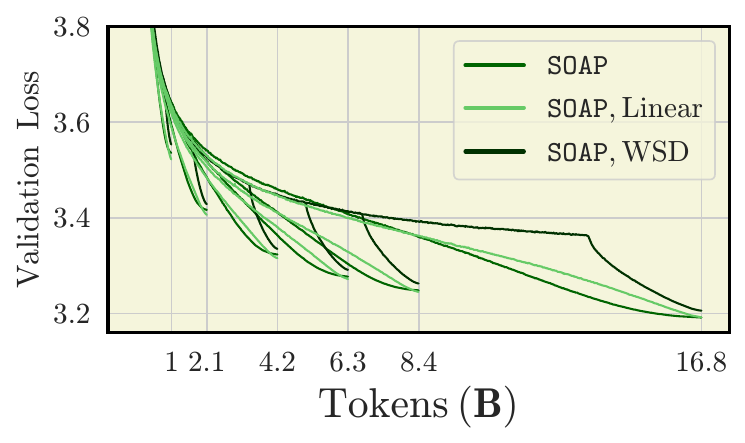}
    \end{minipage}
    \hfill
    \begin{minipage}{0.48\linewidth}
        \includegraphics[width=\linewidth]{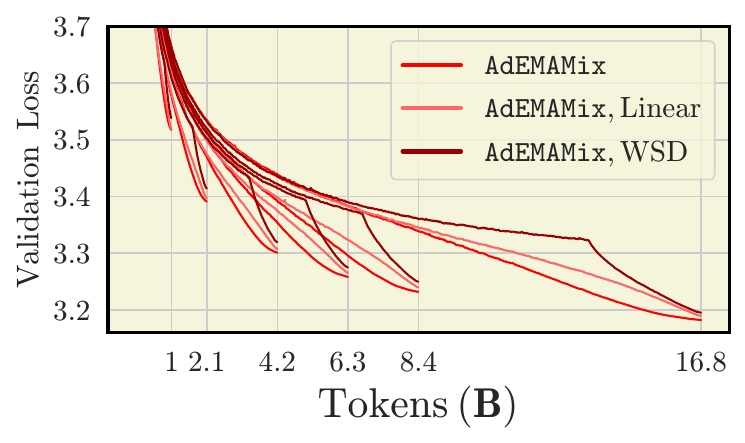}
    \end{minipage}
    \begin{minipage}{0.48\linewidth}
        \includegraphics[width=\linewidth]{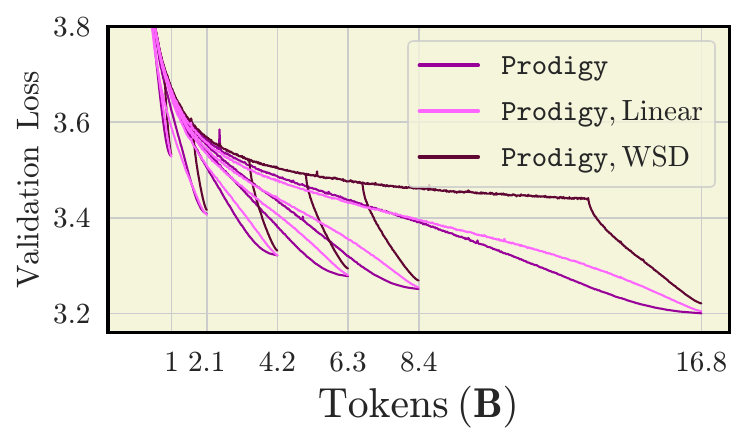}
    \end{minipage}
    \hfill
    \begin{minipage}{0.48\linewidth}
        \includegraphics[width=\linewidth]{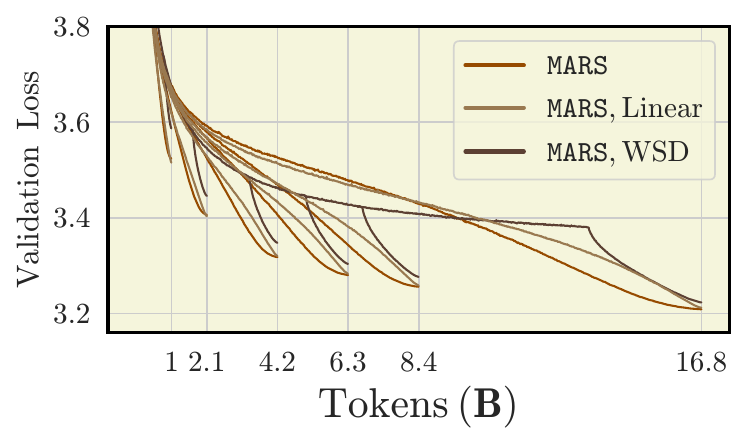}
    \end{minipage}
    \begin{minipage}{0.48\linewidth}
        \includegraphics[width=\linewidth]{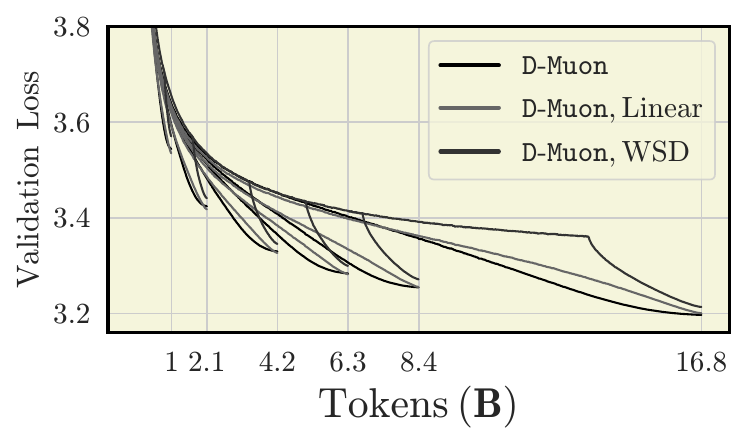}
    \end{minipage}
    \caption{\textbf{Comparisons between cosine, WSD, and the linear schedulers.}
    We complement results in~\cref{fig:wsdvscosine} by extending them to all the optimizers considered in our benchmarking.
    In most cases, the tuned cosine baseline performs similarly to runs using the linear scheduler, with both slightly outperforming WSD.
    However, certain optimizers still tend to ``prefer'' different $\gamma$-schedulers.
    For example, \texttt{Muon} shows a preference in WSD (see~\cref{fig:wsdvscosine}~\textbf{(a)}), \texttt{AdamW} performs better with the cosine scheduler, \texttt{Signum} and \texttt{Lion} appear to favor the linear scheduler.
    While the performance differences are not particularly large, they are still meaningful in the context of benchmarking.
    Therefore, we adopt the cosine scheduler as our default, as even small gaps can substantially impact our setup.
    }
    \label{fig:wsdcosine}
    \vspace{-2em}
\end{figure*}

\textbf{Comparison of learning rate schedulers.}
\label{par:ap_cosinevswsd}
In this part of our ablations, we systematically investigate the impact of $\gamma$-schedulers on optimizers.
As we mention in \S~\ref{sec:setup}, we conduct the majority of experiments on the FineWeb dataset~\cite{penedo2024finewebdatasetsdecantingweb}.
However, here we also present a small ablation on another corpus for LLM pretraining---OpenWebText$2$ (OWT$2$)~\cite{gao2020pile800gbdatasetdiverse}---as the main results of Defazio et al.~\cite{defazio2024roadscheduled} are obtained on the subset of this corpus.
We show our results for two batch size settings: $32\times 512$ for OWT$2$~(\cref{fig:owt2wsdcosine}), and $256\times 512$ for FineWeb~(\cref{fig:wsdcosine}).

In \cref{fig:owt2wsdcosine}, we present our initial results in the small-batch setting on the OWT$2$ dataset~\cite{gao2020pile800gbdatasetdiverse}.
We run the WSD scheduler experiments \textit{without following} the rule of thumb from~\cite{hägele2024scalinglawscomputeoptimaltraining}; instead, use a linear decay shape during the learning rate cooldown and set $\gamma$ to the value that is near-optimal for cosine.
Hence, we use $\gamma_{\max} = 0.001$ with the learning rate decay to $\gamma_\text{end} = 0.01\times\gamma_{\max}$ for both cosine and WSD schedulers.
This is the only experiment where we do not follow the best-practices of using WSD.
Regarding hyperparameter tuning, we observe little shift compared to that found in \cref{sec:ap_tuning} for FineWeb.
We only pose that it may be beneficial to additionally re-tune the gradient clipping threshold, as this depends on the ``cleanliness'' of the dataset.
Our ablations~(\cref{fig:owt2wsdcosine}) reveal that \texttt{SF-AdamW} can potentially outperform the \texttt{AdamW} baseline with the WSD scheduler. 
However, the cosine $\gamma$-scheduler still takes the lead in this setup.

We also report the final validation loss on the FineWeb dataset~\cite{penedo2024finewebdatasetsdecantingweb} for $124\mathbf{M}$ model trained with the batch size of $256\times512$ tokens.
For WSD, we follow the rule of thumb from H\"agele et al.~\cite{hägele2024scalinglawscomputeoptimaltraining}: $20\%$ of the steps for the cooldown, $\left(1 - \sqrt{x}\right)$ decay shape, and the learning rate is half the optimal for cosine, i.e., $0.0005$ if we have the best learning rate $0.001$ for the method.
Additionally, we point out that we do not include stochastic weight averaging~\cite{izmailov2019averagingweightsleadswider} in the comparison, which might potentially enhance the overall performance.
We ran the linear $\gamma$-scheduler with the same learning rates that we found through our tuning for cosine~(\cref{fig:lrsensitivity,fig:ap_lrsensitivity}).
We report our findings in~\cref{fig:wsdcosine}.
All missing optimizers---\texttt{AdamW}, \texttt{Muon}, and \texttt{Sophia}---are in the main part; see~\cref{fig:wsdvscosine}.
    
\begin{figure*}[h]
    \centering
    \begin{minipage}{0.318\linewidth}
        \includegraphics[width=\linewidth]{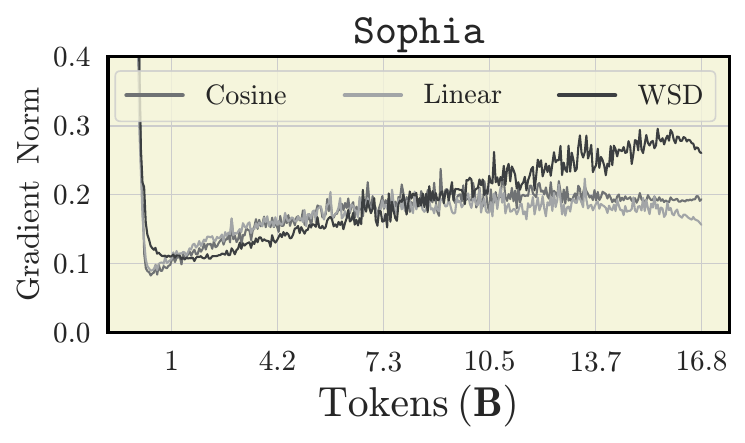}
    \end{minipage}
    \hfill
    \begin{minipage}{0.318\linewidth}
        \includegraphics[width=\linewidth]{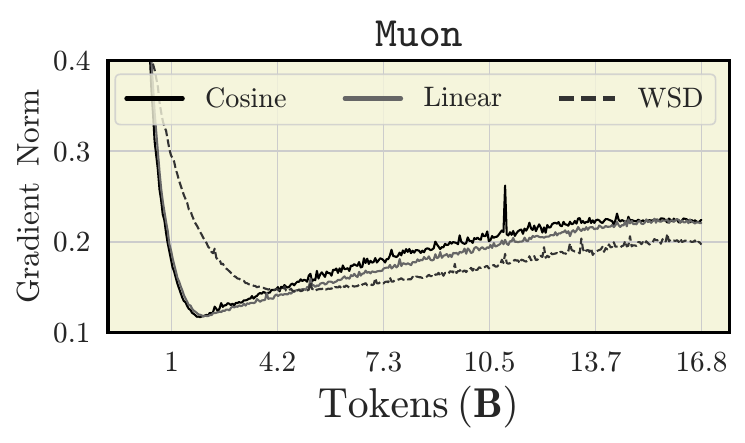}
    \end{minipage}
    \hfill
    \begin{minipage}{0.318\linewidth}
        \includegraphics[width=\linewidth]{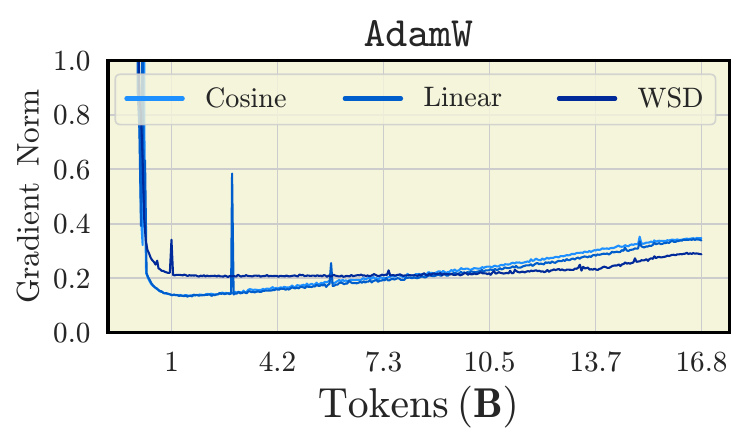}
    \end{minipage}
    \\
    \begin{minipage}{0.318\linewidth}
        \includegraphics[width=\linewidth]{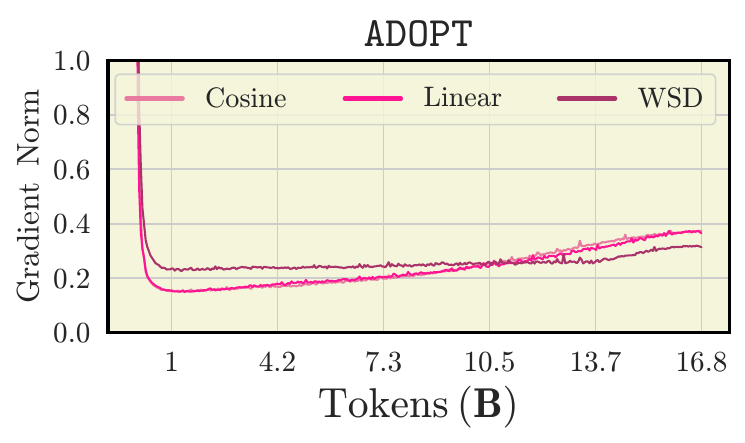}
    \end{minipage}
    \hfill
    \begin{minipage}{0.318\linewidth}
        \includegraphics[width=\linewidth]{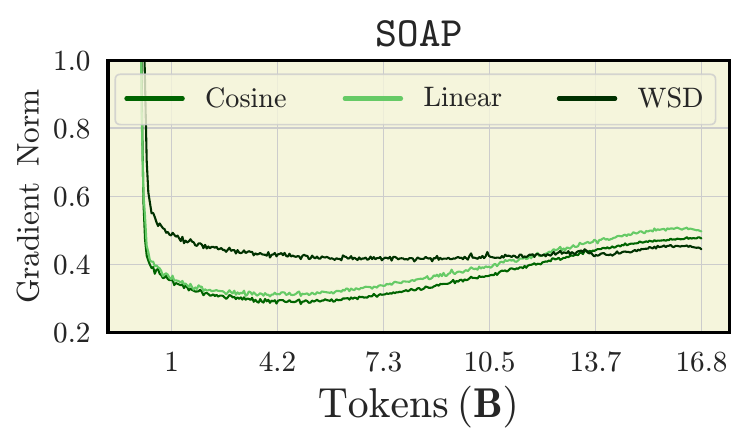}
    \end{minipage}
    \hfill
    \begin{minipage}{0.318\linewidth}
        \includegraphics[width=\linewidth]{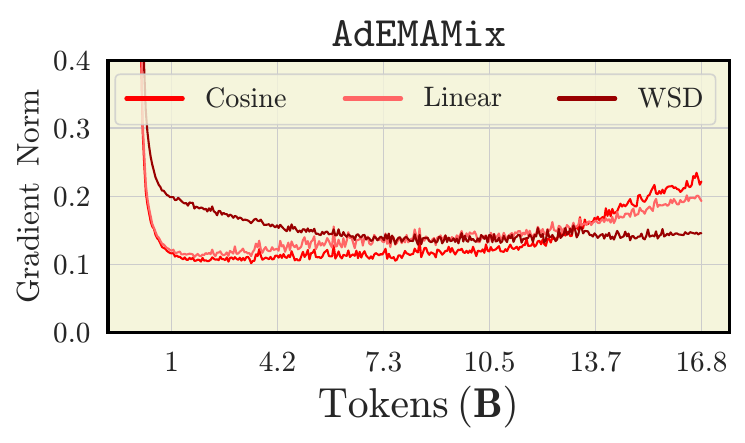}
    \end{minipage}
    \\
    \begin{minipage}{0.318\linewidth}
        \includegraphics[width=\linewidth]{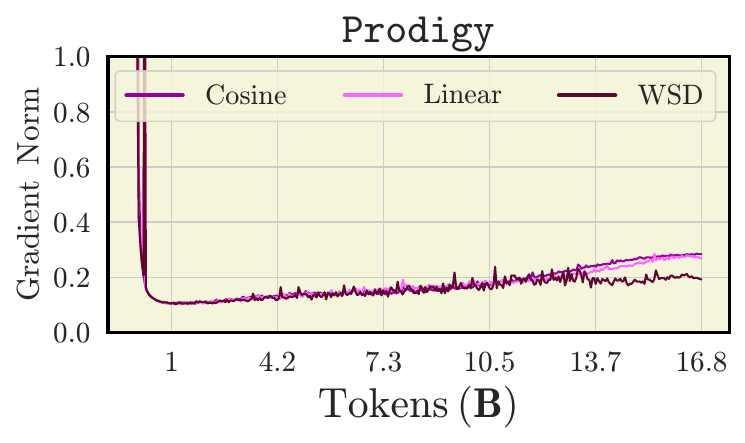}
    \end{minipage}
    \hfill
    \begin{minipage}{0.318\linewidth}
        \includegraphics[width=\linewidth]{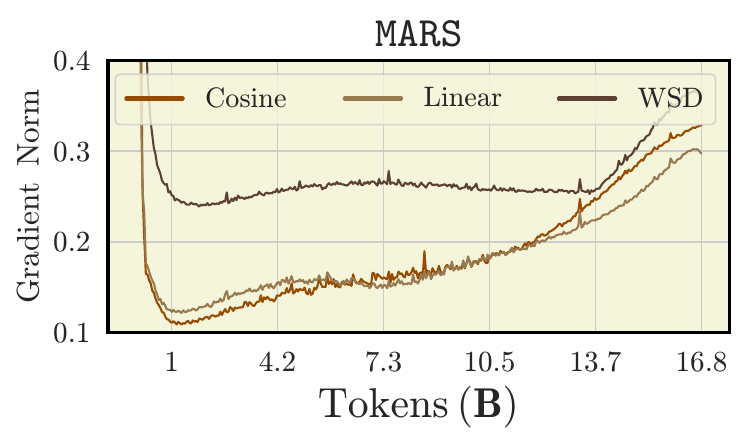}
    \end{minipage}
    \hfill
    \begin{minipage}{0.318\linewidth}
        \includegraphics[width=\linewidth]{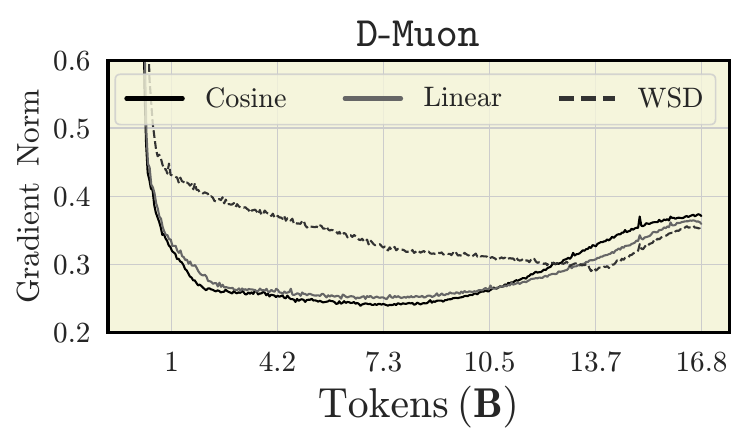}
    \end{minipage}
    \caption{\textbf{Gradient norm patterns for cosine, linear, and WSD $\mathbf{\gamma}$-schedulers.}
    We run all optimizers on $124\mathbf{M}$ models and track the gradient norms (before clipping) for runs using different $\gamma$-schedulers.
    For most optimizers, we see that gradient norms tend to increase over the course of training with cosine and linear schedules.
    In contrast, WSD tends to produce flatter gradient norm trajectories, with consistently lower magnitudes toward the end of training compared to the other schedulers.
    Since the WSD scheduler maintains a constant learning rate until the cooldown phase (the final $20\%$ of the training length), we observe a more stable gradient norm behavior in later stages.
    In this regard, our findings align with prior works~\cite{kosson2024rotationalequilibriumweightdecay,defazio2025gradientsrapidlyincreasenear}, which explore a connection between the learning rate schedule and gradient norm dynamics.
    Interestingly, \texttt{Signum} and \texttt{Lion}---see~\cref{fig:grad-norms-main-part}---exhibit a pronounced drop in gradient norm during the cooldown phase, setting them apart from the other optimizers.
    }
    \label{fig:ap_grad_norms_all}
    \vspace{-1em}
\end{figure*}

\begin{figure*}[h]
    \centering
    \begin{minipage}{0.318\linewidth}
        \includegraphics[width=\linewidth]{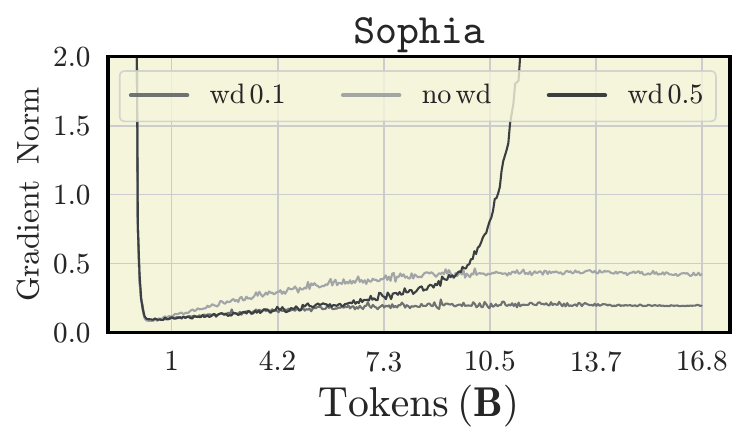}
    \end{minipage}
    \hfill
    \begin{minipage}{0.318\linewidth}
        \includegraphics[width=\linewidth]{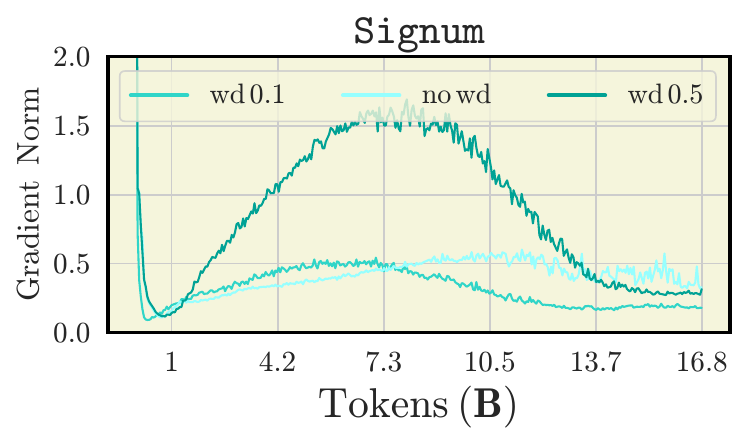}
    \end{minipage}
    \hfill
    \begin{minipage}{0.318\linewidth}
        \includegraphics[width=\linewidth]{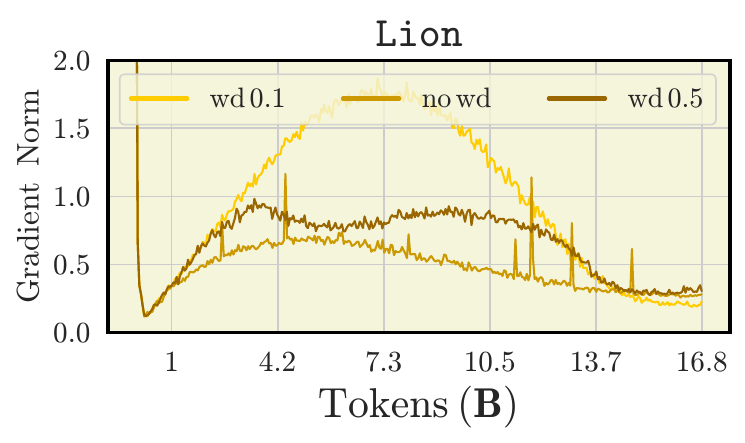}
    \end{minipage}
    \\
    \begin{minipage}{0.318\linewidth}
        \includegraphics[width=\linewidth]{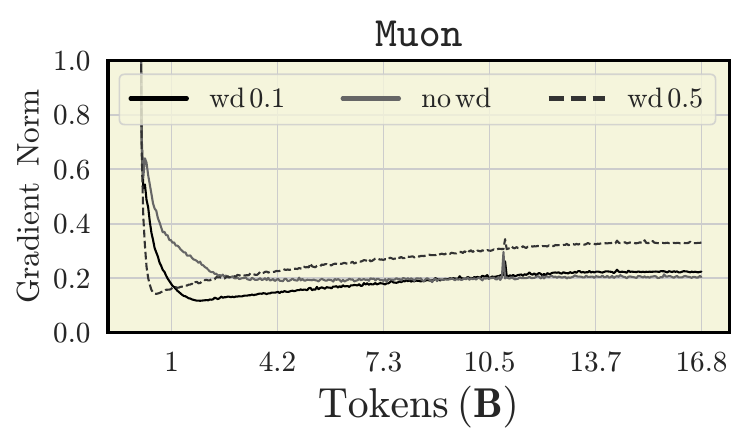}
    \end{minipage}
    \hfill
    \begin{minipage}{0.318\linewidth}
        \includegraphics[width=\linewidth]{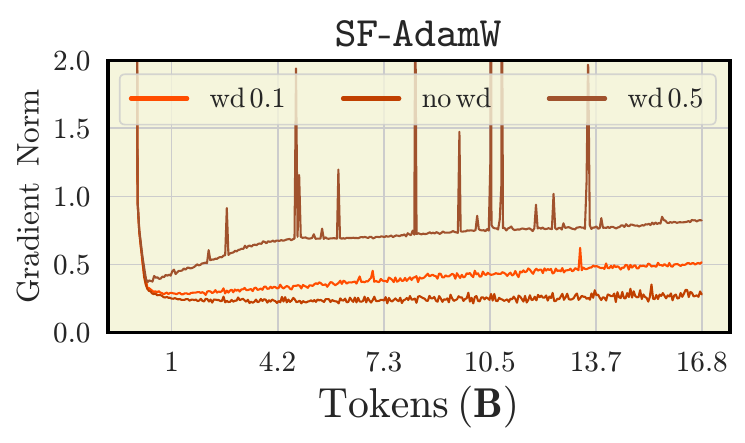}
    \end{minipage}
    \hfill
    \begin{minipage}{0.318\linewidth}
        \includegraphics[width=\linewidth]{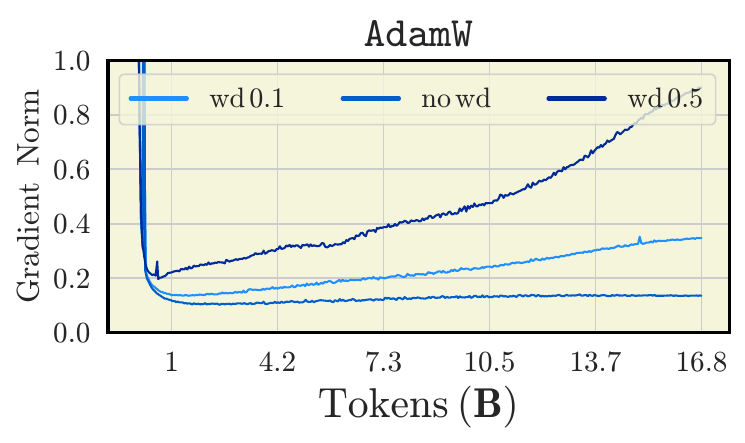}
    \end{minipage}
    \\
    \begin{minipage}{0.318\linewidth}
        \includegraphics[width=\linewidth]{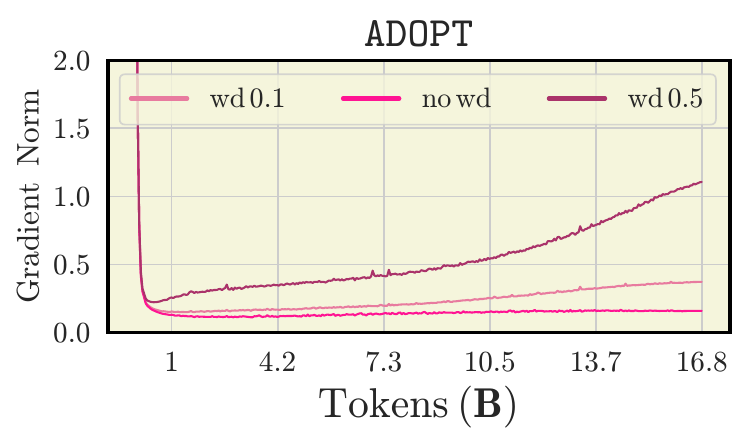}
    \end{minipage}
    \hfill
    \begin{minipage}{0.318\linewidth}
        \includegraphics[width=\linewidth]{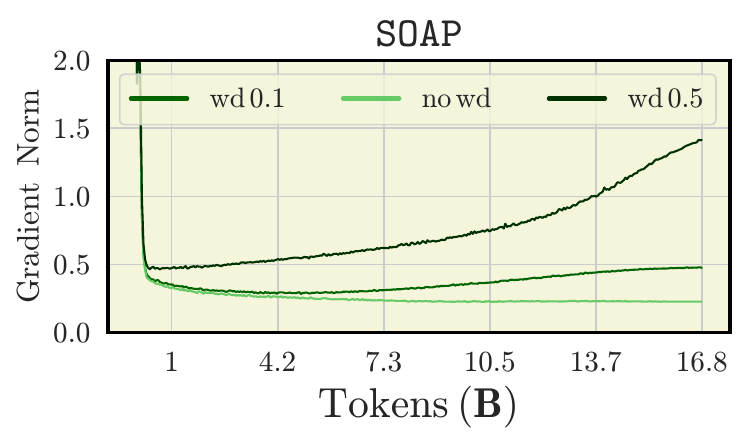}
    \end{minipage}
    \hfill
    \begin{minipage}{0.318\linewidth}
        \includegraphics[width=\linewidth]{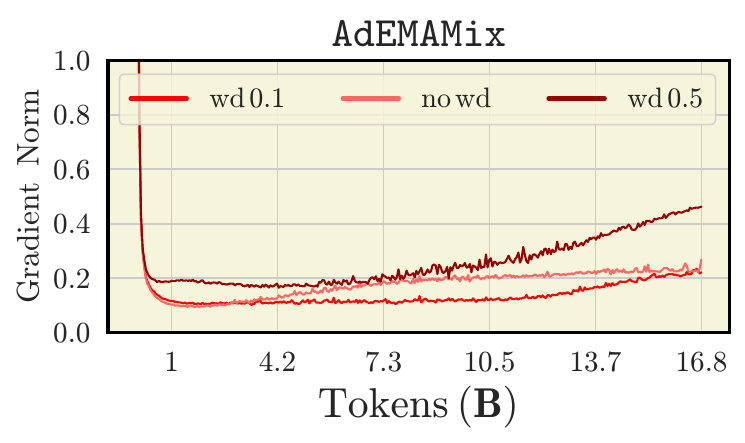}
    \end{minipage}
    \\
    \begin{minipage}{0.318\linewidth}
        \includegraphics[width=\linewidth]{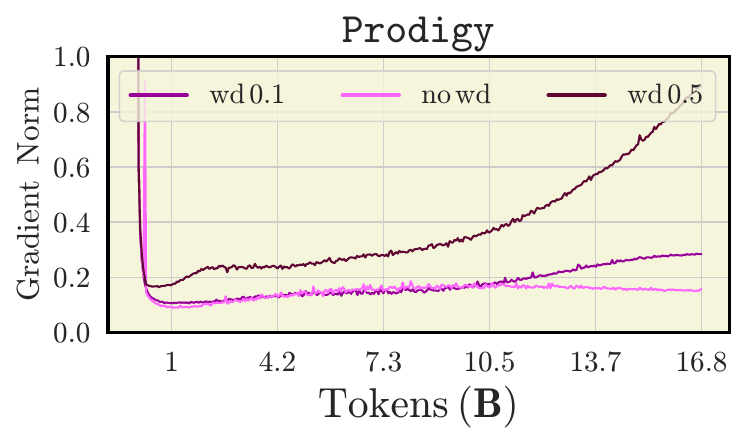}
    \end{minipage}
    \hfill
    \begin{minipage}{0.318\linewidth}
        \includegraphics[width=\linewidth]{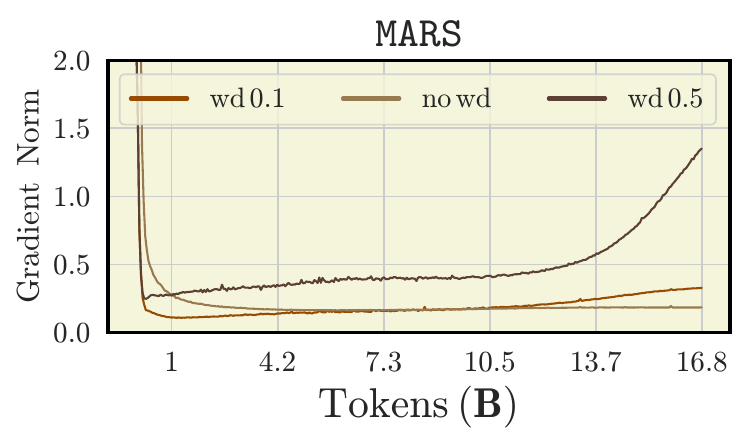}
    \end{minipage}
    \hfill
    \begin{minipage}{0.318\linewidth}
        \includegraphics[width=\linewidth]{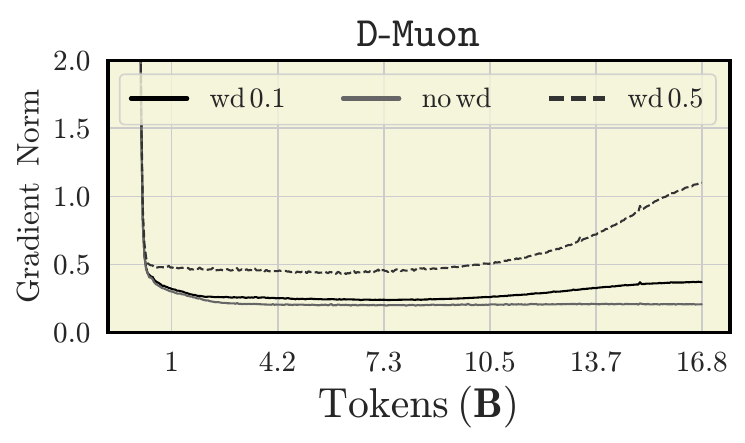}
    \end{minipage}
    \caption{\textbf{Gradient norm patterns for weight decay sweep.}
    We complement our weight decay ablation~(\cref{fig:wdablation_main,fig:ap_wdablation}) by tracking the gradient norms for all the optimizers studied in our benchmark.
    To highlight the effect of changing the weight decay, we use the same cosine $\gamma$-scheduler for all optimizers and keep the other best hyperparameters found, sweeping only the weight decay values as described in~\S~\ref{sec:setup}---i.e., we fix the maximum learning rate and only change the weight decay.
    For \texttt{Muon}, we only sweep the weight decay for $\{\mathtt{embeds}, \mathtt{scalar\_params}, \mathtt{lm\_head}\}$ (as in the initial implementation, the weight decay has not been applied to matrix parameters), while for \texttt{MARS}, we only sweep the weight decay of $2$D parameters.
    Our observations reveal that, regardless of optimizer used, runs with a larger weight decay result in higher gradient norms. 
    For \texttt{Muon}, \texttt{AdEMAMix}, \texttt{Sophia}, and sign-based methods, runs with moderate $\lambda=0.1$ result in the most flattened and smallest gradient norms in magnitude.
    While for \texttt{AdamW}-like methods, \texttt{D-Muon}, \texttt{SOAP}, \texttt{Prodigy}, and \texttt{SF-AdamW}, this holds for $\lambda=0$.
    We attribute the discrepancies between \texttt{D-Muon} and \texttt{Muon} to the latter's absence of weight decay for matrix parameters.
    As shown in~\cref{fig:wdablation_main,fig:ap_wdablation}, \texttt{AdEMAMix} can benefit from large weight decay for longer training durations.
    Runs of \texttt{AdEMAMix} with $\lambda=0.5$ are still outperform those with $\lambda=0.1$.
    Interestingly, this is reflected in the gradient norms, as the absolute values corresponding to $\lambda=0.5$ are much smaller than those of the respective runs of other \texttt{AdamW}-like optimizers.
    }
    \label{fig:ap_grad_norms_all_wd}
    \vspace{-2em}
\end{figure*}

\begin{figure*}[h]
    \centering
        \begin{minipage}{0.318\linewidth}
        \includegraphics[width=\linewidth]{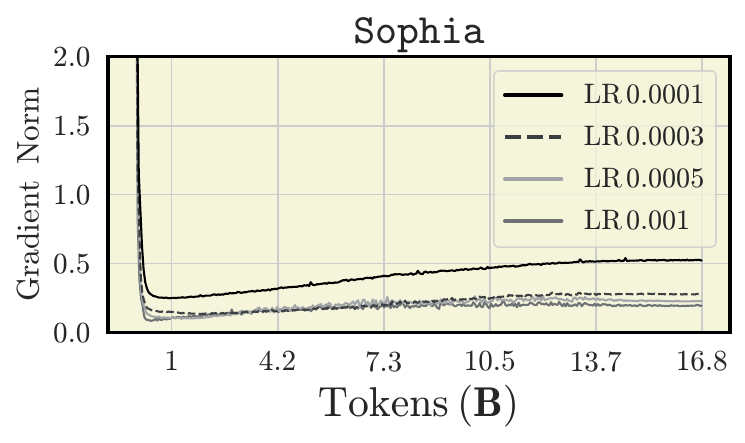}
    \end{minipage}
    \hfill
    \begin{minipage}{0.318\linewidth}
        \includegraphics[width=\linewidth]{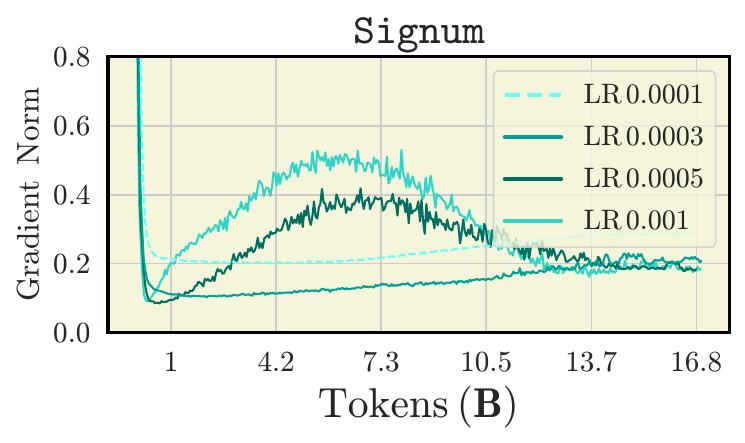}
    \end{minipage}
    \hfill
    \begin{minipage}{0.318\linewidth}
        \includegraphics[width=\linewidth]{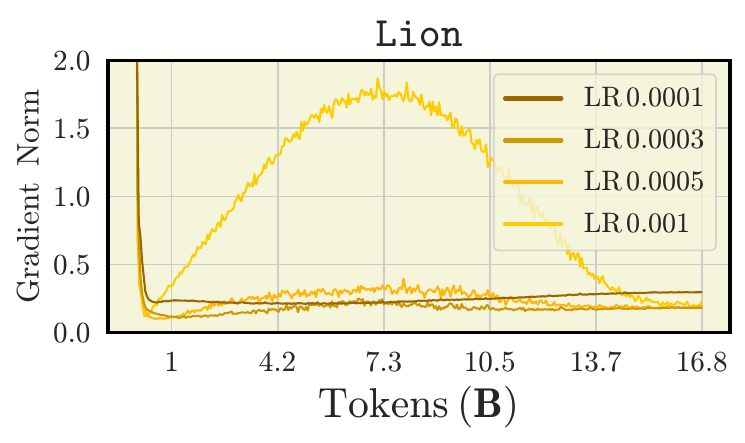}
    \end{minipage}
    \\
    \begin{minipage}{0.318\linewidth}
        \includegraphics[width=\linewidth]{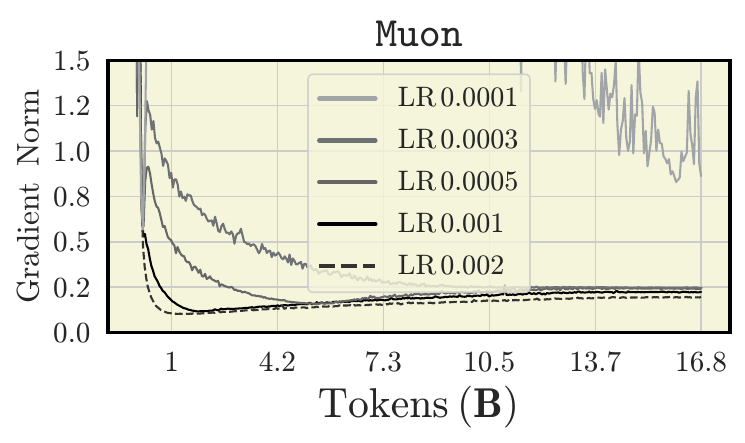}
    \end{minipage}
    \hfill
    \begin{minipage}{0.318\linewidth}
        \includegraphics[width=\linewidth]{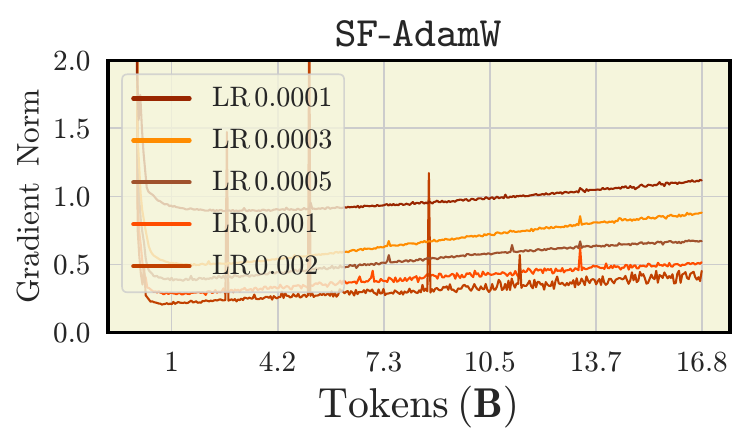}
    \end{minipage}
    \hfill
    \begin{minipage}{0.318\linewidth}
        \includegraphics[width=\linewidth]{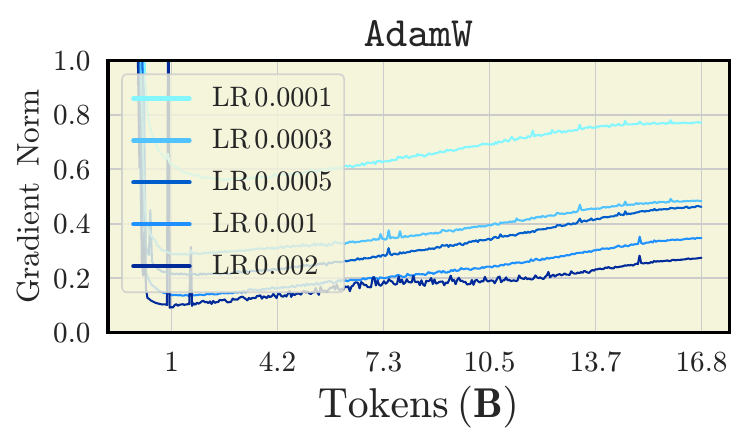}
    \end{minipage}
    \\
    \begin{minipage}{0.318\linewidth}
        \includegraphics[width=\linewidth]{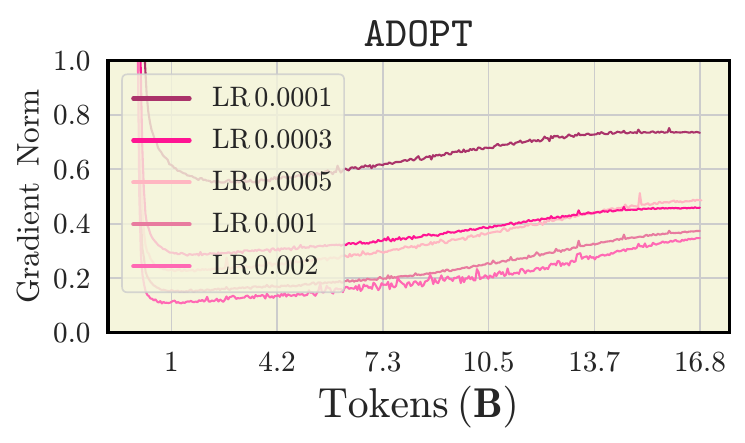}
    \end{minipage}
    \hfill
    \begin{minipage}{0.318\linewidth}
        \includegraphics[width=\linewidth]{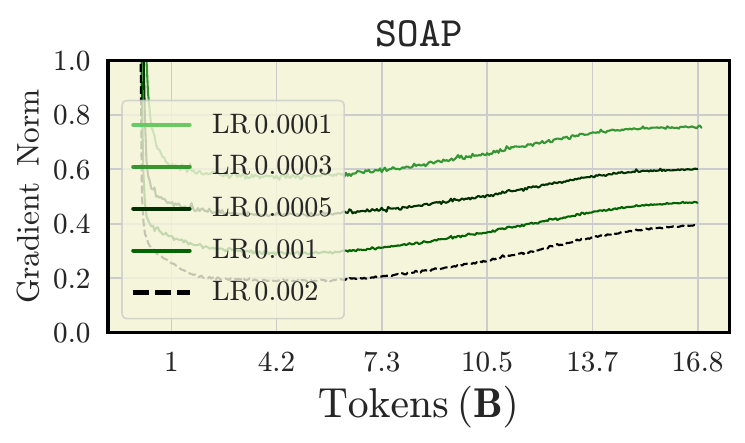}
    \end{minipage}
    \hfill
    \begin{minipage}{0.318\linewidth}
        \includegraphics[width=\linewidth]{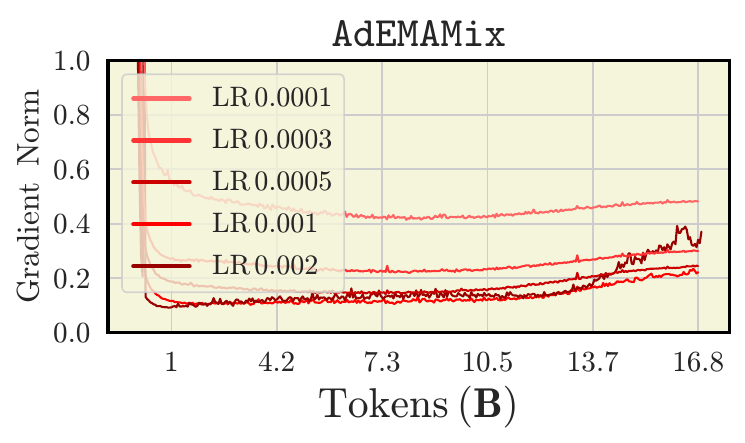}
    \end{minipage}
    \\
    \centering 
    \begin{minipage}{0.318\linewidth}
        \includegraphics[width=\linewidth]{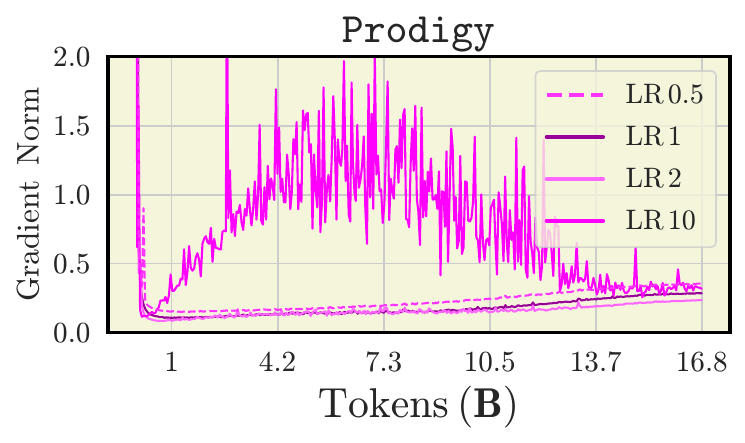}
    \end{minipage}
    \hfill
    \begin{minipage}{0.318\linewidth}
        \includegraphics[width=\linewidth]{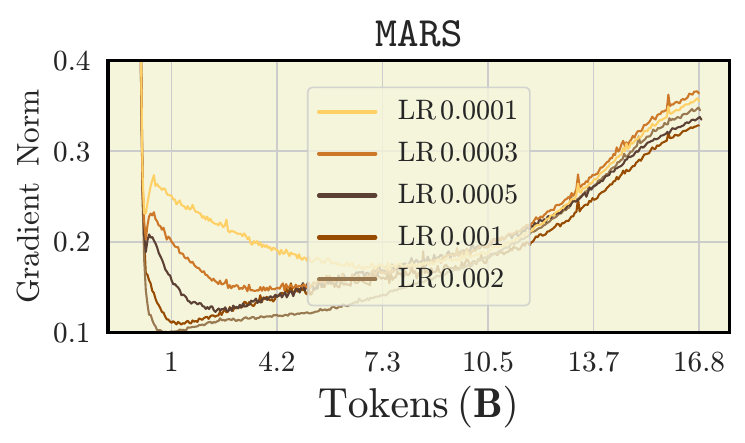}
    \end{minipage}
    \hfill
    \begin{minipage}{0.318\linewidth}
        \includegraphics[width=\linewidth]{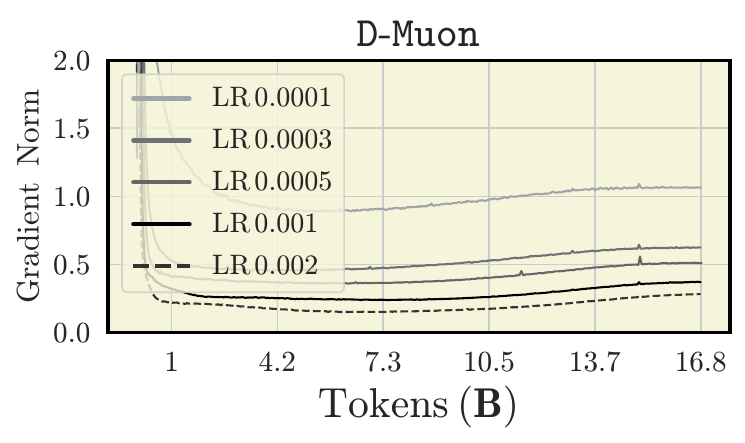}
    \end{minipage}
    \caption{\textbf{Gradient norm patterns for learning rate sweep.}
    In this experiment, we complement the result on the learning rate sweep for optimizers~(\cref{fig:lrsensitivity,fig:ap_lrsensitivity}) by tracking the gradient norms.
    We follow the same setup as for the $\gamma$-sensitivity ablation, varying the learning rates while training $124\mathbf{M}$ language models for $16.8\mathbf{B}$ tokens using a cosine $\gamma$-scheduler with $\gamma_\text{end} = 0.01\times\gamma_{\max}$.
    Except for \texttt{Lion} and \texttt{Signum}, we see that smaller $\gamma_{\max}$ leads to larger magnitude of the gradient norms---unless the learning rate is high enough to nearly lead to divergence, e.g., $\gamma_{\max} = 10$ for \texttt{Prodigy}.
    Interestingly, we connect the ``bump'' shape of the gradient norms for sign-based methods with the fact that $\gamma_{\max} = 0.001$, used for them, is close to the ``critical'' value, an increase of which also leads to divergence---and our experiments with these optimizers on larger models support this, as we were able to decrease $\gamma$ in order to train properly.
    }
    \label{fig:ap_grad_norms_all_lr_sens}
    \vspace{-2.3em}
\end{figure*}

\textbf{Gradient norm patterns.}
We systematically track the evolution of gradient norms across weight decay ($\lambda$), maximum learning rate ($\gamma_{\max}$), and learning rate scheduler sweeps in~\cref{fig:ap_grad_norms_all,fig:ap_grad_norms_all_wd,fig:ap_grad_norms_all_lr_sens}.
This analysis spans all optimizers in our benchmark, providing insight into how these hyperparameters influence gradient magnitude and stability. 
Our goal is to determine whether the gradient norm dynamics correlate with improved convergence and whether these trends are optimizer-specific or general. 
We also investigate whether deviations from expected patterns (e.g., premature flattening or explosive growth) can serve as indicators of suboptimal configuration, potentially informing better tuning heuristics.

Firstly, we study the dynamics of gradient norms while sweeping the learning rate schedulers---see~\cref{fig:ap_grad_norms_all}.
This result complements the one in~\cref{fig:grad-norms-main-part}.
In general, Defazio et al.~\cite{defazio2024optimallineardecaylearning} argue that there exists an interdependence between the learning rate schedule and observed gradient norm patterns, proposing a schedule refinement for optimization algorithms.
The observation that $\gamma$-scheduler can tract the gradient norm pattern and vice versa encourages us to expand experimental observations to optimizers studied in our benchmark.

Prior works~\cite{kosson2024rotationalequilibriumweightdecay,defazio2025gradientsrapidlyincreasenear} study the connection between gradient norm patterns, weight decay, and learning rate.
Kosson et al.~\cite{kosson2024rotationalequilibriumweightdecay} explore how weight decay influences the update behavior of individual neurons in deep neural networks. 
The authors show that weight decay causes the weight norm to reach a stable equilibrium magnitude.
At this equilibrium point, the opposing effects of gradient updates (which increase the norm) and weight decay (which reduce it) cancel each other out.
Importantly, this study highlights the effectiveness of the decoupled weight decay for optimization over $\ell_2$ regularization, noting that the gradient norm varies between neurons or layers for $\ell_2$ regularization which is not the case for decoupled weight decay.
In our experiments, we study how (decoupled) weight decay influences gradient norms---see~\cref{fig:ap_grad_norms_all_wd}.
We use the cosine learning scheduler and the best other hyperparameters found for optimizers, sweeping the weight decay across three critic values: $0$, the standard one of $0.1$, and the ``large'' weight decay of $0.5$.
Basically, these gradient norms were tracked during weight decay ablation and correspond to~\cref{fig:wdablation_main,fig:ap_wdablation}.
We observe that runs without weight decay typically result in gradient norm curves that are more flattened and with a smaller magnitude compared to runs with $\lambda \in \{0.1, 0.5\}$.
Exceptions are sign-based methods, \texttt{Muon}, \texttt{AdEMAMix}, and \texttt{Sophia}.
Using the large weight decay term of $0.5$ results in a dramatic increase in the gradient norms towards the end of the training.
Nevertheless, we present figures for long training runs of $~7\times$ Chinchilla optimal duration for $124\mathbf{M}$ models~(resp. $16.8\mathbf{B}$ tokens and $128\mathbf{k}$ steps)---where runs with $\lambda=0.1$ outperform ones with $\lambda \in \{0, 0.5\}$---we emphasize that the same patterns of the gradient norms are also observed in shorter runs where $\lambda=0.5$ still demonstrates the best performance.

Another key factor influencing the gradient norms is the learning rate.
As with previous ablations on gradient norms~(\cref{fig:ap_grad_norms_all,fig:ap_grad_norms_all_wd}), we follow our benchmarking setup~(\S~\ref{sec:setup}).
During the learning rate sweep~(\cref{fig:lrsensitivity,fig:ap_lrsensitivity}), we track the gradient norms presented in~\cref{fig:ap_grad_norms_all_lr_sens}.
Notably, smaller learning rates result in larger gradient norm magnitudes, with exceptions for sign-based \texttt{Signum} and \texttt{Lion}.
We also observe a dramatic increase in gradient norms for \texttt{Muon} with $\gamma_{\max}=0.0001$, which we attribute to the large difference between learning rates for $1$D and $2$D parameters, the latter typically set around $0.01$~(see the ``Learning rate \texttt{Muon}'' row in~\cref{tab:124m_muon_hyperparams}).
For \texttt{Prodigy} with $\gamma_{\max}=10$ the explosion in gradient norms might be caused by the critical value of the learning rate, which leads to divergence if increased.

\prettybox{
\takeaway{tkw:grad_norms_patterns}(\rom{1}) The WSD scheduler produces stable, flat gradient norm trajectories, contrasting with the increasing norms from cosine and linear schedules.
(\rom{2}) The impact of weight decay is optimizer-specific, with no single value (e.g., $0$ or $0.1$) universally yielding optimal stability; larger decay often increases norms late in training.
(\rom{3}) Smaller learning rates typically lead to larger gradient norms, a trend from which sign-based methods notably deviate.
}

\begin{figure*}[h]
    \centering
    \subfigure[]{
        \includegraphics[width=0.31\linewidth]{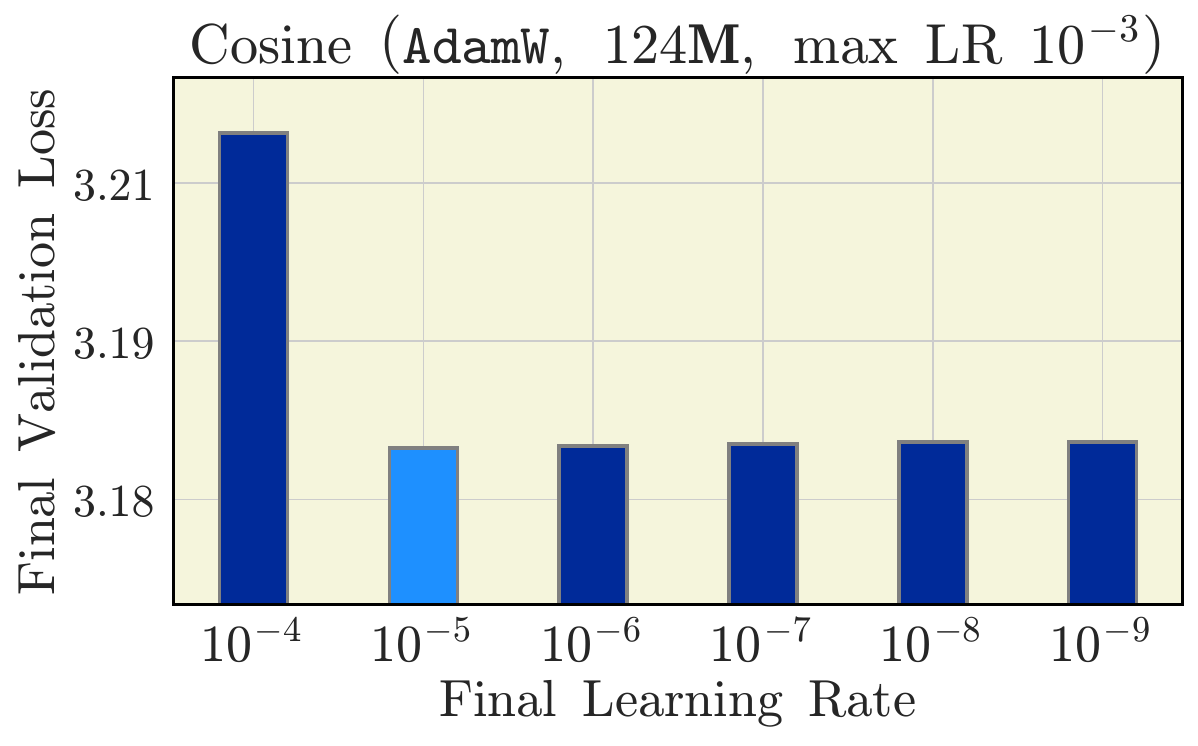}
    }
    \hfill
    \subfigure[]{
        \includegraphics[width=0.31\linewidth]{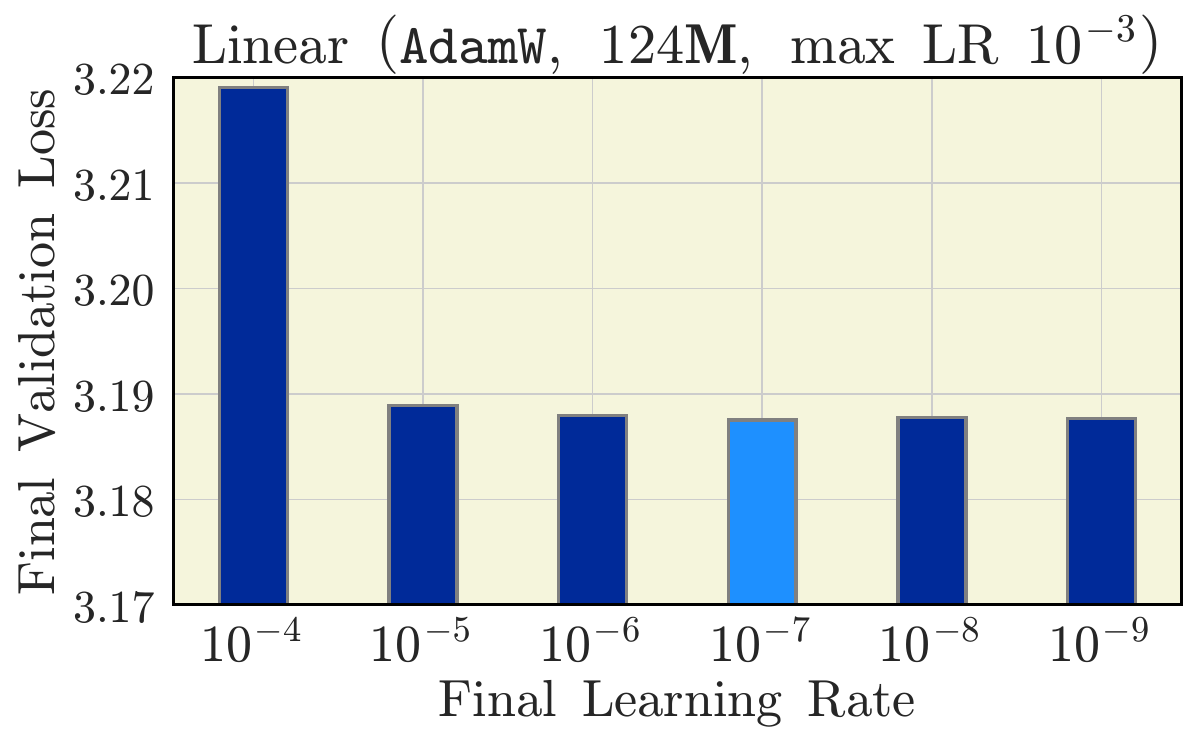}
    }
    \hfill
    \subfigure[]{
        \includegraphics[width=0.31\linewidth]{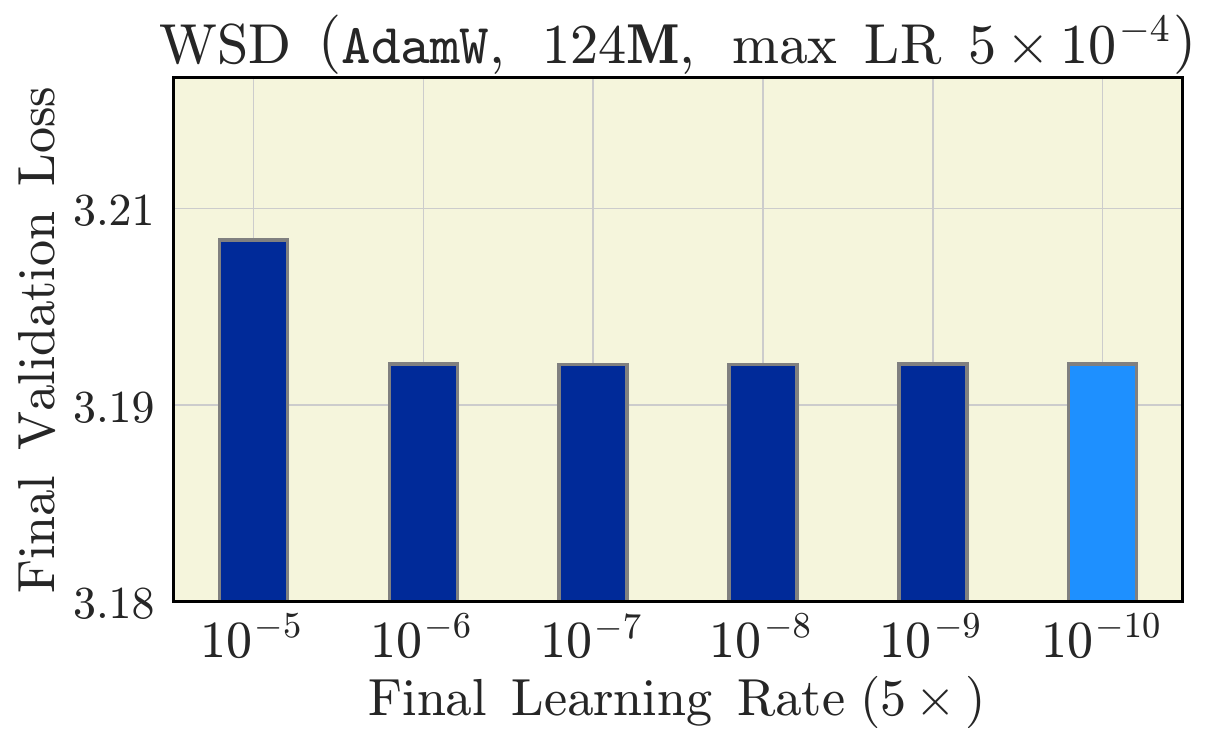}
    }
    \vspace{-2mm}
    \caption{\textbf{Do not decay the learning rate down to $\boldsymbol{10\%}$: ablation on $\mathbf{124M}$ models.}
    We extend our ablation on learning rate decay from~\cref{fig:lrdecay} to studying the impact of the change of model parameters---reducing it from $210\mathbf{M}$ to $124\mathbf{M}$.
    Consistent with our previous results, decaying the learning rate only down to $10\%$ of the maximum results in significantly worse final performance, indicating the need for further decay.
    Notably, for the linear~\textbf{(b)} and WSD~\textbf{(c)} schedulers, the best choice of $\gamma_\text{end}$ differs from that observed at $210\mathbf{M}$.
    For linear, the optimal setting shifts to $\gamma_\text{end}=10^{-4}\times \gamma_{\max}$ (vs. $10^{-3}\times \gamma_{\max}$ at $210\mathbf{M}$), and for WSD to $\gamma_\text{end}=10^{-6}\times \gamma_{\max}$ (vs. $10^{-2}\times \gamma_{\max}$ at $210\mathbf{M}$); see~\cref{fig:lrdecay}~\textbf{(b,c)}.
    Overall, while the differences in final performance beyond $0.1\times \gamma_{\max}$ are relatively small, these results highlight that the optimal $\gamma_\text{end}$ depends on model size, which in turn influences the appropriate learning rate schedule.
    }
    \vspace{-1.5em}
    \label{fig:lrdecay-124m-appendix}
\end{figure*}

\textbf{Learning rate decaying for $124\mathbf{M}$ model.}
Prior ablation studies on $210\mathbf{M}$ models~(\cref{fig:lrdecay}) demonstrated that decaying the learning rate down to $10\%$ of its maximum value underperforms compared to ${0.01, 0.001} \times \gamma_{\max}$.
To generalize this finding, we conduct the same ablation on a smaller $124\mathbf{M}$ model.
As before, we use three $\gamma$-schedulers---cosine, linear, and WSD, utilizing the best hyperparameters for \texttt{AdamW} at this scale, training for $16.8\mathbf{B}$ tokens with the batch size of $256\times512$ tokens.
We $\gamma_{\max}=0.001$---a robust and well-adopted value---and sweep the final learning rate $\gamma_\text{end}$ across $\{10^{-1}, 10^{-2}, 10^{-3}, 10^{-4}, 10^{-5}, 10^{-6}\}\times \gamma_{\max}$.
We present the results of this ablation in~\cref{fig:lrdecay-124m-appendix}.
Recently, the question of the learning rate decaying has been an interesting topic of discussion~\cite{bergsma2025straightzerolinearlydecaying,schaipp2025surprisingagreementconvexoptimization,hägele2024scalinglawscomputeoptimaltraining}, with works focusing on the explanations of the WSD scheduler pointing to the possible impact of decaying $\gamma$ to zero (or very small magnitudes).
Importantly, our ablations for models of two scales---$124\mathbf{M}$ and $210\mathbf{M}$---suggest that the optimal choice of $\gamma_\text{end}$ may depend on the model scale.
For example, $\gamma_\text{end}=0.01\times \gamma_{\max}$ delivers the best performance for $210\mathbf{M}$ model trained with WSD, while for $124\mathbf{M}$ model $\gamma_\text{end}=10^{-6}\times\gamma_{\max}$ takes the lead, which is closer to decaying to zero, as in prior works~\cite{hägele2024scalinglawscomputeoptimaltraining,schaipp2025surprisingagreementconvexoptimization}.
We also highlight that increasing the model size decreases the optimal learning rate for the model, thus the very small values of $\gamma_\text{end}$ might not affect the final performance much, while slowing the training at the latest stage, which is undesirable for modern large-scale pretraining tasks.
Furthermore, we do not conduct the learning rate decaying ablations for different optimizers, utilizing only \texttt{AdamW}.
Thus, we point out that it is possible for $\gamma_\text{end}$ to depend on the optimizer choice as well---this is an interesting branch of the research on optimizers to explore in future work.

\begin{wrapfigure}{r}{0.48\linewidth}
    \centering
    \begin{minipage}{\linewidth}
        \includegraphics[width=\linewidth]{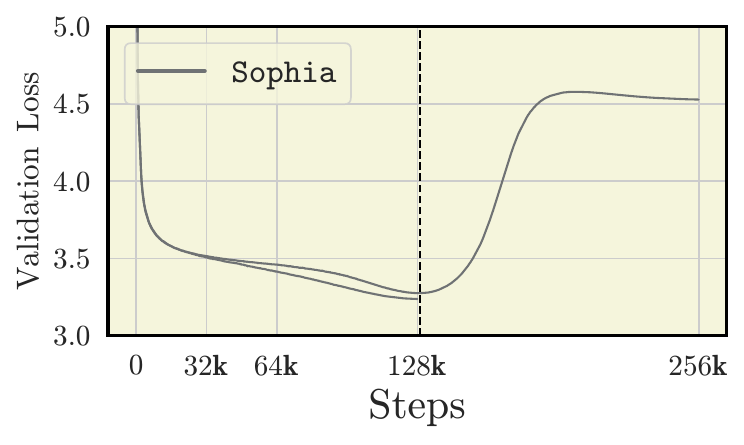}
    \end{minipage}
    \caption{\textbf{\texttt{Sophia} diverges in the large-batch setup, when training for many iterations.}
    In the small-batch setup, we observed that \texttt{Sophia} exhibited convergence issues.
    With batch size $256\times512$, \texttt{Sophia} initially converges reliably across all training durations for $124\mathbf{M}$ models used in our benchmarking. 
    However, when extending training beyond $16.8\mathbf{B}$ tokens, divergence reappears. 
    To clearly visualize so, we present the best stable run ($T=128\mathbf{k}$ steps, $16.8\mathbf{B}$ tokens) with the unstable one ($T=256\mathbf{k}$ steps, $33.6\mathbf{B}$ tokens), using identical hyperparameters. 
    The dashed line marks the iteration $t=129720$ where divergence begins.
    This instability raises serious concerns about the practicality of \texttt{Sophia} for long training runs at scale.
    }
    \label{fig:failofsophia-largebs}
    \vspace{-0.8em}
\end{wrapfigure}

\begin{figure*}[h]
    \vspace{-1em}
    \centering
    \subfigure[Divergence after $\sim2.1\mathbf{B}$ tokens.]{
        \includegraphics[width=0.31\linewidth]{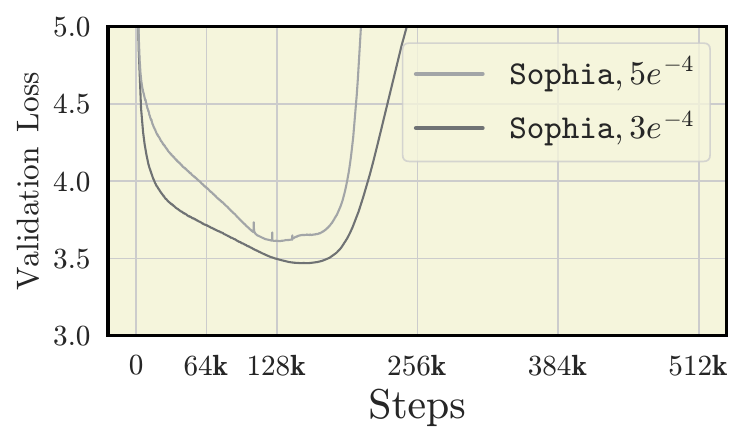}
    }
    \hfill
    \subfigure[Next-token prediction accuracy.]{
        \includegraphics[width=0.31\linewidth]{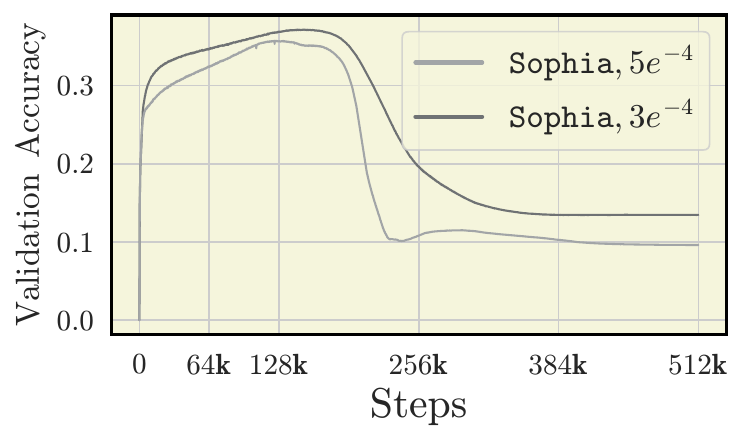}
    }
    \hfill
    \subfigure[Exploded gradient norms.]{
        \includegraphics[width=0.31\linewidth]{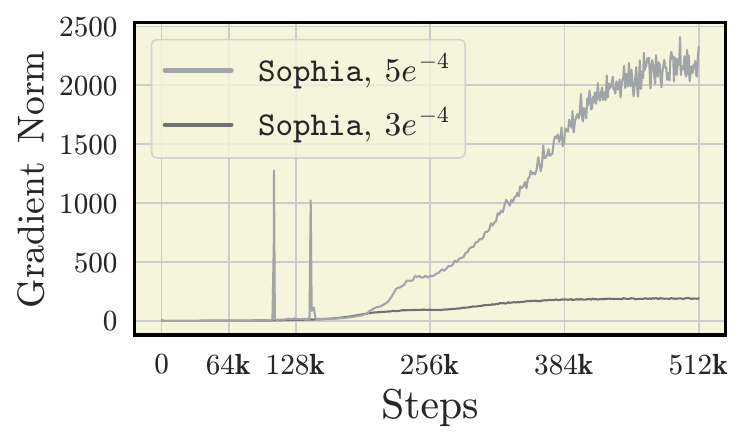}
    }
    \caption{\textbf{\texttt{Sophia} diverges in the small-batch setting even with sufficiently small learning rate.}
    We train $124\mathbf{M}$ Llama models with batch size $32\times512$ tokens for $T\in\{64,128,256,384,512,1024\}\mathbf{k}$ iterations. 
    \texttt{Sophia} diverges with the typical learning rate $\gamma_{\max}=0.001$, and even at smaller values (e.g., $3e^{-4}, 5e^{-4}$) it still fails shortly after $2.1\mathbf{B}$ tokens ($\equiv128\mathbf{k}$ steps). 
    Figures~\textbf{(a–c)} show loss, next-token prediction accuracy, and gradient norms, respectively. 
    For both reported $\gamma_{\max}$ values, divergence occurs at nearly the same iteration (within $10\mathbf{k}$ steps, $\sim164\mathbf{M}$ tokens). 
    We do not attribute this instability to implementation bugs, since \texttt{Sophia} converges on larger MoE models for longer horizons~(\cref{fig:benchmarking-moe-losses}). 
    Whether this instability is related to the Chinchilla optimal horizon remains unclear; however, with a larger batch size ($256\times512$), \texttt{Sophia} again fails once training exceeds $16.8\mathbf{B}$ tokens (see~\cref{fig:failofsophia-largebs}).
    }
    \label{fig:failofsophia}
    \vspace{-1em}
\end{figure*}

\textbf{Fail of \texttt{Sophia}.}
Another striking effect we observed throughout our benchmarking experiments is the convergence issues of the \texttt{Sophia} optimizer.
In the main text (see \cref{tkw:32bs_ranking}), we reported that ``\texttt{Sophia} diverges in the small-batch setting when trained beyond the Chinchilla optimal horizon, even with sufficiently small learning rates.''
Later, we also noted that in the large-batch regime ``\texttt{Sophia} exhibits convergence issues when extending the training run, diverging shortly after $130\mathbf{k}$ steps.''  
These phenomena are particularly puzzling, since \texttt{Sophia} does converge in long runs of $336\mathbf{k}$ steps on MoE models.
\cref{fig:failofsophia} demonstrates loss curves of $124\mathbf{M}$ Llama model trained with a small batch size of $32\times512$ tokens and using the cosine $\gamma$-scheduler.
Initially, we used $\gamma_{\max}=0.001$, which proved too large for this setup, so we switched to $\gamma_{\max}\in\{1e^{-4}, 3e^{-4}, 5e^{-4}\}$.
For runs up to $T=64\mathbf{k}$ steps, training converged properly. However, increasing the number of steps beyond this point led to divergence (see~\cref{fig:failofsophia}~\textbf{(a)}). Interestingly, the divergence onset occurred at almost the same iteration for both $3e^{-4}$ and $5e^{-4}$ learning rate values.  
For reference, training with $T=128\mathbf{k}$ steps in the small-batch setup results in $\sim2.1\mathbf{B}$ tokens, while the Chinchilla optimal horizon for this model is about $2.5\mathbf{B}$.
Thus, \texttt{Sophia} fails to converge with such a small batch size even before reaching the optimal horizon.
When switching to a larger batch size of $256\times512$, we initially observed stable convergence across training durations from $1\mathbf{B}$ to $16.8\mathbf{B}$ tokens (see~\cref{fig:benchmark-124}~\textbf{(b)}).
The same held true for an even larger batch size of $512\times512$ tokens, where \texttt{Sophia} converged for $64\mathbf{k}$ iterations, i.e., $16.8\mathbf{B}$ tokens (see~\cref{fig:scalebs_scaletokens}, \textit{left}).  
However, doubling the training steps with the $256\times512$ batch size again led to divergence (see~\cref{fig:failofsophia-largebs} and \cref{fig:scalebs_scaletokens}, \textit{right}).
Using the same hyperparameters that worked well for $16.8\mathbf{B}$ tokens, we extended training to $33.6\mathbf{B}$ tokens ($\equiv256\mathbf{k}$ iterations).
Strangely, shortly after reaching $16.8\mathbf{B}$ tokens, \texttt{Sophia} diverged, with the failure occurring precisely at $t=129720$ (marked by the dashed line).
We do not attribute these issues to implementation bugs, since \texttt{Sophia} converges in much longer runs ($336\mathbf{k}$ steps) with larger $520\mathbf{M}$ models (see~\cref{fig:benchmarking-moe-losses}).
Instead, we caution practitioners against relying on \texttt{Sophia} in its current form and emphasize that there remains substantial room for improvement. 
We also note that previous benchmarking work~\cite{kaddour2023traingainrevisitingefficient} evaluated \texttt{Sophia} only on BERT~\cite{devlin2019bertpretrainingdeepbidirectional} and T$5$~\cite{raffel2023exploringlimitstransferlearning} pretraining tasks (encoder-only and encoder-decoder architectures, respectively).

\prettybox{
\takeaway{tkw:failofsophia}(\rom{1}) \texttt{Sophia} diverges in the small-batch setting, even with sufficiently small learning rate.
(\rom{2}) When training with an increased batch size, \texttt{Sophia} starts to diverge after exceeding some limit in iterations---nearly $7\times$ Chinchilla optimal horizon in our experiments.
}

\begin{figure*}[h]
    \vspace{-1em}
    \centering
    \subfigure[Disabling clipping causes instability to \texttt{SF-AdamW}.]{
        \includegraphics[width=0.476\linewidth]{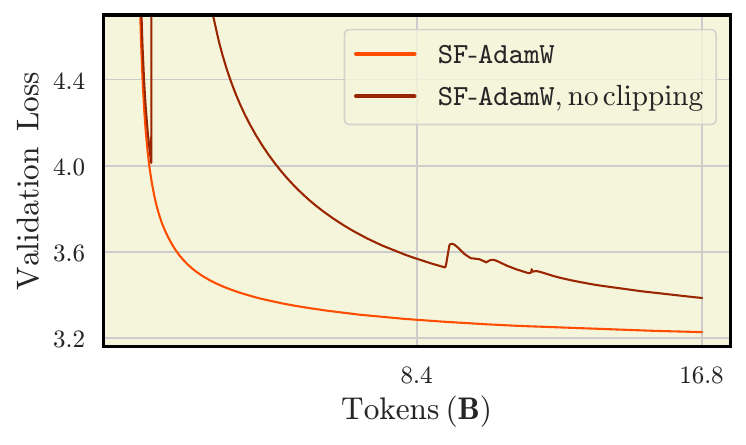}
    }
    \hfill
    \subfigure[Reducing $\gamma$ helps, but the clipped baseline is better.]{
        \includegraphics[width=0.4865\linewidth]{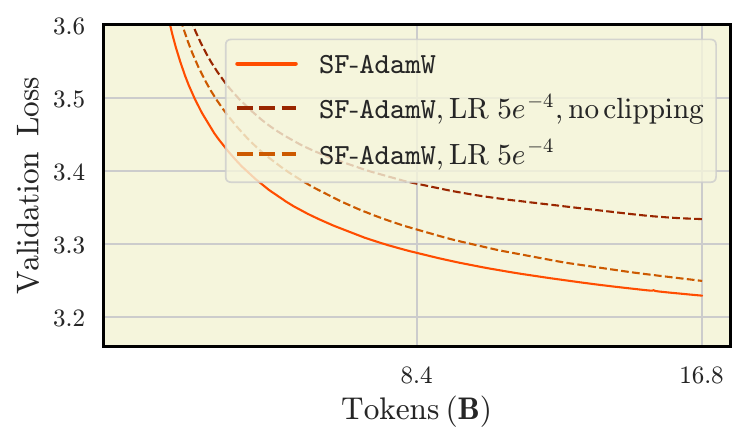}
    }
    \caption{\textbf{Clipping is significant for \texttt{Schedule-Free}}.
    Contrary to the claims of Defazio et al.~\cite{defazio2024roadscheduled}, we find that gradient clipping remains a critical hyperparameter for \texttt{SF-AdamW}.
    As shown in~\textbf{(a)}, disabling clipping causes severe training instabilities. 
    To mitigate these undesired loss dynamics, we reduced the learning rate from $0.001$ to $0.0005$, which stabilized training~\textbf{(b)}. 
    However, even under this adjustment, the clipped runs still outperform those without clipping.
    }
    \label{fig:sfclipping}
\end{figure*}

\textbf{Clipping \& \texttt{SF-AdamW}.}
Defazio et al.~\cite{defazio2024roadscheduled}, when introducing the schedule-free concept, emphasized that gradient clipping should be disabled for \texttt{SF-AdamW}. 
Motivated by this claim, we paid particular attention to clipping during tuning. 
Following our setup~(\S~\ref{sec:setup}), we trained $124\mathbf{M}$ models with a batch size of $256\times512$ tokens for up to $128\mathbf{k}$ steps ($\equiv16.8\mathbf{B}$ tokens). 
While sweeping the main hyperparameters of \texttt{SF-AdamW}---($\beta_1$, $\beta_2$), $\gamma$, $\lambda$, $T_\text{warmup}$---we also varied the gradient clipping threshold across $\{0.5, 1\}$ and tested runs without clipping, as suggested in the original paper.
Our results, summarized in~\cref{fig:sfclipping}, show a clear discrepancy with prior claims. 
Disabling clipping consistently produced unstable training, with highly spiky loss curves (\cref{fig:sfclipping}\textbf{a}). 
To mitigate this, we reduced the learning rate from $0.001$ to $0.0005$, which largely stabilized the runs (\cref{fig:sfclipping}\textbf{b}). 
However, even under this adjustment, the best clipped run---with the clipping threshold of $0.5$---still outperformed the no-clipping alternative. 
Thus, contrary to Defazio et al.~\cite{defazio2024roadscheduled}, we find gradient clipping to be a critical hyperparameter for the stability of \texttt{SF-AdamW}.

\prettybox{
\takeaway{tkw:sfclipping}Gradient clipping is crucial for stability of \texttt{Schedule-Free AdamW}. 
}

\begin{wrapfigure}{l}{0.48\linewidth}
    \vspace{-2em}
    \centering
    \begin{minipage}{\linewidth}
        \includegraphics[width=\linewidth]{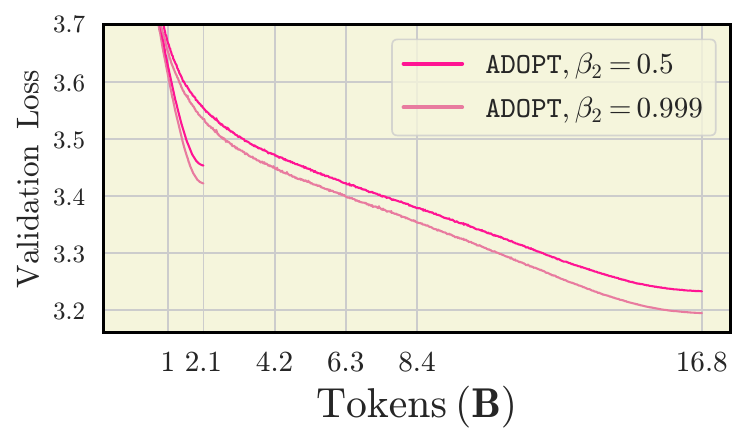}
    \end{minipage}
    \vspace{-1.2em}
    \caption{\textbf{\texttt{ADOPT} still needs $\beta_2$.}
    One of the main theoretical claims of Taniguchi et al.~\cite{taniguchi2024adoptmodifiedadamconverge}---that \texttt{ADOPT} converges with any $\beta_2$.
    The authors verify those on a toy problem motivated by Reddi et al.~\cite{reddi2019convergenceadam}.
    However, in LLM training, the choice of $\beta_2$ still matters significantly.
    Our results demonstrate that, despite theoretical guarantees, performance strongly depends on tuning $\beta_2$ in practice.
    }
    \label{fig:adopt-beta2}
    \vspace{-0.45em}
\end{wrapfigure}

\begin{figure}[h]
    \vspace{-1.1em}
    \centering
    \subfigure[Minor change in $\beta_2$ leads to divergence.]{
        \includegraphics[width=0.48\linewidth]{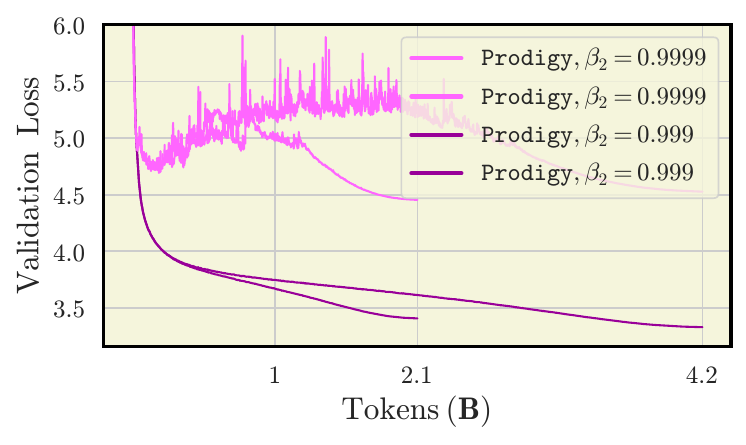}
    }
    \hfill
    \subfigure[Bias correction resolves issues with divergence.]{
        \includegraphics[width=0.48\linewidth]{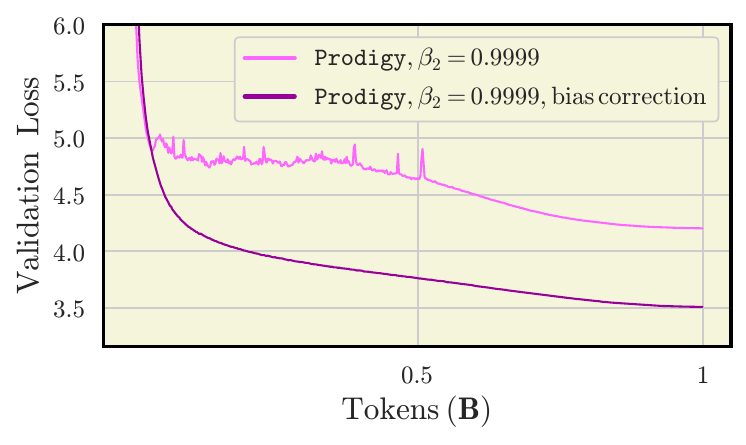}
    }
    \caption{\textbf{\texttt{Prodigy} is sensitive to beta parameters in the small-batch setting}.
    In this experiment, we follow our setup~(\S~\ref{sec:setup}) with a small batch size of $32\times512$ tokens, training $124\mathbf{M}$ models with the best hyperparameters while sweeping $\beta_2$.
    Although $\beta_2=0.999$ yields the best results in this setting~(see~\cref{tab:124m_prodigy_hyperparams}), even a slight change to $\beta_2=0.9999$ causes divergence.
    This occurs because ($\beta_1$, $\beta_2$) directly affect the internal statistics $r_t$, $\ss_t$, which determine the optimizer’s effective learning rate.
    As shown in~\textbf{(b)}, enabling bias correction (see \texttt{line 7} of~\cref{alg:prodigy}) effectively resolves this instability.
    }
    \vspace{-1em}
    \label{fig:prodigy_betas}
\end{figure}

\begin{figure}[h]
    \begin{minipage}{\linewidth}
        \centering
        \subfigure[$\beta_2$ sweep with fixed $\beta_1=0.9$.]{
            \includegraphics[width=0.45\linewidth]{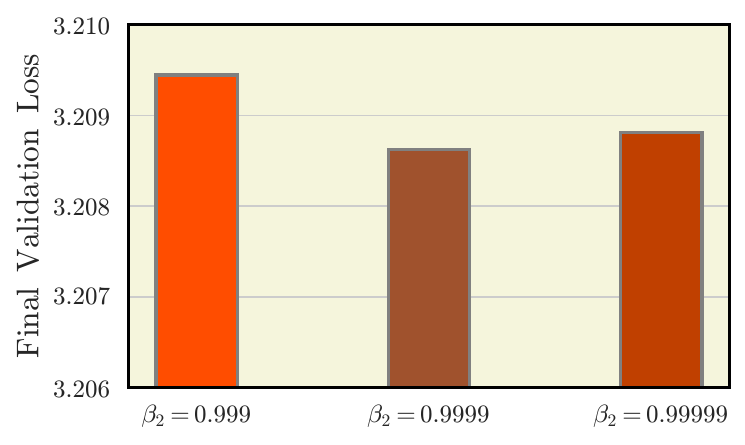}
        }
        \subfigure[Comparison with Defazio~\cite{defazio2024roadscheduled} and H\"agele~\cite{hägele2024scalinglawscomputeoptimaltraining}.]{
            \includegraphics[width=0.45\linewidth]{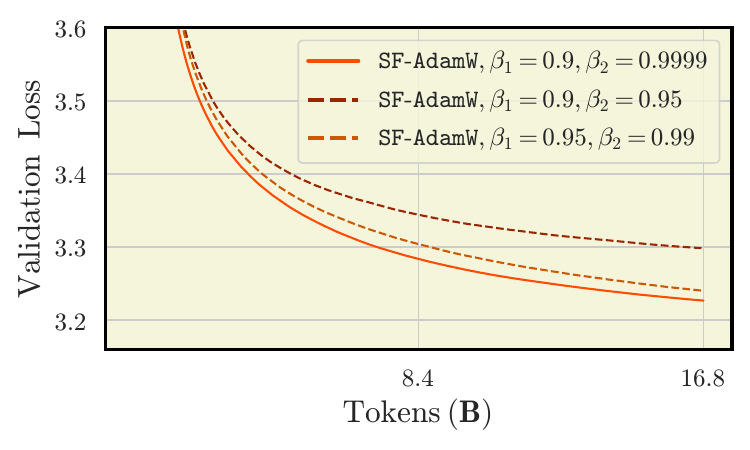}
        }
        \caption{\textbf{Impact of beta parameters on \texttt{Schedule-Free}}.
        We elaborate further on the question of the sensitivity of \texttt{SF-AdamW} to $\beta_2$.
        For language modeling, Defazio et al.~\cite{defazio2024roadscheduled} initially suggested using default ($\beta_1=0.9$, $\beta_2=0.95$).
        Then, H\"agele et al.~\cite{hägele2024scalinglawscomputeoptimaltraining} revisited hyperparameter tuning of the schedule-free optimizer, proposing to apply ($\beta_1=0.95$, $\beta_2=0.99$), which improved the performance a lot.
        Based on our tuning, we claim that ($\beta_1=0.9$, $\beta_2=0.9999$) achieves the best performance at this scale---see \textbf{(b)}.
        In addition, we fix $\beta_1=0.9$ and report the result with sweep of $\beta_2\in\{0.999, 0.9999, 0.99999\}$, showing that the large and unconventional value of $\beta_2=0.9999$ is indeed the best in schedule-free runs.
        We also notice that \texttt{SF-AdamW} requires a slightly larger optimal $\beta_2$, compared to all other optimizers.
        }
        \label{fig:sf_betas}
        \vspace{-2.3em}
    \end{minipage}
\end{figure}

\textbf{Betas sensitivity.}
The impact of the beta parameters on optimizers---especially in \texttt{Adam}-like methods---has been studied both theoretically~\cite{reddi2019convergenceadam,défossez2022simpleconvergenceproofadam,NEURIPS2018_90365351} and empirically~\cite{pagliardini2024ademamixoptimizerbetterfaster,marek2025smallbatchsizetraining,busbridge2023scaleema}. 
However, many large-scale works in industry~\cite{deepseekai2024deepseekv3technicalreport,brown2020languagemodelsfewshotlearners,touvron2023llamaopenefficientfoundation,olmo20242olmo2furious,jaghouar2024intellect1technicalreport} either do not tune the betas at all or simply adopt conventional defaults ($\beta_1=0.9$, $\beta_2=0.95$).
Earlier in this manuscript~(\cref{tkw:retuning_betas}), we argued that betas should be tuned in tandem with training duration---a conclusion supported by extensive ablations and hyperparameter sweeps.
Here, we demonstrate the most striking effects of tuning beta parameters, with a particular focus on $\beta_2$.
Our ablation focuses on ``parameter-free'' methods such as \texttt{Prodigy}~(\cref{fig:prodigy_betas}), \texttt{SF-AdamW}~(\cref{fig:sf_betas}), and the \texttt{Adam}-like optimizer \texttt{ADOPT}~(\cref{fig:adopt-beta2}).

We highlight that: (\rom{1}) despite the theoretical convergence guarantees of \texttt{ADOPT} for any $\beta_2$, in practice the performance gap between the best and a poorly chosen $\beta_2$ remains substantial; (\rom{2}) when the batch size is small ($32\times512$ tokens), \texttt{Prodigy} is very sensitive to $\beta_2$, even diverging when changing it from $0.999$ to $0.9999$, however, applying the bias correction---see \texttt{line 7} of~\cref{alg:prodigy}---fixes this issue; (\rom{3}) prior works~\cite{hägele2024scalinglawscomputeoptimaltraining,song2025riverunderstandingbenefitschedulefree} question a sensitivity of \texttt{SF-AdamW} to $\beta_2$, which also studied by Defazio et al.~\cite{defazio2024roadscheduled} on image classification tasks, we confirm that changes in betas, especially $\beta_2$, highly affects the overall performance, in~\cref{fig:sf_betas}~\textbf{(b)} we compare our best found ($\beta_1=0.9$, $\beta_2=0.9999$) hyperparameters with default ($\beta_1=0.9$, $\beta_2=0.95$) used by Defazio et al.~\cite{defazio2024roadscheduled}, and ($\beta_1=0.95$, $\beta_2=0.99$) noticed by H\"agele et al.~\cite{hägele2024scalinglawscomputeoptimaltraining}.

\begin{center}
\begin{minipage}{\textwidth}
\prettybox{
\takeaway{tkw:betas_sensitivity}(\rom{1}) \texttt{Prodigy} diverges with a minor change in $\beta_2$, when the batch size is small.
Using bias correction should resolve this issue.
(\rom{2}) \texttt{SF-AdamW} is sensitive to ($\beta_1$, $\beta_2$); we find that typically large $\beta_2$ values (e.g., $0.9999$) are beneficial for schedule-free runs.
(\rom{3}) Despite the established convergence theory for any $\beta_2$, \texttt{ADOPT} still requires careful tuning of this hyperparameter.
}
\end{minipage}
\end{center}

\begin{figure}[h]
    \vspace{-1.5em}
    \centering       
        \includegraphics[width=.4\linewidth]{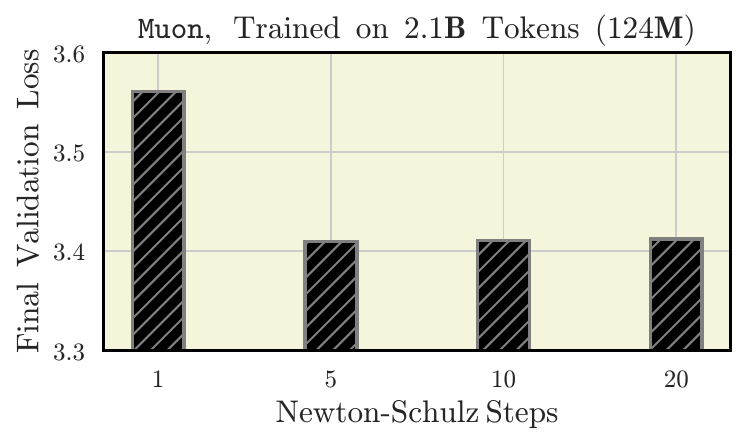}
    \caption{\textbf{\texttt{Muon}'s dependence on the number of Newton-Schulz iterations.}
    We perform a short ablation targeting the final loss of \texttt{Muon}~(\cref{alg:muon}) by varying the number of Newton-Schulz iterations. 
    Training is done for $16\mathbf{k}$ steps with a batch size of $256\times512$ tokens, sweeping $T_{\mathrm{NS}} \in \{1, 5, 10, 20\}$. 
    We find that increasing $T_{\mathrm{NS}}$ beyond $5$ does not improve performance, while unnecessarily increasing wall-clock time.
    }
    \label{fig:muon_ns}
    \vspace{-1.5em}
\end{figure}

\textbf{\texttt{Muon}'s Newton-Schulz iterations.}
We briefly study the impact of Newton-Schulz iterations on the \texttt{Muon} optimizer, focusing on the first version of \texttt{Muon} with weight decay applied only to $1$D parameters.
Recent research~\cite{ahn2025dioncommunicationefficientoptimizerlarge,amsel2025polarexpressoptimalmatrix,grishina2025acceleratingnewtonschulziterationorthogonalization} has extensively explored the Newton-Schulz orthogonalization procedure, examining its impact on the wall-clock speed, communication efficiency on many GPUs, and numerical precision formats. 
Additionally, the theoretical implications of orthogonalization procedures on optimizer convergence have been investigated in~\cite{kovalev2025understandinggradientorthogonalizationdeep,riabinin2025gluon}.
In this ablation, we focus solely on the final loss performance of \texttt{Muon}, setting aside other considerations such as computational efficiency or wall-clock time.
Following the tuning setup~(\S~\ref{sec:setup}) for smaller $124\mathbf{M}$ parameter models with batch size of $256\times512$ tokens, we train for $2.1\mathbf{B}$ tokens ($\equiv16\mathbf{k}$ steps), slightly below the Chinchilla optimal training horizon.
Once the main hyperparameters of \texttt{Muon} are properly tuned, we sweep the number of Newton-Schulz iterations $T_{\mathrm{NS}} \in \{1, 5, 10, 20\}$.
The default setting for both \texttt{Muon}~(\cref{alg:muon}) and \texttt{D-Muon}~\cite{liu2025muonscalablellmtraining} is $T_{\mathrm{NS}}=5$.
Our results indicate that $T_{\mathrm{NS}}\in\{5, 10\}$, and $20$ yield comparable performance, with $T_{\mathrm{NS}}=5$ slightly outperforming the others.
However, setting $T_{\mathrm{NS}}=1$ significantly degrades performance.
These findings are summarized in~\cref{fig:muon_ns}.
Importantly, we always use Nesterov momentum, when running \texttt{Muon}-like methods.

\textbf{\texttt{Signum} configurations.}
We consider the \texttt{Signum} optimizer~(\cref{alg:signumtorch}), which, perhaps unexpectedly, demonstrates strong performance at a small scale and competes effectively with \texttt{AdamW} when batch sizes are large~(\cref{fig:scalebs_scaletokens}~(\textit{left})).
A key factor contributing to this performance is the decoupled weight decay.
However, a fixed weight decay alone does not fully account for \texttt{Signum}’s efficiency.
Another important ingredient is the momentum mechanism.
In this ablation, we study two momentum configurations: Nesterov momentum~\cite{Nesterov1983AMF} (our default, in \cref{alg:signumtorch}) and dampening, which is commonly used in PyTorch’s implementation of \texttt{SGD}.
We also compare both with the ``plain'' \texttt{Signum}, which uses conventional momentum without Nesterov.
To give a better understanding of these concepts, we provide a brief algorithmic description in~\cref{sec:ap_signbased} and below:

1. Dampening update:
\begin{align*}
    \begin{cases}
         \mm_t \leftarrow \beta \mm_{t-1} + (1 - \tau) \bg_t, \\
         \xx_{t+1} \leftarrow \xx_t - \gamma_t \left(\mathtt{sign}\left(\mm_t\right) + \lambda \xx_t\right).
    \end{cases}
\end{align*}

2. The ``plain'' update of \texttt{Signum} without Nesterov momentum:
\begin{align*}
    \begin{cases}
         \mm_t \leftarrow \beta \mm_{t-1} + \bg_t, \\
         \xx_{t+1} \leftarrow \xx_t - \gamma_t \left(\mathtt{sign}\left(\mm_t\right) + \lambda \xx_t\right).
    \end{cases}
\end{align*}

\begin{wrapfigure}{l}{0.48\linewidth}
    \vspace{-2.1em}
    \centering
    \begin{minipage}{\linewidth}
        \includegraphics[width=\linewidth]{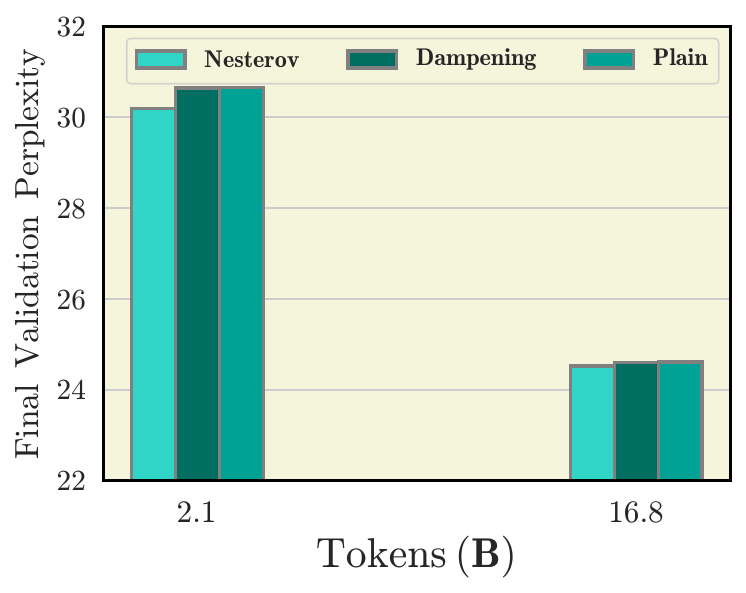}
    \end{minipage}
    \vspace{-1em}
    \caption{\textbf{Comparison of different update rules for \texttt{Signum}.}
    We evaluate three variants of the \texttt{Signum} update: Nesterov (our default), dampening---which resembles an EMA of $\mm_t$ when the dampening parameter $\tau$ equals the momentum $\beta$---and the ``plain'' \texttt{Signum} without Nesterov momentum or dampening. 
    Validation perplexity is reported for two training horizons in  ($256\times512$) batch size setting.
    The Nesterov variant corresponds to the runs included in our main benchmarking results~(\cref{fig:benchmark-124,fig:benchmarking-124m-losses}). 
    While Nesterov style momentum consistently achieves the best performance, the relative perplexity gap compared to the other variants decreases as the training horizon increases.
    }
    \label{fig:signum-configurations}
    \vspace{-2em}
\end{wrapfigure}

That is, the dampening update rule with $\tau = \beta$ resembles the basic EMA we used to see in methods such as \texttt{AdamW}---\texttt{line 5} of~\cref{alg:adamw}.
And the ``plain'' \texttt{Signum} follows the conventional momentum style of \texttt{SGD} used in its PyTorch implementation~\footnote{\href{https://docs.pytorch.org/docs/stable/generated/torch.optim.SGD.html}{torch.optim.SGD}}.

The results of the comparison are shown in~\cref{fig:signum-configurations}.
We ran three variations of the method for $2.1\mathbf{B}$ and $16.8\mathbf{B}$ in the ``large'' batch setup, and reported the final perplexity (PPL).
For the Nesterov momentum version (our default), we use $\beta=0.95$ found through careful tuning.
For the damping version, we found that $\tau=0.9$ is the best, i.e. the explicit momentum update at each iteration results in $\mm_t \leftarrow 0.95 \cdot \mm_{t-1} + 0.1 \cdot \bg_t$; we found this configuration to be slightly better than $\tau=\beta=0.95$.
The same $\beta=0.95$ is used in the ``plain'' \texttt{Signum} configuration.
In all cases, the method with Nesterov momentum leads with a significant margin (for LLM pre-training) of $\sim 0.45$ PPL for $2.1\mathbf{B}$ tokens run and $\sim 0.11$ PPL for long $16.8\mathbf{B}$ tokens training over dampening and plain \texttt{Signum} variations.
Interestingly, these margins vanish with the increased training horizon.
We highlight the importance of Nesterov momentum for \texttt{Signum} runs in~\cref{tkw:signum_configurations}.
We also notice that Nesterov momentum slowdowns training, but not significantly, as our wall-clock time ablation reveals that \texttt{Signum}, with Nesterov momentum, is still the fastest method in various scenarios.

\prettybox{
\takeaway{tkw:signum_configurations}\texttt{Signum} with Nesterov momentum~(our PyTorch implementation) consistently outperforms both the dampening variant~(EMA-like) and the basic version without Nesterov.
}
\vspace{0.5em}

\begin{figure*}[h]
    \vspace{-1.2em}
    \centering
    \subfigure[\texttt{MARS} family of methods.]{
        \includegraphics[width=0.316\linewidth]{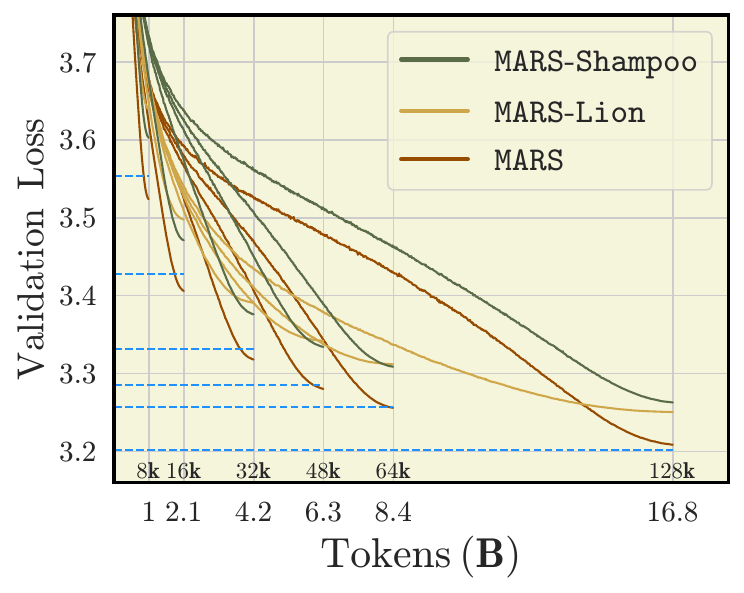}
    }
    \hfill
    \subfigure[Sensitivity to $\gamma$-schedulers.]{
        \includegraphics[width=0.316\linewidth]{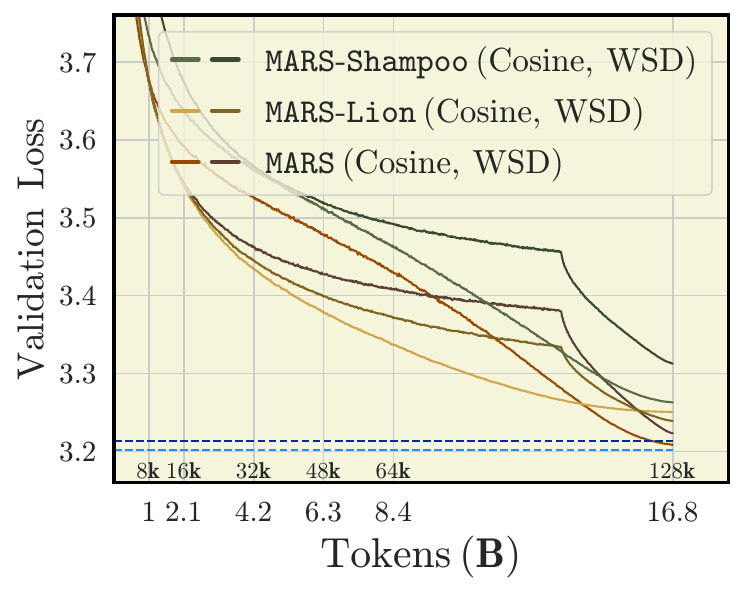}
    }
    \hfill
    \subfigure[Warmup ablation.]{
        \includegraphics[width=0.316\linewidth]{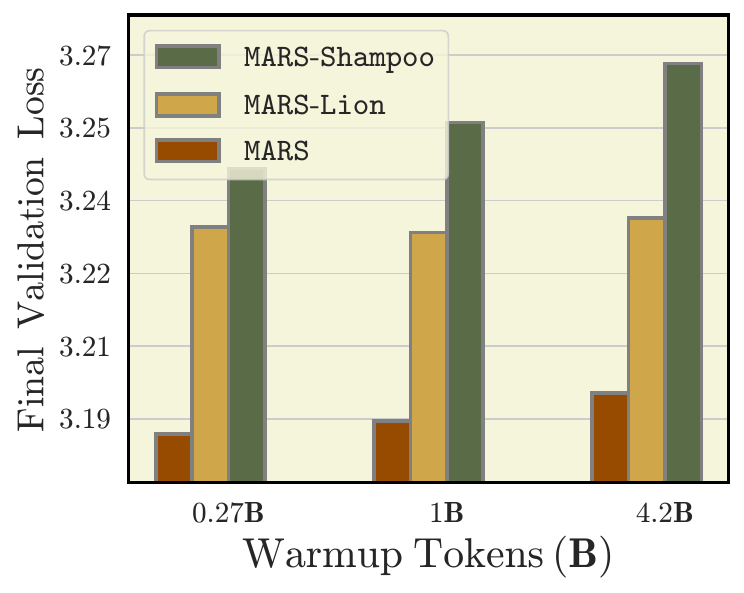}
    }
    \caption{\textbf{\texttt{MARS} family of optimizers.}
    We study three \texttt{MARS}-based algorithms: \texttt{MARS-AdamW} (just \texttt{MARS} in our work), \texttt{MARS-Lion}, and \texttt{MARS-Shampoo}.
    In this ablation, our goal is to complement our \texttt{MARS} runs with experiments for other similar methods, and support findings for these optimizer with our previous experience in tuning \texttt{Lion}.
    We replicate the setup from \S~\ref{sec:setup} and train with the batch size of $256\times512$ for the same training durations as in \cref{sec:smallscalebench}.
    In \textbf{(a)}, we show that, indeed, \texttt{MARS-AdamW} outperforms other alike methods, as reported in~\cite{yuan2024marsunleashingpowervariance} regarding the \texttt{MARS-Lion} optimizer.
    Interestingly, in~\textbf{(b)}, we show that the choice of $\gamma$-scheduler for \texttt{MARS}-based methods also depends on optimizer, as such, WSD runs of \texttt{MARS-Lion} outperform itself with cosine.
    Dashed blue and dark blue lines correspond to the \texttt{AdamW} baseline with cosine and WSD schedulers, respectively.
    Furthermore, in the same way as \texttt{Lion} benefits from larger warmup (\cref{fig:warmup}), \texttt{MARS-Lion} also improves with $8\mathbf{k}$ steps~($\equiv1\mathbf{B}$ tokens) warmup, however, this improvement is not as dramatic~\textbf{(c)}.
    }
    \label{fig:mars_types}
\end{figure*}

\textbf{\texttt{MARS} types.}
In addition to the \texttt{MARS} optimizer that leverages~\cref{alg:marsadamw} to optimize $2$D parameters, and \texttt{AdamW} to optimize $1$D parameters and $\mathtt{lm\_head}$, we also study \texttt{MARS-Lion} and \texttt{MARS-Shampoo} methods---\cref{alg:marslion,alg:marsshampoo} respectively.
Before delving into the experimental details, we note that it is possible to use \texttt{MARS}-like methods for all parameters of LLM, however, this would be inefficient and in the original codebase\footnote{\href{https://github.com/AGI-Arena/MARS}{https://github.com/AGI-Arena/MARS}}, the default choice is to optimize all $1$D parameters with \texttt{AdamW}.
Therefore, we do the same in our experiments.
For this ablation, we utilize $124\mathbf{M}$ model and train for $T\in\{8, 16, 32, 48, 64, 128\}\mathbf{k}$ with batch size of $256\times512$ (we report plots only for this batch setting), varying $\gamma$-schedulers and $T_\text{warmup}$. 
We observe similar patterns regarding the impact of weight decay on these methods---for the majority of the training the loss curves with $\lambda=0$ look ``convex'' and lie below the curves corresponding to $\lambda=0.1$, but then runs with the non-zero weight decay take the lead. 
Regarding tuning \texttt{MARS-Lion} and \texttt{MARS-Shampoo}, we found interesting observations related to our previous experience in hyperparameter tuning.
\texttt{MARS-Lion}, despite optimizing only $1$D parameters with \texttt{Lion}, is also sensitive to warmup, as the latter method, and benefits from longer warmup durations---see \cref{fig:mars_types}~\textbf{(c)}.
Similarly to \texttt{Lion}, \texttt{MARS-Lion} also prefers the WSD scheduler (\cref{fig:mars_types}~\textbf{(b)}) that outperforms the corresponding runs with the cosine baseline.
Notably, the best ($\beta_1$, $\beta_2$) parameters of \texttt{MARS-Lion} coincide with those found for \texttt{Lion} in \cref{tab:124m_lion_hyperparams} and in~\cite{chen2023symbolicdiscoveryoptimizationalgorithms}.
Of all the \texttt{MARS} versions, \texttt{MARS-Shampoo} performs the worst.
We also note that this variant of \texttt{MARS} is not included in the original paper's~\cite{yuan2024marsunleashingpowervariance} experiments on LLMs.
In our setup with batch size of $256\times512$ both \texttt{MARS} (\texttt{MARS-AdamW}) and \texttt{MARS-Lion} do not outperform the \texttt{AdamW} baseline.
However, this may be due to the smaller batch size: in the original work, the authors use $480\times1024$~($\equiv492\mathbf{M}$ tokens) batch size, and our experiments with the larger batch size of $1984\times512$~($\equiv1\mathbf{B}$ tokens)---see~\cref{fig:benchmark-720}---reveal that both \texttt{MARS-AdamW} and \texttt{Lion} greatly benefit from the increased batch size. Therefore, we highlight that it may be the case that \texttt{MARS-Lion} can outperform \texttt{AdamW} in some cases.

\vspace{-0.3em}
\prettybox{
\takeaway{tkw:mars_types}Among current \texttt{MARS} variants, \texttt{MARS-AdamW} is the best.
Notably, other modifications---\texttt{MARS-Shampoo} and \texttt{MARS-Lion} are differently affected by $\gamma$-schedulers and warmup.
\texttt{MARS-Lion} prefers the WSD scheduler over cosine, and shows the greatest stability to warmup sweep among all \texttt{MARS}-based methods.
}

\textbf{On learning rates of \texttt{Prodigy}.}
Throughout our benchmarking results~(\cref{fig:benchmarking-124m-losses,fig:benchmarking-210m-losses,fig:benchmark-720}), \texttt{Prodigy} consistently ranks among the top $6$ optimizers, performing close to \texttt{AdamW} at smaller scales and maintaining strong performance even when applied to MoE architectures.
Interestingly, when training $124\mathbf{M}$ models with an increased batch size of $512\times512$ tokens, \texttt{Prodigy} outperforms the \texttt{AdamW} baseline, suggesting that its critical batch size~\cite{epoch2024datamovement,zhang2024doescriticalbatchsize,hoffmann2022trainingcomputeoptimallargelanguage} may be larger than that of \texttt{AdamW}.
While highly efficient, \texttt{Prodigy} is generally easy to tune, except for its sensitivity to $\beta_2$~(\cref{fig:prodigy_betas}) in the small-batch setup.
This robustness is attributed to its adaptive learning rate mechanism, which relies on two exponential moving average sequences
\begin{align*}
    \begin{cases}
        r_t \leftarrow \sqrt{\beta_2} r_{t-1} + (1 - \sqrt{\beta_2})\gamma_t d^2_t\langle \bg_t, \xx_0 - \xx_t\rangle, \\
        \boldsymbol{s}_t \leftarrow \sqrt{\beta_2}\boldsymbol{s}_{t-1} + (1 - \sqrt{\beta_2})\gamma_t d^2_t\bg_t,
    \end{cases}
\end{align*}
that control the learning rate magnitude with a multiplier of
\begin{align*}
    d_{t+1} \leftarrow \max\left\{d_t, \frac{r_t}{\|\boldsymbol{s}_t\|_1}\right\}.
\end{align*}
At first, we define the effective learning rate of \texttt{Prodigy} as:
\begin{align}
\label{eq:prodigy_eff_lr}
    \mathtt{\gamma^\text{eff}_{t+1}} \coloneqq \gamma_t d_{t},
\end{align}
thus, when bias correction is applied---which we found necessary to ensure stability for small batches---\cref{eq:prodigy_eff_lr} becomes:
\begin{align*}
    \mathtt{\gamma^\text{eff}_{t+1}} = \gamma d_t \frac{\sqrt{1 - \beta_2^t}}{1 - \beta_1^t},
\end{align*}
where $\gamma$---the learning rate---is usually set to $1$ for \texttt{Prodigy}, which we confirmed to work the best through the sweeps in~\cref{fig:lrsensitivity,fig:ap_lrsensitivity}.

\begin{figure*}[h]
    \vspace{-1.4em}
    \centering
    \subfigure[Effective learning rate of \texttt{Prodigy}.]{
        \includegraphics[width=0.48\linewidth]{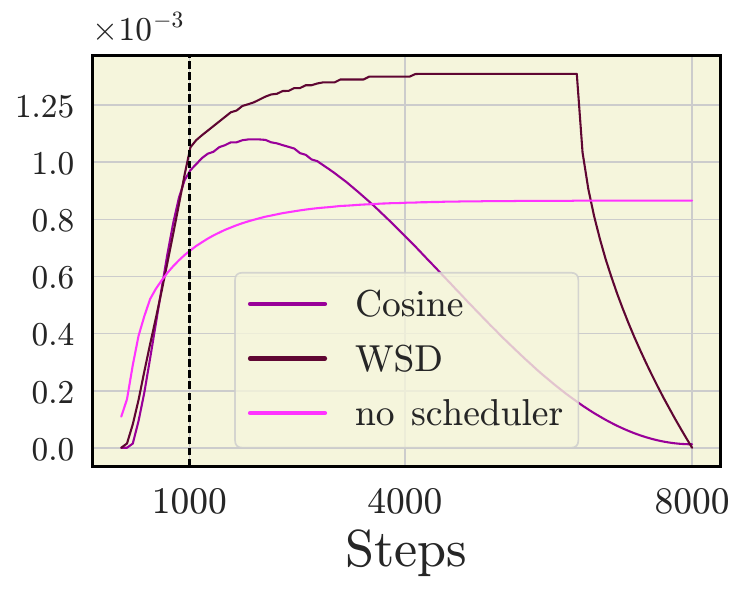}
    }
    \hfill
    \subfigure[Learning rate of \texttt{Prodigy}.]{
        \includegraphics[width=0.446\linewidth]{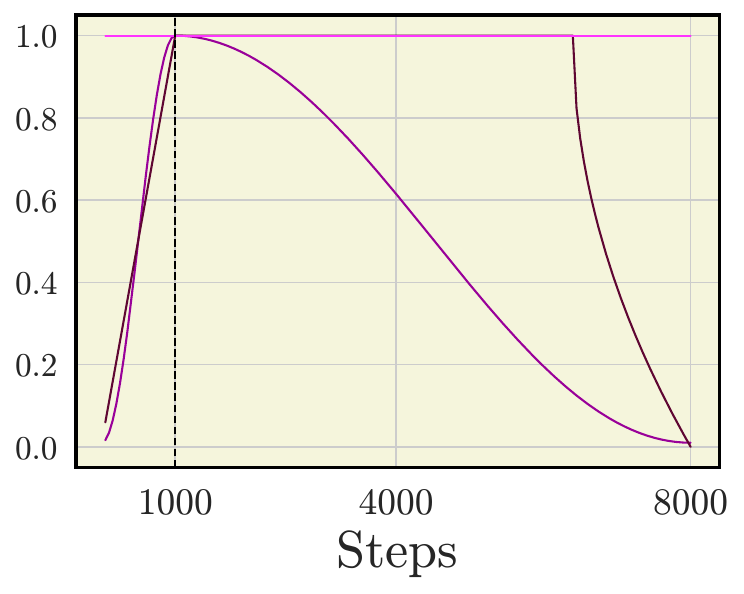}
    }
    \vspace{-0.2em}
    \caption{\textbf{EMA sequences of \texttt{Prodigy} result in the effective learning rate that emulates the dynamics of learning rate that we used to observe for \texttt{AdamW}.}
    Fixing the peak learning rate at $\gamma=1$ (following Mishchenko et al.~\cite{mishchenko2024prodigyexpeditiouslyadaptiveparameterfree}), the EMA sequences $r_t$ and $\ss_t$ (\texttt{lines 8, 9} of~\cref{alg:prodigy}) result in the effective learning rate shown in~\textbf{(a)}.
    The dashed line indicates the warmup duration.
    Across all schedulers and the run without a $\gamma$-scheduler, the warmup of $\gamma^\text{eff}_t$~\textbf{(a)} is consistently longer than that of $\gamma_t$~\textbf{(b)}, providing an implicit warmup.
    With cosine and WSD schedulers, the peak $\gamma^\text{eff}_t$ exceeds that of the run without a scheduler.
    Notably, the peak effective learning rates, especially for the cosine scheduler, are very close to the default value $0.001$ used for \texttt{AdamW} at this model scale.
    This demonstrates that \texttt{Prodigy} may guide practitioners in tuning learning rates for \texttt{Adam}-like optimizers.
    }
    \label{fig:ap_prodigy_effective_lr}
    \vspace{-1.2em}
\end{figure*}

When introducing the concept of the effective learning rate $\gamma^\text{eff}$, we note that it depends on the momentum parameters ($\beta_1$, $\beta_2$), the base learning rate $\gamma$, and the EMA sequences $\ss_t$, $r_t$.
Moreover, applying a $\gamma$-scheduler further influences how $\gamma$ evolves over iterations.
To study these interactions, we examine the dynamics of the effective learning rate~(\cref{eq:prodigy_eff_lr}) under different schedulers (\cref{eq:prodigy_eff_lr}).
For this purpose, we train a small $124\mathbf{M}$ model with the batch size of $256\times512$ tokens.
The training horizon is short---$8\mathbf{k}$ steps---with a warmup of $1000$ steps, and $\beta_2=0.999$---the best for \texttt{Prodigy} in this setup according to our tuning.
We also set the learning rate of \texttt{Prodigy} to $1$ as in our best benchmarking runs.
As in previous experiments, we apply WSD and cosine $\gamma$-schedulers, and additionally show a run without any scheduler.
For the WSD scheduler in this ablation, we do not rescale $\gamma$ to half the optimal value for cosine, as we are interested in the dynamics of $\gamma^\text{eff}_t$ rather than the final performance; observing it without rescaling provides a clearer picture.
\cref{fig:ap_prodigy_effective_lr}~\textbf{(a)} shows the dynamics of the effective learning rate $\gamma^\text{eff}_t$, while \textbf{(b)} illustrates the effect of applying scheduling to $\gamma=1$.
The starting points of the curves differ slightly due to variations in the final learning rate---cosine decays $\gamma_t$ down to $0.01$, whereas WSD decays it to zero using the $\left(1 - \sqrt{x}\right)$ decay pattern---however, those differences do not affect the qualitative shape of the figures obtained.

Interestingly, across all schedulers, we observe a common pattern---the effective learning rate warmup is longer than $T_\text{warmup}=1000$ steps---meaning that \texttt{Prodigy} experiences an ``implicit warmup'' beyond the explicitly set value.
Another notable observation is that when using the cosine scheduler with $\gamma=1$, the maximal effective learning rate reaches $\gamma^\text{eff}_{\max} \sim 1.08 \times$ larger than the learning rate of \texttt{AdamW} we use in a similar setting ($0.001$).
Consequently, setting \texttt{Prodigy}’s learning rate to the default value of $1$ produces dynamics closely matching those of \texttt{AdamW}.
This insight could be useful for practitioners as a proxy for tuning \texttt{Adam}-like optimizers: one can launch \texttt{Prodigy} with $\gamma=1$, track the effective learning rate~(\cref{eq:prodigy_eff_lr}), and then set the \texttt{AdamW} peak learning rate to $\gamma^\text{eff}_{\max}$. 
We highlight this one more time in~\cref{tkw:prodigy_effective_lr_2}.

\prettybox{
\takeaway{tkw:prodigy_effective_lr}We explain the effectiveness of \texttt{Prodigy} in ``learning rate-free'' training through the concept of the effective learning rate~(\cref{eq:prodigy_eff_lr}). Determined by two EMA sequences, the effective learning rate mimics the behavior and magnitude of the learning rate in \texttt{AdamW}-like methods.
Importantly: (\rom{1}) the magnitudes of the effective learning rate are close to those of \texttt{AdamW}; (\rom{2}) effective learning rate ensures an implicit warmup that is longer than initially set.
}
\prettybox{
\takeaway{tkw:prodigy_effective_lr_2}We point out that it might be interesting for researchers to try \texttt{Prodigy} as a proxy for learning rate tuning of \texttt{Adam}-like methods, e.g., (\rom{1}) tune betas of \texttt{Prodigy}, (\rom{2}) set $\gamma=1$, (\rom{3}) track $\gamma^\text{eff}_t$, and (\rom{4}) look at the $\gamma^\text{eff}_{\max}$ and set the learning rate of the \texttt{Adam}-like method to this value.
}

\subsection{Ablations for $210\mathbf{M}$ model}
\label{sec:ap_210mablations}

In this section, we complement our ablations from the main part with experiments specifically targeting $210\mathbf{M}$ models.
Compared to $124\mathbf{M}$ ablations~(\S~\ref{sec:ap_124mablations}), we perform fewer studies here.
We focus on two aspects: the sensitivity of \texttt{ADOPT} to its $\varepsilon$ hyperparameter, and the impact of weight initialization in LLMs and its interaction with the warmup.

\begin{wrapfigure}{r}{0.48\linewidth}
    \vspace{-1.4em}
    \centering
    \begin{minipage}{\linewidth}
        \includegraphics[width=\linewidth]{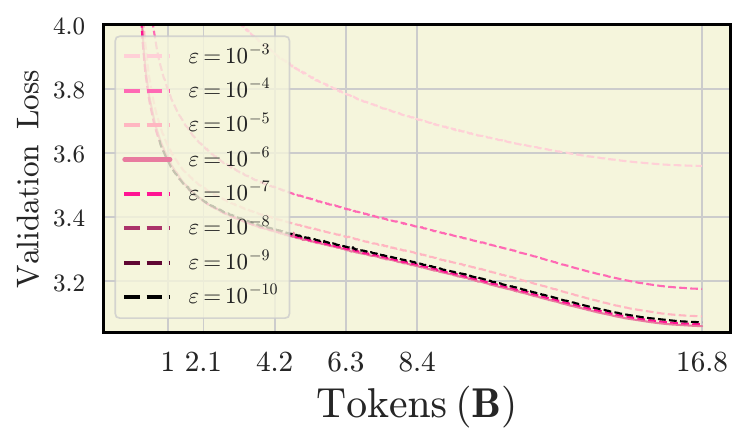}
    \end{minipage}
    \caption{\textbf{\texttt{ADOPT}'s sensitivity to $\varepsilon$.}
    Interestingly, the suggested by the authors $\varepsilon = 10^{-6}$ is the best hyperparameter for this method.
    There is not a noticeable difference in convergence for $\varepsilon = \{10^{-6}, 10^{-7}, 10^{-8}, 10^{-9}, 10^{-10}\}$, but the values of $10^{-5}$ and above give a much morse results.
    }
    \label{fig:adopt-eps}
\end{wrapfigure}

\textbf{\texttt{ADOPT} is sensitive to the epsilon hyperapameter, but the suggested $\varepsilon = 10^{-6}$ is the best.}
Among the many important hyperparameters, some receive less attention despite their influence.
For \texttt{Adam}-like methods, one such parameter is $\varepsilon$ in the denominator of the update rule.
While the default and widely accepted value for \texttt{AdamW} is $10^{-8}$, there is ongoing discussion in the community regarding other values that can significantly degrade training~\cite{groeneveld2024olmoacceleratingsciencelanguage,he2024understandingminimisingoutlierfeatures}.
The \texttt{ADOPT} optimizer also includes this hyperparameter---see \texttt{line 6} of~\cref{alg:adopt}.
Interestingly, the authors recommend using a larger value of $\varepsilon = 10^{-6}$, which is higher than the conventional choice for \texttt{AdamW}.
We perform a sweep over $\varepsilon$, keeping all other hyperparameters at their best values, and report the results in~\cref{fig:adopt-eps}.
As suggested by Taniguchi et al.~\cite{taniguchi2024adoptmodifiedadamconverge}, $\varepsilon = 10^{-6}$ outperforms all other tested values, with a noticeable margin for $\varepsilon \leq 10^{-5}$.

\begin{figure*}[h]
    \centering
    \subfigure{
        \includegraphics[width=0.48\linewidth]{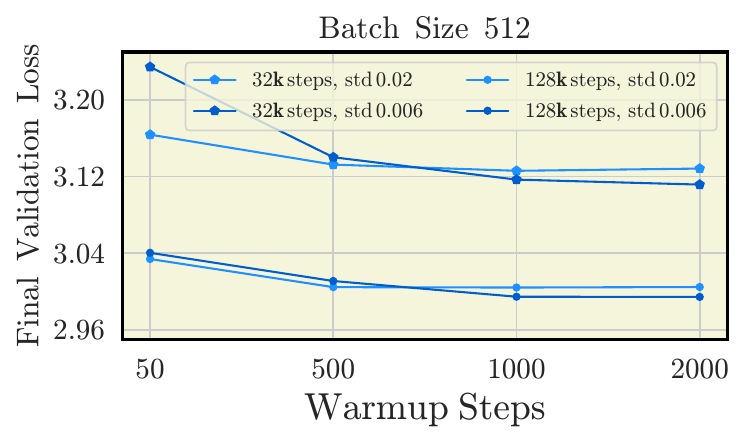}
    }
    \hfill
    \subfigure{
        \includegraphics[width=0.48\linewidth]{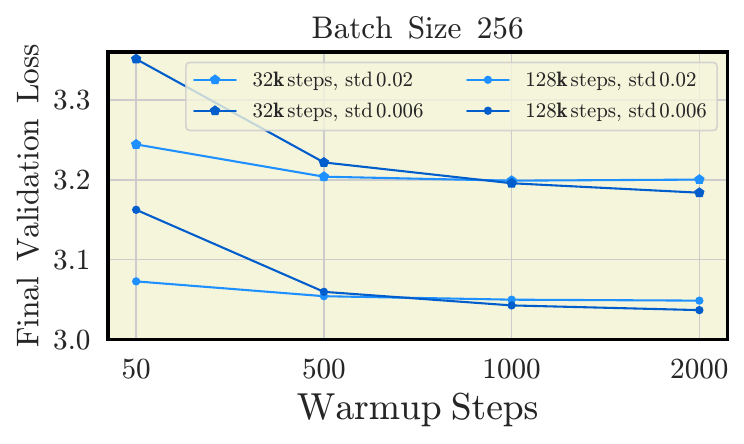}
    }
    \caption{\textbf{Weight initialization with smaller std prefers longer warmup.}
    We compare final loss of models trained with \texttt{AdamW} using two weight initializations: the conventional $\text{std}=0.02$ and a smaller $\text{std}=0.006$ as in DeepSeek.
    We vary the training horizon, warmup duration, and batch size (without changing the number of iterations).  
    Our results indicate that smaller initialization benefits from longer warmup, leading to better performance compared to $\text{std}=0.02$.  
    However, with very short warmup, the conventional initialization outperforms the smaller one.  
    Interestingly, increasing the batch size reduces the performance gap between the two initializations for longer training runs.
    }
    \label{fig:weight_init}
    \vspace{-0.5em}
\end{figure*}

\textbf{Changing weight initialization and the effect on warmup.}
A common approach to weight initialization in LLMs is the truncated Gaussian distribution with a predefined standard deviation (std).  
In popular codebases for scalable training~\cite{shoeybi2020megatronlmtrainingmultibillionparameter,deepspeed2020,groeneveld2024olmoacceleratingsciencelanguage}, the default std is $0.02$.  
Notably, in DeepSeek-V$3$~\cite{deepseekai2024deepseekv3technicalreport}, the default std is reduced to $0.006$.
Previously established connections between weight initialization and warmup report twofold results: ones~\cite{pmlr-v119-huang20f,zhu2021gradinitlearninginitializeneural} state that with a smaller std, one can reduce or even eliminate the need for warmup, while others~\cite{kalra2024warmuplearningrateunderlying,kosson2024analyzingreducingneed} highlight the importance of warmup for small weight initializations.
In our experiments, we investigate how both initialization styles interact with the warmup duration and the batch size scaling.  
Specifically, we compare the DeepSeek style initialization ($\text{std}=0.006$) with the conventional initialization ($\text{std}=0.02$).
We use two batch size settings: $512\times512$ tokens and $256\times512$ tokens, training Llama-based models for two horizons $T\in\{32, 128\}\mathbf{k}$ steps and sweeping $T_\text{warmup} \in \{50, 500, 1000, 2000\}$ iterations.
For this ablation, we use only \texttt{AdamW} with all other hyperparameters set to the best values identified from tuning of $210\mathbf{M}$ models.
We report the results in~\cref{fig:weight_init}.
Overall, we observe that smaller weight initialization favors longer warmup durations and performs significantly worse with short warmup.
Increasing the batch size reduces this gap for shorter warmups, suggesting an interplay between initialization scale, warmup duration, and batch size.

\prettybox{
\takeaway{tkw:weight_init}Weight initialization with smaller standard deviation, as in DeepSeek, benefits from longer warmup but underperforms with very short warmup. 
Increasing the batch size reduces the performance gap between small and conventional initializations.
}

\subsection{Wall-clock performance of optimizers across models of different scale}
\label{sec:ap_walltime}

We complement the wall-clock performance analysis from the main part~(\cref{fig:wall_time}) by presenting complete results for all optimizers.  
The experimental setup is simple and consistent: we use a batch size of $16$~($16\times512$ tokens), run for $100$ iterations on a single GPU, without gradient accumulation, and we do not apply \texttt{torch.compile}.  
Precise model configurations for all scales ($30\mathbf{M}$--$1\mathbf{B}$) are reported in~\cref{tab:llama_models_walltime}.

\begin{table*}[ht]
    \centering
    \caption{\textbf{Configurations for our Llama-like models for the wall-clock experiments.}}
    \label{tab:llama_models_walltime}
    \makebox[\linewidth]{
    \resizebox{1.0\textwidth}{!}{ 
    \begin{tabular}{|c|c|c|c|c|c|c|c|c|c|}
    \hline
    \textbf{\# Parameters} & $\mathbf{30}\mathbf{M}$ & $\mathbf{52}\mathbf{M}$ & $\mathbf{80}\mathbf{M}$ & $\mathbf{124}\mathbf{M}$ & $\mathbf{150}\mathbf{M}$ & $\mathbf{210}\mathbf{M}$ & $\mathbf{360}\mathbf{M}$ & $\mathbf{720}\mathbf{M}$ & $\mathbf{1026}\mathbf{M}$ \\ 
    \hline
    Hidden size & $384$ & $512$ & $768$ & $768$ & $768$ & $768$ & $1024$ & $2048$ & $1792$ \\ 
    \hline
    \# Attention heads & $6$  & $8$ & $6$ & $12$ & $12$ & $12$ & $16$ & $16$ & $14$  \\ 
    \hline
    \# Layers & $8$ & $8$ & $6$ & $12$ & $16$ & $24$ & $24$ & $12$ & $24$  \\ 
    \hline
     Init. std & $0.02$ & $0.02$ & $0.02$ & $0.02$ & $0.02$ & $0.02$ & $0.02$ & $0.02$ & $0.02$ \\
    \hline
    Use bias & no & no & no & no & no & no & no & no & no \\
    \hline
    RMSNorm epsilon & $0.00001$ & $0.00001$ & $0.00001$ & $0.00001$ & $0.00001$ & $0.00001$ & $0.00001$ & $0.00001$ & $0.00001$ \\
    \hline
    Positional encoding & RoPE & RoPE & RoPE & RoPE & RoPE & RoPE & RoPE & RoPE & RoPE  \\
    \hline
    \end{tabular}
    }
    }
\end{table*}

\begin{figure*}[h]
    \centering
    \begin{minipage}{0.8\linewidth}
        \includegraphics[width=\linewidth]{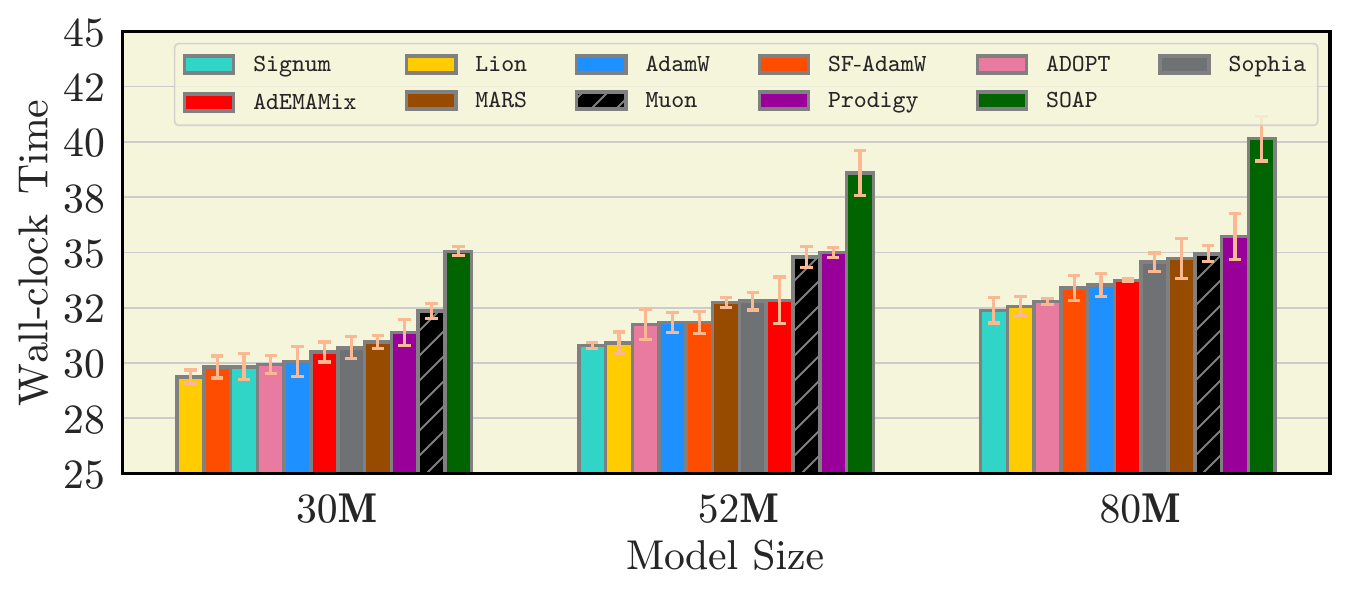}
    \end{minipage}
    \\
    \begin{minipage}{0.8\linewidth}
        \includegraphics[width=\linewidth]{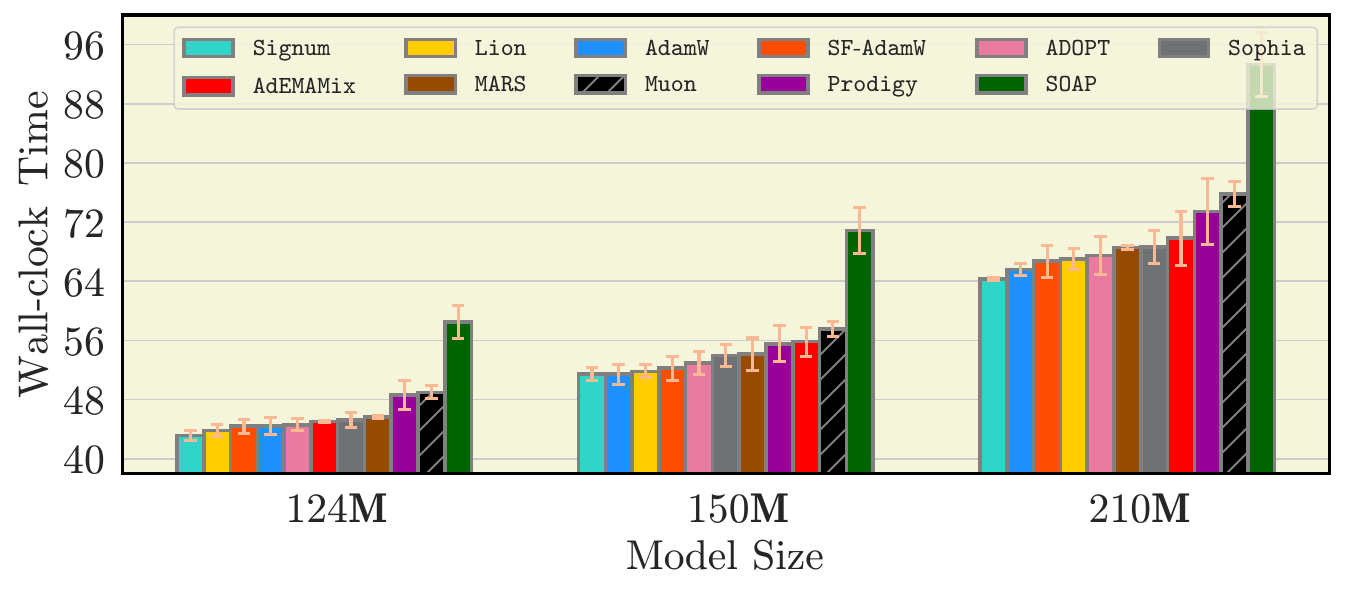}
    \end{minipage}
    \\
    \begin{minipage}{0.8\linewidth}
        \includegraphics[width=\linewidth]{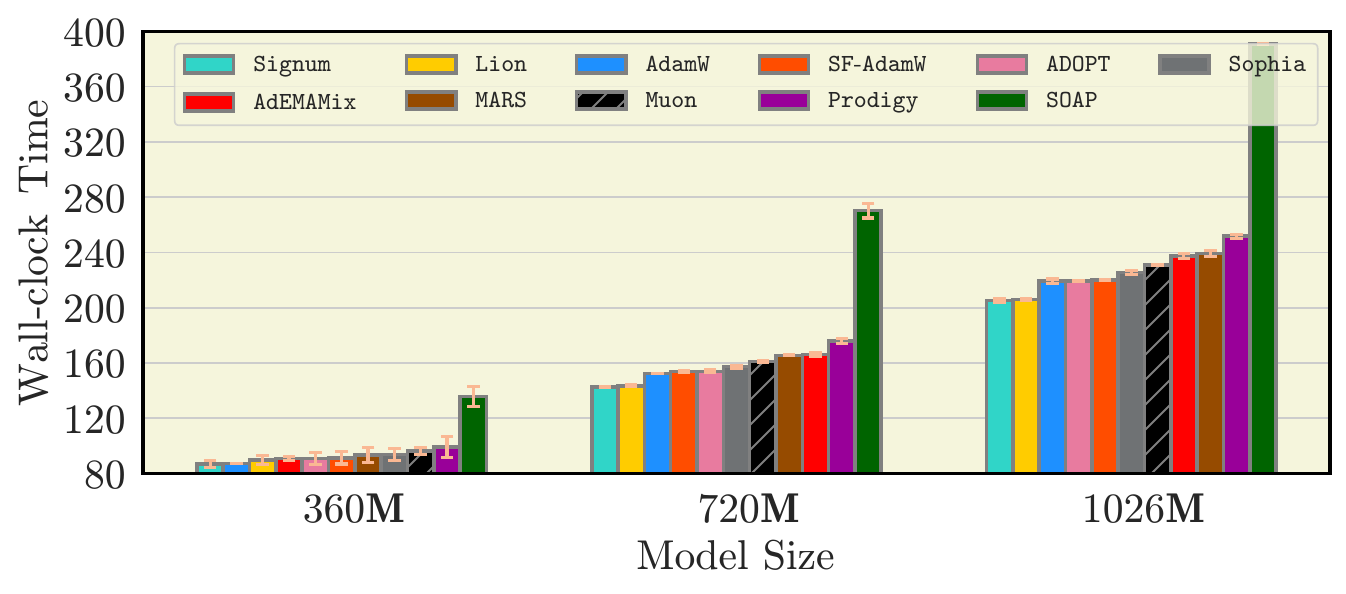}
    \end{minipage}
    \caption{\textbf{Wall-clock time performance: gathered.}
    We report the wall-clock time (in seconds) for training each model for $100$ iterations using a small batch size of $16\times512$ tokens on a single GPU, without gradient accumulation or \texttt{torch.compile}.  
    Bars show the ranking of optimizers from fastest (\texttt{Signum}) to slowest (\texttt{SOAP}) gathered across all model scales.  
    While the differences between most optimizers are small, \texttt{SOAP} is consistently slower.
    The absolute times may vary depending on the hardware, but the relative patterns remain consistent.
    }
    \label{fig:ap_walltime}
    \vspace{-2em}
\end{figure*}

\begin{figure*}[h]
    \centering
    \begin{minipage}{0.48\linewidth}
        \includegraphics[width=\linewidth]{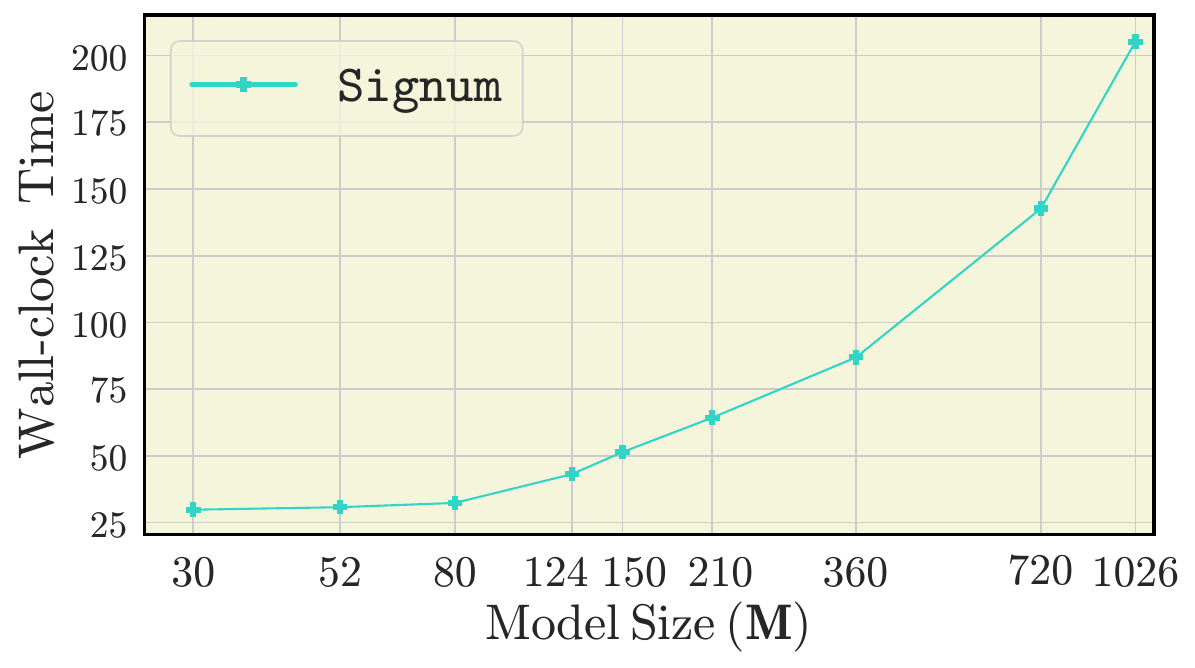}
    \end{minipage}
    \hfill
    \begin{minipage}{0.48\linewidth}
        \includegraphics[width=\linewidth]{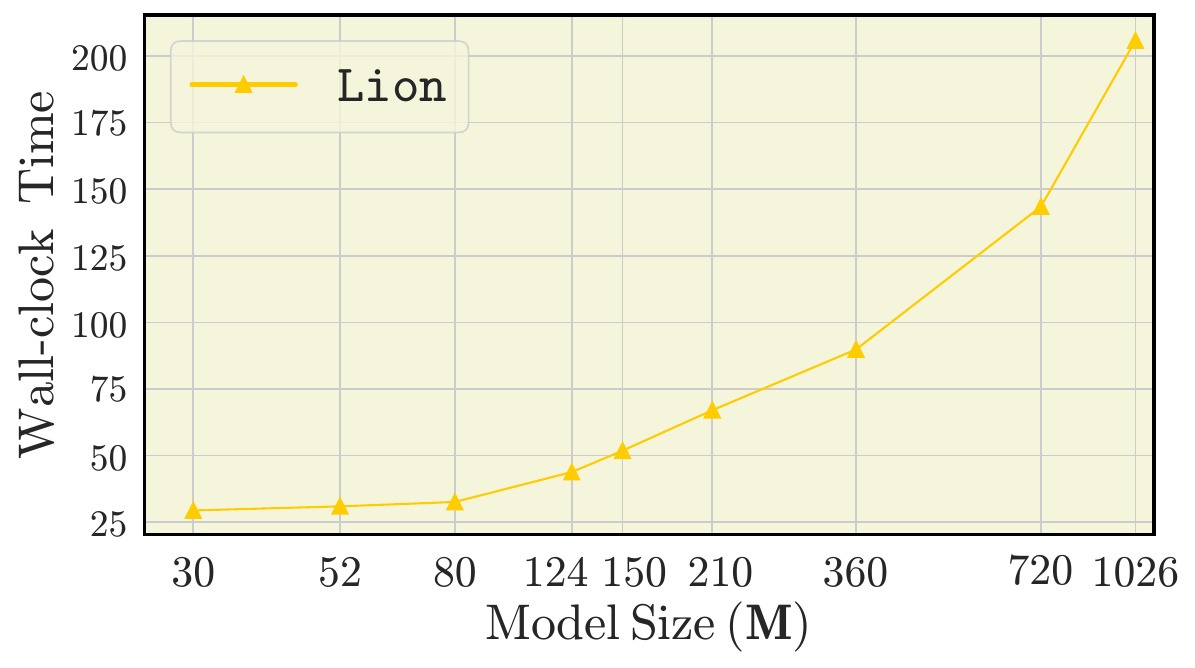}
    \end{minipage}
    \\
    \begin{minipage}{0.48\linewidth}
        \includegraphics[width=\linewidth]{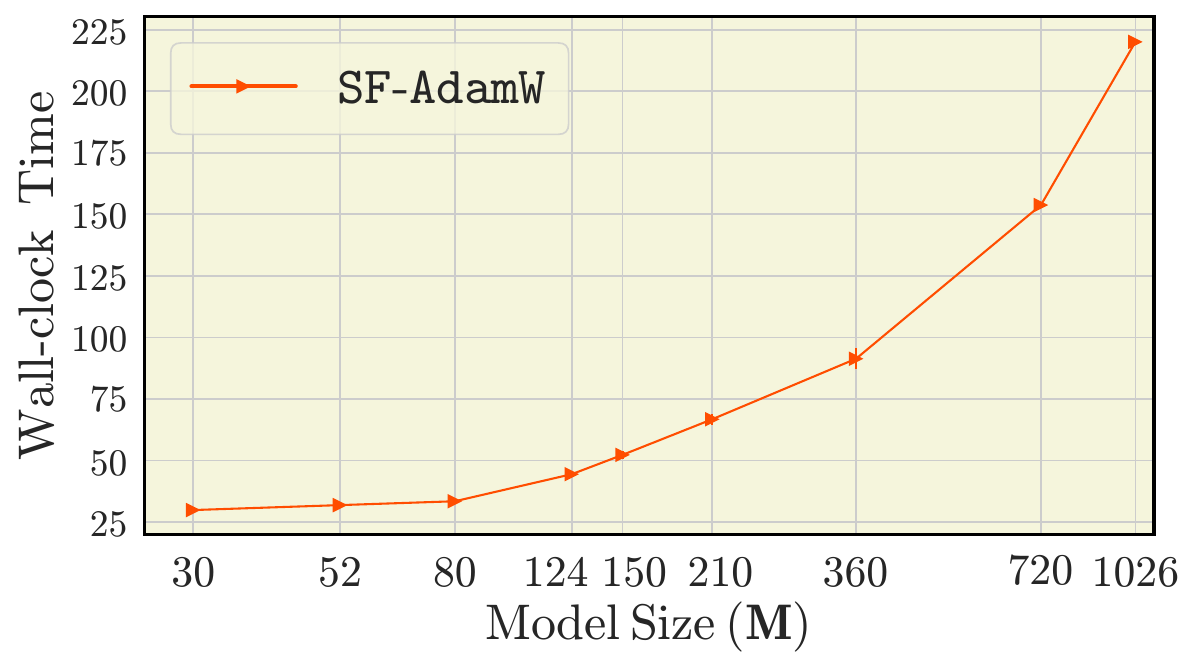}
    \end{minipage}
    \hfill
    \begin{minipage}{0.48\linewidth}
        \includegraphics[width=\linewidth]{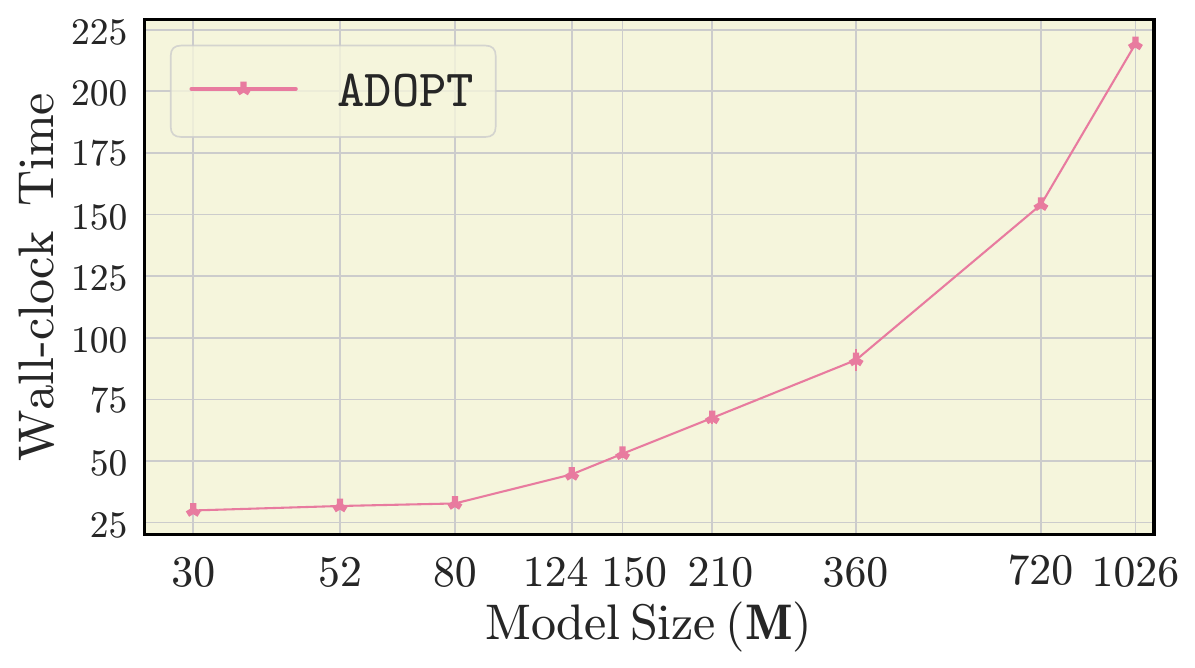}
    \end{minipage}
    \\
    \begin{minipage}{0.48\linewidth}
        \includegraphics[width=\linewidth]{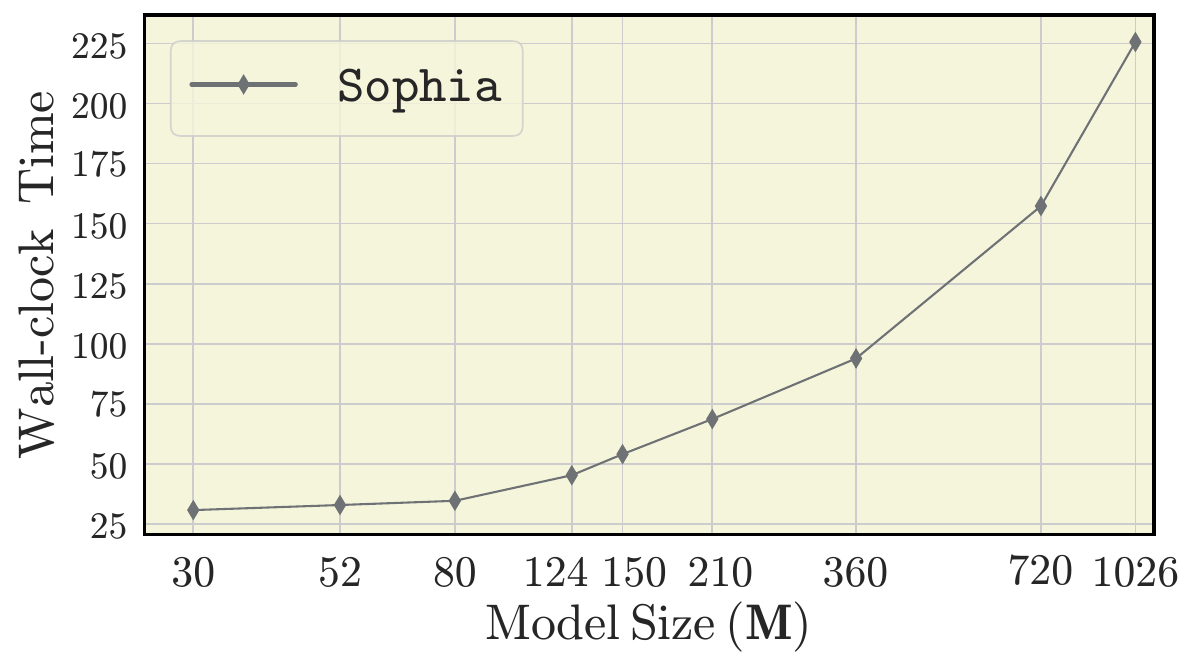}
    \end{minipage}
    \hfill
    \begin{minipage}{0.48\linewidth}
        \includegraphics[width=\linewidth]{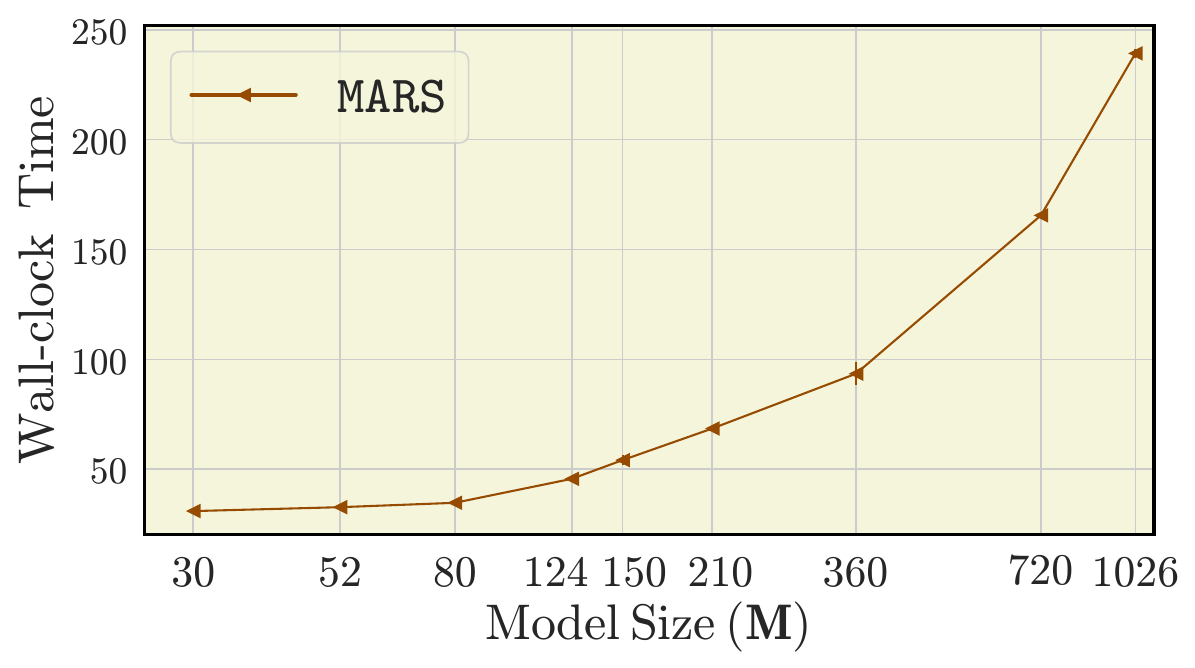}
    \end{minipage}
    \\
    \begin{minipage}{0.48\linewidth}
        \centering
        \includegraphics[width=\linewidth]{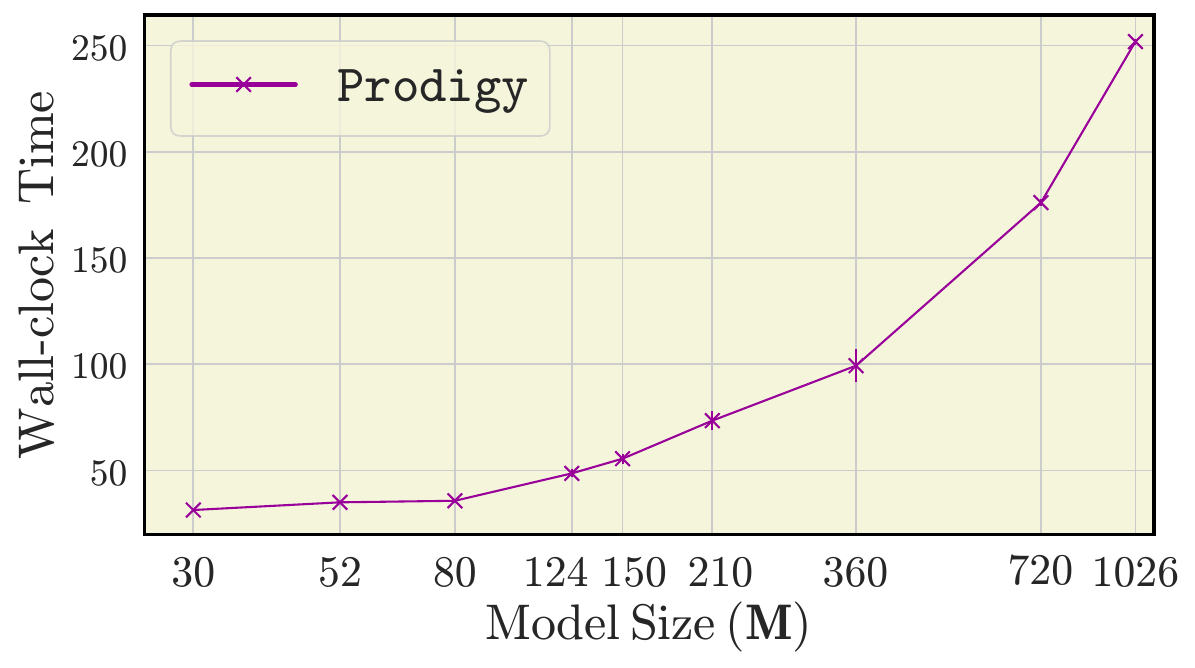}
    \end{minipage}
    \caption{\textbf{Wall-clock time performance: individual.}
    Complementing~\cref{fig:wall_time,fig:ap_walltime}, this figure shows the evolution of wall-clock time per $100$ iterations for each optimizer as model size increases.  
    Optimizers already shown in the main part are omitted.  
    To improve visualization, the abscissa is re-scaled to highlight the increase in wall-clock time with model size.
    }
    \label{fig:ap_wall_time_individual}
    \vspace{-2.2em}
\end{figure*}

\cref{fig:ap_walltime} shows a bar plot summarizing wall-clock time comparisons for all optimizers.  
Additionally, \cref{fig:ap_wall_time_individual} visualizes the per-optimizer behavior when scaling model size, omitting \texttt{SOAP}, \texttt{AdEMAMix}, \texttt{Muon}, and \texttt{AdamW}, as their results are already presented in the main part---see~\cref{fig:wall_time}.

\newpage

\section{Hyperparameter tuning}
\label{sec:ap_tuning}
\vspace{-0.8em}

\textbf{How do we tune hyperparameters?}
We perform systematic hyperparameter tuning for all algorithms, starting with smaller models ($124\mathbf{M}$, $210\mathbf{M}$) and extrapolating to larger, $583\mathbf{M}$ and $720\mathbf{M}$ models.
Our tuning process for $124\mathbf{M}$ model focused on two primary settings: \textbf{``small'' batch setting} ($32$ batch size) and \textbf{``large'' batch setting} ($256$ batch size).
For both settings, we use a sequence length of $512$ tokens, resulting in $16\mathbf{k}$ and $130\mathbf{k}$ tokens per batch, respectively. 
If the batch cannot fit into memory, we use gradient accumulation steps, while maintaining the effective batch size.

We also include ablations on even larger batch size for $124\mathbf{M}$ models, where we train on $512$ batch size ($260\mathbf{k}$ tokens correspondingly).
We train $583\mathbf{M}$ models on the batch size of $3936$, preserving the basic sequence length of $512$, that is, $\sim2\mathbf{M}$ tokens.
And the larger models for benchmarking purposes---of $720\mathbf{M}$---were trained on the batch size of $1984$, resulting in $\sim1\mathbf{M}$ tokens.

We first run multiple experiments, greed searching hyperparameters, on near Chinchilla optimal training length using \textit{cosine learning rate scheduler} (except for \texttt{SF-AdamW}):

$\bullet$ for $124\mathbf{M}$ models we tune at $2.1\mathbf{B}$ tokens for both ``small'' ($32$) and ``large'' ($256$) batch size setting (see \cref{sec:ap_124mtuning}),

$\bullet$ for $210\mathbf{M}$ models we replicate training runs with the best hyperparameters found at $124\mathbf{M}$ scale, except for the learning rate (see \cref{sec:ap_210mtuning}),

$\bullet$ at $583\mathbf{M}$ scale, we only ablate the effect of the $z$-loss regularizer while training with \texttt{AdamW} and \texttt{SOAP} on a near-Chinchilla optimal number of tokens (see \cref{sec:ap_600mtuning}),

$\bullet$ for $720\mathbf{M}$ models we tune at $16\mathbf{B}$ tokens (see \cref{sec:ap_720mtuning}),

$\bullet$ our MoE setting we discuss in-depth in \cref{sec:ap_520moetuning}.

We present the configurations for different training horizons in
Tables~\ref{tab:training_horizons_smallbs},~\ref{tab:training_horizons_largebs},~\ref{tab:training_horizons_1mbs},~\ref{tab:training_horizons_2mbs}.

\begin{table*}[h!]
    \centering
    \caption{\textbf{Lengths of training for \textbf{``Small'' batch settings} ($\mathbf{32 \times 512}$).}}
    \label{tab:training_horizons_smallbs}
    \makebox[\linewidth]{
    \resizebox{1.0\textwidth}{!}{ 
    \begin{tabular}{|c|c|c|c|c|c|c|c|}
    \hline
    \textbf{\# Parameters} & \multicolumn{6}{c|}{\textbf{Tokens (Iterations)}} & \textbf{Chinchilla Tokens} \\
    \hline
    $124\mathbf{M}$ & $1\mathbf{B}$ ($64\mathbf{k}$) & $2.1\mathbf{B}$ ($128\mathbf{k}$) & $4.2\mathbf{B}$ ($256\mathbf{k}$) & $6.3\mathbf{B}$ ($384\mathbf{k}$) & $8.4\mathbf{B}$ ($512\mathbf{k}$) & $16.8\mathbf{B}$ ($1024\mathbf{k}$) & $2.5\mathbf{B}$\\
    \hline
    $210\mathbf{M}$ & $1\mathbf{B}$ ($64\mathbf{k}$) & $2.1\mathbf{B}$ ($128\mathbf{k}$) & $4.2\mathbf{B}$ ($256\mathbf{k}$) & $6.3\mathbf{B}$ ($384\mathbf{k}$) & $8.4\mathbf{B}$ ($512\mathbf{k}$) & $16.8\mathbf{B}$ ($1024\mathbf{k}$) & $4.2\mathbf{B}$\\
    \hline
    \end{tabular}
    }
    }
\end{table*}

\begin{table*}[h!]
    \centering
    \caption{\textbf{Lengths of training for \textbf{``Large'' batch settings} ($\mathbf{256 \times 512}$).}}
    \label{tab:training_horizons_largebs}
    \makebox[\linewidth]{
    \resizebox{1.0\textwidth}{!}{
    \begin{tabular}{|c|c|c|c|c|c|c|c|}
    \hline
    \textbf{\# Parameters} & \multicolumn{6}{c|}{\textbf{Tokens (Iterations)}} & \textbf{Chinchilla Tokens}\\
    \hline
    $124\mathbf{M}$ & $1\mathbf{B}$ ($8\mathbf{k}$) & $2.1\mathbf{B}$ ($16\mathbf{k}$) & $4.2\mathbf{B}$ ($32\mathbf{k}$) & $6.3\mathbf{B}$ ($48\mathbf{k}$) & $8.4\mathbf{B}$ ($64\mathbf{k}$) & $16.8\mathbf{B}$ ($128\mathbf{k}$) & $2.5\mathbf{B}$ \\
    \hline
    $210\mathbf{M}$ & $1\mathbf{B}$ ($8\mathbf{k}$) & $2.1\mathbf{B}$ ($16\mathbf{k}$) & $4.2\mathbf{B}$ ($32\mathbf{k}$) & $6.3\mathbf{B}$ ($48\mathbf{k}$) & $8.4\mathbf{B}$ ($64\mathbf{k}$) & $16.8\mathbf{B}$ ($128\mathbf{k}$) & $4.2\mathbf{B}$\\
    \hline
    \end{tabular}
    }
    }
\end{table*}

\begin{table*}[h!]
    \centering
    \caption{\textbf{Lengths of training for $\mathbf{2M}$ ($\mathbf{3936 \times 512}$) batch size setting.}}
    \label{tab:training_horizons_2mbs}
    \begin{tabular}{|c|c|c|}
    \hline
    \textbf{\# Parameters} & \multicolumn{1}{c|}{\textbf{Tokens (Iterations)}} & \textbf{Chinchilla Tokens}\\
    \hline
    $583\mathbf{M}$ & $13\mathbf{B}$ ($6.5\mathbf{k}$) &  $11.7\mathbf{B}$ \\
    \hline
    \end{tabular}
\end{table*}

\begin{table*}[h!]
    \centering
    \caption{\textbf{Lengths of training for $\mathbf{1M}$ ($\mathbf{1984 \times 512}$) batch size setting.}}
    \label{tab:training_horizons_1mbs}
    \begin{tabular}{|c|c|c|c|c|}
    \hline
    \textbf{\# Parameters} & \multicolumn{3}{c|}{\textbf{Tokens (Iterations)}} & \textbf{Chinchilla Tokens}\\
    \hline
    $720\mathbf{M}$ & $8\mathbf{B}$ ($8\mathbf{k}$) & $16\mathbf{B}$ ($16\mathbf{k}$) & $48\mathbf{B}$ ($48\mathbf{k}$) &  $14.4\mathbf{B}$ \\
    \hline
    \end{tabular}
\end{table*}

Important to note, for larger models, we mostly kept the best hyperparameters found for the $124\mathbf{M}$ model and re-tuned the learning rate, beta parameters, and gradient clipping.
For dense LLMs, summarize this process in Appendices~\ref{sec:ap_124mtuning},~\ref{sec:ap_210mtuning},~\ref{sec:ap_600mtuning},~\ref{sec:ap_720mtuning}, and cover the MoE setup in~\cref{sec:ap_520moetuning}.

Additionally, when we report the effect of a particular hyperparameter, we assume that the remaining hyperparameters of the algorithm have already been tuned.  
Thus, the results isolate and highlight only the impact of the chosen hyperparameter on overall performance.

\textbf{Hyperparameters used in our experiments with learning rate schedulers.}
Once we found the best setting for each method using cosine learning rate scheduler, we are ready to obtain the optimal performance of our method with WSD~\cite{hu2024minicpmunveilingpotentialsmall} and linear schedulers.
For the latter one, we use the same hyperparameters as for the cosine scheduler.
However, for WSD, we follow the rule of thumb from \cite{hägele2024scalinglawscomputeoptimaltraining}:

$\bullet$ use half the optimal learning rate for the cosine scheduler,

$\bullet$ use $20\%$ of iterations for cooldown phase,

$\bullet$ use $\left(1 - \sqrt{x}\right)$ decay shape for the cooldown phase,

the only difference is that we do not employ stochastic weight averaging \cite{izmailov2019averagingweightsleadswider}.

Therefore, we maintain most hyperparameters across optimizers, only re-tuning the learning rate. 
For \texttt{Muon} and \texttt{MARS}, we reduce both \texttt{AdamW}'s learning rate and the learning rate for non-$1$D parameters. 
This approach ensures a fair comparison while accounting for the unique properties of each optimizer.

Importantly, the rule of thumb~\cite{hägele2024scalinglawscomputeoptimaltraining} for using the decay shape $\left(1 - \sqrt{x}\right)$ works better in our setting.
We use exactly this shape during the cooldown phase of the WSD scheduler for all optimizers.

We report a series of comparisons between different schedulers in~\cref{fig:wsdcosine,fig:wsdvscosine,fig:owt2wsdcosine}.

It has been shown~\cite{hägele2024scalinglawscomputeoptimaltraining,bergsma2025straightzerolinearlydecaying} that annealing the learning rate to smaller values than $10\%$ of the maximum learning rate improves performance.
We consider three mentioned schedulers, and report the ablation on the learning rate decay for the $210\mathbf{M}$ models in \cref{fig:lrdecay}, and for the $124\mathbf{M}$ models in \cref{fig:lrdecay-124m-appendix}.
In the tables that show the greed-search across hyperparameters we mention the learning rate decay factor (Final learning rate $\mathrm{X} \times \text{max LR}$) only for those optimizers, where we performed the corresponding ablation for.
If this field is omitted from the table, we use $0.01 \times \gamma_{\max}$ for this method regardless of the learning rate scheduler applied.

\newpage

\subsection{$\mathbf{124M}$ parameters model}
\label{sec:ap_124mtuning}

Below, we provide tables with complete information regarding hyperparameter tuning for $124\mathbf{M}$ models including the important sweeps~(weight decay, warmup, etc.) conducted for our ablations.

\begin{table}[h]
    \centering
    \caption{\textbf{\texttt{AdamW} hyperparameter tuning for our $\mathbf{124M}$ parameter large language models.}
    Bold hyperparameters are the best.}
    \label{tab:124m_adamw_hyperparams}
    \begin{center}
    \makebox[\linewidth]{
    \resizebox{1.0\textwidth}{!}{ 
    \begin{tabular}{|c|c|c|}
    \hline
    \textbf{Hyperparameter} & \textbf{``Small'' batch setting} & \textbf{``Large'' batch setting}  \\ 
    \hline
    Learning rate & $0.0001,\mathbf{0.0005},0.0008,0.001,0.002$  & $0.0001, 0.0003, 0.0005,\mathbf{0.001}, 0.002$ \\ 
    \hline
    Batch size & $32$  & $256$ \\ 
    \hline
    Sequence length & $512$ & $512$ \\ 
    \hline
    Number of warmup steps & $\mathbf{3000},5000,8000$ & $500,1000,\mathbf{2000},3000, 8000, 32000$ \\
    \hline
    Weight decay & $0.1$ & no, $\mathbf{0.1}, 0.5, 0.7$ \\
    \hline
    Learning rate decay scheduler & WSD, \textbf{cosine} & WSD, \textbf{cosine}, linear \\
    \hline
    Gradient clipping & no, $\mathbf{0.5}, 1, 1.5$ & no, $\mathbf{0.5}, 1$ \\
    \hline
    \texttt{AdamW} $\beta_1$ & $0.5,\mathbf{0.8}, 0.9$ & $\mathbf{0.8}, 0.9$ \\
    \hline
    \texttt{AdamW} $\beta_2$ & $0.95,\mathbf{0.999}$ & $0.95, 0.99, \mathbf{0.999}, 0.9999$ \\
    \hline
    Final learning rate $\mathrm{X} \times \text{max cosine LR}$ & \text{---}  & $10^{-1}$, $\mathbf{10^{-2}}$, $10^{-3}$, $10^{-4}$, $10^{-5}$, $10^{-6}$ \\
    \hline
    Final learning rate $\mathrm{X} \times \text{max WSD LR}$ & \text{---} & $10^{-1}$, $10^{-2}$, $10^{-3}$, $10^{-4}$, $10^{-5}$, $\mathbf{10^{-6}}$ \\
    \hline
    Final learning rate $\mathrm{X} \times \text{max linear LR}$ & \text{---} & $10^{-1}$, $10^{-2}$, $10^{-3}$, $\mathbf{10^{-4}}$, $10^{-5}$, $10^{-6}$ \\
    \hline
    \end{tabular}
    }
    }
    \end{center}
\end{table}

\begin{table}[h]
    \centering
    \caption{\textbf{\texttt{ADOPT} hyperparameter tuning for our $\mathbf{124M}$ parameter large language models.}
    Bold hyperparameters are the best.}
    \label{tab:124m_adopt_hyperparams}
    \begin{center}
    \begin{tabular}{|c|c|c|}
    \hline
    \textbf{Hyperparameter} & \textbf{``Small'' batch setting} & \textbf{``Large'' batch setting}  \\ 
    \hline
    Learning rate & $0.001$ & $0.0001, 0.0003, 0.0005, \mathbf{0.001}, 0.002$ \\ 
    \hline
    Batch size & $32$  & $256$ \\ 
    \hline
    Sequence length & $512$ & $512$ \\ 
    \hline
    Number of warmup steps & $\mathbf{3000}, 8000$ & $\mathbf{2000}, 8000, 32000$ \\
    \hline
    Weight decay & $0.1$ & no, $\mathbf{0.1}, 0.5$ \\
    \hline
    Learning rate decay scheduler & WSD, \textbf{cosine} & WSD, \textbf{cosine}, linear \\
    \hline
    Gradient clipping & $0.5$ & no, $\mathbf{0.5}, 1$ \\
    \hline
    \texttt{ADOPT} $\beta_1$ & $0.9$ & $0.8, \mathbf{0.9}$ \\
    \hline
    \texttt{ADOPT} $\beta_2$ & $\mathbf{0.999}, 0.9999$ & $0.5, \mathbf{0.999}, 0.9999$\\
    \hline
    \texttt{ADOPT} $\varepsilon$ & $10^{-6}$ & $10^{-6}$ \\
    \hline
    \end{tabular}
    \end{center}
\end{table}

\begin{table}[h]
    \centering
    \caption{\textbf{\texttt{AdEMAMix} hyperparameter tuning for our $\mathbf{124M}$ parameter large language models.}
    Bold hyperparameters are the best.}
    \label{tab:124m_ademamix_hyperparams}
    \begin{center}
    \makebox[\linewidth]{
    \resizebox{1.0\textwidth}{!}{ 
    \begin{tabular}{|c|c|c|}
    \hline
    \textbf{Hyperparameter} & \textbf{``Small'' batch setting} & \textbf{``Large'' batch setting}  \\ 
    \hline
    Learning rate & $0.0001,\mathbf{0.0005},0.0008,0.001,0.002$ & $0.0001,0.0003,0.0005,\mathbf{0.001}, 0.002$  \\ 
    \hline
    Batch size & $32$ & $256$  \\ 
    \hline
    Sequence length & $512$ & $512$ \\ 
    \hline
    Number of warmup steps & $\mathbf{3000}, 8000$ & $\mathbf{2000}, 8000, 32000$ \\
    \hline
    Weight decay & $0.1$ & no, $\mathbf{0.1}, 0.5, 0.7$ \\
    \hline
    Learning rate decay scheduler & WSD, \textbf{cosine} & WSD, \textbf{cosine}, linear \\
    \hline
    Gradient clipping & no, $\mathbf{0.5}, 1, 1.5$ & no, $\mathbf{0.5}, 1$ \\
    \hline
    \texttt{AdEMAMix} $\beta_1$ & $0.5,\mathbf{0.8},0.9$ & $0.8,\mathbf{0.9}$ \\
    \hline
    \texttt{AdEMAMix} $\beta_2$ & $0.999$ & $\mathbf{0.999}, 0.9999$ \\
    \hline
    \texttt{AdEMAMix} $\beta_3$ & $0.999,\mathbf{0.9999},0.99995$ & $\mathbf{0.999}, 0.9999$ \\
    \hline
    \texttt{AdEMAMix} $\alpha$ & $5,\mathbf{8},12$ & $8$ \\
    \hline
    \end{tabular}
    }
    }
    \end{center}
\end{table}

\newpage

\begin{table}[ht]
    \centering
    \caption{\textbf{\texttt{Lion} hyperparameter tuning for our $\mathbf{124M}$ parameter large language models.}
    Bold hyperparameters are the best.}
    \label{tab:124m_lion_hyperparams}
    \begin{center}
    \begin{tabular}{|c|c|c|}
    \hline
    \textbf{Hyperparameter} & \textbf{``Small'' batch setting} & \textbf{``Large'' batch setting}    \\ 
    \hline
    Learning rate & $0.00005,\mathbf{0.0001},0.0005,0.001$ & $0.0001,0.0005,\mathbf{0.001}, 0.002$ \\ 
    \hline
    Batch size & $32$ & $256$ \\ 
    \hline
    Sequence length & $512$ & $512$ \\ 
    \hline
    Number of warmup steps & $3000$ & $2000, 8000, \mathbf{32000}$ \\
    \hline
    Weight decay & no, $0.1,0.2,\mathbf{0.5}$ & no, $\mathbf{0.1},0.5,0.7$ \\
    \hline
    Learning rate decay scheduler & WSD, \textbf{cosine} & WSD, cosine, \textbf{linear} \\
    \hline
    Gradient clipping & $0.5$ & no, $\mathbf{0.5}, 1$ \\
    \hline
    \texttt{Lion} $\beta_1$ & $0.7,\mathbf{0.9},0.99$ & $0.5,\mathbf{0.9}$ \\
    \hline
    \texttt{Lion} $\beta_2$ & $0.9,\mathbf{0.99},0.999$ & $\mathbf{0.99},0.999$ \\
    \hline
    \end{tabular}
    \end{center}
\end{table}

\begin{table}[ht]
    \centering
    \caption{\textbf{\texttt{Signum} hyperparameter tuning for our $\mathbf{124M}$ parameter large language models.}
    Bold hyperparameters are the best.}
    \label{tab:124m_signum_hyperparams}
    \begin{center}
    \makebox[\linewidth]{
    \resizebox{1.0\textwidth}{!}{ 
    \begin{tabular}{|c|c|c|}
    \hline
    \textbf{Hyperparameter} & \textbf{``Small'' batch setting} & \textbf{``Large'' batch setting}  \\ 
    \hline
    Learning rate & $0.0003, 0.0005, \mathbf{0.001}$  & $0.0001, 0.0003 0.0005, 0.0003, \mathbf{0.001}, 0.002$\\ 
    \hline
    Batch size & $32$ & $256$ \\ 
    \hline
    Sequence length & $512$ & $512$ \\ 
    \hline
    Number of warmup steps & $2000, \mathbf{3000}$ & $2000, \mathbf{8000}, 32000$\\
    \hline
    Weight decay & no, $0, \mathbf{0.1}, 0.5$ & no, $0, \mathbf{0.1}, 0.5, 0.7$\\
    \hline
    Learning rate decay scheduler & WSD, \textbf{cosine} & WSD, cosine, \textbf{linear}\\
    \hline
    Gradient clipping & no, $\mathbf{0.5}, 1$ & no, $\mathbf{0.5}, 1$ \\
    \hline
    Momentum & no, $0.9, \mathbf{0.95}$ & no, $0.9, \mathbf{0.95}, 0.99$\\
    \hline
    Nesterov momentum & no, \textbf{yes} & no, \textbf{yes} \\
    \hline
    \end{tabular}
    }
    }
    \end{center}
\end{table}

\begin{table}[h!]
    \centering
    \caption{\textbf{\texttt{Muon} hyperparameter tuning for our $\mathbf{124M}$ parameter large language models.}
    Bold hyperparameters are the best.}
    \label{tab:124m_muon_hyperparams}
    \begin{center}
    \makebox[\linewidth]{
    \resizebox{1.0\textwidth}{!}{ 
    \begin{tabular}{|c|c|c|}
    \hline
    \textbf{Hyperparameter} & \textbf{``Small'' batch setting} & \textbf{``Large'' batch setting}  \\ 
    \hline
    Learning rate \texttt{AdamW} & $0.0001, 0.0003, 0.0005, \mathbf{0.001}, 0.002$  & $0.0001, 0.0003, 0.0005, \mathbf{0.001}, 0.002$\\ 
    \hline
    Learning rate \texttt{Muon} & $0.001, \mathbf{0.01}, 0.02$  & $0.001, \mathbf{0.01}, 0.02$\\ 
    \hline
    Batch size & $32$  & $256$ \\ 
    \hline
    Sequence length & $512$ & $512$ \\ 
    \hline
    Number of warmup steps & $\mathbf{3000}, 8000$ & $\mathbf{2000}, 8000, 32000$ \\
    \hline
    Weight decay & no, $\mathbf{0.1}, 0.5$ & no, $\mathbf{0.1}, 0.5$ \\
    \hline
    Learning rate decay scheduler & \textbf{WSD}, cosine & \textbf{WSD}, cosine, linear \\
    \hline
    Gradient clipping & no, $\mathbf{0.5}$ & no, $\mathbf{0.5}, 1.0$ \\
    \hline
    Momentum \texttt{Muon} & $0.9, 0.95, \mathbf{0.99}$ & $\mathbf{0.95}, 0.99$ \\
    \hline
    Optimizer for $1$D layers  & \texttt{AdamW} & \texttt{AdamW} \\
    \hline
    Optimizer for $1$D layers, $\beta_1$ & $\mathbf{0.8}, 0.9$ & $\mathbf{0.8}, 0.9$ \\
    \hline
    Optimizer for $1$D layers, $\beta_2$ & $0.99, \mathbf{0.999}, 0.9999$ & $0.99, \mathbf{0.999}, 0.9999$ \\
    \hline
    Newton-Schulz a & $3.4445$ & $3.4445$ \\
    \hline
    Newton-Schultz b & $-4.7750$ & $-4.7750$ \\
    \hline
    Newton-Schultz c & $2.0315$ & $2.0315$ \\
    \hline
    Newton-Schultz iterations & $5$ & $1, \mathbf{5}, 10, 20$ \\
    \hline
    Nesterov momentum & no, \textbf{yes} & no, \textbf{yes} \\
    \hline
    \end{tabular}
    }
    }
    \end{center}
\end{table}

\newpage

\begin{table}[h!]
    \centering
    \caption{\textbf{\texttt{D-Muon} hyperparameter tuning for our $\mathbf{124M}$ parameter large language models.}
    Bold hyperparameters are the best.}
    \label{tab:124m_dmuon_hyperparams}
    \begin{center}
    \makebox[\linewidth]{
    \resizebox{1.0\textwidth}{!}{ 
    \begin{tabular}{|c|c|c|}
    \hline
    \textbf{Hyperparameter} & \textbf{``Small'' batch setting} & \textbf{``Large'' batch setting}  \\ 
    \hline
    Learning rate & $0.001$  & $0.0001, 0.0003, 0.0005, 0.001, \mathbf{0.002}$\\ 
    \hline
    Batch size & $32$  & $256$ \\ 
    \hline
    Sequence length & $512$ & $512$ \\ 
    \hline
    Number of warmup steps & $3000$ & $\mathbf{2000}, 8000, 32000$ \\
    \hline
    Weight decay & $0.1$ & no, $\mathbf{0.1}, 0.5$ \\
    \hline
    Learning rate decay scheduler & WSD, \textbf{cosine} & WSD, \textbf{cosine}, linear \\
    \hline
    Gradient clipping & no, $\mathbf{0.5}$ & no, $\mathbf{0.5}, 1.0$ \\
    \hline
    Momentum \texttt{D-Muon} & $0.95$ & $0.95$ \\
    \hline
    Optimizer for $1$D layers  & \texttt{AdamW} & \texttt{AdamW} \\
    \hline
    Optimizer for $1$D layers, $\beta_1$ & $\mathbf{0.8}, 0.9$ & $\mathbf{0.8}, 0.9$ \\
    \hline
    Optimizer for $1$D layers, $\beta_2$ & $0.99, \mathbf{0.999}, 0.9999$ & $0.99, \mathbf{0.999}, 0.9999$ \\
    \hline
    Newton-Schulz a & $3.4445$ & $3.4445$ \\
    \hline
    Newton-Schultz b & $-4.7750$ & $-4.7750$ \\
    \hline
    Newton-Schultz c & $2.0315$ & $2.0315$ \\
    \hline
    Newton-Schultz iterations & $5$ & $5$ \\
    \hline
    Nesterov momentum & yes & yes \\
    \hline
    \end{tabular}
    }
    }
    \end{center}
\end{table}

\begin{table}[h!]
    \centering
    \caption{\textbf{\texttt{SOAP} hyperparameter tuning for our $\mathbf{124M}$ parameter large language models.}
    Bold hyperparameters are the best.}
    \label{tab:124m_soap_hyperparams}
    \begin{center}
    \makebox[\linewidth]{
    \resizebox{1.0\textwidth}{!}{ 
    \begin{tabular}{|c|c|c|}
    \hline
    \textbf{Hyperparameter} & \textbf{``Small'' batch setting} & \textbf{``Large'' batch setting}  \\ 
    \hline
    Learning rate & $0.005, \mathbf{0.001}$ &  $0.0001, 0.0003, 0.0005, 0.001, \mathbf{0.002}$ \\ 
    \hline
    Batch size & $32$ & $256$ \\ 
    \hline
    Sequence length & $512$ & $512$ \\ 
    \hline
    Number of warmup steps & $\mathbf{3000}, 8000$ & $\mathbf{2000}, 4000, 8000, 12000, 16000, 32000$ \\
    \hline
    Weight decay & $0.1$ & no, $\mathbf{0.1}, 0.5$ \\
    \hline
    Learning rate decay scheduler & WSD, \textbf{cosine} & WSD, \textbf{cosine}, linear \\
    \hline
    Gradient clipping & $0.5$ & no, $\mathbf{0.5}, 1$ \\
    \hline
        Preconditioner dimension & $10000$ & $10000$ \\
    \hline
    Preconditioning frequency & $1, 5, \mathbf{10}$ & $1, 5, \mathbf{10}$ \\
    \hline
    \texttt{SOAP} $\beta_1$ & $0.8, \mathbf{0.9}$ & $0.8, \mathbf{0.9}, 0.95$ \\
    \hline
    \texttt{SOAP} $\beta_2$ & $0.95, 0.99, \mathbf{0.999}, 0.9999$ & $0.95, 0.99, \mathbf{0.999}, 0.9999$\\
    \hline
    \end{tabular}
    }
    }
    \end{center}
\end{table}

\begin{table}[h!]
    \centering
    \caption{\textbf{\texttt{Sophia} hyperparameter tuning for our $\mathbf{124M}$ parameter large language models.}
    Bold hyperparameters are the best.}
    \label{tab:124m_sophia_hyperparams}
    \begin{center}
    \makebox[\linewidth]{
    \resizebox{1.0\textwidth}{!}{ 
    \begin{tabular}{|c|c|c|}
    \hline
    \textbf{Hyperparameter} & \textbf{``Small'' batch setting} & \textbf{``Large'' batch setting}  \\ 
    \hline
    Learning rate & $0.0001, \mathbf{0.0003}, 0.0005, 0.001, 0.002$ & $0.0001, 0.0003, 0.0005, \mathbf{0.001}, 0.002, 0.01$\\ 
    \hline
    Batch size & $32$ & $256$ \\ 
    \hline
    Sequence length & $512$ & $512$ \\ 
    \hline
    Number of warmup steps & $\mathbf{2000}, 3000$ & $2000, 8000, \mathbf{32000}$\\
    \hline
    Weight decay & $0.1$ & no, $\mathbf{0.1}, 0.5$ \\
    \hline
    Learning rate decay scheduler & WSD, \textbf{cosine} & WSD, \textbf{cosine}, linear \\
    \hline
    Gradient clipping & $0.5$ & no, $\mathbf{0.5}, 1$ \\
    \hline
    Estimator & Gauss-Newton-Bartlett & Gauss-Newton-Bartlett \\
    \hline
    Estimator frequency & $10$ & $10$ \\
    \hline
    \texttt{Sophia} $\beta_1$ & $0.9$ & $0.8, \mathbf{0.9}$ \\
    \hline
    \texttt{Sophia} $\beta_2$ & $0.95, \mathbf{0.999}, 0.9999, 0.99999$ & $0.95, \mathbf{0.999}, 0.9999, 0.99999$ \\
    \hline
    \texttt{Sophia} $\rho$ & $0, 0.03, \mathbf{0.04}$ & $0, 0.03, \mathbf{0.04}$ \\
    \hline
    \end{tabular}
    }
    }
    \end{center}
\end{table}

\newpage

\begin{table}[h!]
    \centering
    \caption{\textbf{\texttt{Schedule-Free AdamW} hyperparameter tuning for our $\mathbf{124M}$ parameter large language models.}
    Bold hyperparameters are the best.}
    \label{tab:124m_sfadamw_hyperparams}
    \makebox[\linewidth]{
    \resizebox{1.0\textwidth}{!}{ 
    \begin{tabular}{|c|c|c|}
    \hline
    \textbf{Hyperparameter} & \textbf{``Small'' batch setting} & \textbf{``Large'' batch setting}  \\ 
    \hline
    Learning rate & $0.0001, 0.0003, 0.0005, \mathbf{0.001}, 0.005$ & $0.0001, 0.0003, 0.0005, 0.001, \mathbf{0.002}, 0.005$  \\ 
    \hline
    Batch size & $32$ & $256$ \\ 
    \hline
    Sequence length & $512$ & $512$ \\ 
    \hline
    Number of warmup steps & $\mathbf{3000}, 8000$ & $2000, 4000, \mathbf{8000}, 12000, 16000, 32000$ \\
    \hline
    Weight decay & no, $0.05, \mathbf{0.1}, 0.5$ & no, $0.05, \mathbf{0.1}, 0.5$ \\
    \hline
    Learning rate decay scheduler & no & no \\
    \hline
    Gradient clipping & no, $\mathbf{0.5}$ & no, $\mathbf{0.5}, 1$ \\
    \hline
    \texttt{Schedule-Free AdamW} $\beta_1$ & $\mathbf{0.9}, 0.95, 0.98$ & $\mathbf{0.9}, 0.95, 0.98$\\
    \hline
    \texttt{Schedule-Free AdamW} $\beta_2$ & $0.95, 0.99, 0.999, \mathbf{0.9999}, 0.99999$& $0.95, 0.99, 0.999, \mathbf{0.9999}, 0.99999$ \\
    \hline
    \end{tabular}
    }
    }
\end{table}

\begin{table}[h!]
    \centering
    \caption{\textbf{\texttt{Prodigy} hyperparameter tuning for our $\mathbf{124M}$ parameter large language models.}
    Bold hyperparameters are the best.}
    \label{tab:124m_prodigy_hyperparams} 
    \begin{center}
    \makebox[\linewidth]{
    \resizebox{1.0\textwidth}{!}{
    \begin{tabular}{|c|c|c|}
    \hline
    \textbf{Hyperparameter} & \textbf{``Small'' batch setting} & \textbf{``Large'' batch setting}  \\ 
    \hline
    Learning rate & $0.5, \mathbf{1}$  & $0.5, \mathbf{1}, 2, 10, 100$ \\ 
    \hline
    Batch size & $32$  & $256$ \\ 
    \hline
    Sequence length & $512$ & $512$ \\ 
    \hline
    Number of warmup steps & $\mathbf{3000}, 8000$ & $\mathbf{2000}, 4000, 8000, 12000, 16000, 32000$ \\
    \hline
    Weight decay & no, $\mathbf{0.1}, 0.5$ & no, $\mathbf{0.1}, 0.5$ \\
    \hline
    Learning rate decay scheduler & no, WSD, \textbf{cosine} & no, WSD, \textbf{cosine}, linear \\
    \hline
    Gradient clipping & no, $\mathbf{0.5}, 1$ & no, $\mathbf{0.5}, 1$ \\
    \hline
    \texttt{Prodigy} $\beta_1$ & $0.9$ & $0.8, \mathbf{0.9}$ \\
    \hline
    \texttt{Prodigy} $\beta_2$ & $0.99, \mathbf{0.999}, 0.9999$ & $\mathbf{0.999}, 0.9999$\\
    \hline
    \texttt{Prodigy} bias correction & no, \textbf{yes} & no, \textbf{yes} \\
    \hline
    \end{tabular}
    }
    }
    \end{center}
\end{table}

\begin{table}[h!]
    \centering
    \caption{\textbf{\texttt{MARS} (\texttt{MARS-AdamW}) hyperparameter tuning for our $\mathbf{124M}$ parameter large language models.}
    Bold hyperparameters are the best.}
    \label{tab:124m_mars_hyperparams}
    \begin{center}
    \begin{tabular}{|c|c|c|}
    \hline
    \textbf{Hyperparameter} & \textbf{``Small'' batch setting} & \textbf{``Large'' batch setting}  \\ 
    \hline
    Learning rate \texttt{AdamW} & $0.0001, 0.0005, \mathbf{0.001}, 0.003$  & $0.0001, 0.0005, \mathbf{0.001}, 0.003$ \\ 
    \hline
    Learning rate \texttt{MARS} & $0.001, \mathbf{0.003}$  & $0.001, \mathbf{0.003}$ \\ 
    \hline
    Batch size & $32$  & $256$ \\ 
    \hline
    Sequence length & $512$ & $512$ \\ 
    \hline
    Number of warmup steps & $2000, \mathbf{3000}$ & $\mathbf{2000}, 8000, 32000$ \\
    \hline
    Weight decay \texttt{MARS} & no, $\mathbf{0.1}$ & no, $\mathbf{0.1}, 0.5$ \\
    \hline
    Weight decay for $1$D layers & $0.1$ & $0.1$\\
    \hline
    Learning rate decay scheduler & WSD, \textbf{cosine} & WSD, \textbf{cosine}, linear \\
    \hline
    Gradient clipping & $0.5$ & $0.5$ \\
    \hline
    Optimizer for $1$D layers & \texttt{AdamW} & \texttt{AdamW} \\
    \hline
    Optimizer for $1$D layers $\beta_1$ & $\mathbf{0.8}, 0.9$ & $\mathbf{0.8}, 0.9, 0.95$ \\
    \hline
    Optimizer for $1$D layers $\beta_2$ & $0.95, 0.99, \mathbf{0.999}$ & $0.95, 0.99, \mathbf{0.999}$ \\
    \hline
    \texttt{MARS} $\beta_1$ & $0.9, \mathbf{0.95}$ & $0.9, \mathbf{0.95}$ \\
    \hline
    \texttt{MARS} $\beta_2$ & $0.95, \mathbf{0.99}$ & $0.95, \mathbf{0.99}$ \\
    \hline
    VR scaling factor $\eta$ & $0.023, 0.024, \mathbf{0.025}$ & $0.023, 0.024, \mathbf{0.025}$ \\
    \hline
    \end{tabular}
    \end{center}
\end{table}

\newpage

\begin{table}[h!]
    \centering
    \caption{\textbf{\texttt{MARS-Lion} hyperparameter tuning for our $\mathbf{124M}$ parameter large language models.}
    Bold hyperparameters are the best.}
    \label{tab:124m_marslion_hyperparams}
    \begin{center}
    \begin{tabular}{|c|c|c|}
    \hline
    \textbf{Hyperparameter} & \textbf{``Small'' batch setting} & \textbf{``Large'' batch setting}  \\ 
    \hline
    Learning rate \texttt{Lion} & $\mathbf{0.0001}, 0.0005, 0.001, 0.003$  & $\mathbf{0.0001}, 0.0005, 0.001, 0.003$ \\ 
    \hline
    Learning rate \texttt{MARS} & $\mathbf{0.0001}, 0.001, 0.003$  & $\mathbf{0.0001}, 0.001, 0.003$ \\ 
    \hline
    Batch size & $32$  & $256$ \\ 
    \hline
    Sequence length & $512$ & $512$ \\ 
    \hline
    Number of warmup steps & $2000, \mathbf{3000}$ & $\mathbf{2000}, 8000, 32000$ \\
    \hline
    Weight decay \texttt{MARS} & no, $\mathbf{0.1}$ & no, $\mathbf{0.1}, 0.5$ \\
    \hline
    Weight decay for $1$D layers & $0.1$ & $0.1$\\
    \hline
    Learning rate decay scheduler & \textbf{WSD}, cosine & \textbf{WSD}, cosine \\
    \hline
    Gradient clipping & $0.5$ & $0.5$ \\
    \hline
    Optimizer for $1$D layers & \texttt{Lion} & \texttt{Lion} \\
    \hline
    Optimizer for $1$D layers $\beta_1$ & $0.8, \mathbf{0.9}$ & $0.8, \mathbf{0.9}, 0.95$ \\
    \hline
    Optimizer for $1$D layers $\beta_2$ & $0.95, \mathbf{0.99}, 0.999$ & $0.95, \mathbf{0.99}, 0.999$ \\
    \hline
    \texttt{MARS} $\beta_1$ & $0.9, \mathbf{0.95}$ & $0.9, \mathbf{0.95}$ \\
    \hline
    \texttt{MARS} $\beta_2$ & $0.95, \mathbf{0.99}$ & $0.95, \mathbf{0.99}$ \\
    \hline
    VR scaling factor $\eta$ & $0.024, \mathbf{0.025}$ & $0.024, \mathbf{0.025}$ \\
    \hline
    \end{tabular}
    \end{center}
\end{table}

\begin{table}[h!]
    \centering
    \caption{\textbf{\texttt{MARS-Shampoo} hyperparameter tuning for our $\mathbf{124M}$ parameter large language models.}
    Bold hyperparameters are the best.}
    \label{tab:124m_marsshampoo_hyperparams}
    \begin{center}
    \begin{tabular}{|c|c|c|}
    \hline
    \textbf{Hyperparameter} & \textbf{``Small'' batch setting} & \textbf{``Large'' batch setting}  \\ 
    \hline
    Learning rate \texttt{Shampoo} & $0.0001, 0.0005, \mathbf{0.001}, 0.003$  & $0.0001, 0.0005, \mathbf{0.001}, 0.003$ \\ 
    \hline
    Learning rate \texttt{MARS} & $0.001, \mathbf{0.003}$  & $0.001, \mathbf{0.003}$ \\ 
    \hline
    Batch size & $32$  & $256$ \\ 
    \hline
    Sequence length & $512$ & $512$ \\ 
    \hline
    Number of warmup steps & $2000, \mathbf{3000}$ & $\mathbf{2000}, 8000, 32000$ \\
    \hline
    Weight decay \texttt{MARS} & no, $\mathbf{0.1}$ & no, $\mathbf{0.1}, 0.5$ \\
    \hline
    Weight decay for $1$D layers & $0.1$ & $0.1$\\
    \hline
    Learning rate decay scheduler & WSD, \textbf{cosine} & WSD, \textbf{cosine} \\
    \hline
    Gradient clipping & $0.5$ & $0.5$ \\
    \hline
    Optimizer for $1$D layers & \texttt{Shampoo} & \texttt{Shampoo} \\
    \hline
    Optimizer for $1$D layers $\beta_1$ & $0.8, \mathbf{0.9}$ & $0.8, \mathbf{0.9}, 0.95$ \\
    \hline
    Optimizer for $1$D layers $\beta_2$ & $0.95, 0.99, \mathbf{0.999}$ & $0.95, 0.99, \mathbf{0.999}$ \\
    \hline
    \texttt{MARS} $\beta_1$ & $0.9, \mathbf{0.95}$ & $0.9, \mathbf{0.95}$ \\
    \hline
    \texttt{MARS} $\beta_2$ & $0.95, \mathbf{0.99}$ & $0.95, \mathbf{0.99}$ \\
    \hline
    VR scaling factor $\eta$ & $0.024, \mathbf{0.025}$ & $0.024, \mathbf{0.025}$ \\
    \hline
    \end{tabular}
    \end{center}
\end{table}

\newpage
\subsection{$\mathbf{210M}$ parameters model}
\label{sec:ap_210mtuning}

For $210\mathbf{M}$ models we perform training runs only with the batch size of $256 \times 512$ tokens, utilizing the same training durations as for $124\mathbf{M}$ model with this batch size, i.e., $\{8\mathbf{k}, 16\mathbf{k}, 32\mathbf{k}, 48\mathbf{k}, 64\mathbf{k}, 128\mathbf{k}\}$, which corresponds to the following counts in tokens: $\{1\mathbf{B}, 2.1\mathbf{B}, 4.2\mathbf{B}, 6.3\mathbf{B}, 8.4\mathbf{B}, 16.8\mathbf{B}\}$.

We also replicate almost identical hyperparameters to those of the training of the $124\mathbf{M}$ model to verify whether the smooth transition \cref{tkw:smooth_transition} in the final ranking of optimizers and their sensitivity to hyperparameters will be observed. 

\begin{table}[h!]
    \centering
    \caption{\textbf{\texttt{AdamW} hyperparameter tuning for our $\mathbf{210M}$ parameter large language models.}
    Bold hyperparameters are the best.}
    \label{tab:210m_adamw_hyperparams}
    \begin{center}
    \begin{tabular}{|c|c|}
    \hline
    \textbf{Hyperparameter} & \textbf{``Large'' batch setting}  \\ 
    \hline
    Learning rate & $0.001$ \\ 
    \hline
    Batch size & $256$ \\ 
    \hline
    Sequence length & $512$ \\ 
    \hline
    Number of warmup steps & $50, 500,1000,\mathbf{2000}$ \\
    \hline
    Weight decay & no, $\mathbf{0.1}$ \\
    \hline
    Learning rate decay scheduler & WSD, \textbf{cosine}, linear \\
    \hline
    Gradient clipping & $0.5$ \\
    \hline
    \texttt{AdamW} $\beta_1$ & $0.8, \mathbf{0.9}$ \\
    \hline
    \texttt{AdamW} $\beta_2$ & $0.95, 0.99, \mathbf{0.999}, 0.9999$ \\
    \hline
    Final learning rate $\mathrm{X} \times \text{max cosine LR}$ & $10^{-1}$, $\mathbf{10^{-2}}$, $10^{-3}$, $10^{-4}$, $10^{-5}$, $10^{-6}$ \\
    \hline
    Final learning rate $\mathrm{X} \times \text{max WSD LR}$ & $10^{-1}$, $\mathbf{10^{-2}}$, $10^{-3}$, $10^{-4}$, $10^{-5}$, $10^{-6}$ \\
    \hline
    Final learning rate $\mathrm{X} \times \text{max linear LR}$ & $10^{-1}$, $10^{-2}$, $\mathbf{10^{-3}}$, $10^{-4}$, $10^{-5}$, $10^{-6}$ \\
    \hline
    \end{tabular}
    \end{center}
\end{table}

\begin{table}[h!]
    \centering
    \caption{\textbf{\texttt{ADOPT} hyperparameter tuning for our $\mathbf{210M}$ parameter large language models.}
    Bold hyperparameters are the best.}
    \label{tab:210m_adopt_hyperparams}
    \begin{center}
    \begin{tabular}{|c|c|}
    \hline
    \textbf{Hyperparameter} & \textbf{``Large'' batch setting}  \\ 
    \hline
    Learning rate & $0.001$ \\ 
    \hline
    Batch size &  $256$ \\ 
    \hline
    Sequence length & $512$ \\ 
    \hline
    Number of warmup steps &  $2000$ \\
    \hline
    Weight decay & $0.1$ \\
    \hline
    Learning rate decay scheduler &  cosine \\
    \hline
    Gradient clipping & no, $\mathbf{0.5}, 1$ \\
    \hline
    \texttt{ADOPT} $\beta_1$ &  $0.9$ \\
    \hline
    \texttt{ADOPT} $\beta_2$ & $0.5, \mathbf{0.999}, 0.9999$\\
    \hline
    \texttt{ADOPT} $\varepsilon$ & $10^{-3}, 10^{-4}, 10^{-5}, \mathbf{10^{-6}}, 10^{-7}, 10^{-8}, 10^{-9}, 10^{-10}$,  \\
    \hline
    \end{tabular}
    \end{center}
\end{table}

\begin{table}[h!]
    \centering
    \caption{\textbf{\texttt{AdEMAMix} hyperparameter tuning for our $\mathbf{210M}$ parameter large language models.}
    Bold hyperparameters are the best.}
    \label{tab:210m_ademamix_hyperparams}
    \begin{center}
    \begin{tabular}{|c|c|}
    \hline
    \textbf{Hyperparameter} & \textbf{``Large'' batch setting}  \\ 
    \hline
    Learning rate &  $0.001$  \\ 
    \hline
    Batch size & $256$  \\ 
    \hline
    Sequence length & $512$ \\ 
    \hline
    Number of warmup steps & $2000$ \\
    \hline
    Weight decay & $0.1$ \\
    \hline
    Learning rate decay scheduler & cosine \\
    \hline
    Gradient clipping & $0.5$ \\
    \hline
    \texttt{AdEMAMix} $\beta_1$ & $0.9$ \\
    \hline
    \texttt{AdEMAMix} $\beta_2$ &  $0.999$ \\
    \hline
    \texttt{AdEMAMix} $\beta_3$ & $0.999$ \\
    \hline
    \texttt{AdEMAMix} $\alpha$ & $8$ \\
    \hline
    \end{tabular}
    \end{center}
\end{table}

\newpage

\begin{table}[h!]
    \centering
    \caption{\textbf{\texttt{Lion} hyperparameter tuning for our $\mathbf{210M}$ parameter large language models.}
    Bold hyperparameters are the best.}
    \label{tab:210m_lion_hyperparams}
    \begin{center}
    \begin{tabular}{|c|c|}
    \hline
    \textbf{Hyperparameter} & \textbf{``Large'' batch setting}    \\ 
    \hline
    Learning rate &  $0.0001, \mathbf{0.0005}, 0.001$ \\ 
    \hline
    Batch size & $256$ \\ 
    \hline
    Sequence length & $512$ \\ 
    \hline
    Number of warmup steps & $2000$ \\
    \hline
    Weight decay &  $0.1$ \\
    \hline
    Learning rate decay scheduler & cosine \\
    \hline
    Gradient clipping &  $0.5$ \\
    \hline
    \texttt{Lion} $\beta_1$ &  $0.9$ \\
    \hline
    \texttt{Lion} $\beta_2$ & $0.99$ \\
    \hline
    \end{tabular}
    \end{center}
\end{table}

\begin{table}[h!]
    \centering
    \caption{\textbf{\texttt{Signum} hyperparameter tuning for our $\mathbf{210M}$ parameter large language models.}
    Bold hyperparameters are the best.}
    \label{tab:210m_signum_hyperparams}
    \begin{center}
    \begin{tabular}{|c|c|}
    \hline
    \textbf{Hyperparameter} & \textbf{``Large'' batch setting}  \\ 
    \hline
    Learning rate & $0.0001, \mathbf{0.0005}, 0.001$\\ 
    \hline
    Batch size & $256$ \\ 
    \hline
    Sequence length & $512$ \\ 
    \hline
    Number of warmup steps &  $2000$\\
    \hline
    Weight decay & $0.1$\\
    \hline
    Learning rate decay scheduler & cosine\\
    \hline
    Gradient clipping & $0.5$ \\
    \hline
    Momentum & $0.9, \mathbf{0.95}, 0.99$\\
    \hline
    Nesterov momentum & yes \\
    \hline
    \end{tabular}
    \end{center}
\end{table}

\begin{table}[h!]
    \centering
    \caption{\textbf{\texttt{Muon} hyperparameter tuning for our $\mathbf{210M}$ parameter large language models.}
    Bold hyperparameters are the best.}
    \label{tab:210m_muon_hyperparams}
    \begin{center}
    \begin{tabular}{|c|c|}
    \hline
    \textbf{Hyperparameter} & \textbf{``Large'' batch setting}  \\ 
    \hline
    Learning rate \texttt{AdamW} & $0.001$\\ 
    \hline
    Learning rate \texttt{Muon} & $0.01$\\ 
    \hline
    Batch size & $256$ \\ 
    \hline
    Sequence length & $512$ \\ 
    \hline
    Number of warmup steps &  $2000$ \\
    \hline
    Weight decay &  $0.1$ \\
    \hline
    Learning rate decay scheduler & cosine \\
    \hline
    Gradient clipping &  $0.5$ \\
    \hline
    Momentum \texttt{Muon} & $0.95$ \\
    \hline
    Optimizer for $1$D layers  & \texttt{AdamW} \\
    \hline
    Optimizer for $1$D layers, $\beta_1$ & $0.8$ \\
    \hline
    Optimizer for $1$D layers, $\beta_2$ & $0.999$ \\
    \hline
    Newton-Schulz a & $3.4445$ \\
    \hline
    Newton-Schultz b & $-4.7750$ \\
    \hline
    Newton-Schultz c &  $2.0315$ \\
    \hline
    Nesterov momentum &  yes \\
    \hline
    \end{tabular}
    \end{center}
\end{table}

\newpage

\begin{table}[h!]
    \centering
    \caption{\textbf{\texttt{D-Muon} hyperparameter tuning for our $\mathbf{210M}$ parameter large language models.}
    Bold hyperparameters are the best.}
    \label{tab:210m_dmuon_hyperparams}
    \begin{center}
    \begin{tabular}{|c|c|}
    \hline
    \textbf{Hyperparameter} & \textbf{``Large'' batch setting}  \\ 
    \hline
    Learning rate &  $0.001$\\ 
    \hline
    Batch size &  $256$ \\ 
    \hline
    Sequence length &  $512$ \\ 
    \hline
    Number of warmup steps &  $2000$ \\
    \hline
    Weight decay &  $0.1$ \\
    \hline
    Learning rate decay scheduler & cosine \\
    \hline
    Gradient clipping &  $0.5$ \\
    \hline
    Momentum \texttt{D-Muon} & $0.95$ \\
    \hline
    Optimizer for $1$D layers  & \texttt{AdamW} \\
    \hline
    Optimizer for $1$D layers, $\beta_1$ & $\mathbf{0.8}, 0.9$ \\
    \hline
    Optimizer for $1$D layers, $\beta_2$ &  $0.99, \mathbf{0.999}$ \\
    \hline
    Newton-Schulz a &  $3.4445$ \\
    \hline
    Newton-Schultz b &  $-4.7750$ \\
    \hline
    Newton-Schultz c &  $2.0315$ \\
    \hline
    Newton-Schultz iterations & $5$ \\
    \hline
    Nesterov momentum & yes \\
    \hline
    \end{tabular}
    \end{center}
\end{table}

\begin{table}[h!]
    \centering
    \caption{\textbf{\texttt{SOAP} hyperparameter tuning for our $\mathbf{210M}$ parameter large language models.}
    Bold hyperparameters are the best.}
    \label{tab:210m_soap_hyperparams}
    \begin{center}
    \begin{tabular}{|c|c|}
    \hline
    \textbf{Hyperparameter} & \textbf{``Large'' batch setting}  \\ 
    \hline
    Learning rate &   $0.001$ \\ 
    \hline
    Batch size & $256$ \\ 
    \hline
    Sequence length & $512$ \\ 
    \hline
    Number of warmup steps &  $2000$ \\
    \hline
    Weight decay &  $0.1$ \\
    \hline
    Learning rate decay scheduler &  cosine \\
    \hline
    Gradient clipping &  $0.5$ \\
    \hline
        Preconditioner dimension & $10000$ \\
    \hline
    Preconditioning frequency &  $10$ \\
    \hline
    \texttt{SOAP} $\beta_1$ & $0.9$ \\
    \hline
    \texttt{SOAP} $\beta_2$ & $0.999$\\
    \hline
    \end{tabular}
    \end{center}
\end{table}

\begin{table}[h!]
    \centering
    \caption{\textbf{\texttt{Sophia} hyperparameter tuning for our $\mathbf{210M}$ parameter large language models.}
    Bold hyperparameters are the best.}
    \label{tab:210m_sophia_hyperparams}
    \begin{center}
    \begin{tabular}{|c|c|}
    \hline
    \textbf{Hyperparameter} & \textbf{``Large'' batch setting}  \\ 
    \hline
    Learning rate &  $0.001$\\ 
    \hline
    Batch size & $256$ \\ 
    \hline
    Sequence length & $512$ \\ 
    \hline
    Number of warmup steps & $2000$\\
    \hline
    Weight decay &  $0.1$ \\
    \hline
    Learning rate decay scheduler & cosine \\
    \hline
    Gradient clipping & $0.5$ \\
    \hline
    Estimator & Gauss-Newton-Bartlett \\
    \hline
    Estimator frequency & $10$ \\
    \hline
    \texttt{Sophia} $\beta_1$ &  $0.9$ \\
    \hline
    \texttt{Sophia} $\beta_2$ & $0.999$ \\
    \hline
    \texttt{Sophia} $\rho$ & $0.04$ \\
    \hline
    \end{tabular}
    \end{center}
\end{table}

\newpage

\begin{table}[h!]
    \centering
    \caption{\textbf{\texttt{Schedule-Free AdamW} hyperparameter tuning for our $\mathbf{210M}$ parameter large language models.}
    Bold hyperparameters are the best.}
    \label{tab:210m_sfadamw_hyperparams}
    \begin{tabular}{|c|c|}
    \hline
    \textbf{Hyperparameter} & \textbf{``Large'' batch setting}  \\ 
    \hline
    Learning rate &  $0.001$  \\ 
    \hline
    Batch size & $256$ \\ 
    \hline
    Sequence length & $512$ \\ 
    \hline
    Number of warmup steps & $2000, \mathbf{8000}$ \\
    \hline
    Weight decay & $0.1$ \\
    \hline
    Learning rate decay scheduler & no \\
    \hline
    Gradient clipping & $0.5$ \\
    \hline
    \texttt{Schedule-Free AdamW} $\beta_1$ & $0.9$\\
    \hline
    \texttt{Schedule-Free AdamW} $\beta_2$ &  $0.9999$ \\
    \hline
    \end{tabular}
\end{table}

\begin{table}[h!]
    \centering
    \caption{\textbf{\texttt{Prodigy} hyperparameter tuning for our $\mathbf{210M}$ parameter large language models.}
    Bold hyperparameters are the best.}
    \label{tab:210m_prodigy_hyperparams}
    \begin{center}
    \begin{tabular}{|c|c|}
    \hline
    \textbf{Hyperparameter} &  \textbf{``Large'' batch setting}  \\ 
    \hline
    Learning rate &  $1$ \\ 
    \hline
    Batch size &  $256$ \\ 
    \hline
    Sequence length &  $512$ \\ 
    \hline
    Number of warmup steps &  $2000$ \\
    \hline
    Weight decay & $0.1$ \\
    \hline
    Learning rate decay scheduler &  cosine \\
    \hline
    Gradient clipping & $0.5$ \\
    \hline
    \texttt{Prodigy} $\beta_1$ &  $0.9$ \\
    \hline
    \texttt{Prodigy} $\beta_2$ &  $0.999$\\
    \hline
    \texttt{Prodigy} bias correction & yes \\
    \hline
    \end{tabular}
    \end{center}
\end{table}

\begin{table}[h!]
    \centering
    \caption{\textbf{\texttt{MARS} (\texttt{MARS-AdamW}) hyperparameter tuning for our $\mathbf{210M}$ parameter large language models.}
    Bold hyperparameters are the best.}
    \label{tab:210m_mars_hyperparams}
    \begin{center}
    \begin{tabular}{|c|c|}
    \hline
    \textbf{Hyperparameter} & \textbf{``Large'' batch setting}  \\ 
    \hline
    Learning rate \texttt{AdamW} &  $0.001$ \\ 
    \hline
    Learning rate \texttt{MARS} &  $0.003$ \\ 
    \hline
    Batch size &  $256$ \\ 
    \hline
    Sequence length & $512$ \\ 
    \hline
    Number of warmup steps & $2000$ \\
    \hline
    Weight decay \texttt{MARS} &  $0.1$ \\
    \hline
    Weight decay for $1$D layers & $0.1$\\
    \hline
    Learning rate decay scheduler &  cosine \\
    \hline
    Gradient clipping & $0.5$ \\
    \hline
    Optimizer for $1$D layers & \texttt{AdamW} \\
    \hline
    Optimizer for $1$D layers $\beta_1$ &  $\mathbf{0.8}, 0.9$ \\
    \hline
    Optimizer for $1$D layers $\beta_2$ &  $0.999$ \\
    \hline
    \texttt{MARS} $\beta_1$ & $0.95$ \\
    \hline
    \texttt{MARS} $\beta_2$ &  $0.99$ \\
    \hline
    VR scaling factor $\eta$ & $0.024, \mathbf{0.025}$ \\
    \hline
    \end{tabular}
    \end{center}
\end{table}

\newpage
\subsection{$\mathbf{583M}$ parameters model}
\label{sec:ap_600mtuning}

For models of $583\mathbf{M}$ scale, we ablate the difference between our setup and the one from Vyas et al.~\cite{vyas2024soapimprovingstabilizingshampoo}.
The main changes compared to our setup include: learning rate decay down to $10\%$ of the maximum, usage of $z$-loss regularizer in addition to the cross-entropy loss, smaller decoupled weight decay of $0.0001$.
We also point out that \texttt{SOAP} performance in \cite{vyas2024soapimprovingstabilizingshampoo} was measured on the Chinchilla optimal number of tokens and with $2\mathbf{M}$ tokens batch size.
Thus, in \cref{sec:scaling_up} we ablate the differences between our settings on the same training horizons.
A complete list of hyperparameters used for our \texttt{AdamW} and \texttt{SOAP} models in this ablations are presented in \cref{tab:600m_adamw_hyperparams}.

\begin{table}[h!]
    \centering
    \caption{\textbf{\texttt{AdamW} hyperparameter tuning for our $\mathbf{583M}$ parameter large language models.}
    Bold hyperparameters are the best.}
    \label{tab:600m_adamw_hyperparams}
    \begin{center}
    \begin{tabular}{|c|c|}
    \hline
    \textbf{Hyperparameter} & \textbf{$2\mathbf{M}$ batch setting}  \\ 
    \hline
    Learning rate & $0.001, \mathbf{0.005}$ \\ 
    \hline
    Batch size & $3936$ \\ 
    \hline
    Sequence length & $512$ \\ 
    \hline
    Number of warmup steps & $1200$ \\
    \hline
    Weight decay & $0.0001, \mathbf{0.1}$ \\
    \hline
    Learning rate decay scheduler & cosine \\
    \hline
    Gradient clipping & $0.5$ \\
    \hline
    \texttt{AdamW} $\beta_1$ & $0.9, \mathbf{0.95}$ \\
    \hline
    \texttt{AdamW} $\beta_2$ & $\mathbf{0.95}, 0.99$ \\
    \hline
    Final learning rate $\mathrm{X} \times \text{max cosine LR}$ &  $10^{-1}, \mathbf{10^{-2}}$\\
    \hline
    $z$-loss regularization & \textbf{no}, $0.0001$ \\
    \hline
    \end{tabular}
    \end{center}
\end{table}

\begin{table}[h!]
    \centering
    \caption{\textbf{\texttt{SOAP} hyperparameter tuning for our $\mathbf{583M}$ parameter large language models.}
    Bold hyperparameters are the best.}
    \label{tab:600m_soap_hyperparams}
    \begin{center}
    \begin{tabular}{|c|c|}
    \hline
    \textbf{Hyperparameter} & \textbf{$2\mathbf{M}$ batch setting}  \\ 
    \hline
    Learning rate &   $0.001, \mathbf{0.005}$ \\ 
    \hline
    Batch size & $3936$ \\ 
    \hline
    Sequence length & $512$ \\ 
    \hline
    Number of warmup steps &  $1200$ \\
    \hline
    Weight decay &  $0.0001, \mathbf{0.1}$ \\
    \hline
    Learning rate decay scheduler &  cosine \\
    \hline
    Gradient clipping &  $0.5$ \\
    \hline
        Preconditioner dimension & $10000$ \\
    \hline
    Preconditioning frequency &  $10$ \\
    \hline
    \texttt{SOAP} $\beta_1$ & $0.9, \mathbf{0.95}$ \\
    \hline
    \texttt{SOAP} $\beta_2$ & $\mathbf{0.95}, 0.99, 0.999$\\
    \hline
    Final learning rate $\mathrm{X} \times \text{max cosine LR}$ &  $10^{-1}, \mathbf{10^{-2}}$\\
    \hline
    $z$-loss regularization & \textbf{no}, $0.0001$ \\
    \hline
    \end{tabular}
    \end{center}
\end{table}

\newpage
\subsection{$\mathbf{720M}$ parameters model}
\label{sec:ap_720mtuning}

In this section, we provide a complete information about the hyperparameter search for the largest models used in our benchmarking experiments.
Deriving insights from our ablations~(\cref{fig:ap_retuning_betas,fig:sf_betas,fig:prodigy_betas}) on the smaller scale, we suggest to re-tune beta parameters of optimizers as changing the training iterations---see~\cref{tkw:retuning_betas,tkw:betas_sensitivity} for this conclusions.

Tables below cover our tuning outcomes for all methods.
We highlight that, when training with large batches of $1\mathbf{M}$ tokens, we use the smaller number of iterations for our runs: $T \in \{8, 16, 48\}\mathbf{k}(\mathbf{B})$ steps~(tokens)---see~\cref{tab:training_horizons_1mbs}.
Thus, according to~\cref{tkw:retuning_betas}, we find that smaller $\beta_2$ parameter gives better results for \texttt{SOAP}, \texttt{D-Muon} (for $1$D parameters), and \texttt{Prodigy}. 

\begin{table}[h!]
    \centering
    \caption{\textbf{\texttt{AdamW} hyperparameter tuning for our $\mathbf{720M}$ parameter large language models.}
    Bold hyperparameters are the best.}
    \label{tab:720m_adamw_hyperparams}
    \begin{center}
    \begin{tabular}{|c|c|}
    \hline
    \textbf{Hyperparameter} & \textbf{$1\mathbf{M}$ batch setting}  \\ 
    \hline
    Learning rate & $0.0001, 0.0003, 0.0005, \mathbf{0.001}$ \\ 
    \hline
    Batch size & $1984$ \\ 
    \hline
    Sequence length & $512$ \\ 
    \hline
    Number of warmup steps & $2000$ \\
    \hline
    Weight decay & $0.1$ \\
    \hline
    Learning rate decay scheduler & cosine \\
    \hline
    Gradient clipping & $\mathbf{0.1}, 0.5$ \\
    \hline
    \texttt{AdamW} $\beta_1$ & $0.8, \mathbf{0.9}, 0.95$ \\
    \hline
    \texttt{AdamW} $\beta_2$ & $0.95, 0.99, \mathbf{0.999}$ \\
    \hline
    \end{tabular}
    \end{center}
\end{table}

\begin{table}[h!]
    \centering
    \caption{\textbf{\texttt{ADOPT} hyperparameter tuning for our $\mathbf{720M}$ parameter large language models.}
    Bold hyperparameters are the best.}
    \label{tab:720m_adopt_hyperparams}
    \begin{center}
    \begin{tabular}{|c|c|}
    \hline
    \textbf{Hyperparameter} & \textbf{$1\mathbf{M}$ batch setting}  \\ 
    \hline
    Learning rate & $0.001$ \\ 
    \hline
    Batch size &  $1984$ \\ 
    \hline
    Sequence length & $512$ \\ 
    \hline
    Number of warmup steps &  $2000$ \\
    \hline
    Weight decay & $0.1$ \\
    \hline
    Learning rate decay scheduler &  cosine \\
    \hline
    Gradient clipping & $0.1$ \\
    \hline
    \texttt{ADOPT} $\beta_1$ &  $0.9, \mathbf{0.95}$ \\
    \hline
    \texttt{ADOPT} $\beta_2$ & $0.95, 0.99, \mathbf{0.999}$\\
    \hline
    \texttt{ADOPT} $\varepsilon$ & $10^{-6}$  \\
    \hline
    \end{tabular}
    \end{center}
\end{table}

\begin{table}[h!]
    \centering
    \caption{\textbf{\texttt{AdEMAMix} hyperparameter tuning for our $\mathbf{720M}$ parameter large language models.}
    Bold hyperparameters are the best.}
    \label{tab:720m_ademamix_hyperparams}
    \begin{center}
    \begin{tabular}{|c|c|}
    \hline
    \textbf{Hyperparameter} & \textbf{$1\mathbf{M}$ batch setting}  \\ 
    \hline
    Learning rate &  $\mathbf{0.001}, 0.002$  \\ 
    \hline
    Batch size & $1984$  \\ 
    \hline
    Sequence length & $512$ \\ 
    \hline
    Number of warmup steps & $2000$ \\
    \hline
    Weight decay & $0.1$ \\
    \hline
    Learning rate decay scheduler & cosine \\
    \hline
    Gradient clipping & $0.1$ \\
    \hline
    \texttt{AdEMAMix} $\beta_1$ & $0.9$ \\
    \hline
    \texttt{AdEMAMix} $\beta_2$ &  $0.95, \mathbf{0.999}$ \\
    \hline
    \texttt{AdEMAMix} $\beta_3$ & $\mathbf{0.999}, 0.9999$ \\
    \hline
    \texttt{AdEMAMix} $\alpha$ & $8$ \\
    \hline
    \end{tabular}
    \end{center}
\end{table}

\newpage

\begin{table}[h!]
    \centering
    \caption{\textbf{\texttt{Lion} hyperparameter tuning for our $\mathbf{720M}$ parameter large language models.}
    Bold hyperparameters are the best.}
    \label{tab:720m_lion_hyperparams}
    \begin{center}
    \begin{tabular}{|c|c|}
    \hline
    \textbf{Hyperparameter} & \textbf{$1\mathbf{M}$ batch setting}    \\ 
    \hline
    Learning rate &  $0.00005, 0.0001, \mathbf{0.0002}, 0.0003, 0.0005, 0.001$ \\ 
    \hline
    Batch size & $1984$ \\ 
    \hline
    Sequence length & $512$ \\ 
    \hline
    Number of warmup steps & $2000$ \\
    \hline
    Weight decay &  $0.1$ \\
    \hline
    Learning rate decay scheduler & cosine \\
    \hline
    Gradient clipping &  $0.1, \mathbf{1}$ \\
    \hline
    \texttt{Lion} $\beta_1$ &  $0.9$ \\
    \hline
    \texttt{Lion} $\beta_2$ & $0.99$ \\
    \hline
    \end{tabular}
    \end{center}
\end{table}

\begin{table}[h!]
    \centering
    \caption{\textbf{\texttt{Signum} hyperparameter tuning for our $\mathbf{720M}$ parameter large language models.}
    Bold hyperparameters are the best.}
    \label{tab:720m_signum_hyperparams}
    \begin{center}
    \begin{tabular}{|c|c|}
    \hline
    \textbf{Hyperparameter} & \textbf{$1\mathbf{M}$ batch setting}  \\ 
    \hline
    Learning rate & $0.0001, \mathbf{0.0002}, 0.0003, 0.0005, 0.001$\\ 
    \hline
    Batch size & $1984$ \\ 
    \hline
    Sequence length & $512$ \\ 
    \hline
    Number of warmup steps &  $2000$\\
    \hline
    Weight decay & $0.1$\\
    \hline
    Learning rate decay scheduler & cosine\\
    \hline
    Gradient clipping & $\mathbf{0.1}, 1$ \\
    \hline
    Momentum & $0.9, \mathbf{0.95}, 0.99$\\
    \hline
    Nesterov momentum & yes \\
    \hline
    \end{tabular}
    \end{center}
\end{table}

\begin{table}[h!]
    \centering
    \caption{\textbf{\texttt{Muon} hyperparameter tuning for our $\mathbf{720M}$ parameter large language models.}
    Bold hyperparameters are the best.}
    \label{tab:720m_muon_hyperparams}
    \begin{center}
    \begin{tabular}{|c|c|}
    \hline
    \textbf{Hyperparameter} & \textbf{$1\mathbf{M}$ batch setting}  \\ 
    \hline
    Learning rate \texttt{AdamW} & $0.0005, \mathbf{0.001}, 0.002$\\ 
    \hline
    Learning rate \texttt{Muon} & $0.01$\\ 
    \hline
    Batch size & $1984$ \\ 
    \hline
    Sequence length & $512$ \\ 
    \hline
    Number of warmup steps &  $2000$ \\
    \hline
    Weight decay &  $0.1$ \\
    \hline
    Learning rate decay scheduler & cosine \\
    \hline
    Gradient clipping &  $0.1$ \\
    \hline
    Momentum \texttt{Muon} & $0.95$ \\
    \hline
    Optimizer for $1$D layers  & \texttt{AdamW} \\
    \hline
    Optimizer for $1$D layers, $\beta_1$ & $\mathbf{0.8}, 0.9, 0.95$ \\
    \hline
    Optimizer for $1$D layers, $\beta_2$ & $0.95, 0.99, \mathbf{0.999}$ \\
    \hline
    Newton-Schulz a & $3.4445$ \\
    \hline
    Newton-Schultz b & $-4.7750$ \\
    \hline
    Newton-Schultz c &  $2.0315$ \\
    \hline
    Nesterov momentum &  yes \\
    \hline
    \end{tabular}
    \end{center}
\end{table}

\newpage

\begin{table}[h!]
    \centering
    \caption{\textbf{\texttt{D-Muon} hyperparameter tuning for our $\mathbf{720M}$ parameter large language models.}
    Bold hyperparameters are the best.}
    \label{tab:720m_dmuon_hyperparams}
    \begin{center}
    \begin{tabular}{|c|c|}
    \hline
    \textbf{Hyperparameter} & \textbf{$1\mathbf{M}$ batch setting}  \\ 
    \hline
    Learning rate &  $0.0005, \mathbf{0.001}, 0.002, 0.003, 0.005$\\ 
    \hline
    Batch size &  $1984$ \\ 
    \hline
    Sequence length &  $512$ \\ 
    \hline
    Number of warmup steps &  $2000$ \\
    \hline
    Weight decay &  $0.1$ \\
    \hline
    Learning rate decay scheduler & cosine \\
    \hline
    Gradient clipping &  $0.1$ \\
    \hline
    Momentum \texttt{D-Muon} & $0.95$ \\
    \hline
    Optimizer for $1$D layers  & \texttt{AdamW} \\
    \hline
    Optimizer for $1$D layers, $\beta_1$ & $0.8, \mathbf{0.9}, 0.95$ \\
    \hline
    Optimizer for $1$D layers, $\beta_2$ &  0.95, $\mathbf{0.99}, 0.999$ \\
    \hline
    Newton-Schulz a &  $3.4445$ \\
    \hline
    Newton-Schultz b &  $-4.7750$ \\
    \hline
    Newton-Schultz c &  $2.0315$ \\
    \hline
    Newton-Schultz iterations & $5$ \\
    \hline
    Nesterov momentum & yes \\
    \hline
    \end{tabular}
    \end{center}
\end{table}

\begin{table}[h!]
    \centering
    \caption{\textbf{\texttt{SOAP} hyperparameter tuning for our $\mathbf{720M}$ parameter large language models.}
    Bold hyperparameters are the best.}
    \label{tab:720m_soap_hyperparams}
    \begin{center}
    \begin{tabular}{|c|c|}
    \hline
    \textbf{Hyperparameter} & \textbf{$1\mathbf{M}$ batch setting}  \\ 
    \hline
    Learning rate &   $0.001$ \\ 
    \hline
    Batch size & $1984$ \\ 
    \hline
    Sequence length & $512$ \\ 
    \hline
    Number of warmup steps &  $2000$ \\
    \hline
    Weight decay &  $0.1$ \\
    \hline
    Learning rate decay scheduler &  cosine \\
    \hline
    Gradient clipping &  $0.1$ \\
    \hline
        Preconditioner dimension & $10000$ \\
    \hline
    Preconditioning frequency &  $10$ \\
    \hline
    \texttt{SOAP} $\beta_1$ & $0.9, \mathbf{0.95}$ \\
    \hline
    \texttt{SOAP} $\beta_2$ & $\mathbf{0.95}, 0.99, 0.999$\\
    \hline
    \end{tabular}
    \end{center}
\end{table}

\begin{table}[h!]
    \centering
    \caption{\textbf{\texttt{Sophia} hyperparameter tuning for our $\mathbf{720M}$ parameter large language models.}
    Bold hyperparameters are the best.}
    \label{tab:720m_sophia_hyperparams}
    \begin{center}
    \begin{tabular}{|c|c|}
    \hline
    \textbf{Hyperparameter} & \textbf{$1\mathbf{M}$ batch setting}  \\ 
    \hline
    Learning rate &  $0.0001, \mathbf{0.0005}, 0.001$\\ 
    \hline
    Batch size & $1984$ \\ 
    \hline
    Sequence length & $512$ \\ 
    \hline
    Number of warmup steps & $2000$\\
    \hline
    Weight decay &  $0.1$ \\
    \hline
    Learning rate decay scheduler & cosine \\
    \hline
    Gradient clipping & $0.1$ \\
    \hline
    Estimator & Gauss-Newton-Bartlett \\
    \hline
    Estimator frequency & $10$ \\
    \hline
    \texttt{Sophia} $\beta_1$ &  $0.9, \mathbf{0.95}$ \\
    \hline
    \texttt{Sophia} $\beta_2$ & $0.95, 0.99, \mathbf{0.999}$ \\
    \hline
    \texttt{Sophia} $\rho$ & $0.04$ \\
    \hline
    \end{tabular}
    \end{center}
\end{table}

\newpage

\begin{table}[h!]
    \centering
    \caption{\textbf{\texttt{Schedule-Free AdamW} hyperparameter tuning for our $\mathbf{720M}$ parameter large language models.}
    Bold hyperparameters are the best.}
    \label{tab:720m_sfadamw_hyperparams}
    \begin{tabular}{|c|c|}
    \hline
    \textbf{Hyperparameter} & \textbf{$1\mathbf{M}$ batch setting}  \\ 
    \hline
    Learning rate &  $0.001$  \\ 
    \hline
    Batch size & $1984$ \\ 
    \hline
    Sequence length & $512$ \\ 
    \hline
    Number of warmup steps & $2000, \mathbf{8000}$ \\
    \hline
    Weight decay & $0.1$ \\
    \hline
    Learning rate decay scheduler & no \\
    \hline
    Gradient clipping & no, $\mathbf{0.1}$ \\
    \hline
    \texttt{Schedule-Free AdamW} $\beta_1$ & $\mathbf{0.9}, 0.95$\\
    \hline
    \texttt{Schedule-Free AdamW} $\beta_2$ &  $0.95, 0.99, 0.999, \mathbf{0.9999}$ \\
    \hline
    \end{tabular}
\end{table}

\begin{table}[h!]
    \centering
    \caption{\textbf{\texttt{Prodigy} hyperparameter tuning for our $\mathbf{720M}$ parameter large language models.}
    Bold hyperparameters are the best.}
    \label{tab:720m_prodigy_hyperparams}
    \begin{center}
    \begin{tabular}{|c|c|}
    \hline
    \textbf{Hyperparameter} &  \textbf{$1\mathbf{M}$ batch setting}  \\ 
    \hline
    Learning rate &  $0.5, \mathbf{1}, 2$ \\ 
    \hline
    Batch size &  $1984$ \\ 
    \hline
    Sequence length &  $512$ \\ 
    \hline
    Number of warmup steps &  $2000$ \\
    \hline
    Weight decay & $0.1$ \\
    \hline
    Learning rate decay scheduler &  cosine \\
    \hline
    Gradient clipping & $0.1$ \\
    \hline
    \texttt{Prodigy} $\beta_1$ &  $0.9, \mathbf{0.95}$ \\
    \hline
    \texttt{Prodigy} $\beta_2$ &  $0.95, \mathbf{0.99}, 0.999$\\
    \hline
    \texttt{Prodigy} bias correction & yes \\
    \hline
    \end{tabular}
    \end{center}
\end{table}

\begin{table}[h!]
    \centering
    \caption{\textbf{\texttt{MARS} (\texttt{MARS-AdamW}) hyperparameter tuning for our $\mathbf{720M}$ parameter large language models.}
    Bold hyperparameters are the best.}
    \label{tab:720m_mars_hyperparams}
    \begin{center}
    \begin{tabular}{|c|c|}
    \hline
    \textbf{Hyperparameter} & \textbf{$1\mathbf{M}$ batch setting}  \\ 
    \hline
    Learning rate \texttt{AdamW} &  $0.001$ \\ 
    \hline
    Learning rate \texttt{MARS} &  $0.003$ \\ 
    \hline
    Batch size &  $1984$ \\ 
    \hline
    Sequence length & $512$ \\ 
    \hline
    Number of warmup steps & $2000$ \\
    \hline
    Weight decay \texttt{MARS} &  $0.1$ \\
    \hline
    Weight decay for $1$D layers & $0.1$\\
    \hline
    Learning rate decay scheduler &  cosine \\
    \hline
    Gradient clipping & $0.1$ \\
    \hline
    Optimizer for $1$D layers & \texttt{AdamW} \\
    \hline
    Optimizer for $1$D layers $\beta_1$ &  $\mathbf{0.8}, 0.9, 0.95$ \\
    \hline
    Optimizer for $1$D layers $\beta_2$ &  $0.95, 0.99, \mathbf{0.999}$ \\
    \hline
    \texttt{MARS} $\beta_1$ & $0.95$ \\
    \hline
    \texttt{MARS} $\beta_2$ &  $0.99$ \\
    \hline
    VR scaling factor $\eta$ & $0.024, \mathbf{0.025}$ \\
    \hline
    \end{tabular}
    \end{center}
\end{table}

\newpage
\subsection{$\mathbf{520M}$ parameters MoE model}
\label{sec:ap_520moetuning}

We extend our comparison of optimizers beyond dense models to include Mixture of Experts (MoE) architectures.  
Starting from our Llama-like transformer with tied embeddings, we construct an MoE variant following the Switch-Transformer implementation~\cite{fedus2022switchtransformersscalingtrillion}.  
The model employs classical linear gating with softmax and top-$k$ routing ($k=2$) over $8$ experts.  
We retain the SwiGLU activation functions~\cite{shazeer2020gluvariantsimprovetransformer}, RMSNorm layers~\cite{zhang2019rootmeansquarelayer}, and RoPE embeddings~\cite{su2023roformerenhancedtransformerrotary} exactly as in our dense LLMs.  
Keeping the same hidden size, number of layers, and attention heads as the $124\mathbf{M}$ dense model, this results in a $\sim 520\mathbf{M}$ parameter MoE architecture.  
A detailed specification of this model is provided in \cref{tab:moe_models}.

\begin{table}[h!]
    \centering
    \caption{\textbf{Configurations for our Llama-based MoE model.}}
    \label{tab:moe_models}
    \begin{tabular}{|c|c|}
    \hline
    \textbf{\# Parameters} & $\mathbf{520}\mathbf{M}$  \\ 
    \hline
    Hidden size & $768$ \\ 
    \hline
    \# Attention heads & $12$ \\ 
    \hline
    \# Layers & $12$  \\ 
    \hline
     Init. std & $0.02$ \\
    \hline
    Use bias & no \\
    \hline
    RMSNorm epsilon & $0.00001$  \\
    \hline
    Positional encoding & RoPE  \\
    \hline
    MoE router loss & load balancing loss~\cite{fedus2022switchtransformersscalingtrillion}~(Eq. $4$) \& router $z$-loss~\cite{zoph2022stmoedesigningstabletransferable}~(Eq. $5$)\\
    \hline
    \# Experts per layer & $8$ \\
    \hline
    \# Shared experts & $0$ \\
    \hline
    Top-$k$ routing ($k$) & $2$ \\
    \hline
    MoE softmax order & top-$k$ $\rightarrow$ softmax \\
    \hline
    \end{tabular}
\end{table}

For training, we use a batch size of $256 \times 512$.  
Optimizer hyperparameters are taken directly from \cref{sec:ap_210mtuning}, with one adjustment: the learning rate for \texttt{Sophia} is set to $0.0005$ instead of $0.001$.  
The purpose of this ablation is to evaluate how optimizers, tuned on dense models, perform when directly transferred to MoE models.  
In practical scenarios, practitioners often reuse well-established hyperparameters tuned on dense LLMs; hence, we argue that our comparison on the $520\mathbf{M}$ MoE model reflects realistic small-scale deployment settings.

We report our configurations for training runs in \cref{tab:training_horizons_moe}.

\begin{table}[h!]
    \centering
    \caption{\textbf{Lengths of training for the MoE model in ``Large'' batch size setting ($\mathbf{256 \times 512}$).}}
    \label{tab:training_horizons_moe}
    \begin{tabular}{|c|c|c|c|}
    \hline
    \textbf{\# Parameters} & \multicolumn{2}{c|}{\textbf{Tokens (Iterations)}} & \textbf{Chinchilla Tokens}\\
    \hline
    $520\mathbf{M}$ & $5.5\mathbf{B}$ ($42\mathbf{k}$) & $44\mathbf{B}$ ($336\mathbf{k}$) &  $10.4\mathbf{B}$ \\
    \hline
    \end{tabular}
\end{table}


\end{document}